\documentclass[10pt,twocolumn,letterpaper]{article}
\usepackage{cvpr}
\usepackage{times}
\usepackage{epsfig}
\usepackage{graphicx}
\usepackage{amsmath}
\usepackage{amssymb}
\usepackage{subcaption}

\usepackage[pagebackref=true,breaklinks=true,letterpaper=true,colorlinks,bookmarks=false]{hyperref}

\cvprfinalcopy 

\def\cvprPaperID{3991} 

\ifcvprfinal\fi
\begin{document}

\title{ Online Clustering by Penalized Weighted GMM}

\author{Shlomo Bugdary\\
Intel\\
{\tt\small shlomi.bugdary@intel.com}
\and
Shay Maymon\\
Intel\\
{\tt\small shay.maymon@intel.com}
}

\maketitle

%

\begin{abstract}
With the dawn of the  Big Data era, data sets are growing rapidly. Data is streaming from everywhere - from cameras, mobile phones, cars, and other electronic devices. 
Clustering streaming data is a very challenging problem. Unlike the traditional clustering algorithms where the dataset can be stored and scanned multiple times, clustering streaming data has to satisfy constraints such as limit memory size, real-time response, unknown data statistics and an unknown number of clusters. In this paper, we present a novel online clustering algorithm which can be used to cluster streaming data without knowing the number of clusters a priori. Results on both synthetic and real datasets show that the proposed algorithm produces partitions which are close to what you could get if you clustered the whole data at one time.

\end{abstract}

\section{Introduction}
Clustering is the task of grouping objects to structures (also called clusters) that maintain internal homogeneity and external separation. Objects in the same cluster should be similar to each other while objects in different clusters should not.  In past systems, the incoming data was stored on the memory and the clustering task was performed offline. However, modern systems require clustering of streaming data as well as clustering of large datasets which may be treated as streaming data. 

Clustering streaming data is a very challenging problem. Unlike traditional clustering algorithms where the dataset can be stored and scanned multiple times, clustering streaming data has to satisfy constraints such as limited memory, real-time response, unknown data statistics and a dynamic number of clusters. For example, imagine the task of segmentation by clustering of a 4K image (typical resolution of $4096 \times 2160$).  Creating an affinity matrix for traditional clustering algorithms like Normalized Cuts \cite{Shi:2000:NCI:351581.351611} and Affinity Propagation  \cite{Frey07clusteringby} is impossible because  the size of the matrix, $(4096 \times  2160) \times (4096 \times  2160)$,  cannot fit the memory. Applying other traditional algorithms like  K means  \cite{Lloyd82leastsquares}, DBSCAN \cite{ester1996density} and Meanshift \cite{Fukunaga:2006:EGD:2263309.2268796} is also limited since parameters selections for those algorithms require multiple scans of the dataset - a computationally expensive task.

A good clustering algorithm for streaming data should be able to identify the underlying structure of the data in a dynamic environment. It should respond quickly to the changing statistics of the data, be robust to outliers and process the data in real-time. It should also perform as good as if it had the global view of the data \cite{toshniwal2013clustering}. 

There are two approaches to process the incoming data: the incremental learning approach and the two-phase learning approach \cite{nguyen2015survey}. Assuming the data arrives in batches that is, $n_1$ data points arrive at interval $t_1$, $n_2$ at $t_2$ and so on. In the incremental learning approach, the clustering model is incrementally updated after each batch is processed. In this approach, the clustering result is available immediately after the arrival of the last batch but a batch processing is computationally expensive. Unlike the incremental learning approach, the two-phase approach divides the learning process into two phases: online and offline. In the online phase, the arriving data is summarized and only prototypes of each batch are saved to the memory. Then, in the offline phase, after the last batch arrives, clustering is performed on the prototypes of all the batches to create the final clustering model. This approach is able to cluster high-speed data streams.

In this paper, we propose a novel online clustering algorithm based on the two-phase framework. This algorithm, which is based on a Penalized Weighted Gaussian Mixtures Model (PWG), clusters the data as it streams without knowing the number of clusters apriori. By adding a penalty function on the mixing coefficients to the traditional Gaussian Mixtures Model the algorithm can estimate the number of clusters in the dataset dynamically. Moreover, by using a Gaussian Mixtures Model, the algorithm can model any probability density functions to any required level of accuracy.  

One possible application of the algorithm is online segmentation of images in modern cameras. Usually, modern cameras are equipped with an Image Signal Processor (ISP) that processes the raw data from the CCD sensor. Common blocks in such ISP are demosaicing, noise reduction and image sharpening. Every block processes the raw data in parts.  Based on the proposed method,  a segmentation block with hardware and software phases can be designed.  In the first phase, the hardware segments only part of the raw input data by using the PWG model saving weighted prototypes to the DDR. In the second phase, the software clusters the weighted prototypes to create the final segmentation result. 

The rest of this paper is organized as follows.  Section \ref{sec:Related Work}  surveys various online clustering methods that are based on the two-phase framework. Section \ref{sec:opwg} presents the proposed algorithm -  Online Clustering by Penalized Weighted Gaussian Mixtures Model (OPWG). Section \ref{sec:results} presents the algorithm evaluation on both synthetic and real datasets and discusses its result and section \ref{sec:conc} concludes the paper. 

\section{Related Work}
\label{sec:Related Work}

Many algorithms use the two-phase framework to extend traditional clustering algorithms to streaming data. For example, the STREAM \cite{o2002streaming} algorithm is based on the classical k-median algorithm \cite{Jain:1988:ACD:46712}. It consists of two phases. Given the streaming data is divided into buckets, in the first phase, the algorithm finds $K$ clusters in each bucket by the k-median clustering algorithm. For each bucket, it stores the clusters' centroids and their corresponding weights. In the second phase, the weighted centroids are clustered into a small number of clusters, resulting in the final data partition.

Similar to the STREAM algorithm, the Online Fuzzy C-means (OFCM) \cite{hore2008online} is based on the classical algorithm Fuzzy C-means \cite{Bezdek:1981:PRF:539444}  and it also has two phases. Assuming the data arrives in chunks, in the online phase, the arriving data is being clustered by the FCM algorithm. The clustering result is a set of centroids representing each chunk and their corresponding weights. In the post-processing phase, a weighted FCM algorithm is applied to cluster the centroids considering their weights.

Although those algorithms perform well, they have some limitations: STREAM  and OFCM need to be provided with the number of clusters, $K$, which may vary from batch to batch. Choosing a wrong $K$ may affect the algorithm as following - when all data in a given chunk comes from only one class - an over-segmentation is created. Such over segmentation does not result in information loss but increases the algorithm sensitivity to noise and its convergence time. When all the data in a given chunk comes from multiple classes, an under segmentation may occur depending on the value of $K$. Another limitation of those algorithms is the creation of spherical clusters due to the objective function of those algorithms. Creating spherical clusters can cause wrong clustering result since the clusters may have arbitrary shapes. 

A hierarchical approach for clustering streaming data is proposed by the CluStream algorithm \cite{Aggarwal:2003:FCE:1315451.1315460}. The CluStream algorithm extends the notion of a feature vector of the BIRCH \cite{Zhang:1996:BED:235968.233324} method to create micro-clusters in the online phase. In the offline phase, the micro-clusters are clustered into bigger clusters using the K-means algorithm. Like the above methods, the number of clusters should be provided to this algorithm.

A density-based online clustering method is the DenStream \cite{cao2006density}.  The DenStream algorithm extends the notion of core points of the  DBSCAN algorithm to a new concept of micro-clusters. It also has two phases. In the online phase, the algorithm maintains micro clusters structures which approximately capture the density of the data stream. In order to create the final data partition, in the offline phase, a variant of the DBSCAN algorithm is applied to those clusters. The DenStream algorithm should be provided with a cluster radius threshold and data fading rate. Providing the cluster radius is a main drawback of the algorithm. As stated, in streaming data, the data statistics may vary along the time and therefore a single cluster radius may not fit all the data.

\section{Online Clustering by Penalized Weighted GMM}
\label{sec:opwg}

Gaussian Mixtures Model (GMM) is usually the model of choice in distribution-based clustering.  In this approach, the data objects are assumed to be generated according to a mixture of Gaussians and the clusters are defined as objects belonging to the most likely Gaussian. Therefore, finding the clusters of a given data is equivalent to estimating the parameters of the GMM. 

The classical GMM regards the importance of each object in the dataset equally and the clustering result is equally affected by all objects. More generally, different objects may have different impacts on the dataset. An object may have a weight that represents its "mass". The larger the weight the larger the dominance of the object. In such a case, the clustering result will be biased by those objects with high dominance. Such an approach takes place in the second phase of STREAM and OFCM, where the algorithms solve a weighted clustering problem. 

The OPWG algorithm works similarly. Assume a streaming data arrives and is processed in batches so that $n_1$ data points arrive at time $t_1$, $n_2$ at $t_2$ and so on, OPWG clusters the data in two phases. In the first phase, the algorithm clusters each batch based on a degenerated PWG model where all the weights equal to 1, unless there is a prior knowledge on the data. Each batch is clustered to $K < K_{max}$ Gaussians mixture, whose centroids and mixing coefficients are saved. The number of clusters is found dynamically starting from $K_{max}$ and converging to a smaller value $K$. The mixing coefficient acts as a weight - the larger the coefficient - the more dominant its corresponding centroid is. 
In the second phase, the algorithm clusters the weighted centroids from all the batches to get the final clustering results. As in the first phase, this phase also estimates the number of clusters. The PWG model is discussed next.

\subsection{Penalized Weighted GMM} 

The OPWG algorithm is based on a penalized weighted GMM, which will be defined next. Let ${x_i} \in \mathbb{R}^{d}$ be a random vector following a multivariate Gaussian distribution with  mean $\mu \in \mathbb{R}^{d} $ and a covariance matrix $\Sigma \in \mathbb{R}^{d \times d} $. Let $w_i>0$ be a weight indicating the dominanace of an observation ${x_i}$. As suggested in \cite{gebru2015algorithms} the weights $w_i$ are incorporated into the regular GMM model by "observing ${x_i}$, $w_i$ times", resulting with the following PDF of weighted GMM:
\begin{equation}
p(x_i;\Theta ,w_i) = \sum_{k=1}^K {\pi_k N(x_i;\mu_k,\frac{1}{w_i}\Sigma_k) },
\label{eq:gmm}
\end{equation}

where $\Theta={\{\pi_1,...,\pi_K;\theta_1,...,\theta_K\}}$ are the mixtures parameters: $\{\pi_1,...,\pi_k\}$ are the mixing coefficients satisfying $\sum_1^K \pi_k =1$ and $\theta_k=\{\mu_k,\Sigma_k\}$ are the parameters of the $k^{th}$ component. $K$ is the number of components in the model and it corresponds to the number of clusters in the data. Let $X=\{x_1,...,x_N\}$ and $W=\{w_1,...,w_N\}$  be the i.i.d observed data and its corresponding weights, respectively,  the observed data log-likelihood is:
\begin{equation}
l(\Theta) = \sum_{i=1}^N {\ln(\sum_{k=1}^K {\pi_k N(x_i;\mu_k,\frac{1}{w_i}\Sigma_k)})}.
\label{eq:ll}
\end{equation}

Traditional clustering algorithms require the user to provide the number of clusters. In cases where this information is unavailable,  it is common to run the algorithm several times with different values of $K$ and to choose the one that minimizes some information criteria such as AIC \cite{akaike1998information} or BIC \cite{schwarz1978estimating}. In online clustering, such approach of model selection is not applicable. 
We therefore consider a different approach, in which we start with a large number $K_{max}$ of components and shrink insignificant mixing coefficients to zero to preserve a suitable number $K$ of components. This elimination can be obtained by adding a penalty to the objective function on the mixing coefficients. It is shown in \cite{PenHuaZha16} that good results can be achieved  by choosing a penalty of the form of $\ln{\frac{\epsilon + \pi_k}{\epsilon}} = \ln{(\epsilon + \pi_k)} - \ln{\epsilon} $, where $\epsilon$ is a small positive number. This penalty is a monotonically increasing function of $\pi_k$, and it goes to zero when a mixing coefficient, $\pi_k$, goes to zero. 

Considering this penalty, the penalized log-likelihood function becomes:

\begin{equation}
l_p(\Theta) = l(\Theta) - N \lambda D_f \lbrack \sum_{k=1}^K {\ln{(\epsilon + \pi_k)} - \ln{\epsilon}} \rbrack,
\label{eq:llp}
\end{equation}

where $l(\theta)$ is the log-likelihood, $\lambda$ is a tuning parameter and $D_f$ is the number of free parameters of each mixture component. Specifically, for GMM with a full covariance matrix, $D_f = \frac{d(d+1)}{2}+d+1$, and for GMM with a diagonal covariance matrix, $D_f =d+d+1$. Integrating \eqref{eq:ll} and \eqref{eq:llp} results in a penalized weighted GMM log likelihood function:

\begin{equation}
\begin{split}
l_{pw}(\Theta) &=  \sum_{i=1}^N {\ln(\sum_{k=1}^K {\pi_k N(x_i;\mu_k,\frac{1}{w_i}\Sigma_k)})}\\ 
&- N \lambda D_f \lbrack \sum_{k=1}^K {\ln{(\epsilon + \pi_k)} - \ln{\epsilon}} \rbrack.
\end{split}
\label{eq:llpw}
\end{equation}

In order to estimate the model parameters, $\Theta$, we maximize the penalized log-likelihood function.  However, as in mixtures model, a direct maximization of the penalized log-likelihood function is analytically intractable. We therefore use the iterative EM  algorithm \cite{dempster1977maximum} to find a solution. 

\subsection{EM algorithm}

Let us denote by  $Z=\{z_1,...,z_N\}$ a set of hidden variables associated with the observed data $X=\{x_1,...,x_N\}$ such that $z_{ik}=1$ iff $x_i$ is generated by the $k^{th}$ component of the mixture and $z_{ik}=0$ otherwise. Calculating the conditional expectation of the penalized log-likelihood of the complete data given the hidden variables results in:

\begin{equation}
\begin{aligned}
&Q(\Theta,\Theta^t) = E[\ln p(X,Z|\Theta)|X,\Theta^t] = \\ 
&\sum_{i=1}^N {\sum_{k=1}^K {\eta_{ik}^t \big \lbrack \ln{\pi_k^t} -\ln{|\Sigma_k^t|^\frac{1}{2}}   -\frac{w_i}{2} (x_i - \mu_k^t)^T (\Sigma_k^t)^{-1}(x_i-\mu_k^t) \big \rbrack }} \\ 
&  -N\lambda D_f \lbrack \sum_{k=1}^K {\ln{(\epsilon + \pi_k)} - \ln{\epsilon}}  \rbrack,
\end{aligned}
\label{eq:ellpw}
\end{equation}
where the posterior probabilities are updated with:
\begin{equation}
\begin{split}
\eta_{ik}^{t+1} &= p(z_i=k|x_i;w_i,\Theta^t) \\
&= \frac{\pi_k^t N(x_i;\theta^t,w_i)}{\sum_{k=1}^K \pi_k^t  N(x_i;\theta^t,w_i)} \\
&= \frac{\pi_k^t N(x_i;\mu_k^t,\frac{1}{w_i}\Sigma_k^t)}{\sum_{k=1}^K \pi_k^t N(x_i;\mu_k^t,\frac{1}{w_i}\Sigma_k^t)}.
\end{split}
\label{eq:estep}
\end{equation}

In the M-step, the parameters $\Theta={\{\pi_1,...,\pi_K;\mu_1,...,\mu_K;\Sigma_1,...,\Sigma_K\}}$ are updated by maximizing  \eqref{eq:ellpw}:

\begin{equation}
\begin{split}
\Theta^{t+1} = arg \displaystyle\max_\Theta Q(\Theta,\Theta^t).
\end{split}
\end{equation}

The parameters $\{\pi_1,...,\pi_K\}$ and $\{\mu_1,...,\mu_K,\Sigma_1,...,\Sigma_K\}$ can be updated separately since they are not intervened at \eqref{eq:ellpw}. This results in:

\begin{equation}
\mu_k^{t+1} = \frac{\sum_{i=1}^N {w_i \eta_{ik}^{(t+1)}x_i}}{\sum_{i=1}^N {w_i \eta_{ik}^{(t+1)}}},
\label{eq:mstep_mu}
\end{equation} 

\begin{equation}
\Sigma^{t+1}_k = \frac{\sum_{i=1}^N {w_i \eta_{ik}^{(t+1)}(x_i-\mu_k^{t+1})(x_i-\mu_k^{t+1})^T}}{\sum_{i=1}^N {\eta_{ik}^{(t+1)}}}.
\label{eq:mstep_sigma}
\end{equation} 

Incorporating the constraints on $\pi_k$ and using Lagrange multipliers, we obtain:

\begin{equation}
\pi_k^{t+1} = \max \{0,\frac{1}{1-K\lambda D_f} [\frac{1}{N} \sum_{i=1}^N{\eta_{ik}^{t+1}} -\lambda D_f]\}. 
\label{eq:mstep_pik}
\end{equation}

In summary, given predefined weights, $W$, the EM algorithm starts with $K_{max}$ which is an upper bound on the number of clusters in the data. In each iteration, a mixing coefficient may be shrunk to zero by 
\eqref{eq:mstep_pik}, eliminating its corresponding component.  In this way, in each iteration, the algorithm starts with a number of components and may converge to a smaller number. Note that when all the weights are queal to 1's, the algorithm converges to the algorithm proposed by \cite{PenHuaZha16} and will be referred to as Penalized Gaussian Mixtures Model (PGMM).

\subsubsection{Parameters selection}

The PWG algorithm requires three parameters: the tuning parameter $\lambda$, the initial number of clusters $K_{max}$, and the covariance matrix type. 

Equation \eqref{eq:mstep_pik} constrains the choice of the tuning parameter and the number of clusters. In this equation, we want a cluster to be eliminated when its average responsibilities, $\frac{1}{N} \sum_{i=1}^N{\eta_{ik}^{t+1}}$,  is less than the threshold $\lambda D_f$. Therefore,  the term $\frac{1}{1-K_{max}\lambda D_f}$ should be positive. In addition, $\lambda$ should be non-negative from its definition. Therefore :

\begin{equation}
  0  \leq \lambda < \frac{1}{K_{max}D_f}.
\end{equation}

 When $\lambda$ is close to its upper limit, the algorithm will prefer a small number of components. When $\lambda$ approaches zero, the algorithm will converge to the regular Weighted GMM (WGMM) algorithm, proposed by \cite{gebru2015algorithms}. In \cite{PenHuaZha16} it is suggested to choose $\lambda$ by running the algorithm with different values and choosing the one that minimizes the BIC criterion \cite{schwarz1978estimating}:

\begin{equation}
\lambda = \min_{\lambda} BIC(\lambda).
\label{eq:lambda_BIC}
\end{equation}

While the choice of the tuning parameter is constrained, the choice of the type of the covariance matrices is not. Thus, the covariance matrices can be either full or diagonal. However, the best practice is to choose diagonal covariance matrices due to three reasons. First, modeling of a full covariance GMM can be equally achieved by using diagonal covariance GMM with more components. Second, a GMM with diagonal covariance matrices is more computationally efficient. Third, the results of a GMM with diagonal matrices are empirically superior to GMM with full covariance matrices \cite{Bimbot:2004:TTS:1289340.1289376}.


\section{Algorithm Evaluation}
\label{sec:results}

The proposed method has been evaluated by several experiments on both synthetic and real datasets. The synthetic datasets included: a) the data used by scikit package \cite{scikit-learn}  in its clustering algorithm comparison b) simulated Gaussian mixtures PDFs with a different number of components. 
For real datasets evaluation, we used OPWG to segments images from Berkeley BSD dataset \cite{MartinFTM01}.

\subsection{Algorithm Evaluation - Synthetic Datasets}

To evaluate the algorithm on synthetic data we performed 50 experiments on each dataset and averaged the results of the F1 score and the Normalized Mutual Information (NMI) metric, as defined as in \cite{Pfitzner:2009:CES:1554488.1554491}.  

Every dataset consists of $N=100,000$ 2D data samples which is further divided into equal batches of $n_i=1000$ samples each. 
Since the order matters when clustering streaming data, two modes of operations were considered: Mode A in which the data arrives randomly and each batch may contain samples from all the clusters in the dataset and Mode B in which the data in each batch arrives from a subset of the dataset and may not contain samples from all the underlying clusters. To simulate the latter case, the 2D data has been sorted according to x (or y) coordinate and then divided into batches. 

The tuning parameter $\lambda$ was chosen to be $0.005$ in the online stage. Different values of $\lambda$ in the range $[0.003-0.006]$ were tested in the post-processing stage and the one that minimized the BIC of the model was chosen. The covariance matrices were chosen to be diagonal in the online stage and full in the post-processing stage. The maximum number of clusters was chosen to be $K_{max}=25$.

In our evaluation, we compared our proposed algorithm to the following algorithms: OFCM  with a number of clusters equal to that found by OPWG and a batch size of $n_i=1000$ samples, PGMM running on the whole dataset with $\lambda=0.005$, $K_{max}=25$ and a full covariance matrix, and GMM  running on the whole dataset with a number of clusters provided by PGMM and with a full covariance matrix.  

\begin{figure}[htb]
\centering
  \begin{subfigure}[b]{.25\linewidth}
    \centering
    \includegraphics[width=.99\textwidth]{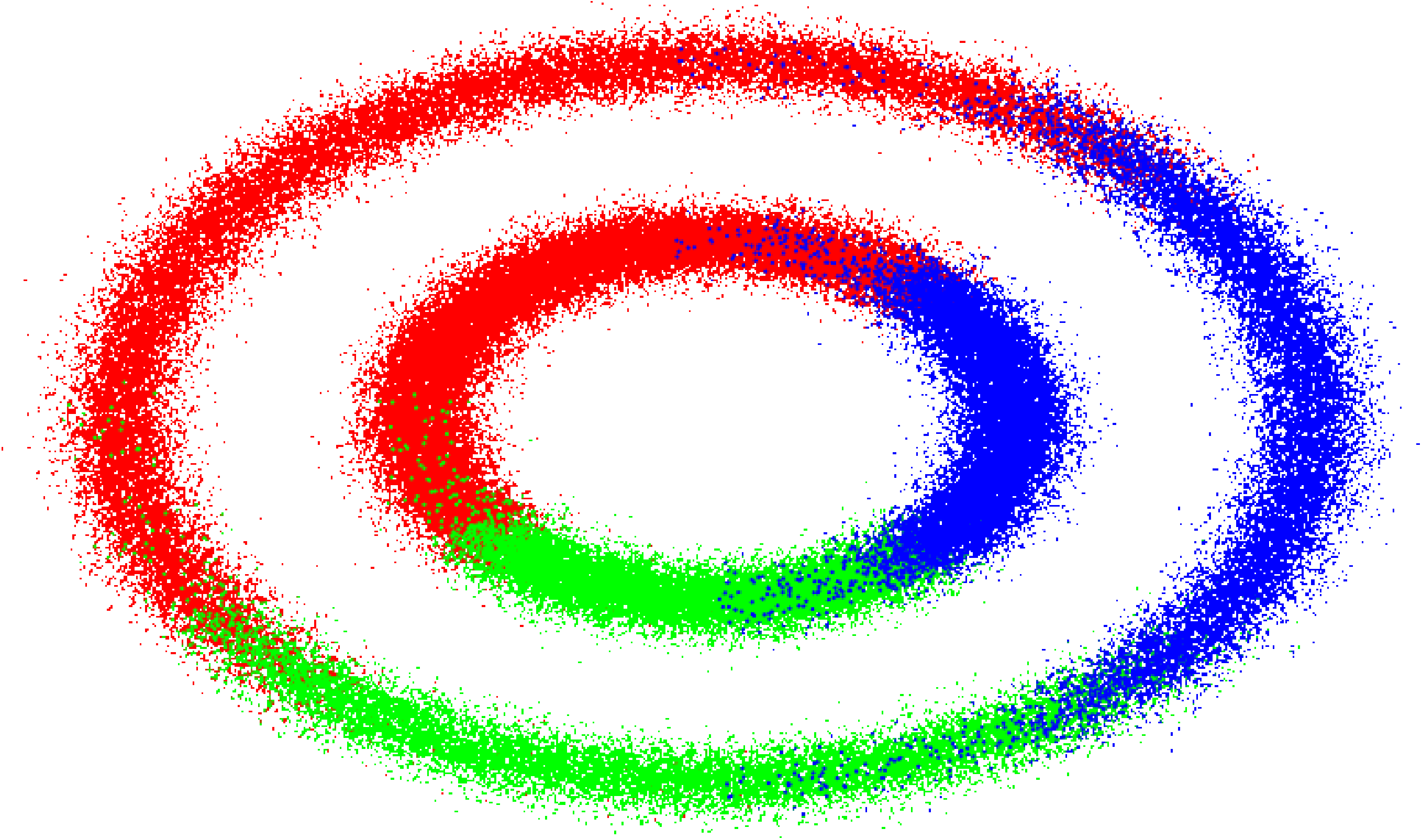}
  \end{subfigure}%
  \begin{subfigure}[b]{.25\linewidth}
    \centering
    \includegraphics[width=.99\textwidth]{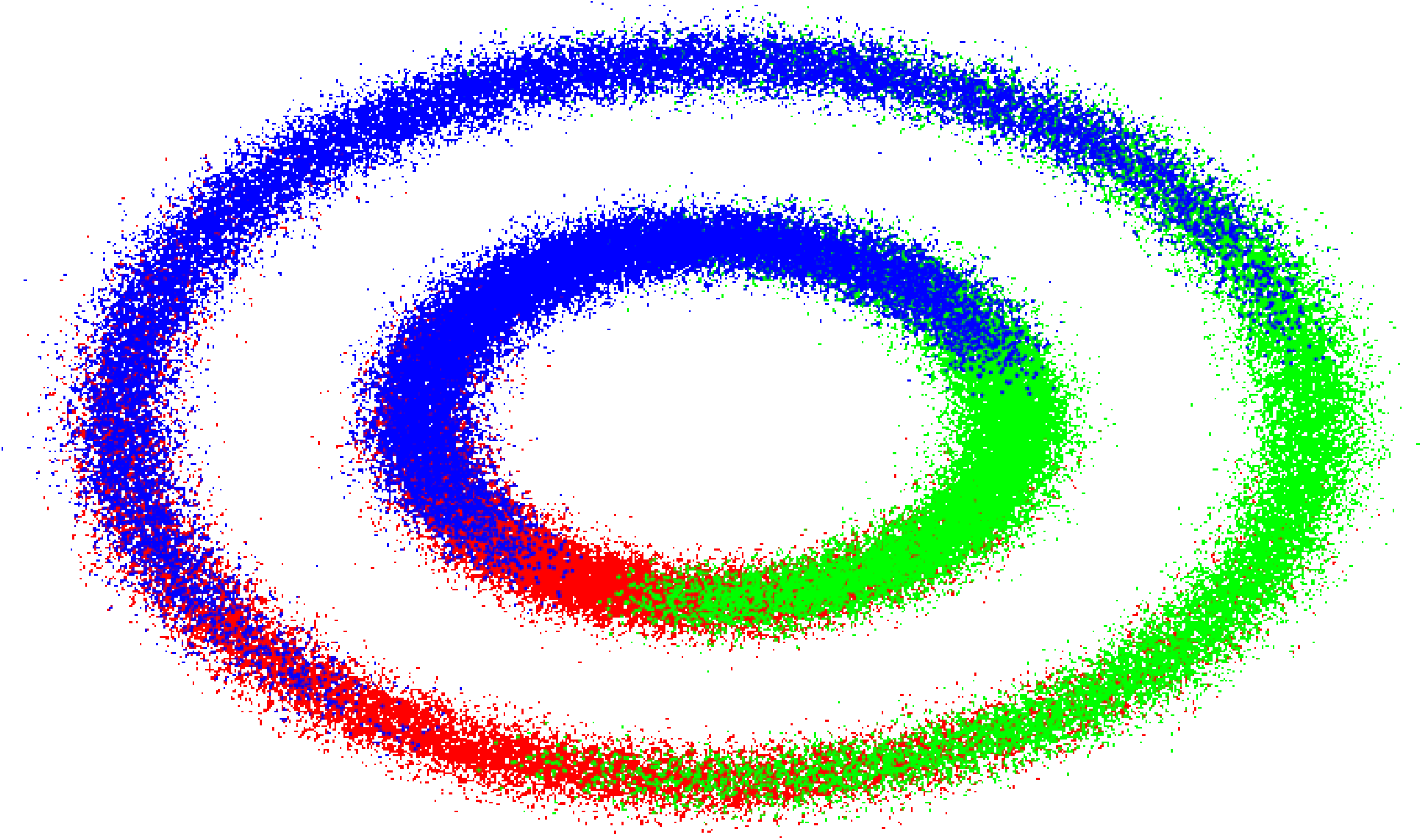}
  \end{subfigure}%
  \begin{subfigure}[b]{.25\linewidth}
    \centering
    \includegraphics[width=.99\textwidth]{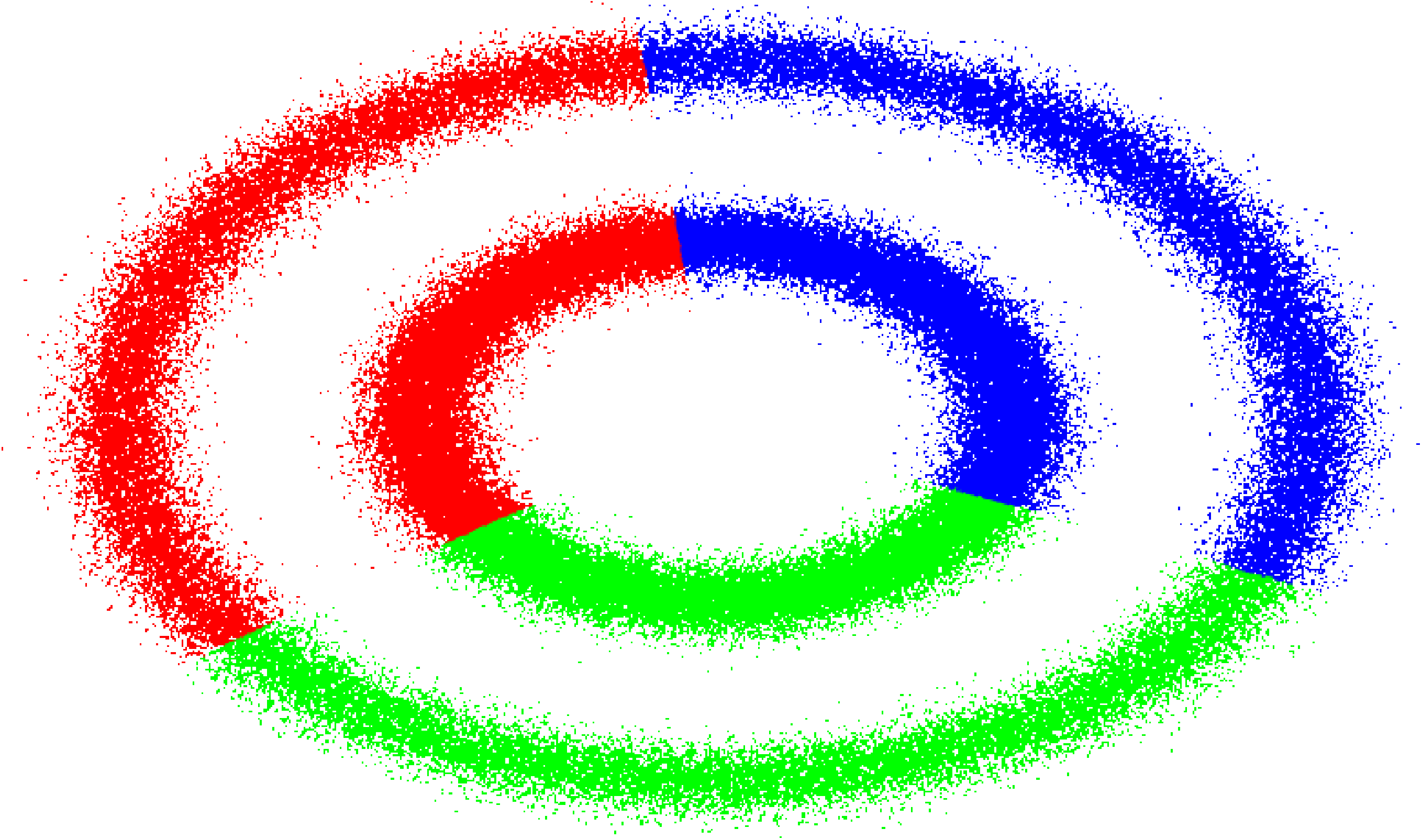}
  \end{subfigure}%
    \begin{subfigure}[b]{.25\linewidth}
    \centering
    \includegraphics[width=.99\textwidth]{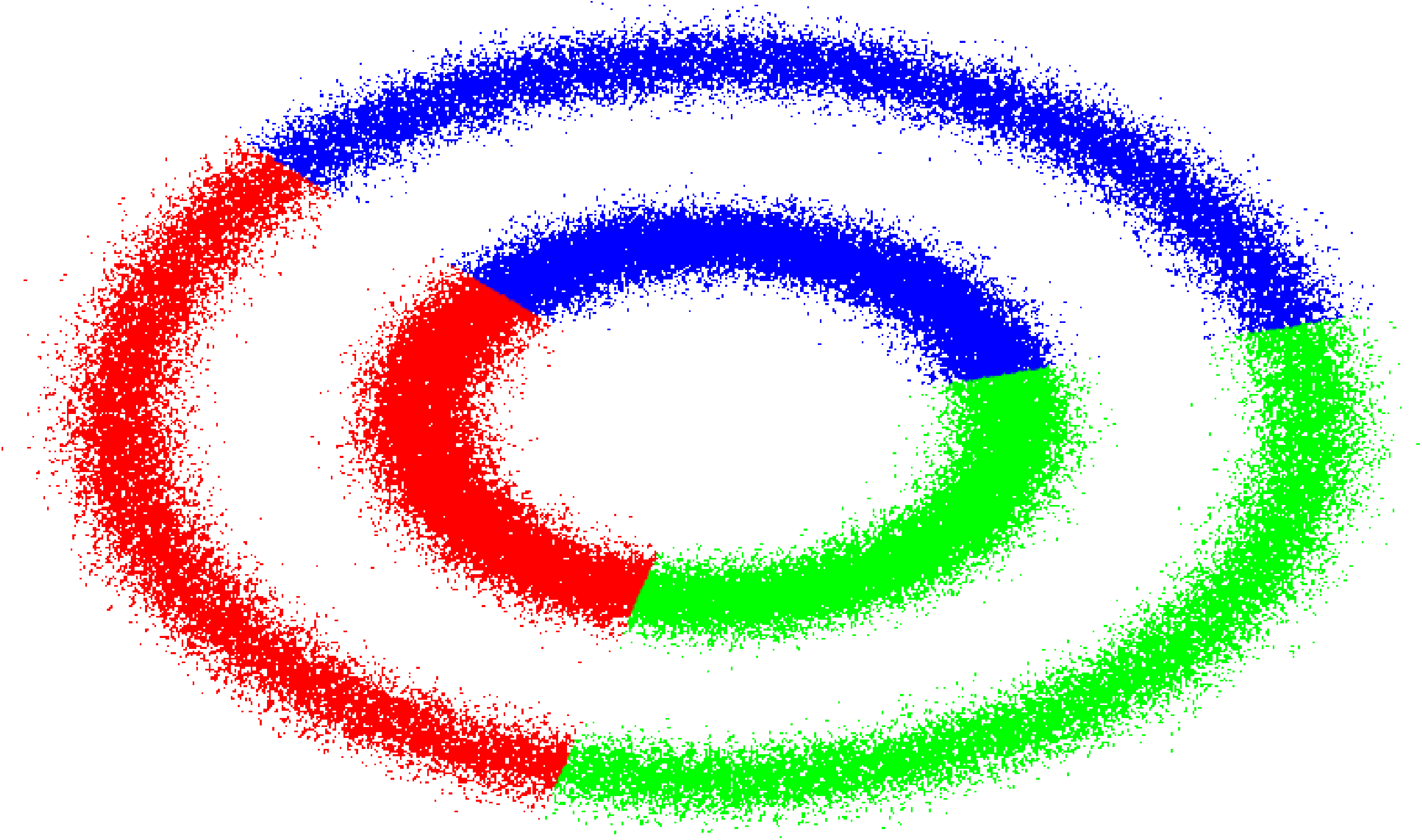}
  \end{subfigure}\\
\begin{subfigure}[b]{.25\linewidth}
    \centering
    \includegraphics[width=.99\textwidth]{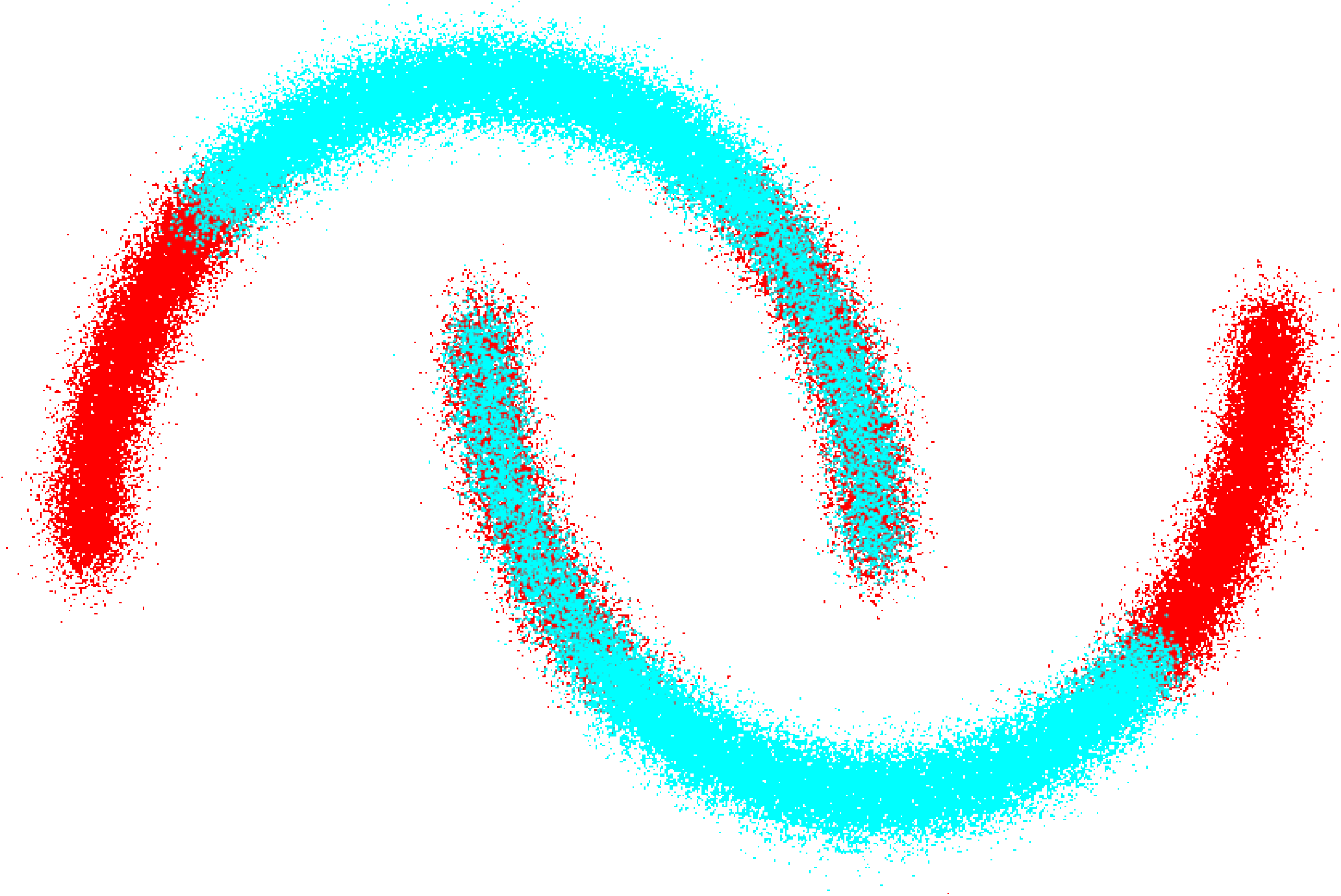}
  \end{subfigure}%
  \begin{subfigure}[b]{.25\linewidth}
    \centering
    \includegraphics[width=.99\textwidth]{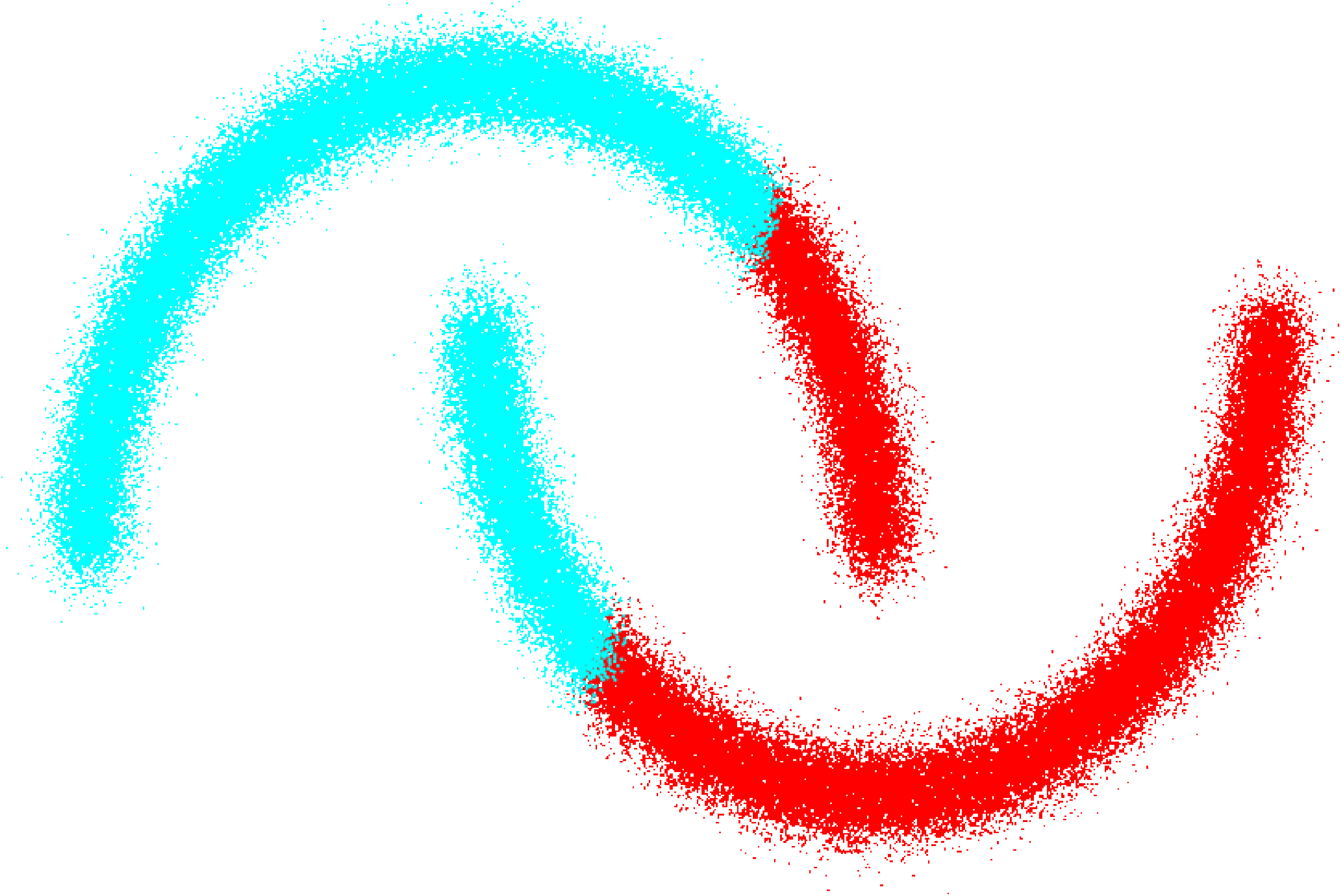}
  \end{subfigure}%
  \begin{subfigure}[b]{.25\linewidth}
    \centering
    \includegraphics[width=.99\textwidth]{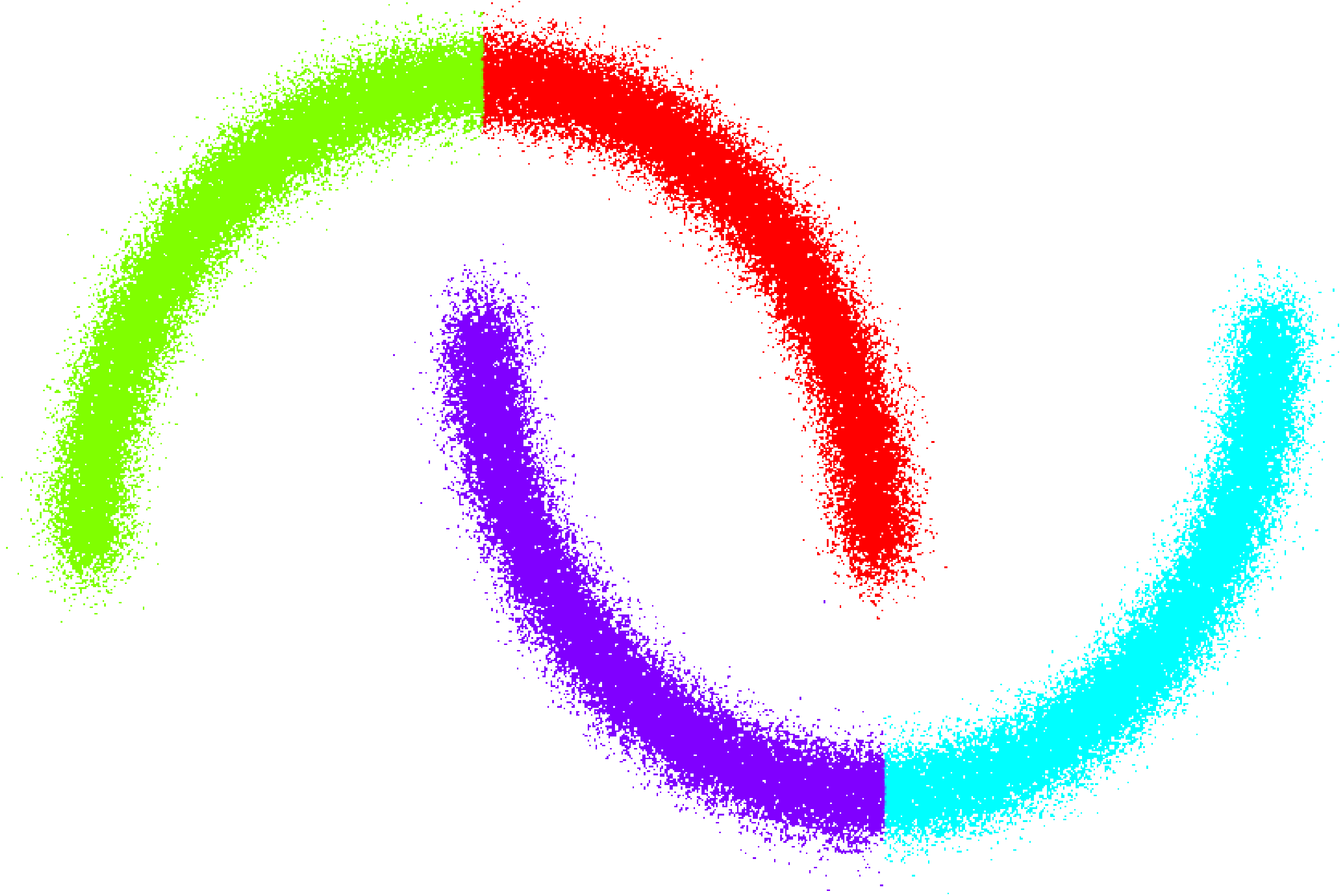}
  \end{subfigure}%
    \begin{subfigure}[b]{.25\linewidth}
    \centering
    \includegraphics[width=.99\textwidth]{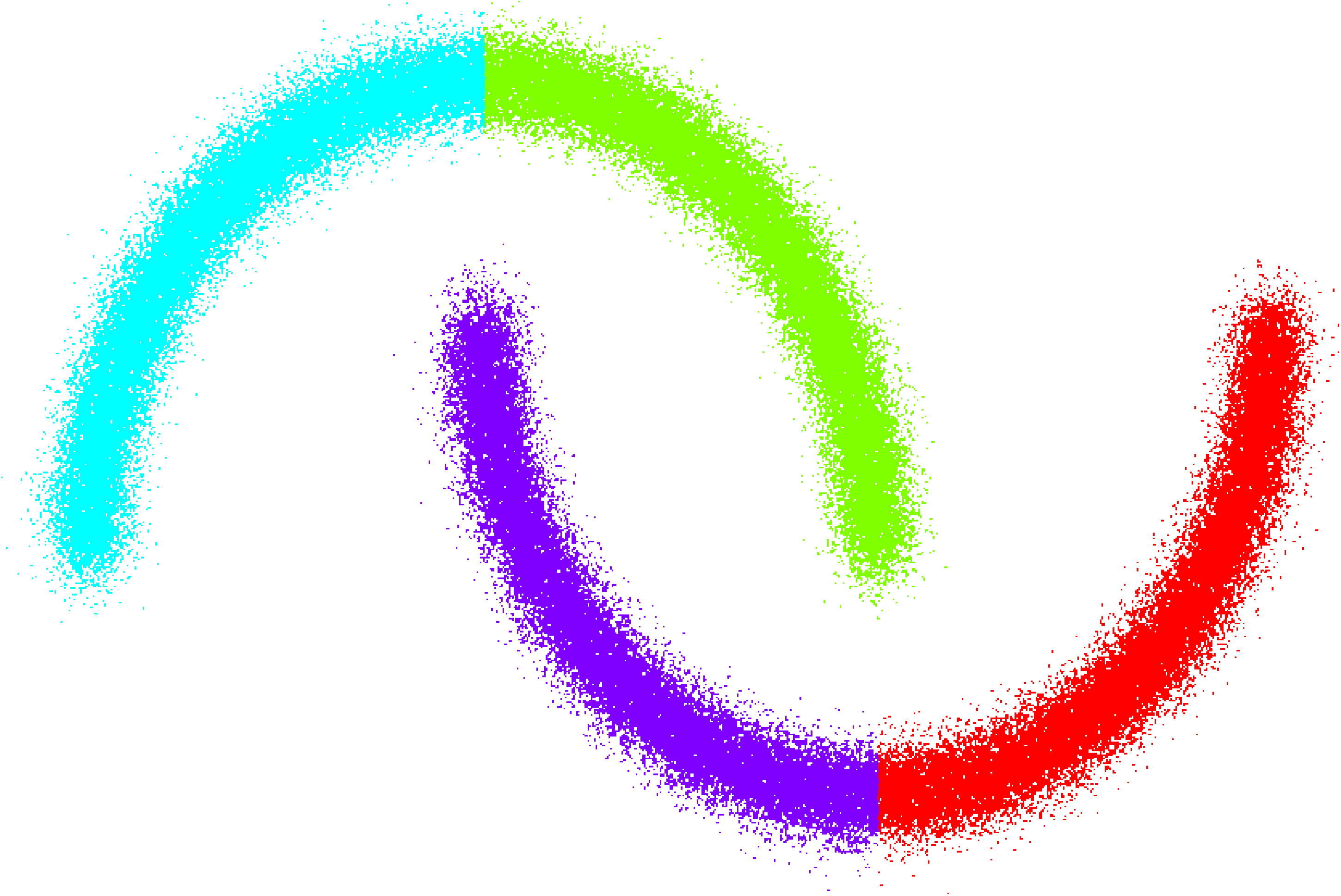}
  \end{subfigure}\\
  \begin{subfigure}[b]{.25\linewidth}
    \centering
    \includegraphics[width=.99\textwidth]{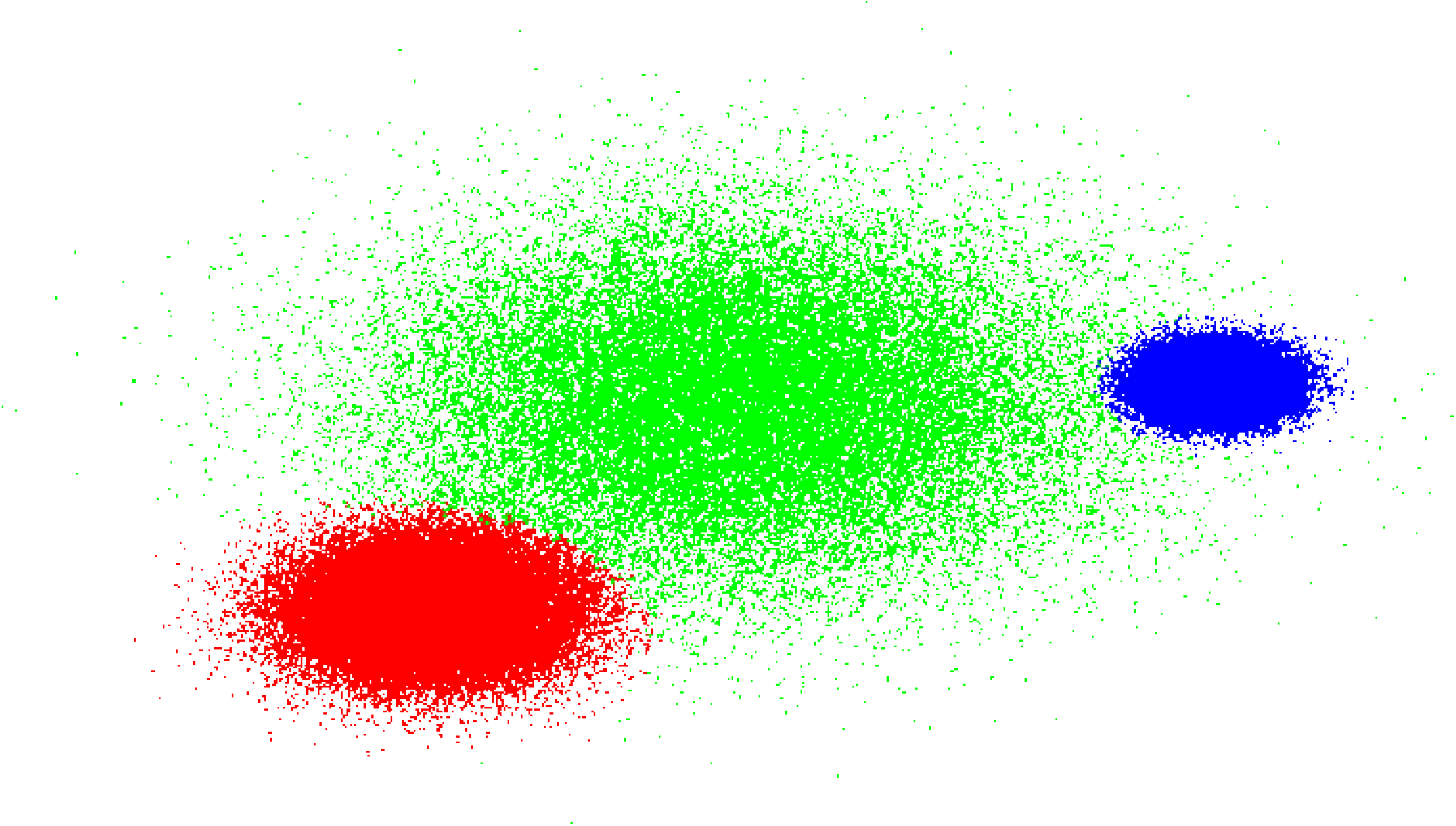}
  \end{subfigure}%
  \begin{subfigure}[b]{.25\linewidth}
    \centering
    \includegraphics[width=.99\textwidth]{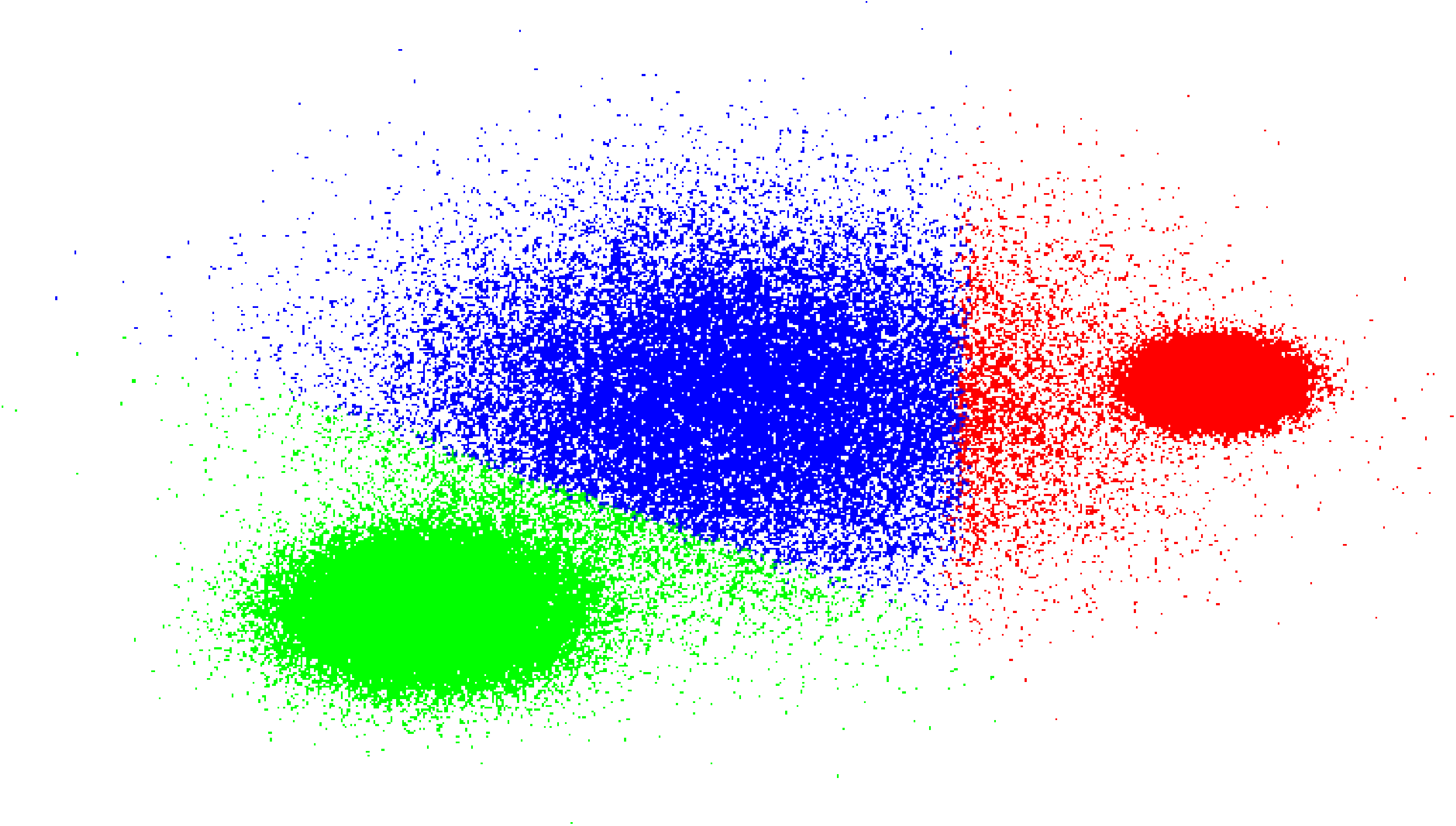}
  \end{subfigure}%
  \begin{subfigure}[b]{.25\linewidth}
    \centering
    \includegraphics[width=.99\textwidth]{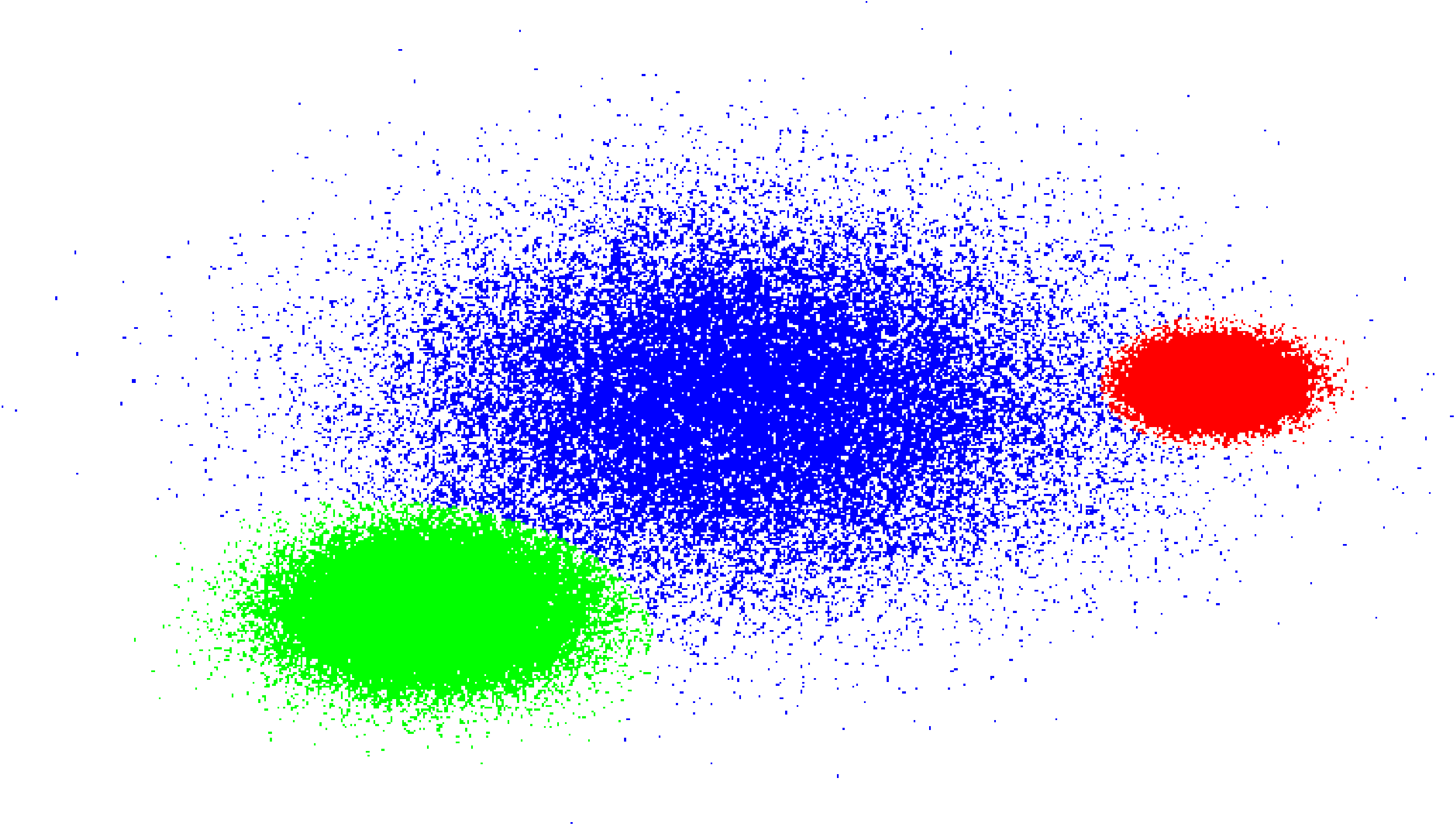}
  \end{subfigure}%
    \begin{subfigure}[b]{.25\linewidth}
    \centering
    \includegraphics[width=.99\textwidth]{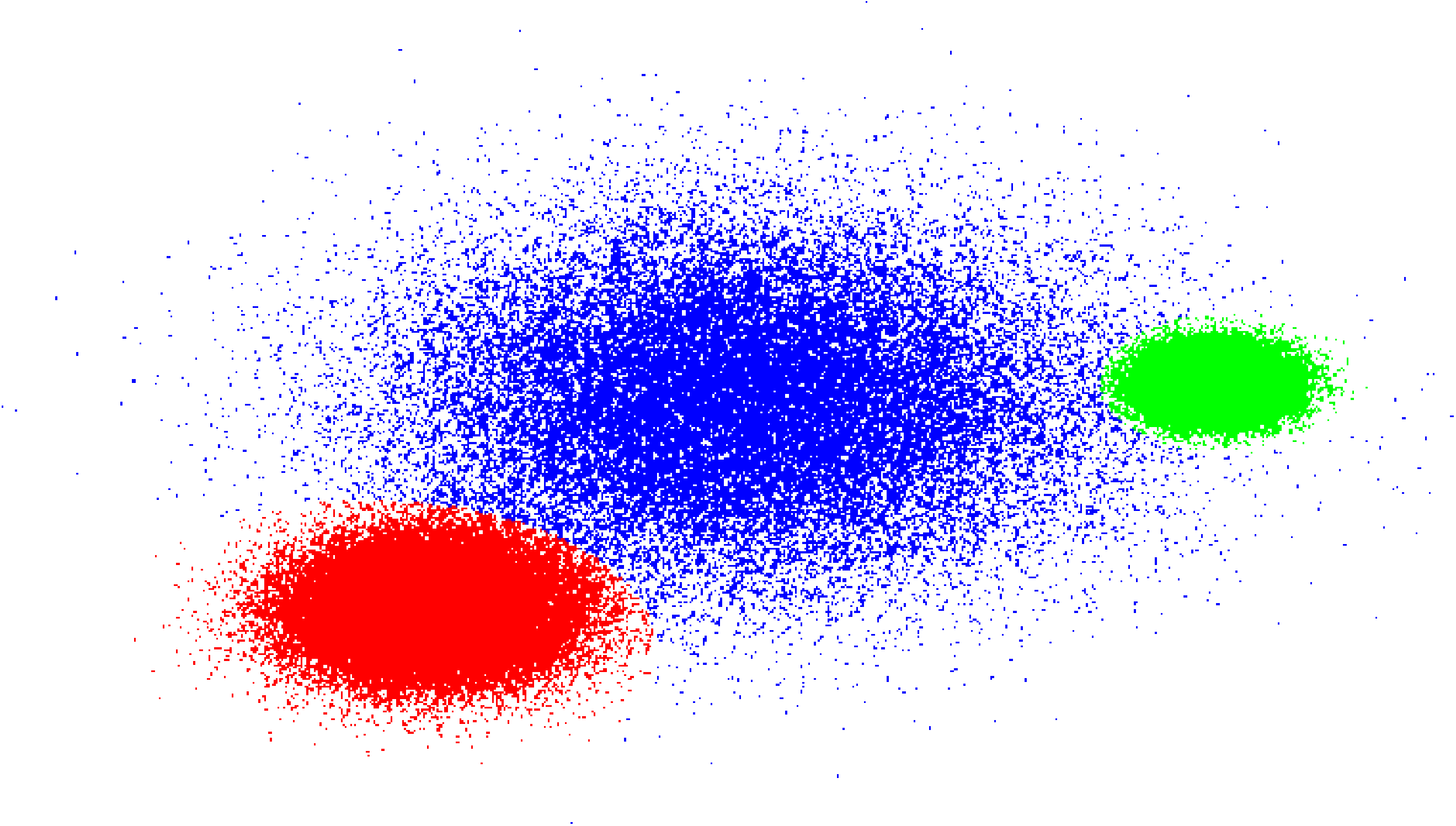}
  \end{subfigure}\\
  \begin{subfigure}[b]{.25\linewidth}
    \centering
    \includegraphics[width=.99\textwidth]{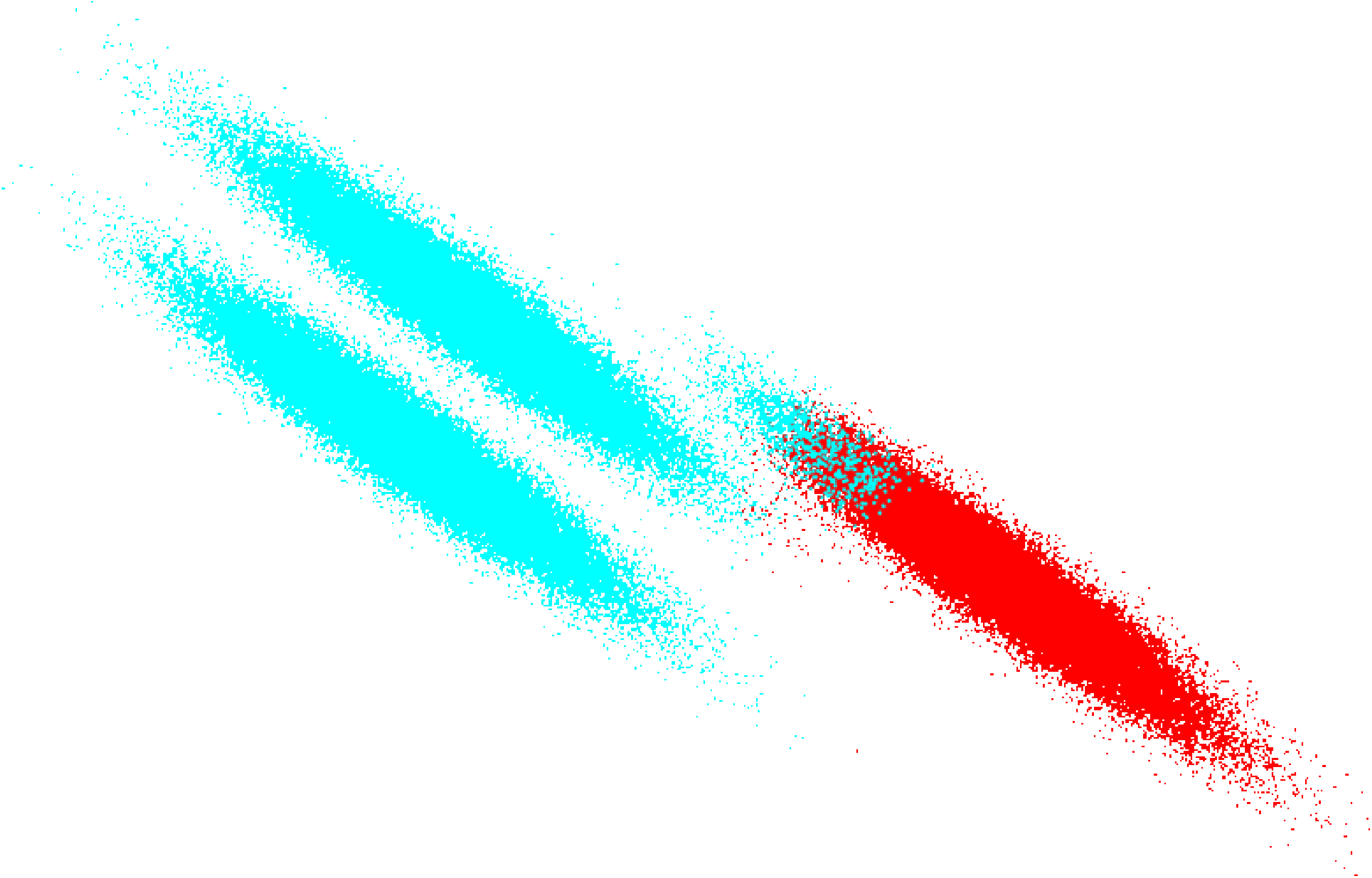}
  \end{subfigure}%
  \begin{subfigure}[b]{.25\linewidth}
    \centering
    \includegraphics[width=.99\textwidth]{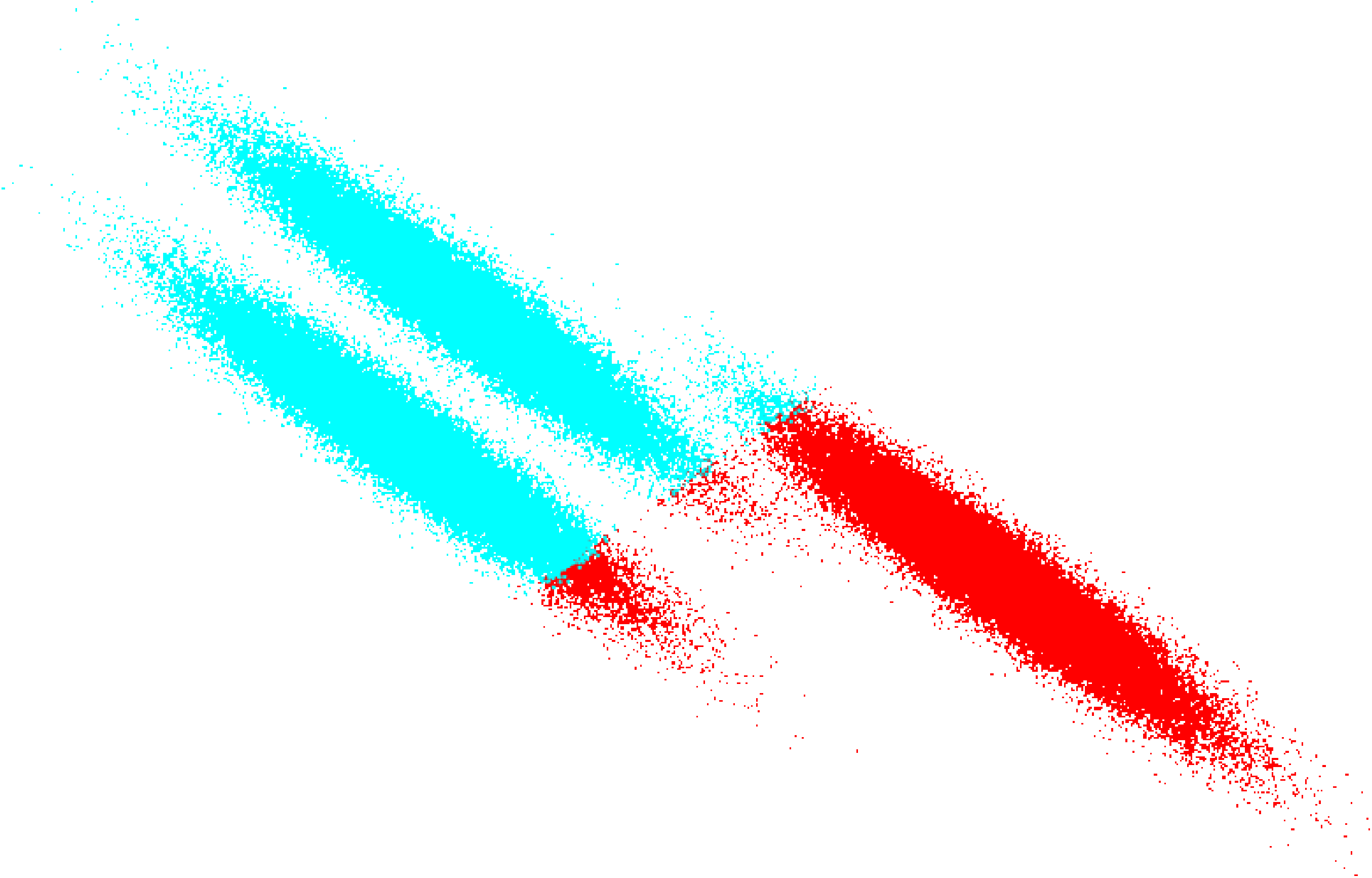}
  \end{subfigure}%
  \begin{subfigure}[b]{.25\linewidth}
    \centering
    \includegraphics[width=.99\textwidth]{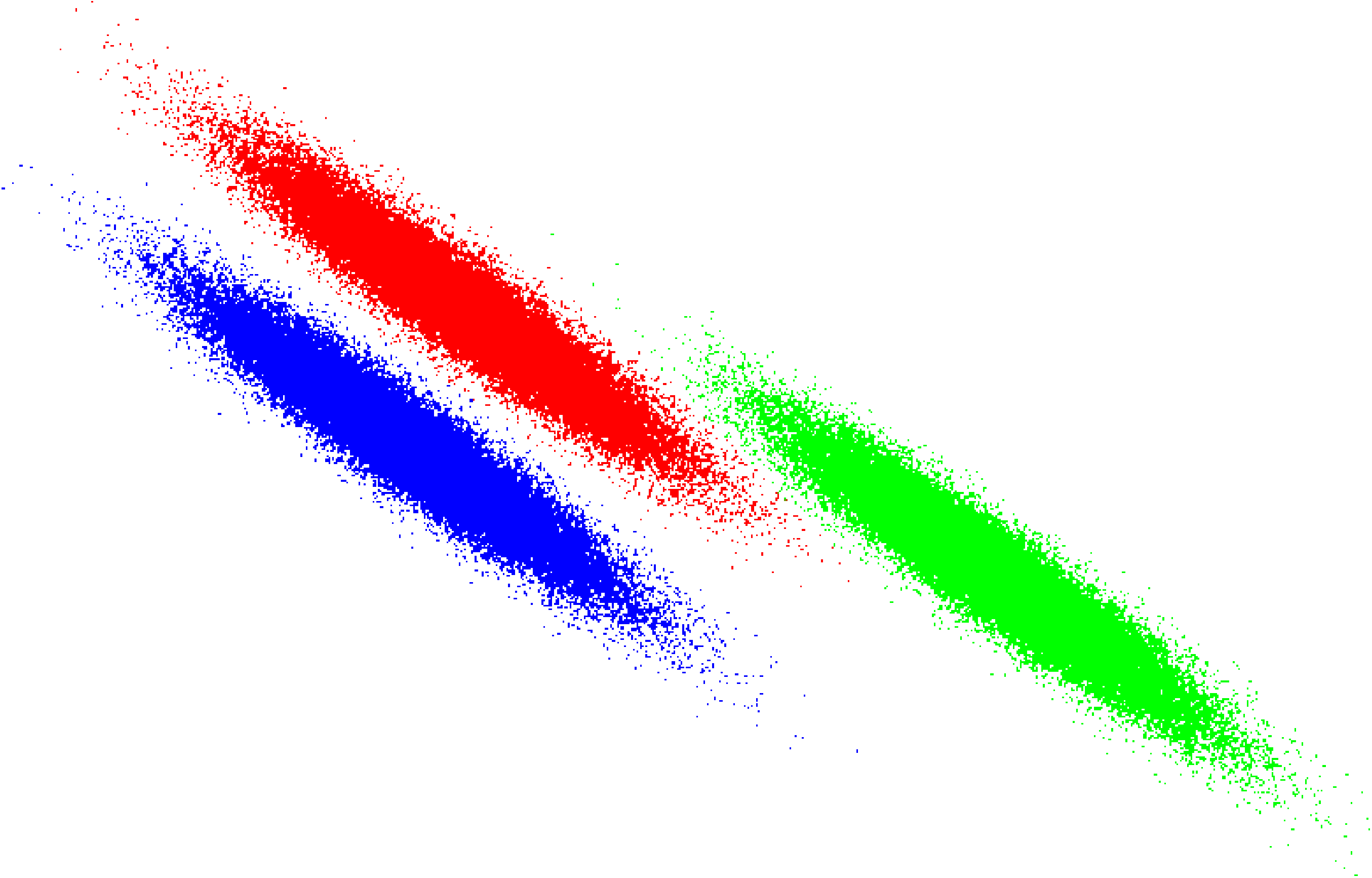}
  \end{subfigure}%
    \begin{subfigure}[b]{.25\linewidth}
    \centering
    \includegraphics[width=.99\textwidth]{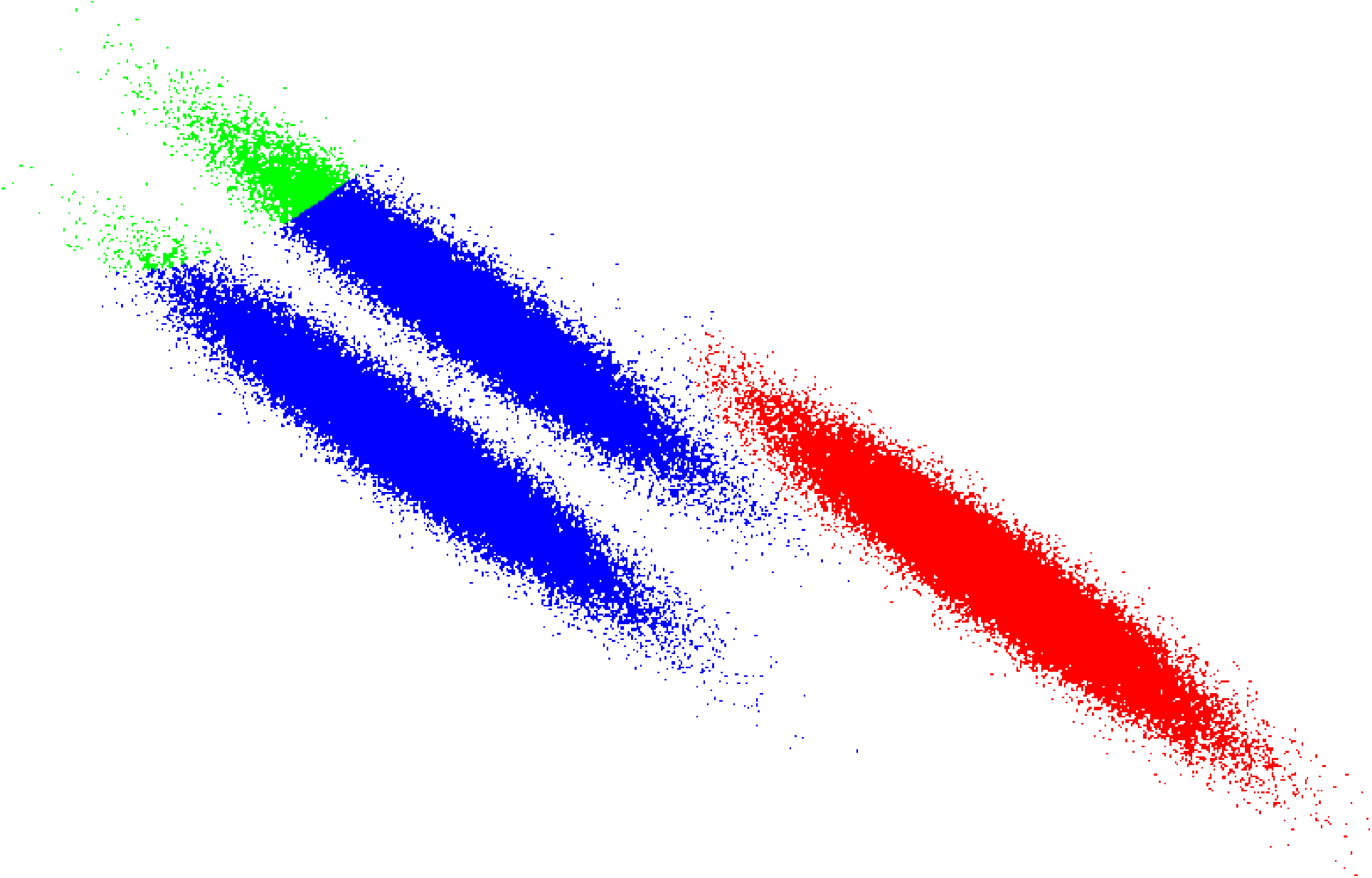}
  \end{subfigure}\\
  \begin{subfigure}[b]{.25\linewidth}
    \centering
    \includegraphics[width=.99\textwidth]{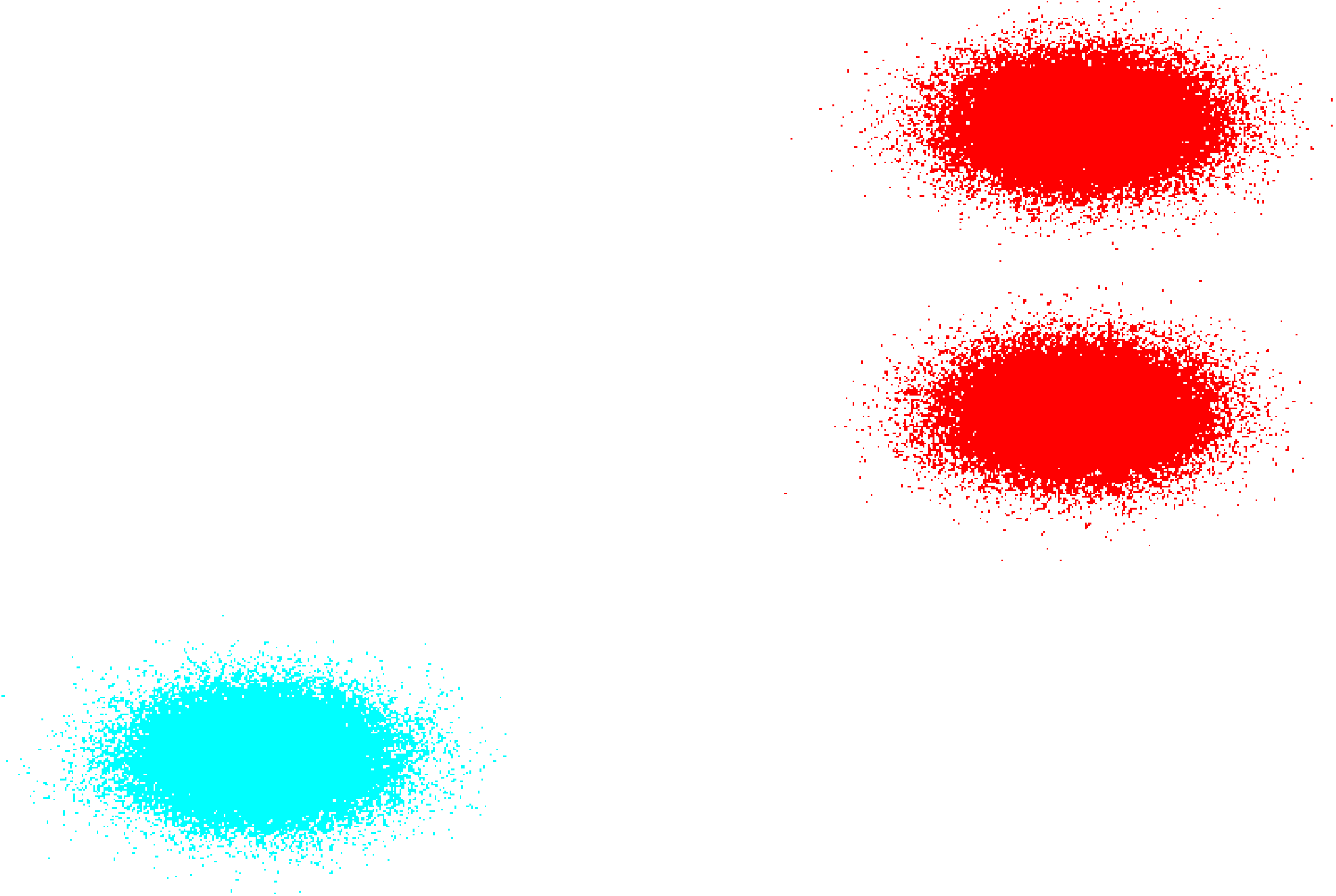}
  \end{subfigure}%
  \begin{subfigure}[b]{.25\linewidth}
    \centering
    \includegraphics[width=.99\textwidth]{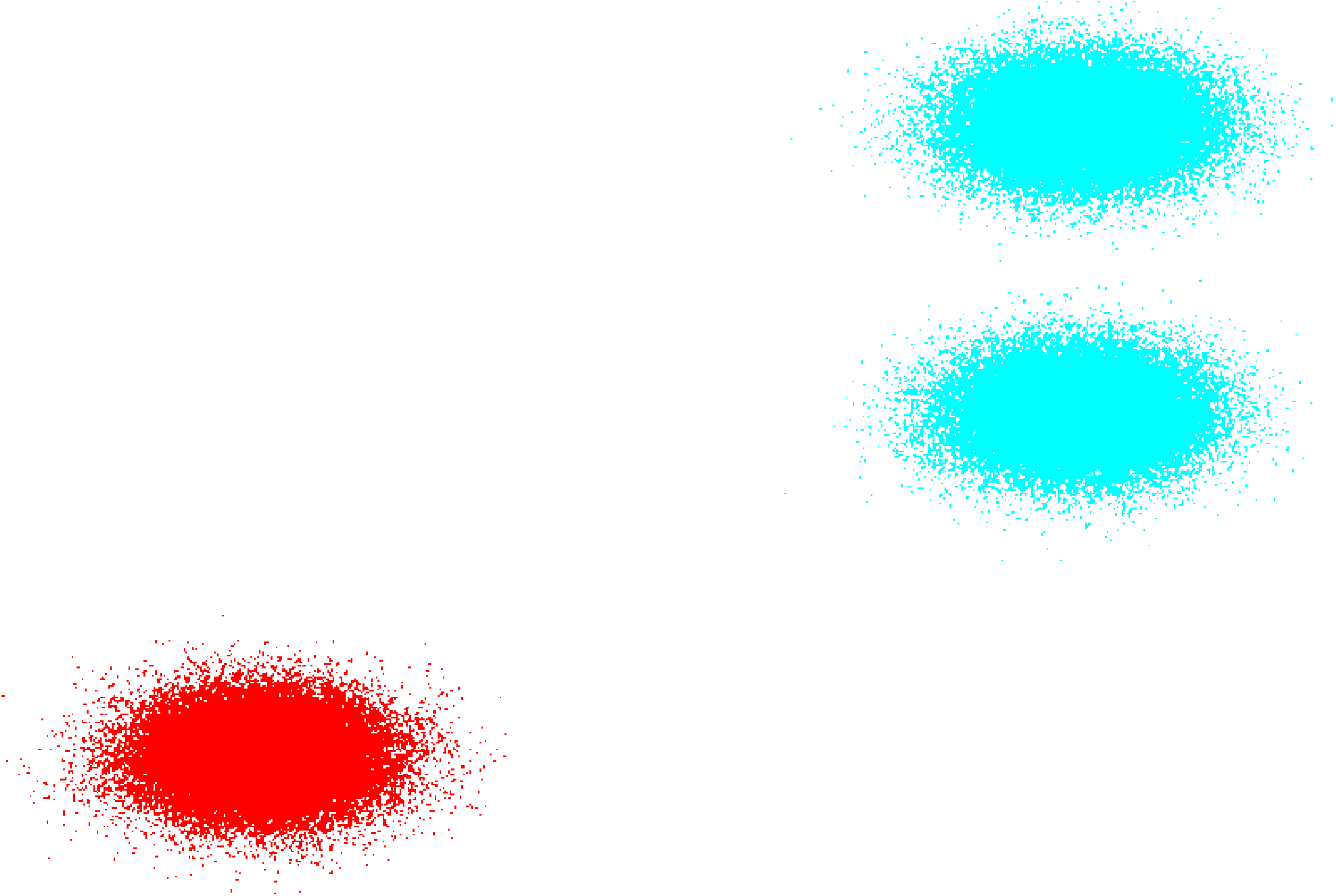}
  \end{subfigure}%
  \begin{subfigure}[b]{.25\linewidth}
    \centering
    \includegraphics[width=.99\textwidth]{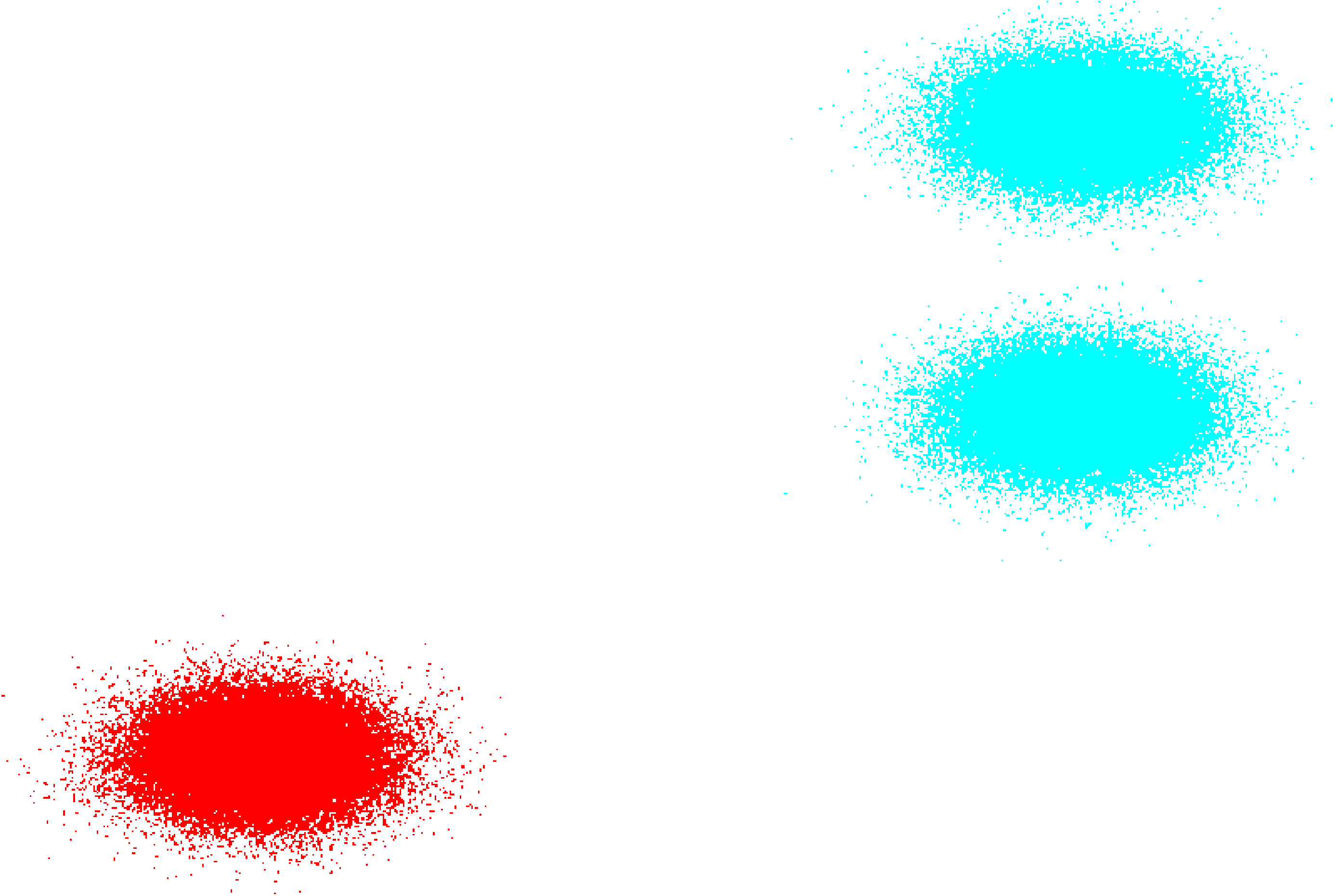}
  \end{subfigure}%
    \begin{subfigure}[b]{.25\linewidth}
    \centering
    \includegraphics[width=.99\textwidth]{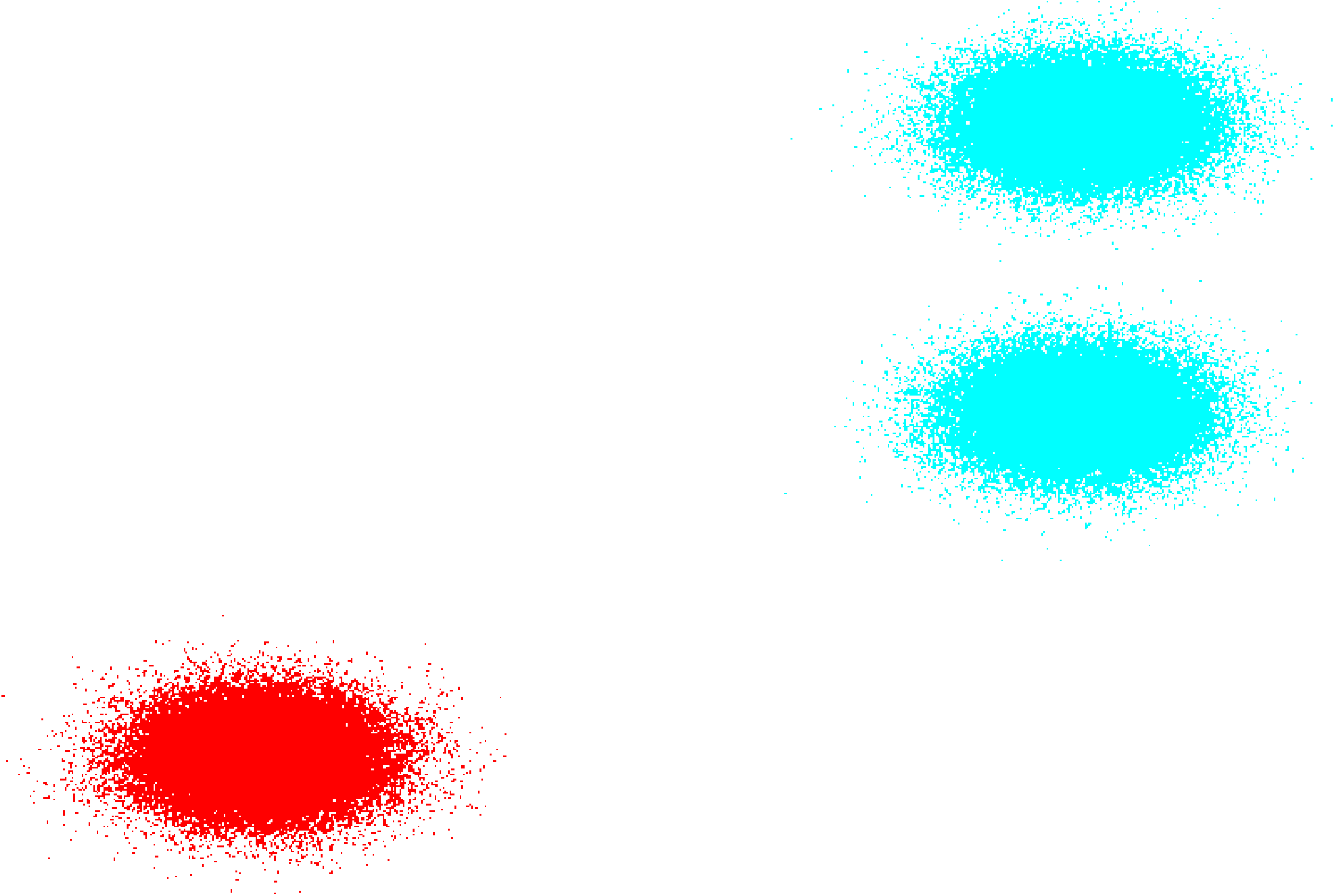}
  \end{subfigure}\\
  \begin{subfigure}[b]{.25\linewidth}
    \centering
    \includegraphics[width=.99\textwidth]{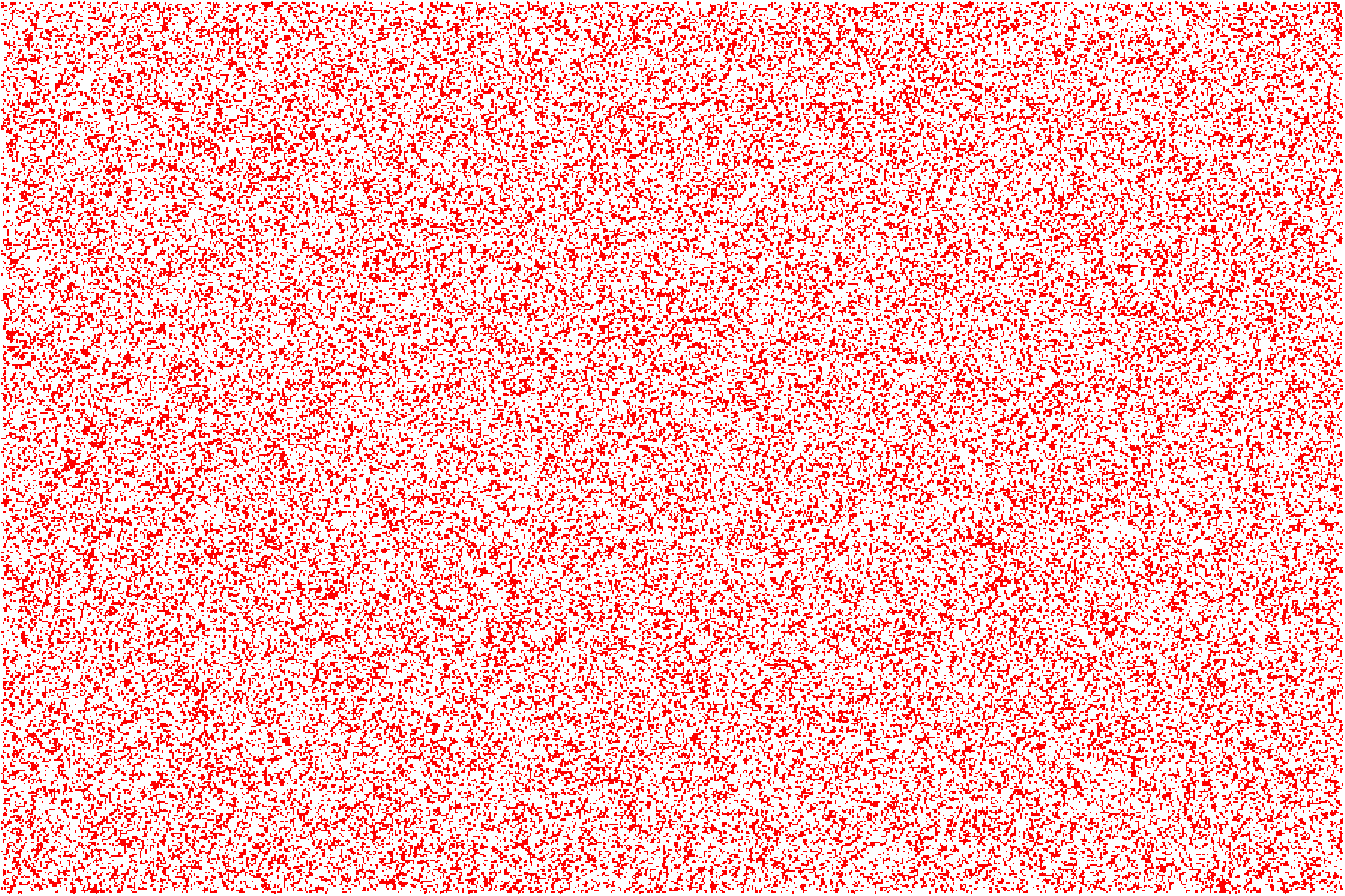}
    \caption{OPWG}
  \end{subfigure}%
  \begin{subfigure}[b]{.25\linewidth}
    \centering
    \includegraphics[width=.99\textwidth]{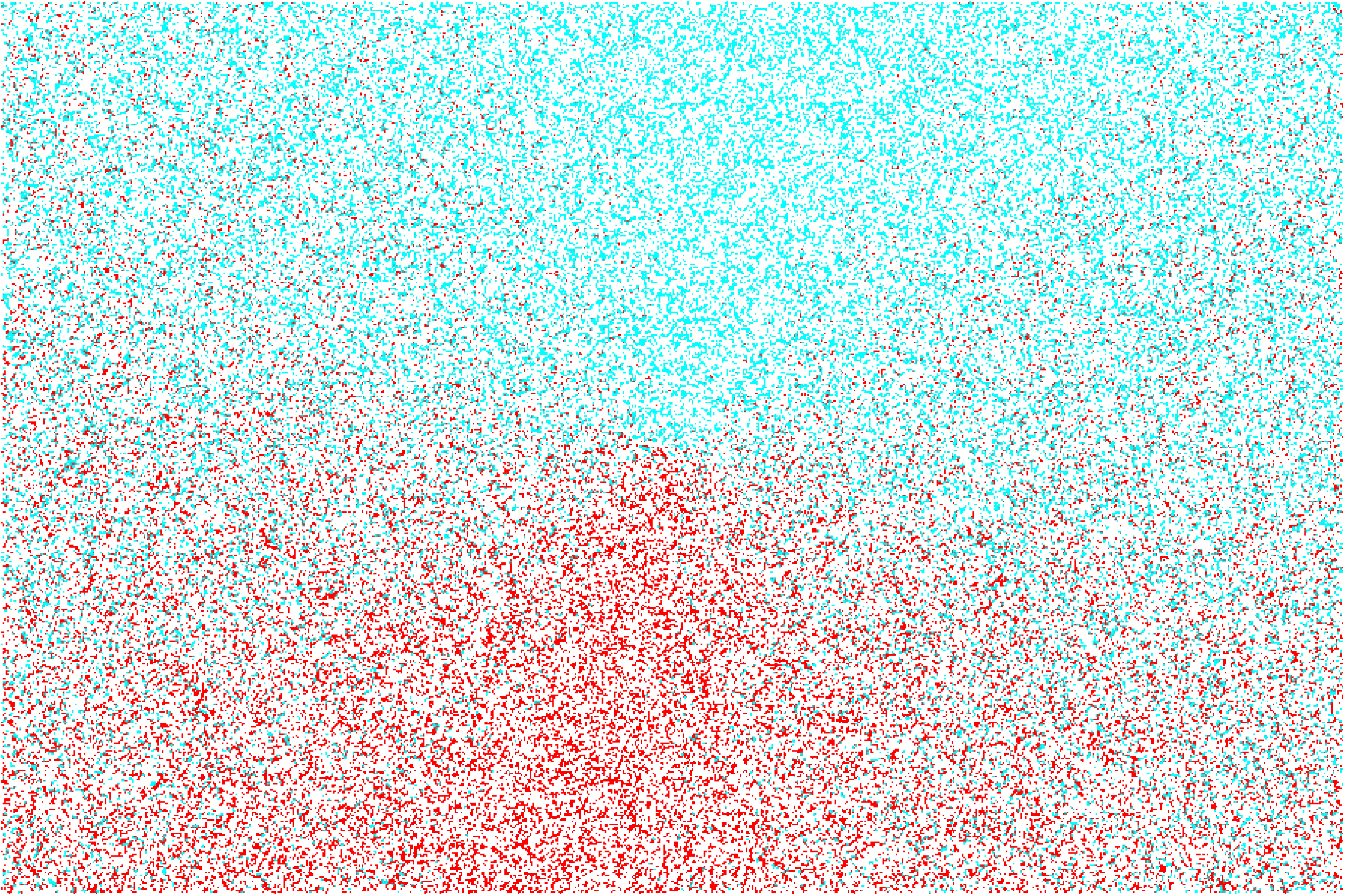}
    \caption{OFCM}
  \end{subfigure}%
  \begin{subfigure}[b]{.25\linewidth}
    \centering
    \includegraphics[width=.99\textwidth]{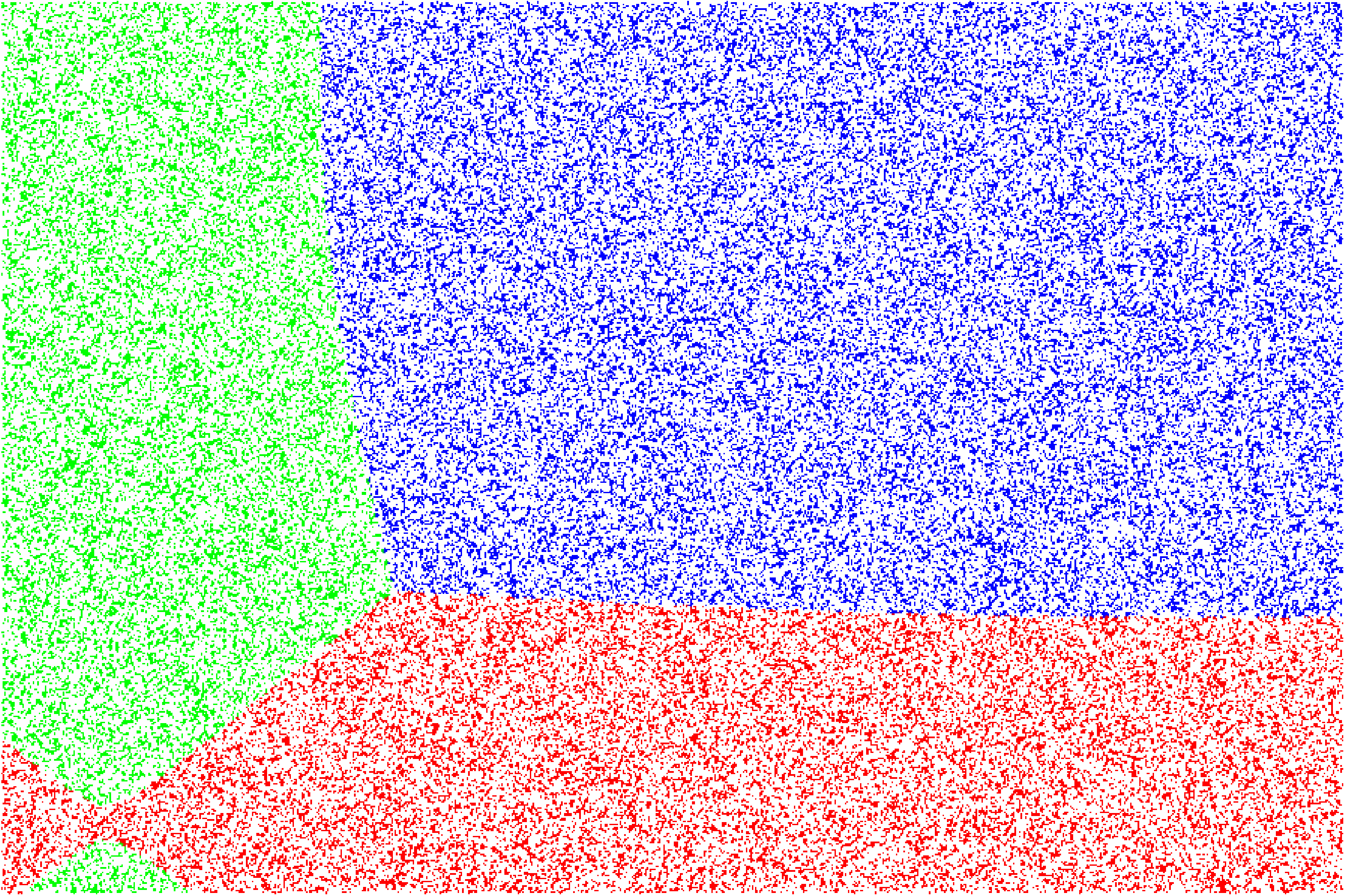}
    \caption{PGMM}
  \end{subfigure}%
    \begin{subfigure}[b]{.25\linewidth}
    \centering
    \includegraphics[width=.99\textwidth]{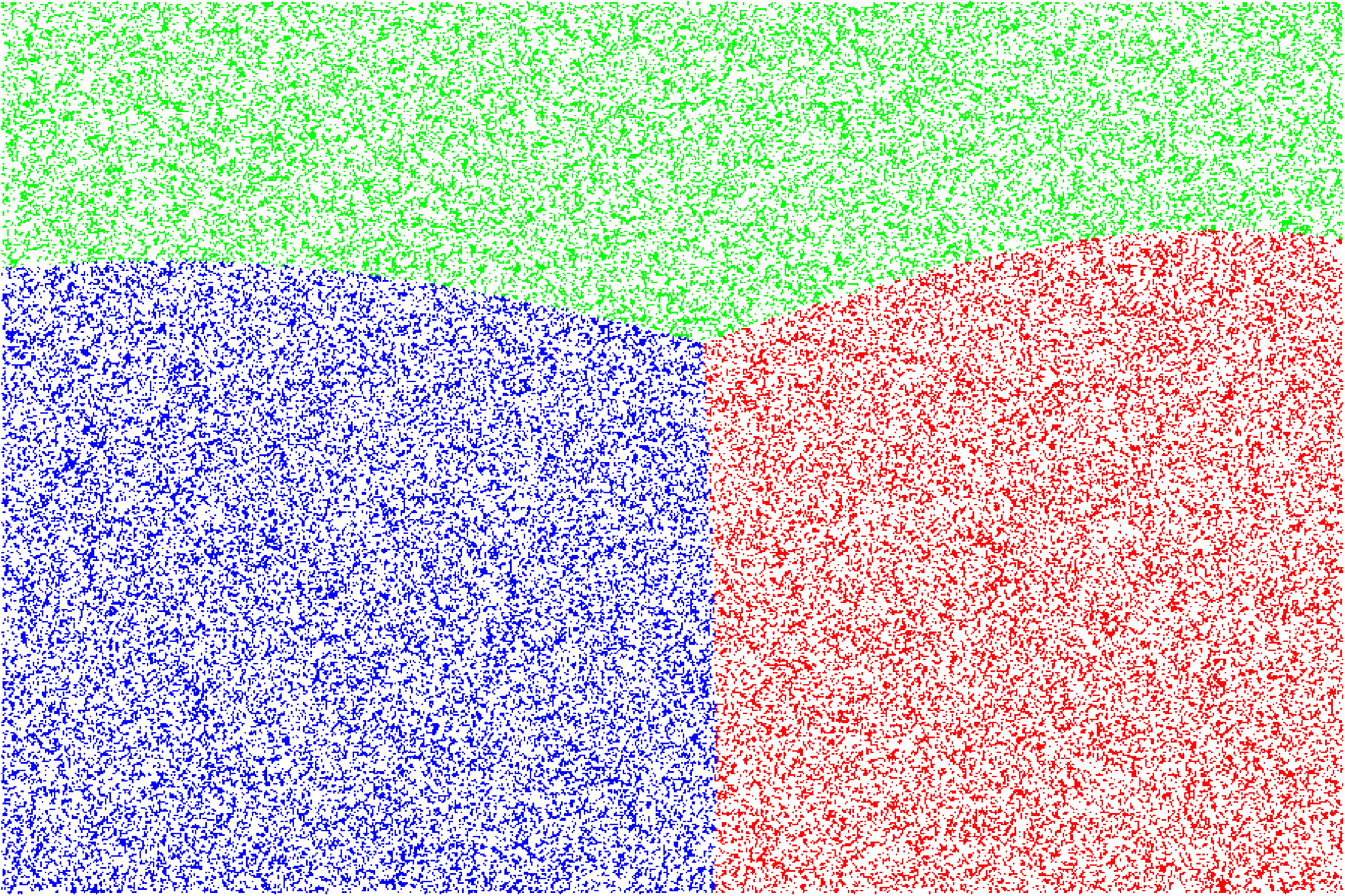}
    \caption{GMM}
  \end{subfigure}\\
\caption
{Mode A.  Each algorithm was evaluated on six different datasets. For OPWG and PGMM, the initial number of clusters was 25. OFCM and GMM were initialized with the number of clusters found in OPWG and PGMM, respectively.}\label{fig:6db_random}
\end{figure}

The results for mode A are reported in Table \ref{table:6db_random} and illustrated in Figure \ref{fig:6db_random}.
As expected, when the full stream is available, as in PGMM and GMM, better results are achieved, whereas, in online mode, the performance deteriorates.
The underlying assumption in online mode that all batches are independent and have an identical distribution yields sub-optimal solution. Further, comparing OPWG with OFCM shows that except for DB2, OPWG outperforms OFCM although its initial number of clusters $K_{max}$ is bigger. In this mode, the average number of clusters that OPWG found, $K=3.18, 1.92, 2.96, 2.48, 2,1 $, was close to the number of clusters that PWG  found - $K= 3.34,5.22,3.04,2.38,2.14,2.76$.

\begin{table*}[h]
\begin{center}
\begin{tabular}{lllllllllll}
\hline
            & \multicolumn{2}{c}{DB1} & \multicolumn{2}{c}{DB2}       & \multicolumn{2}{c}{DB3}       & \multicolumn{2}{c}{DB4}       & \multicolumn{2}{c}{DB5}       \\
            & F1              & NMI   & F1            & NMI           & F1            & NMI           & F1            & NMI           & F1            & NMI           \\ \cline{2-11} 
OPWG  & \textbf{0.46}   & 0.00  & 0.54          & 0.00          & \textbf{0.98} & \textbf{0.92} & 0.75          & 0.63          & 0.77          & 0.73          \\
OFCM & 0.40            & 0.00  & \textbf{0.75} & 0.19          & 0.91          & 0.78          & 0.78          & 0.60          & 0.77          & 0.73          \\
PGMM         & 0.38            & 0.00  & 0.57          & \textbf{0.55} & \textbf{0.98} & \textbf{0.92} & 0.86          & \textbf{0.82} & 0.79          & 0.74          \\
GMM         & 0.38            & 0.00     & 0.56          & 0.54          & \textbf{0.98} & \textbf{0.92} & \textbf{0.85} & 0.81          & \textbf{0.80} & \textbf{0.77} \\ \hline
\end{tabular}
\caption 
{Mode A. Clustering quality measures for each of the algorithms discussed for all datasets considered apart from the last one, which is noise.}\label{table:6db_random}
\end{center}
\end{table*}

\begin{figure}[htb]
\centering
  \begin{subfigure}[b]{.25\linewidth}
    \centering
    \includegraphics[width=.99\textwidth]{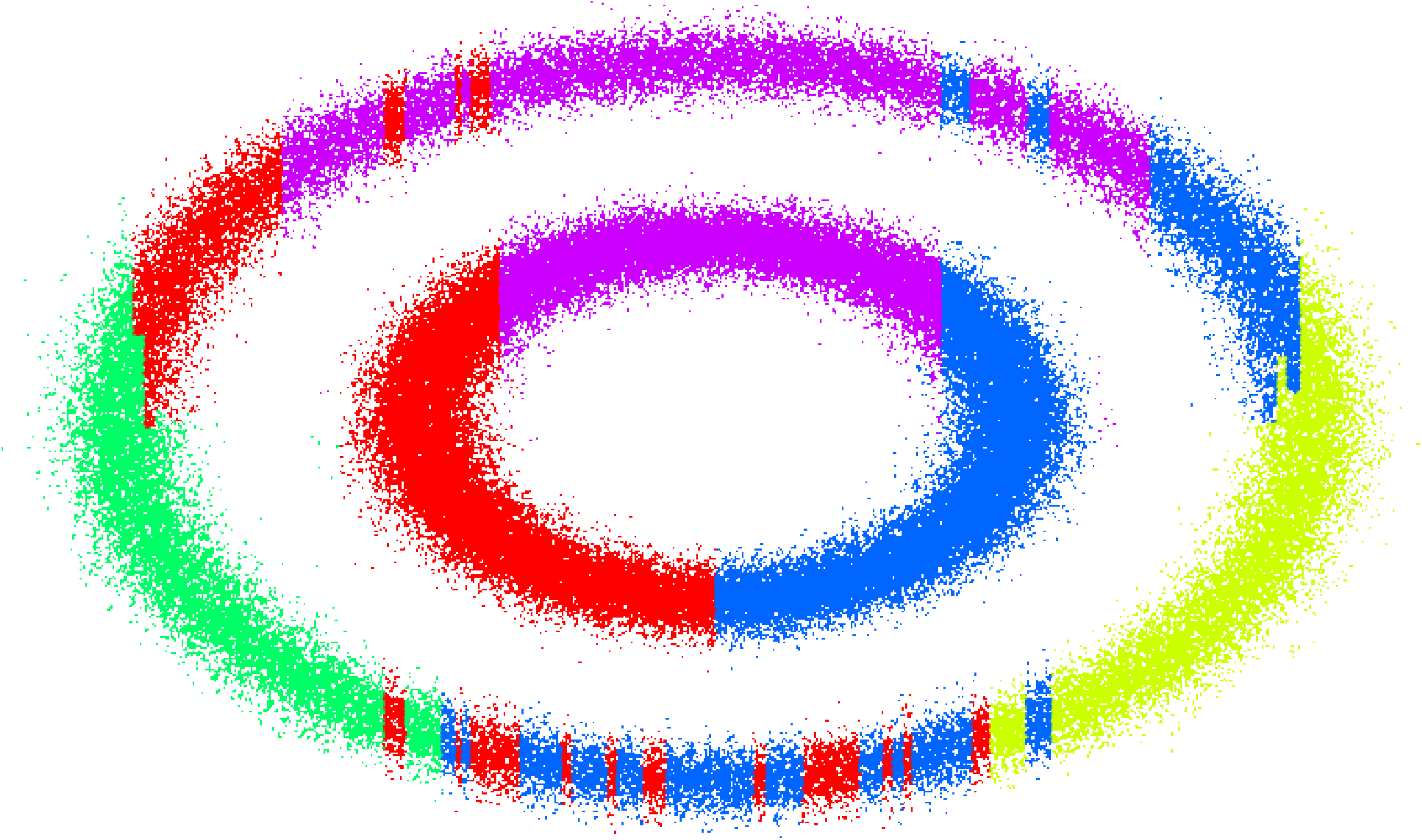}
  \end{subfigure}%
  \begin{subfigure}[b]{.25\linewidth}
    \centering
    \includegraphics[width=.99\textwidth]{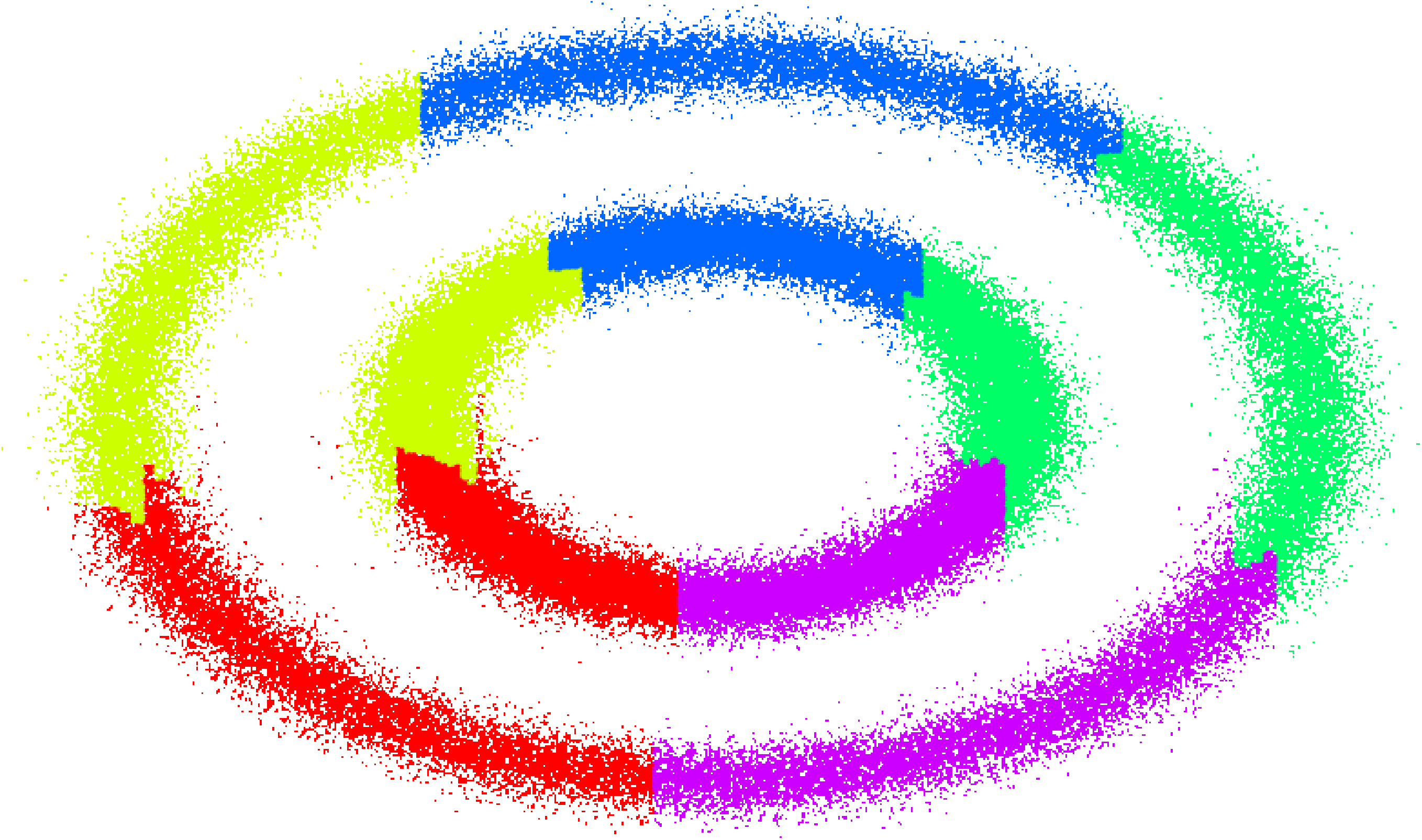}
  \end{subfigure}%
  \begin{subfigure}[b]{.25\linewidth}
    \centering
    \includegraphics[width=.99\textwidth]{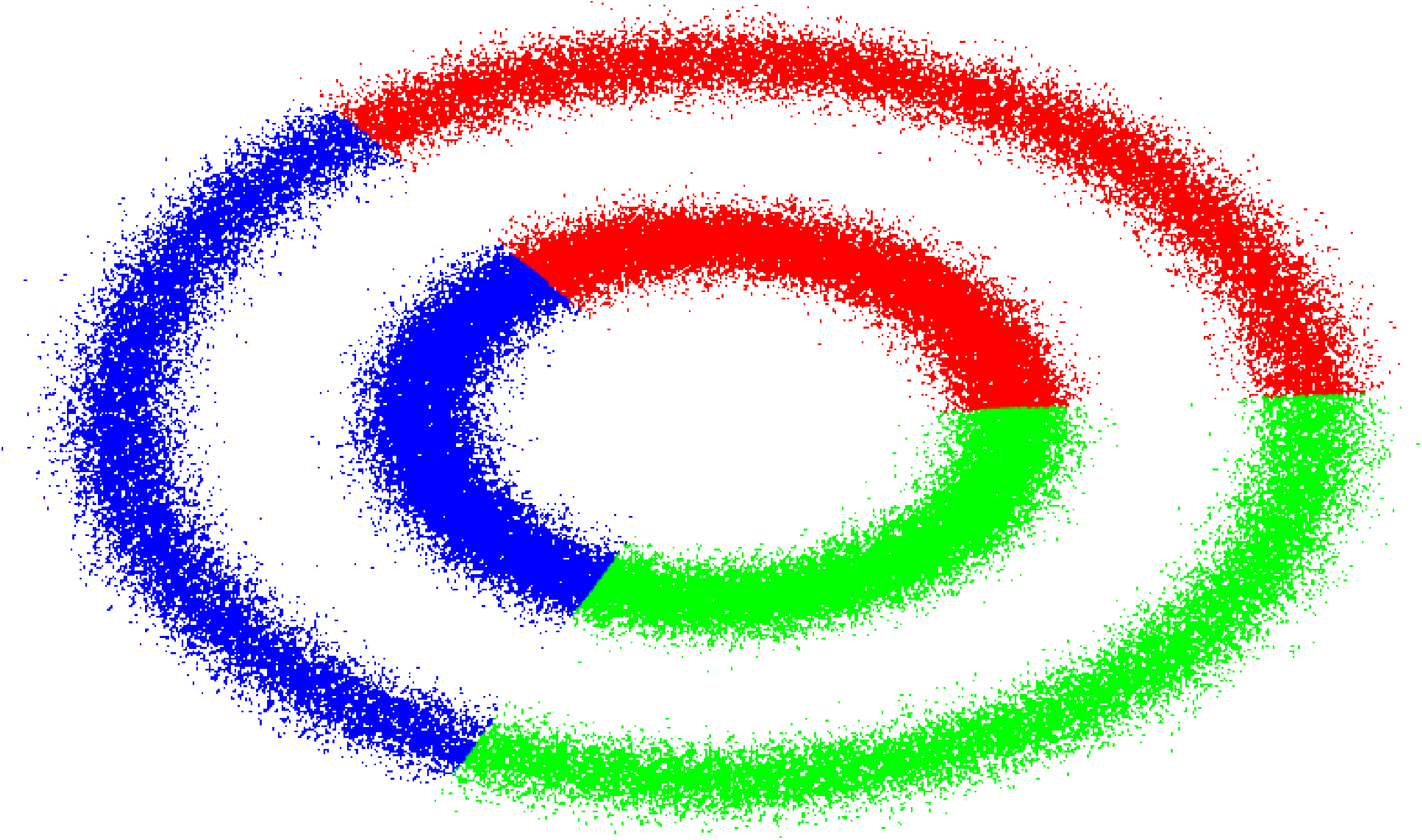}
  \end{subfigure}%
    \begin{subfigure}[b]{.25\linewidth}
    \centering
    \includegraphics[width=.99\textwidth]{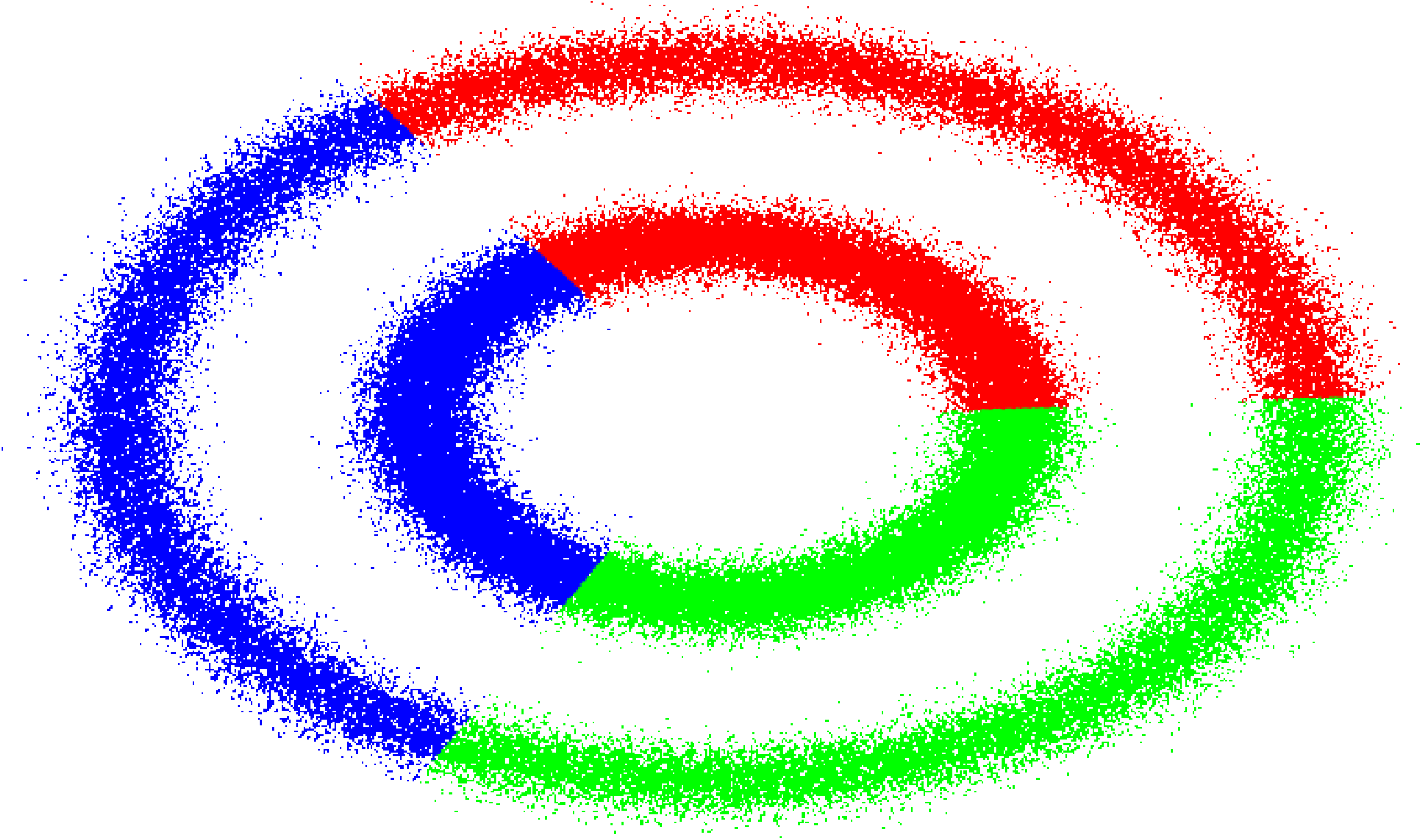}
  \end{subfigure}\\
  \begin{subfigure}[b]{.25\linewidth}
    \centering
    \includegraphics[width=.99\textwidth]{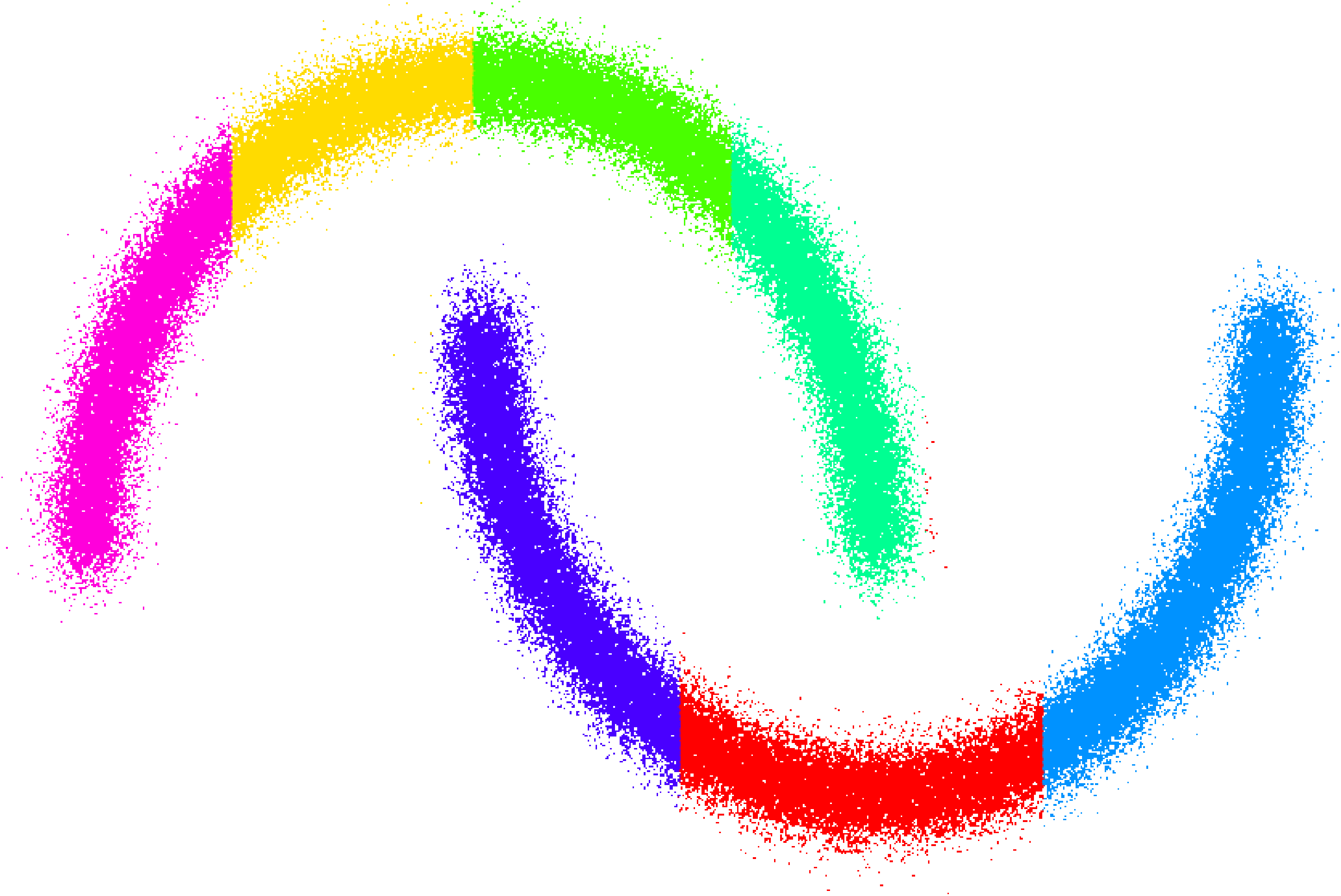}
  \end{subfigure}%
  \begin{subfigure}[b]{.25\linewidth}
    \centering
    \includegraphics[width=.99\textwidth]{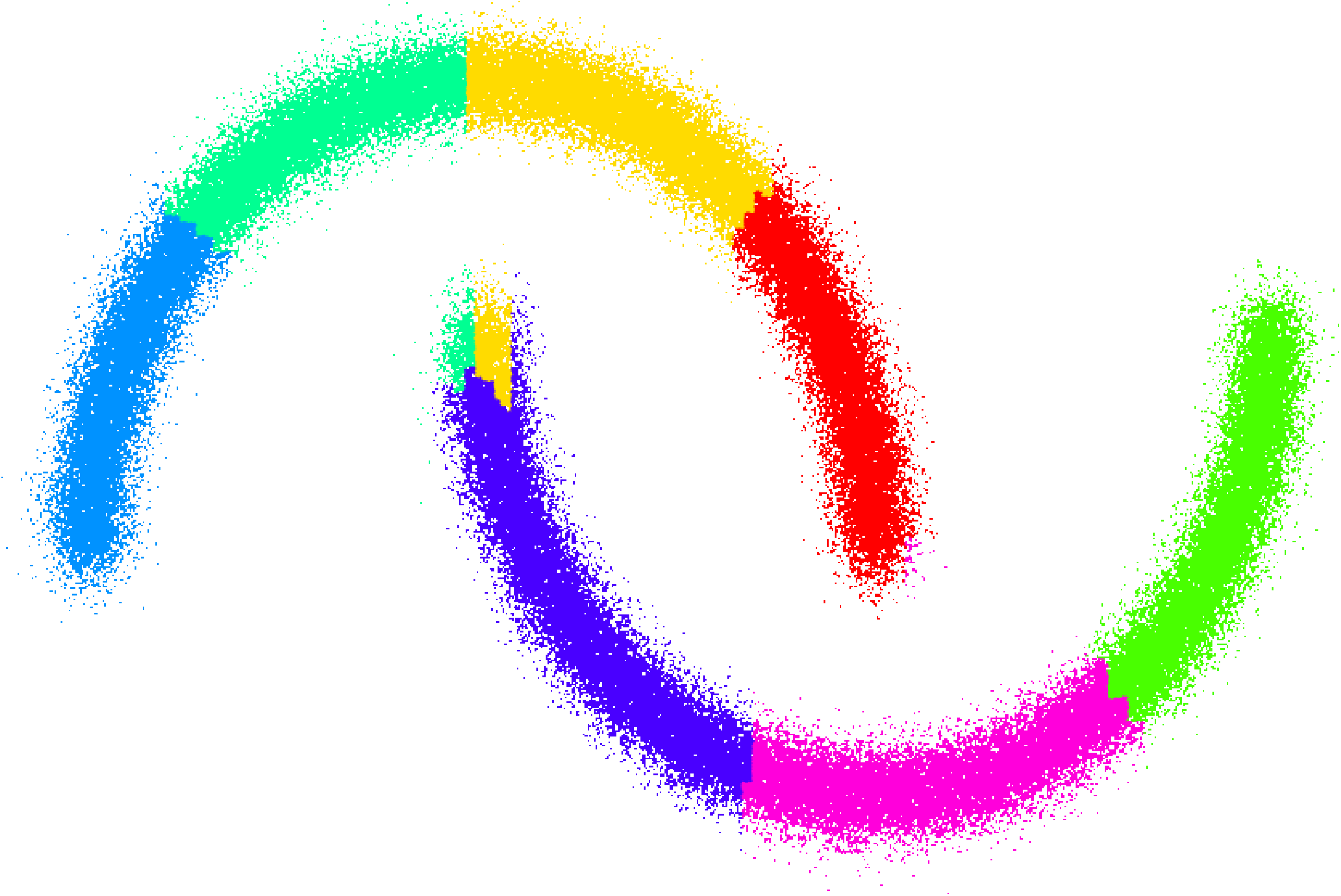}
  \end{subfigure}%
  \begin{subfigure}[b]{.25\linewidth}
    \centering
    \includegraphics[width=.99\textwidth]{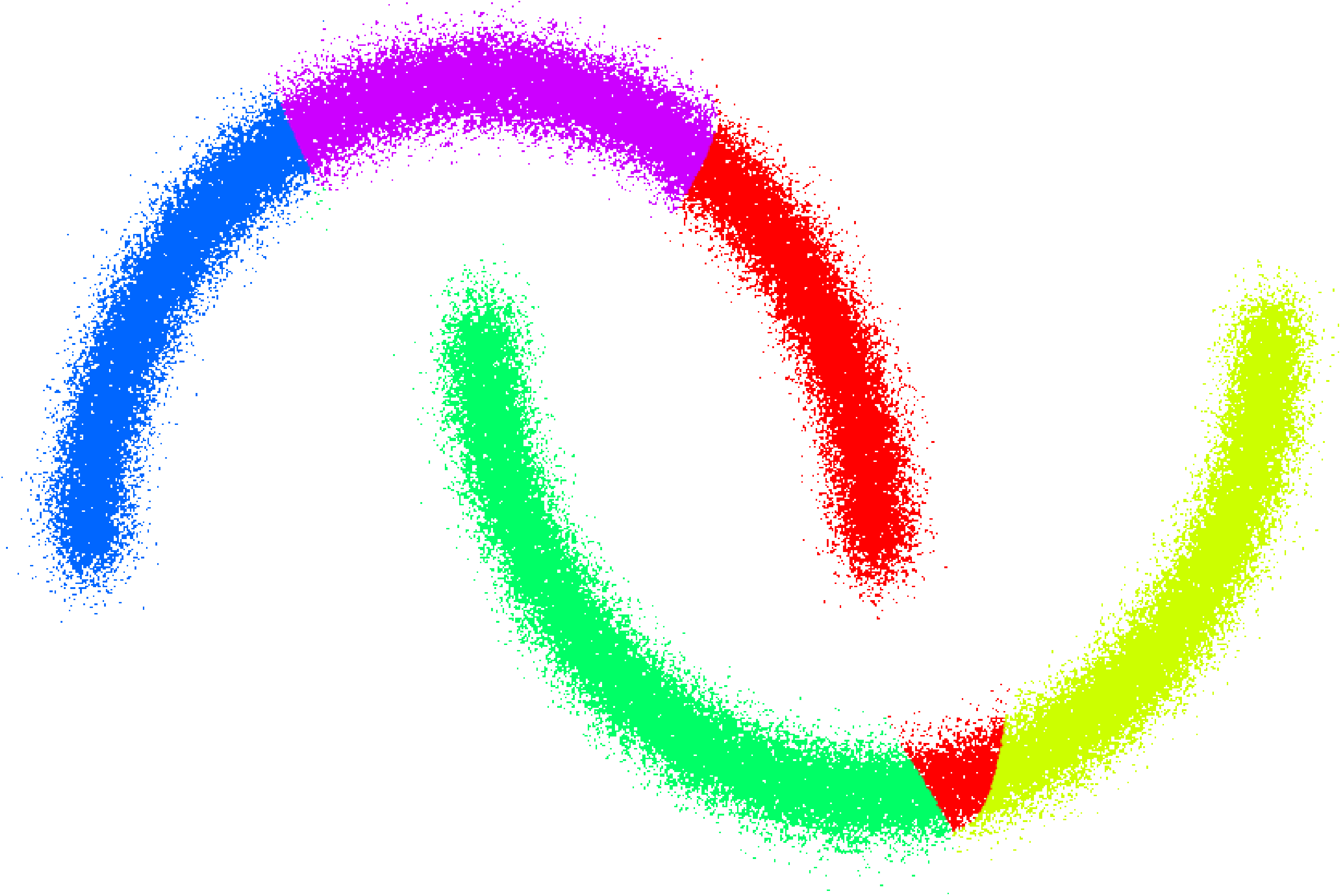}
  \end{subfigure}%
  \begin{subfigure}[b]{.25\linewidth}
    \centering
    \includegraphics[width=.99\textwidth]{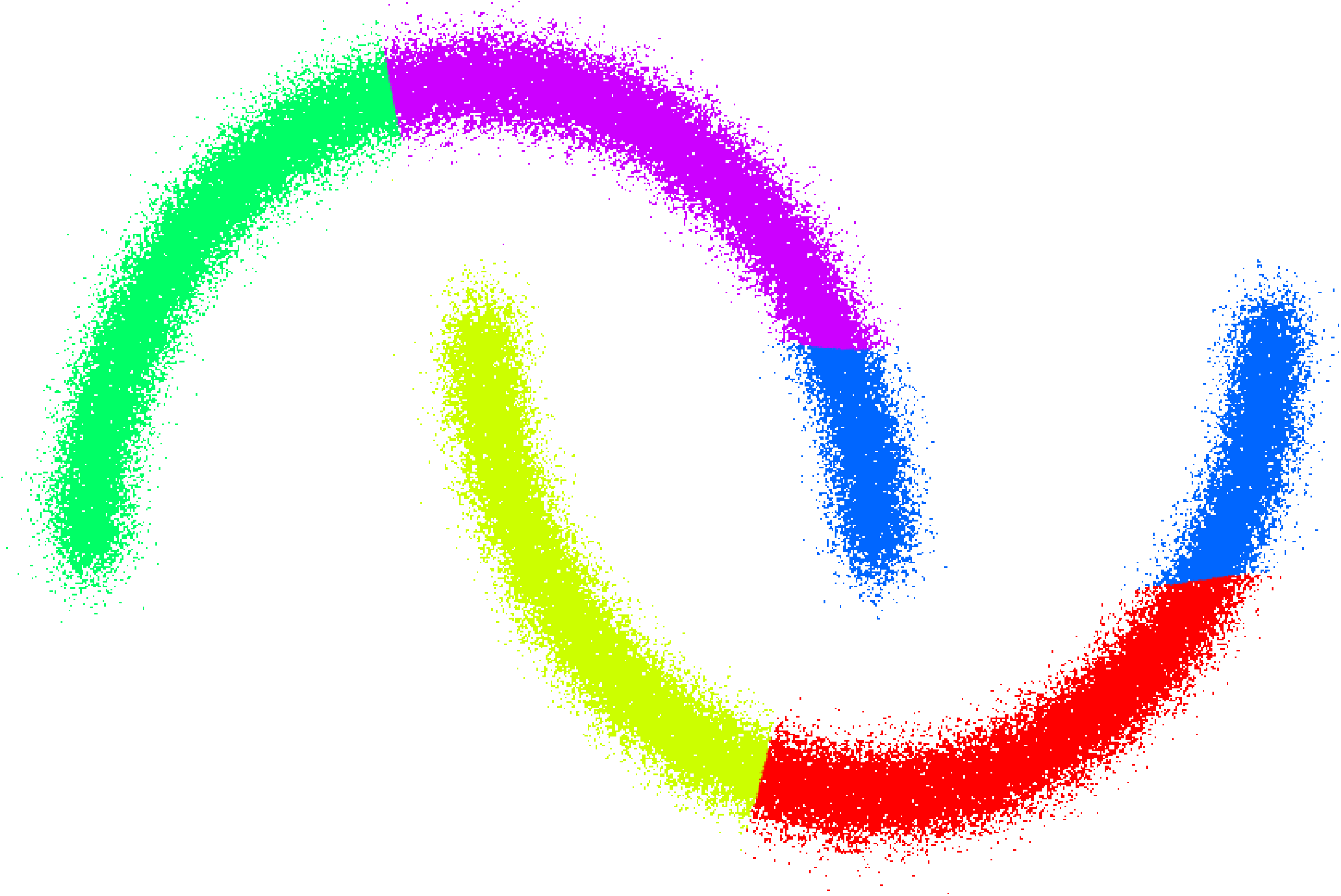}
  \end{subfigure}\\
    \begin{subfigure}[b]{.17\linewidth}
    \centering
    \includegraphics[width=.99\textwidth]{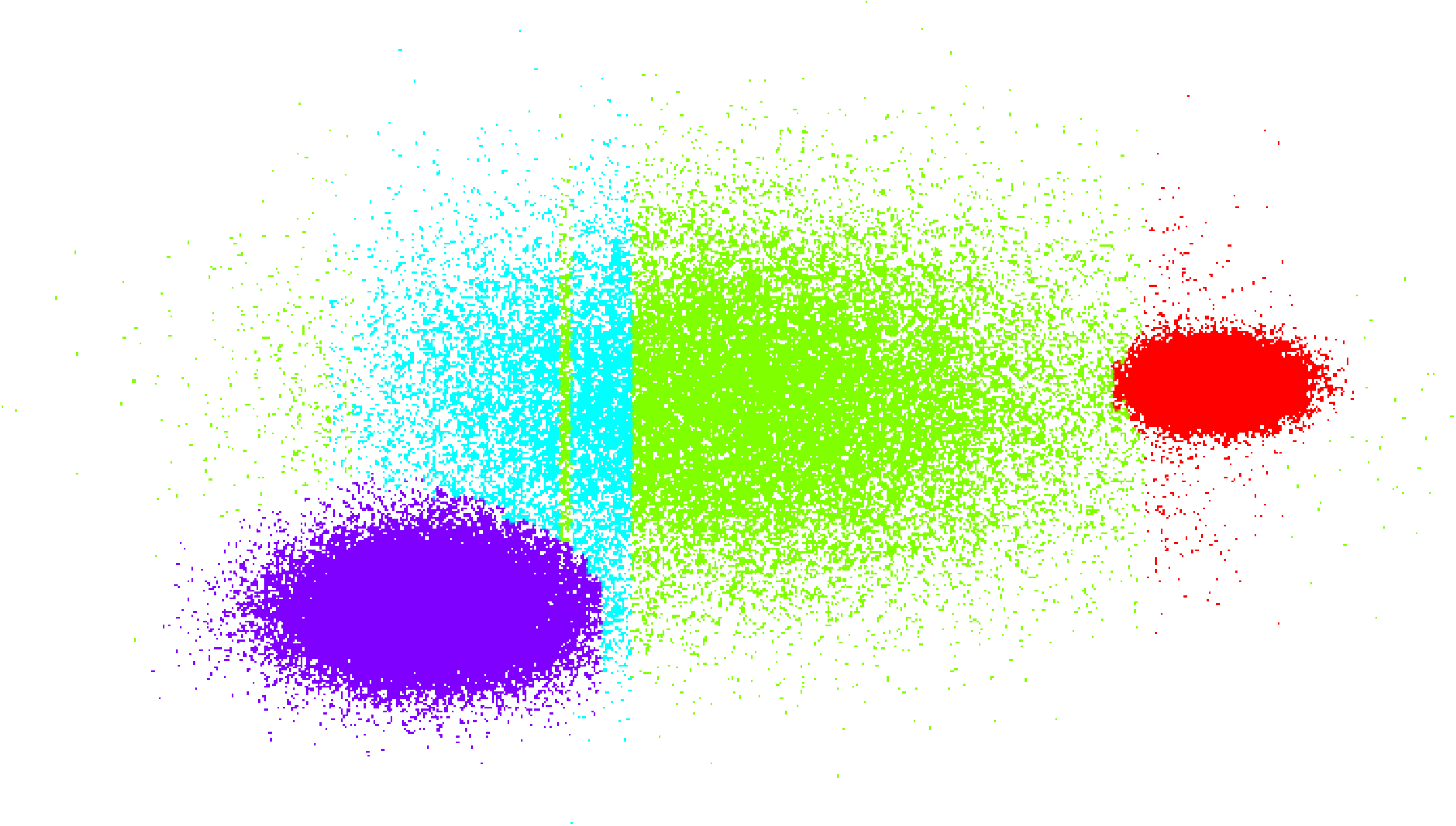}
  \end{subfigure}%
  \begin{subfigure}[b]{.25\linewidth}
    \centering
    \includegraphics[width=.99\textwidth]{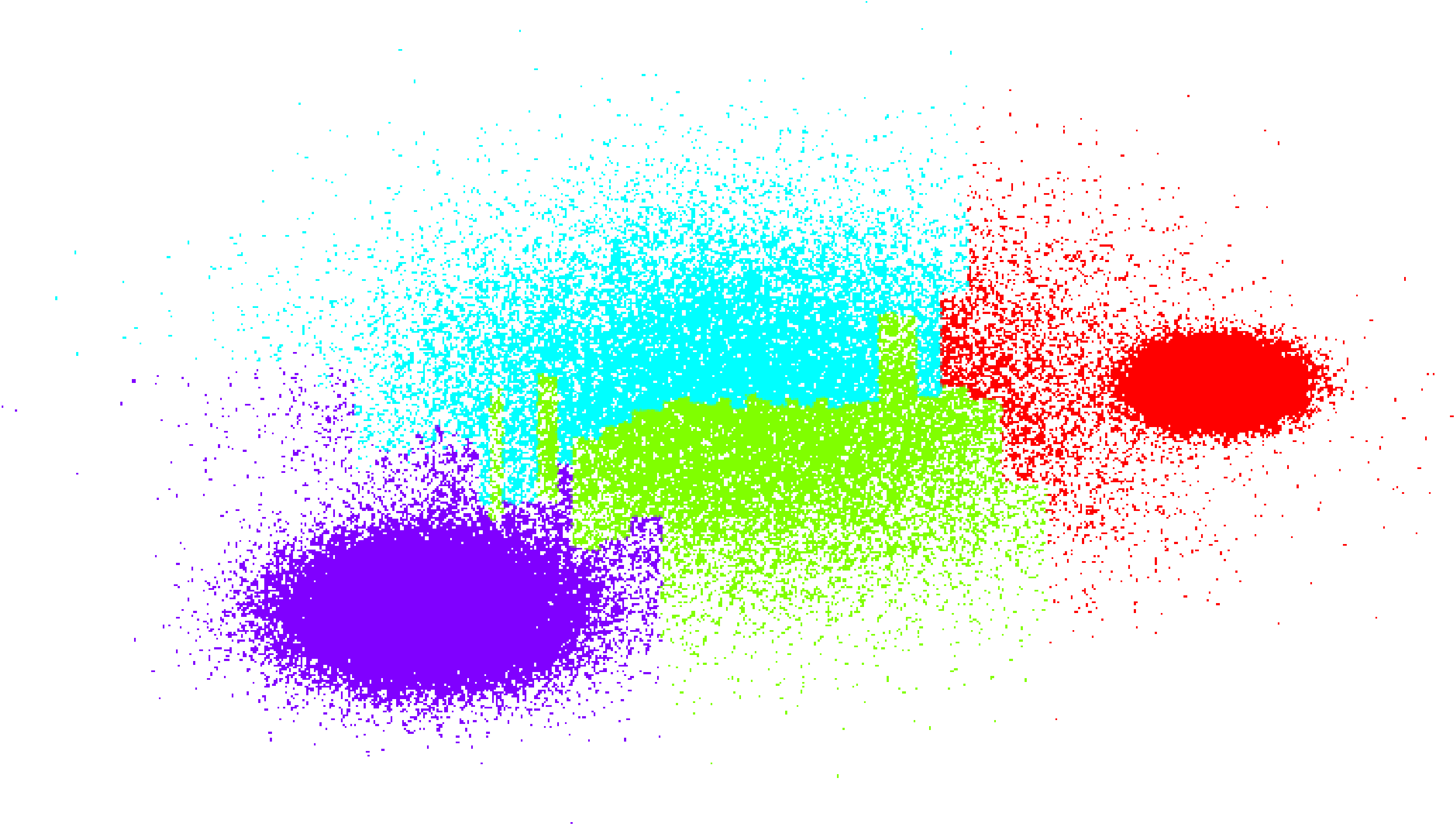}
  \end{subfigure}%
  \begin{subfigure}[b]{.25\linewidth}
    \centering
    \includegraphics[width=.99\textwidth]{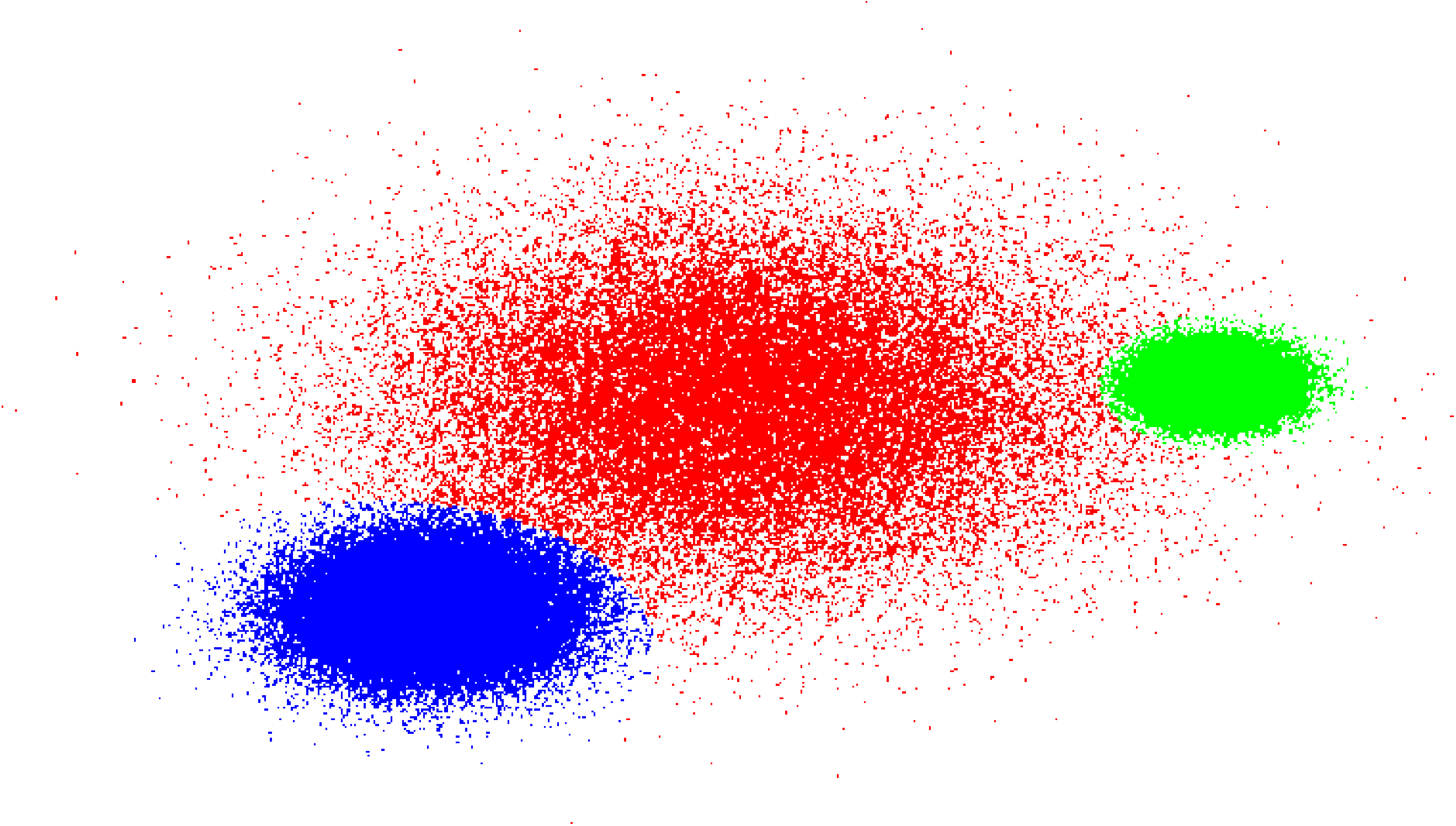}
  \end{subfigure}%
  \begin{subfigure}[b]{.25\linewidth}
    \centering
    \includegraphics[width=.99\textwidth]{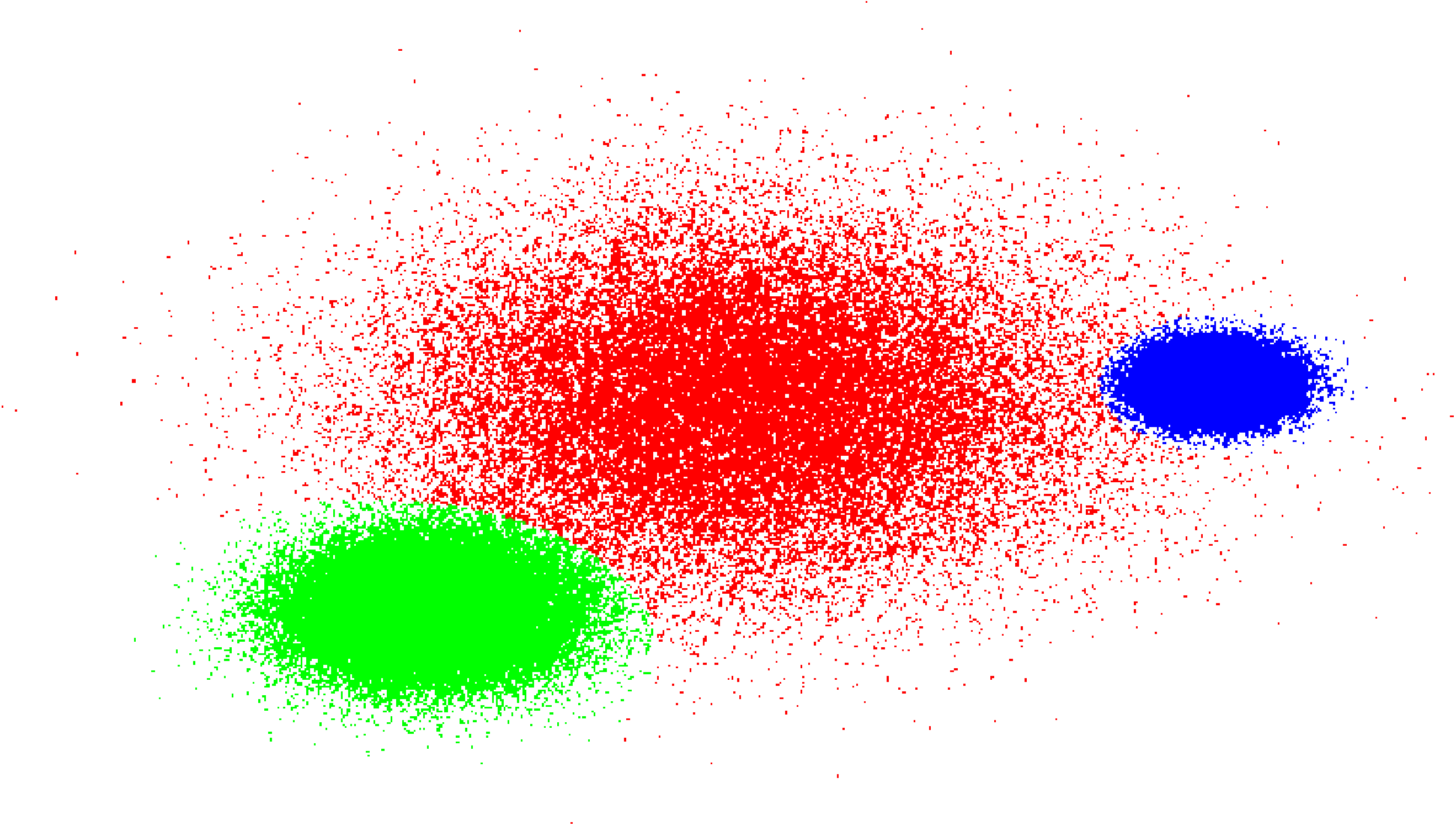}
  \end{subfigure}\\
    \begin{subfigure}[b]{.25\linewidth}
    \centering
    \includegraphics[width=.99\textwidth]{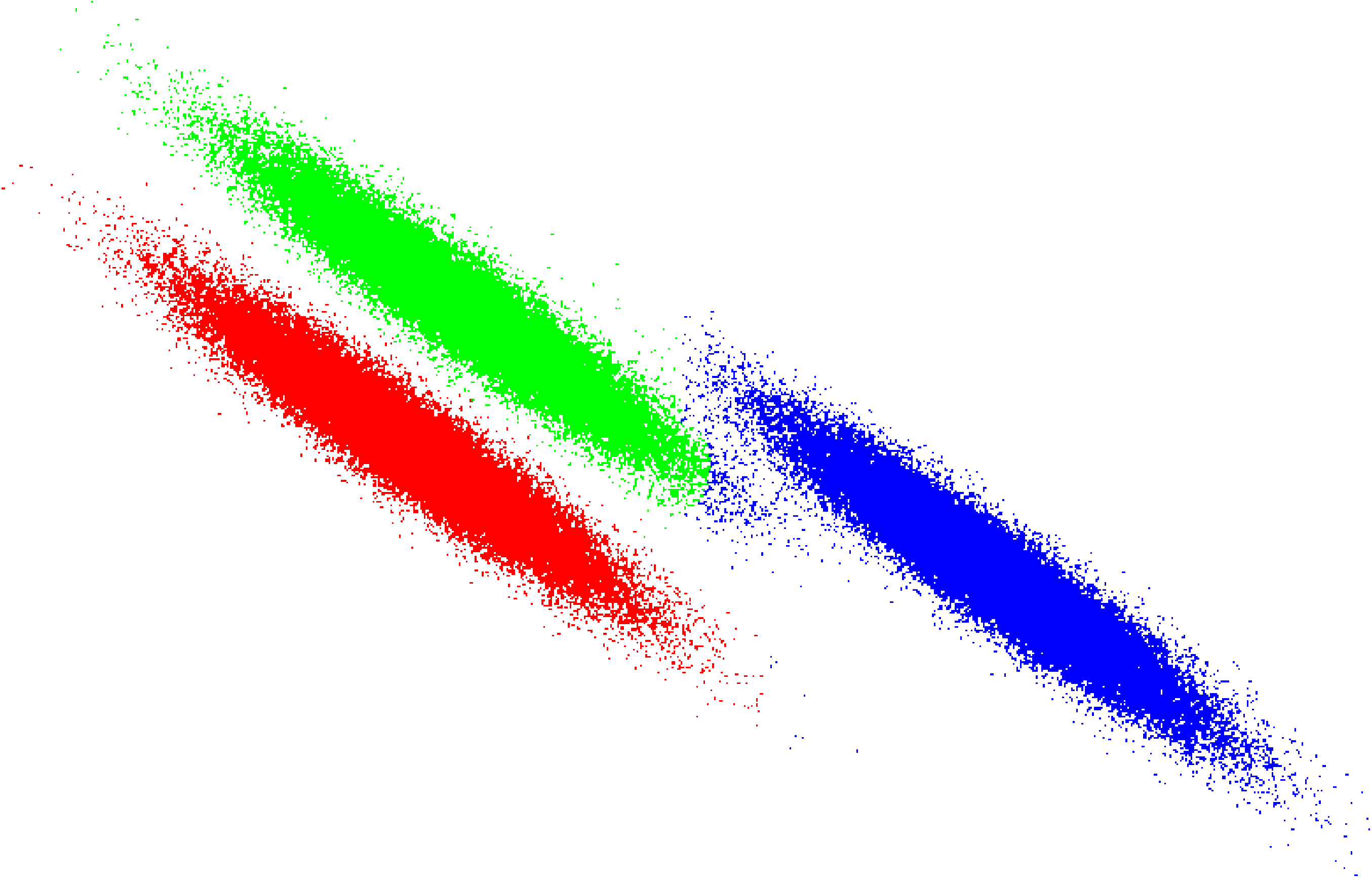}
  \end{subfigure}%
  \begin{subfigure}[b]{.25\linewidth}
    \centering
    \includegraphics[width=.99\textwidth]{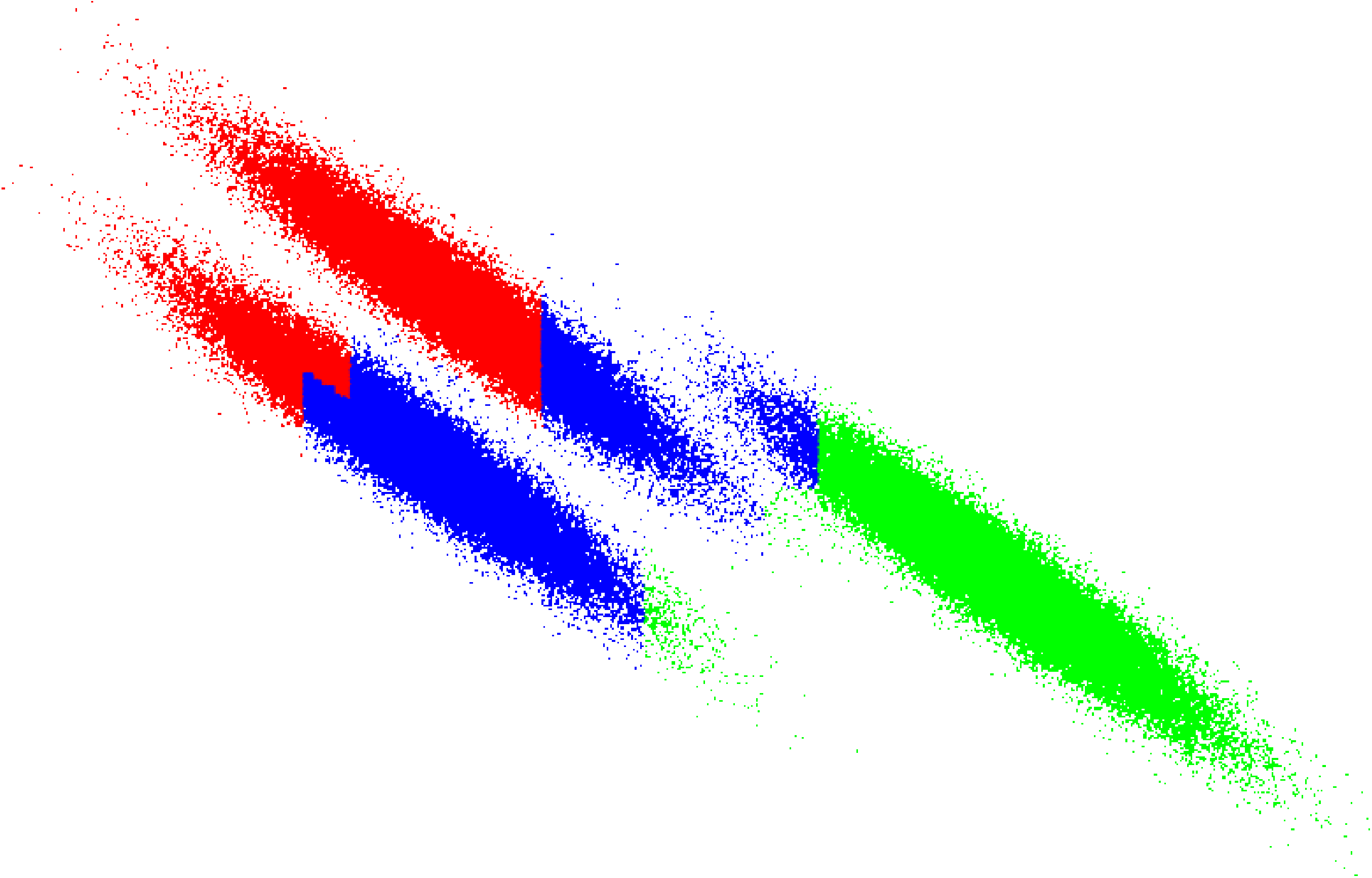}
  \end{subfigure}%
  \begin{subfigure}[b]{.25\linewidth}
    \centering
    \includegraphics[width=.99\textwidth]{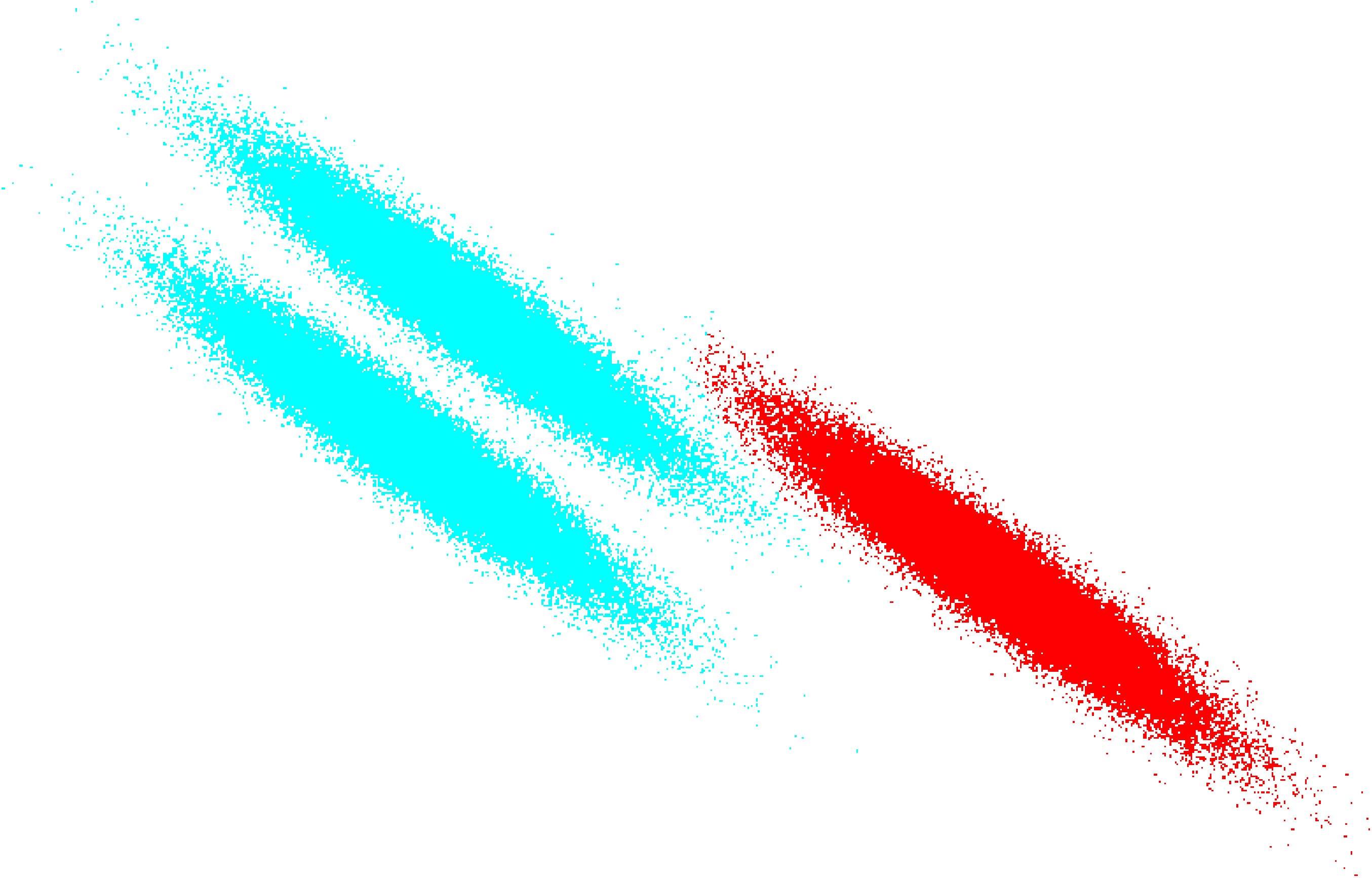}
  \end{subfigure}%
    \begin{subfigure}[b]{.25\linewidth}
    \centering
    \includegraphics[width=.99\textwidth]{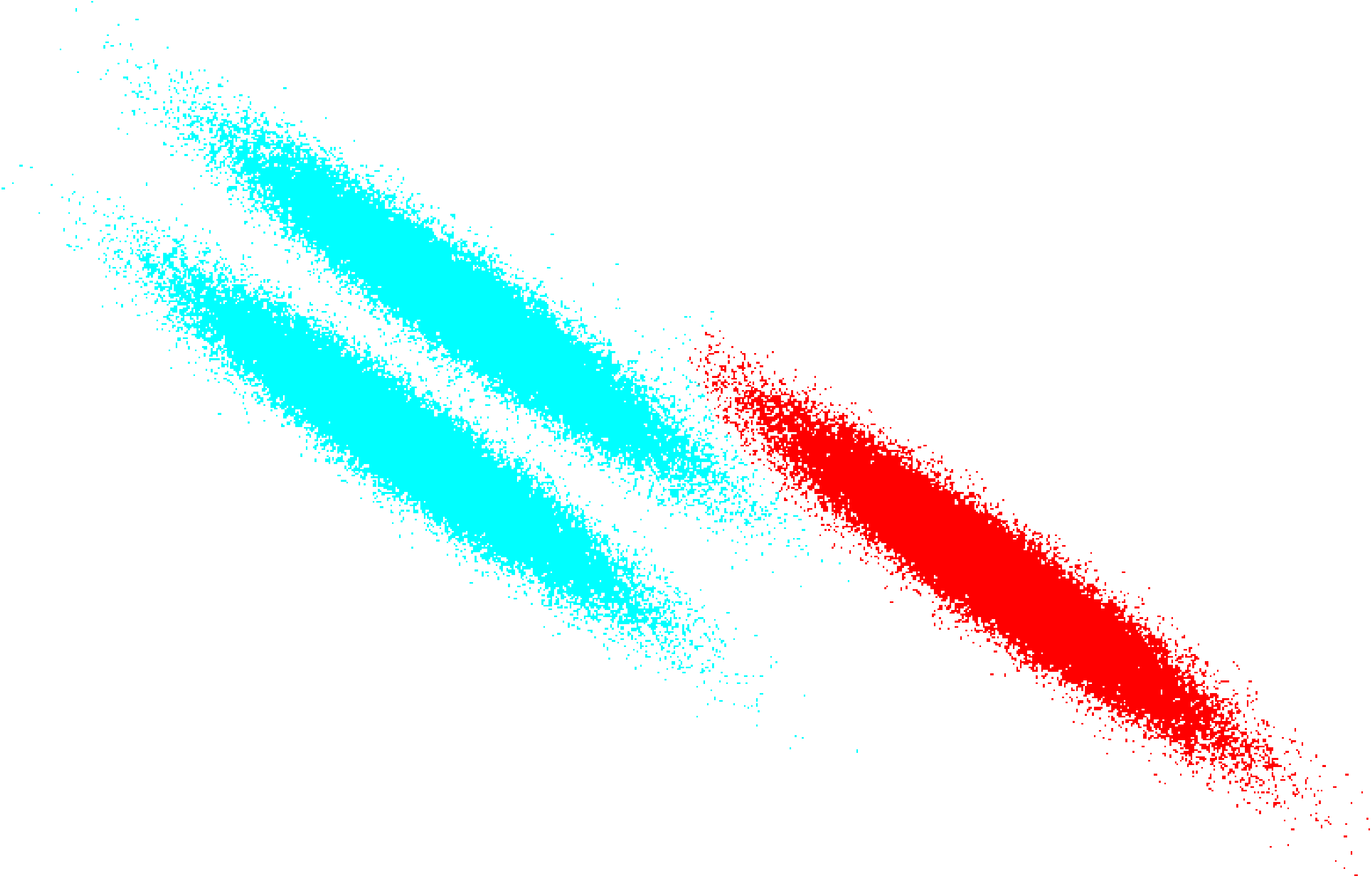}
  \end{subfigure}\\
  \begin{subfigure}[b]{.25\linewidth}
    \centering
    \includegraphics[width=.99\textwidth]{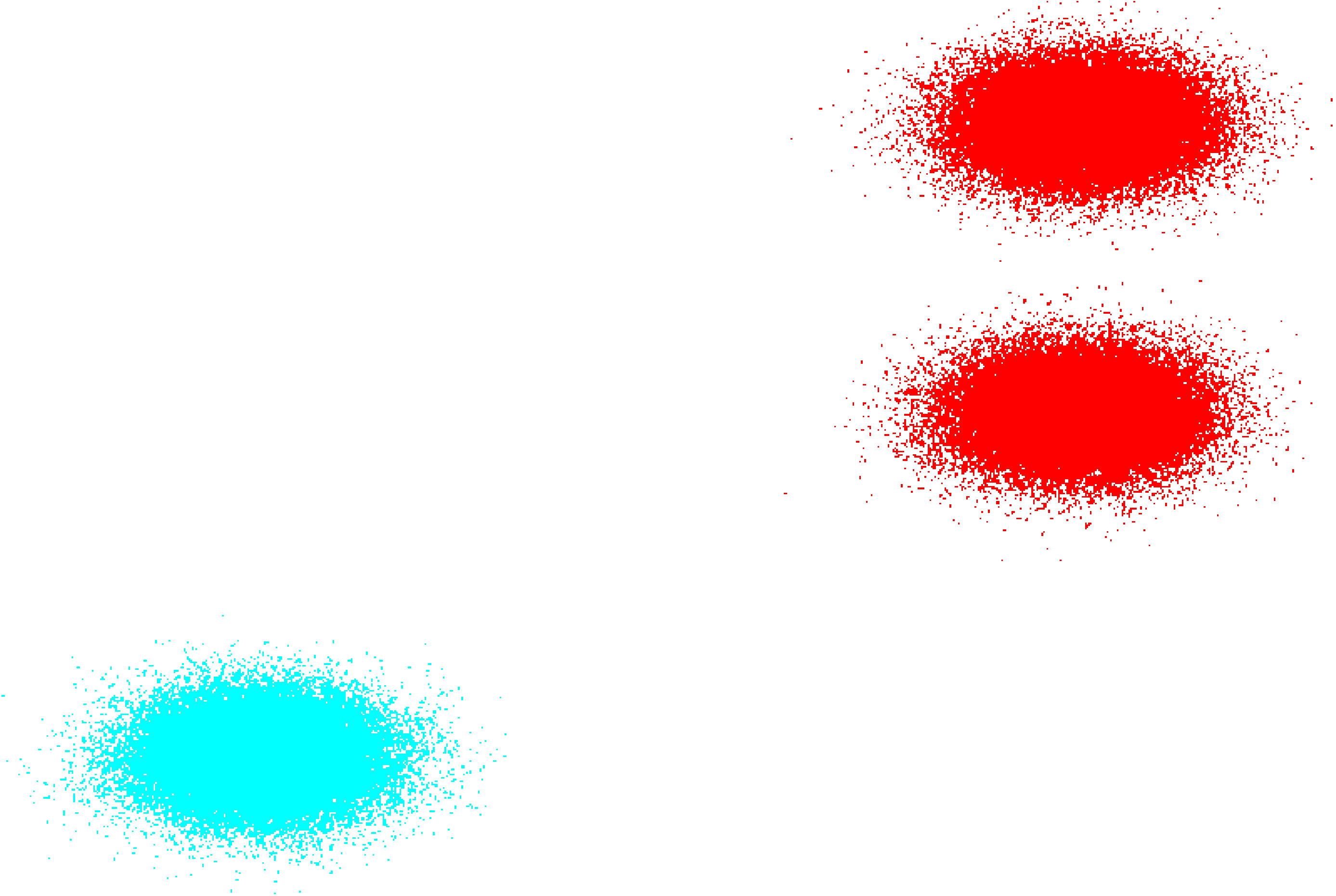}
  \end{subfigure}%
  \begin{subfigure}[b]{.25\linewidth}
    \centering
    \includegraphics[width=.99\textwidth]{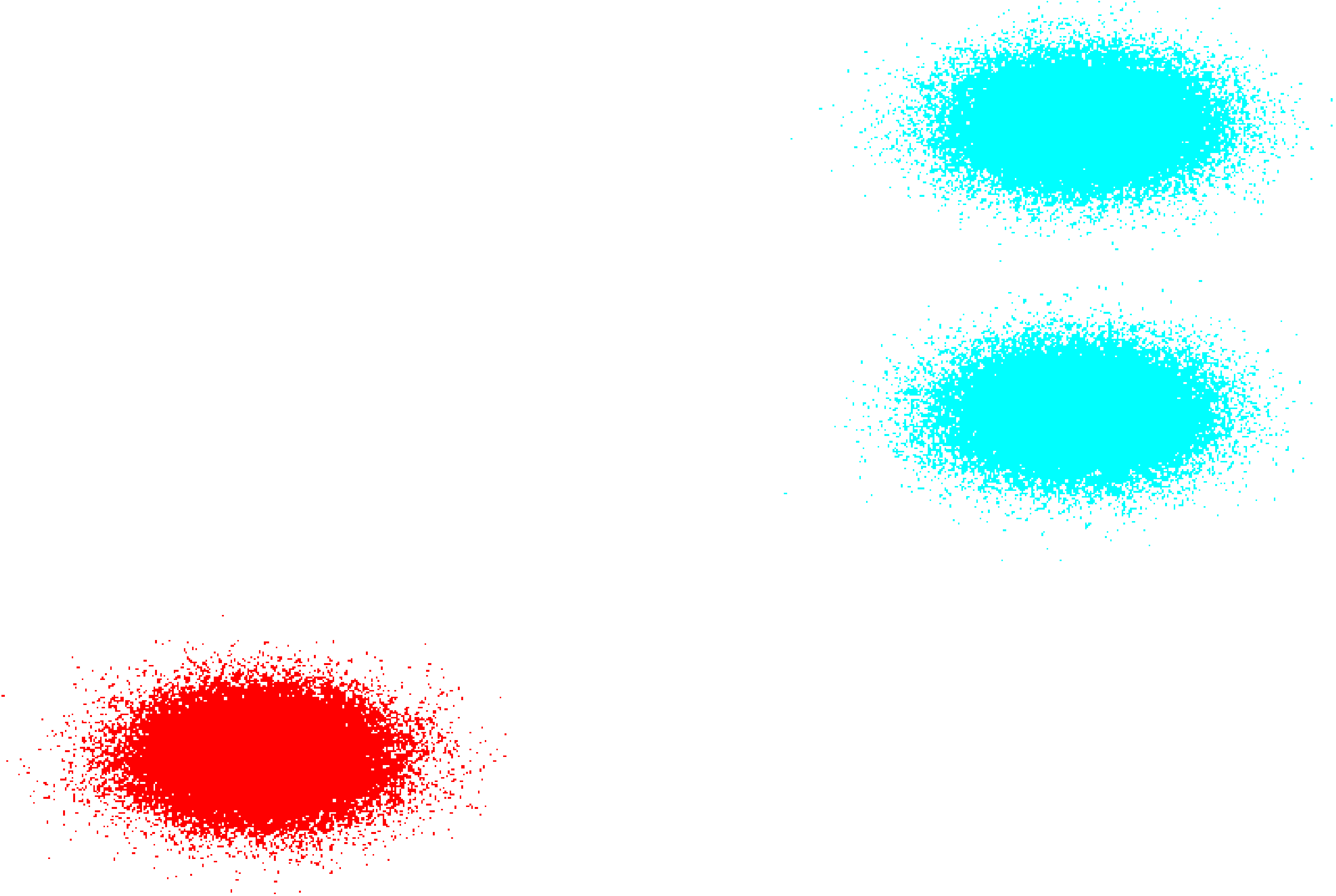}
  \end{subfigure}%
  \begin{subfigure}[b]{.25\linewidth}
    \centering
    \includegraphics[width=.99\textwidth]{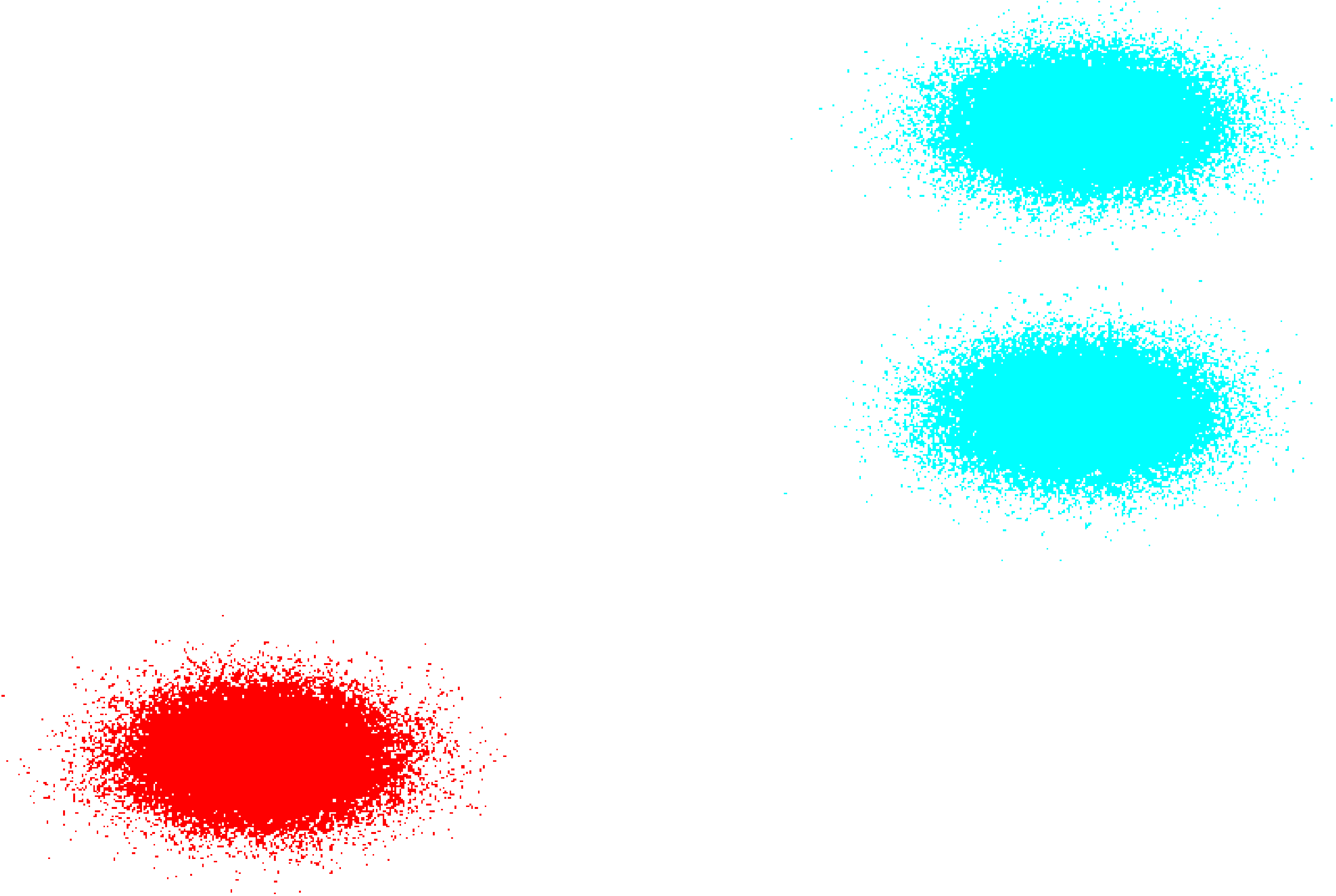}
  \end{subfigure}%
  \begin{subfigure}[b]{.25\linewidth}
    \centering
    \includegraphics[width=.99\textwidth]{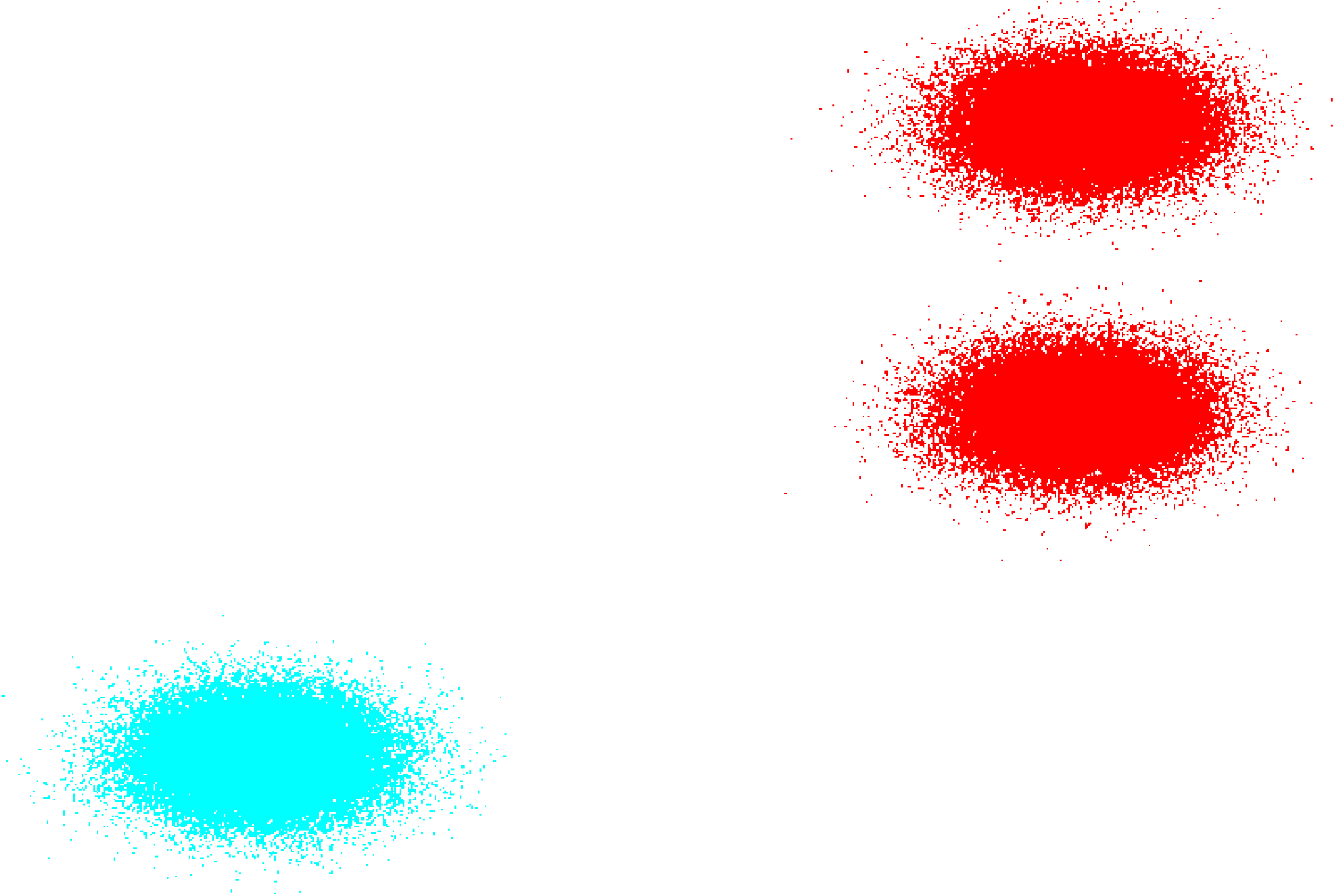}
  \end{subfigure}\\
    \begin{subfigure}[b]{.25\linewidth}
    \centering
    \includegraphics[width=.99\textwidth]{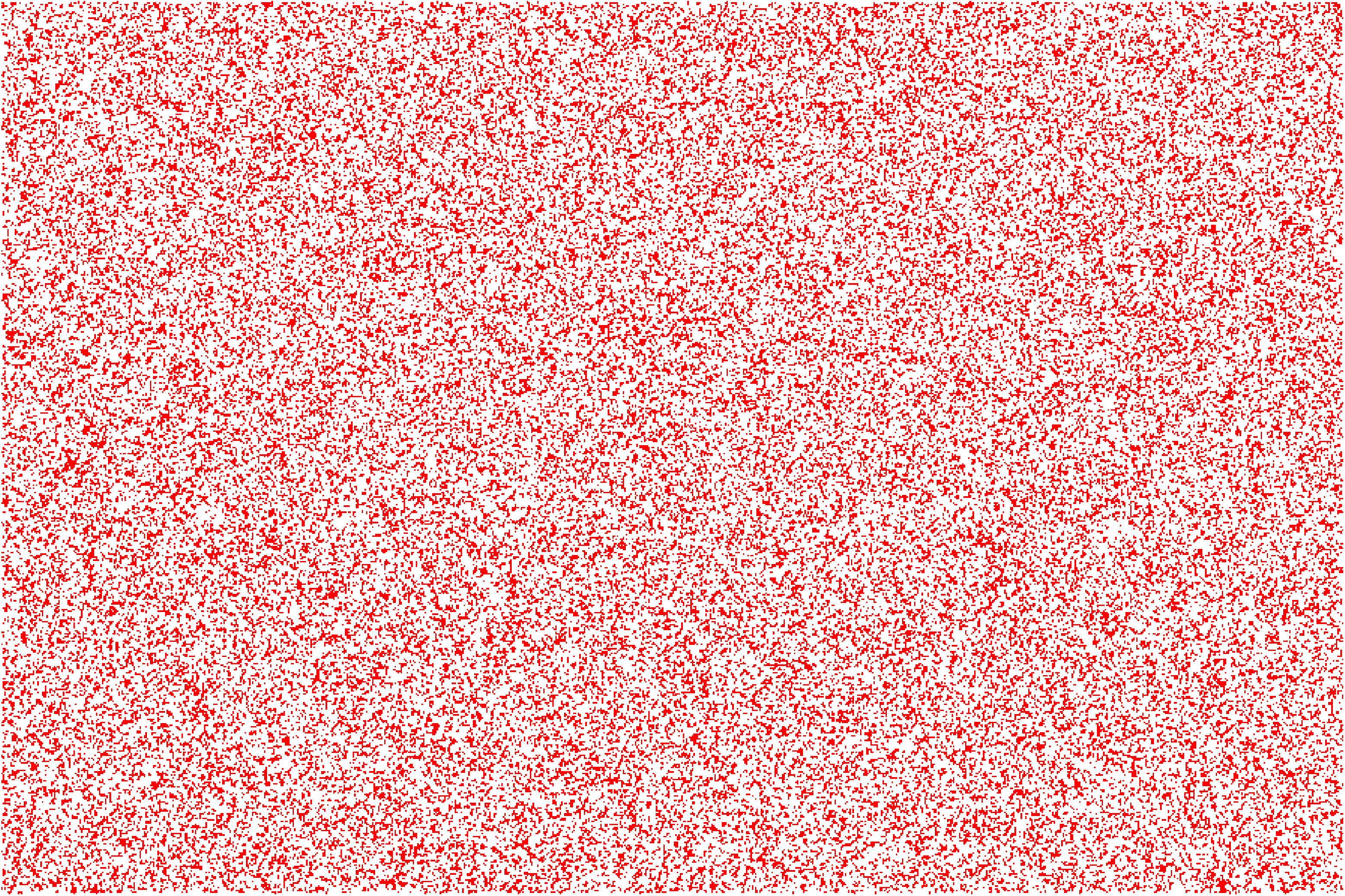}
    \caption{OPWG}
  \end{subfigure}%
  \begin{subfigure}[b]{.25\linewidth}
    \centering
    \includegraphics[width=.99\textwidth]{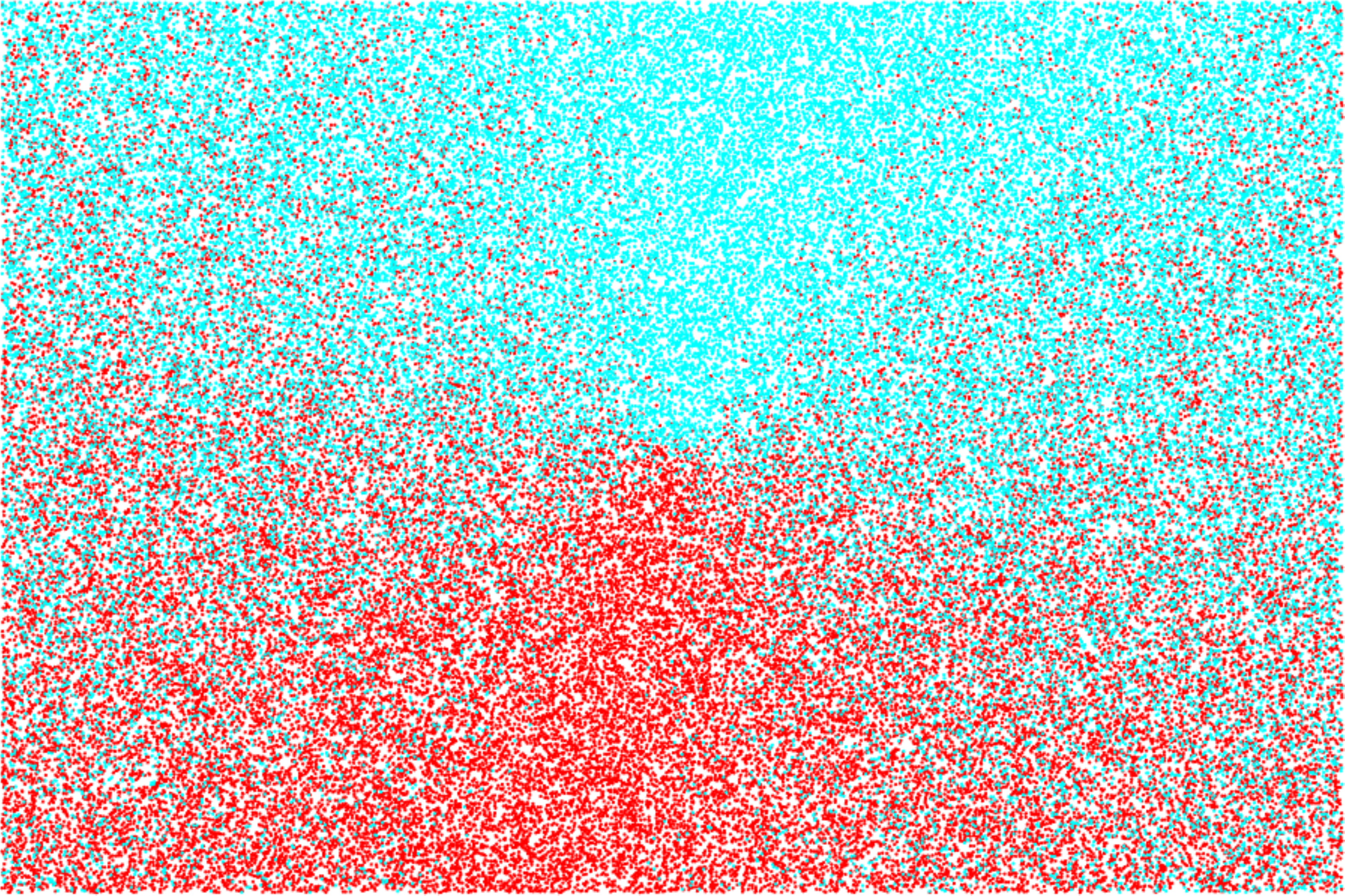}
    \caption{OFCM}
  \end{subfigure}%
  \begin{subfigure}[b]{.25\linewidth}
    \centering
    \includegraphics[width=.99\textwidth]{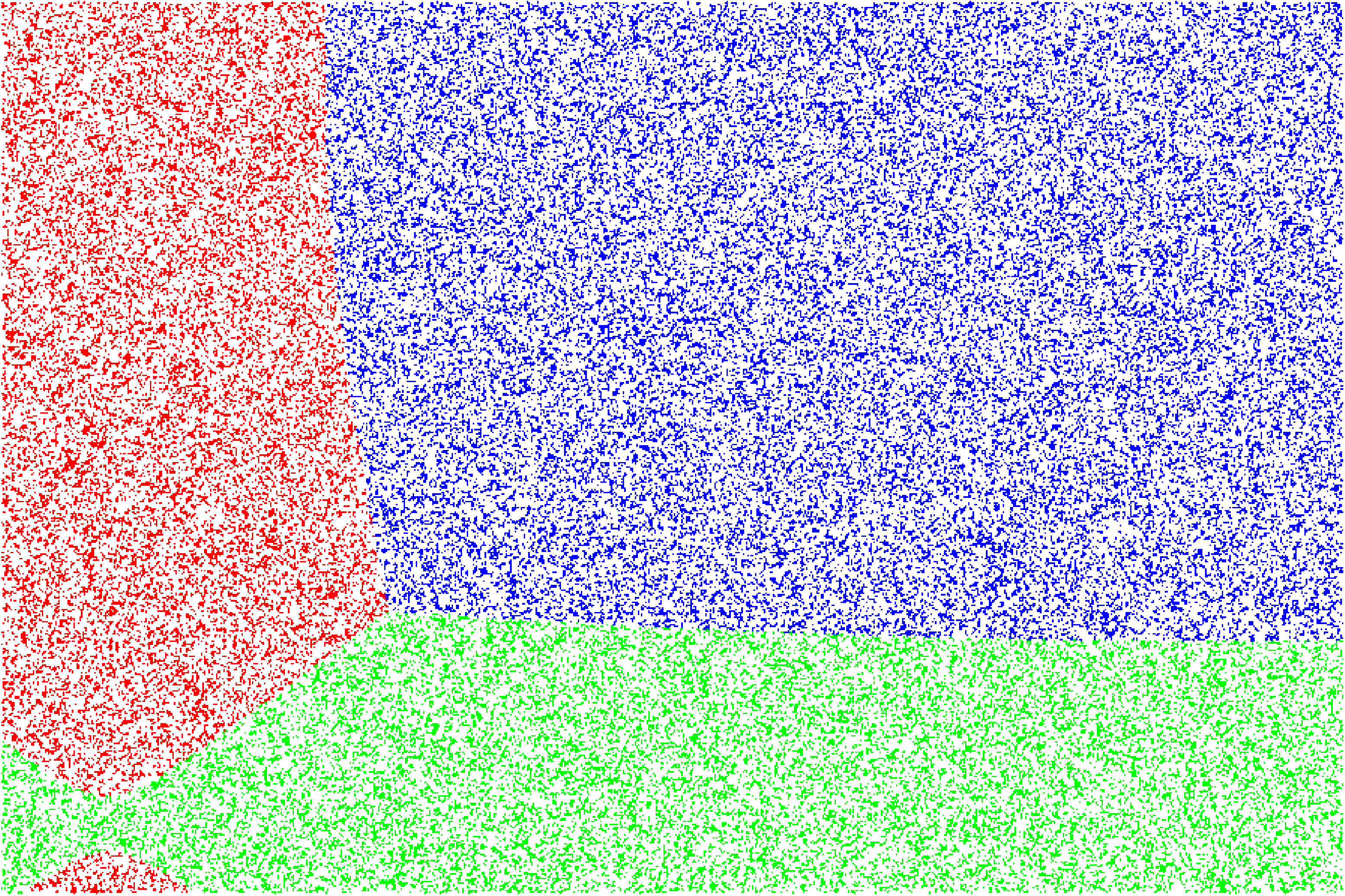}
    \caption{PGMM}
  \end{subfigure}%
  \begin{subfigure}[b]{.25\linewidth}
    \centering
    \includegraphics[width=.99\textwidth]{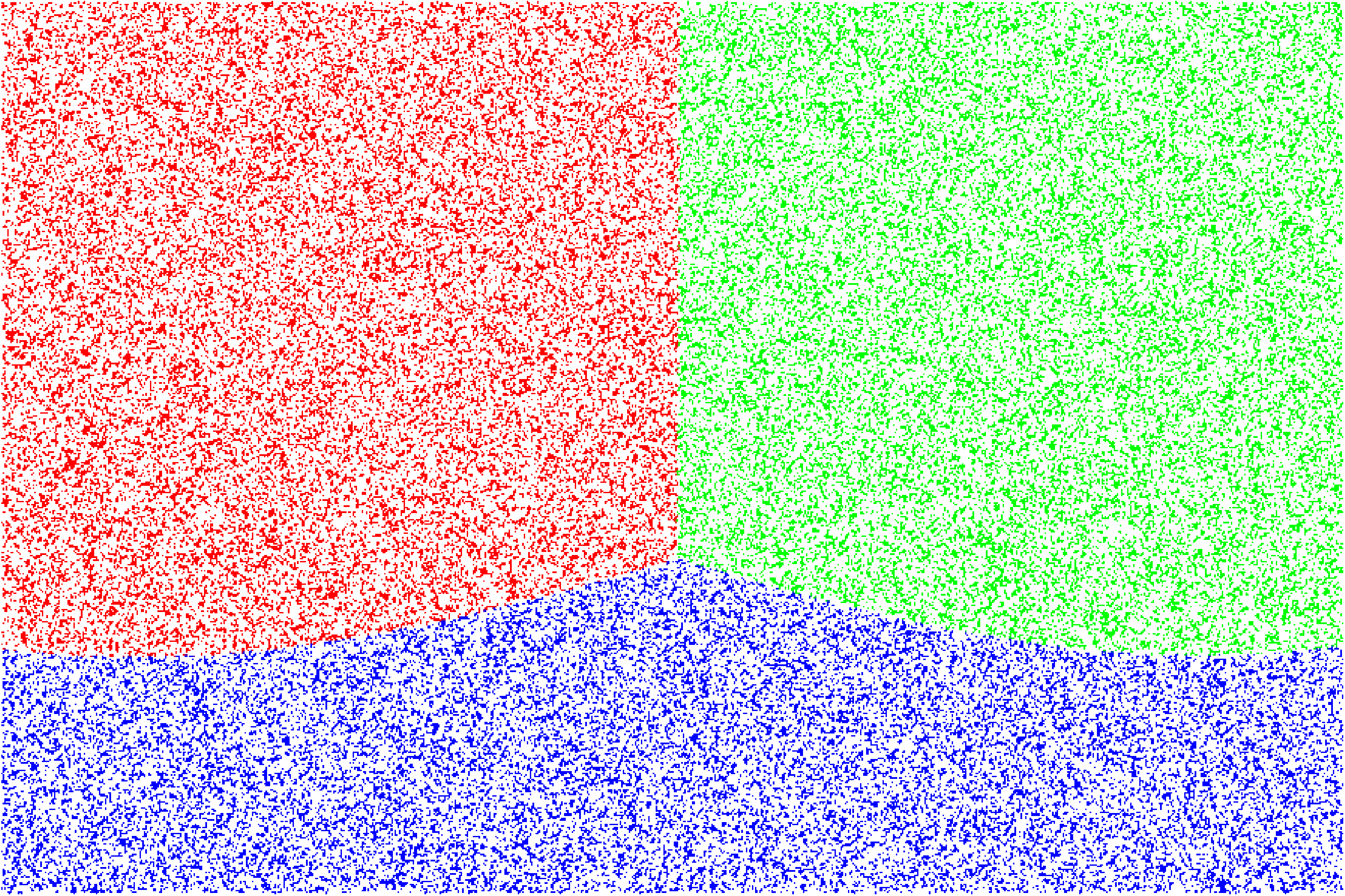}
    \caption{GMM}
  \end{subfigure}\\
\caption
{Mode B. Each algorithm was evaluated on six different datasets. For OPWG and PGMM, the initial number of clusters was 25. OFCM and GMM were initialized with the number of clusters found in OPWG and PGMM, respectively.
}\label{fig:6db_sorted}
\end{figure}

The results for mode B are reported in Table \ref{table:6db_sorted} and illustrated in Figure \ref{fig:6db_sorted}.  In Mode B not all clusters are represented in each batch. While this fact has a little effect on the performance of PGMM and GMM, it significantly affects the online algorithms. However, as can be seen, the OPWG is less affected than the OFCM by this phenomena. Note that, in this mode,  because only subset of the data is represented in each batch, the average number of clusters that the OPWG and PGMM found increased compared to mode A. The average number that OPWG found was $K= 4.84, 7.84, 4.48, 3.04, 2.14, 1.02 $ and the average number of clusters that PGMM found was $K=3.25, 5.4, 3.08, 2.48, 2.32, 2.9$.


\begin{table*}
\begin{center}
\begin{tabular}{lllllllllll}
\hline
            & \multicolumn{2}{c}{DB1}       & \multicolumn{2}{c}{DB2}       & \multicolumn{2}{c}{DB3}       & \multicolumn{2}{c}{DB4}       & \multicolumn{2}{c}{DB5}       \\
            & F1            & NMI           & F1            & NMI           & F1            & NMI           & F1            & NMI           & F1            & NMI           \\ \cline{2-11} 
OPWG  & \textbf{0.41} & \textbf{0.10} & 0.46          & 0.51          & 0.93          & 0.83          & \textbf{0.99} & \textbf{0.98} & 0.78          & 0.73          \\
OFCM & 0.31          & 0.00          & 0.45          & 0.47          & 0.83          & 0.75          & 0.83          & 0.60          & 0.80          & 0.77          \\
PGMM         & 0.39          & 0.00          & 0.54          & \textbf{0.55} & \textbf{0.98} & \textbf{0.93} & 0.87          & 0.84          & 0.82          & 0.78          \\
GMM         & 0.38          & 0.00          & \textbf{0.55} & 0.54          & \textbf{0.98} & \textbf{0.93} & 0.86          & 0.82          & \textbf{0.83} & \textbf{0.79} \\ \hline
\end{tabular}
\caption
{Mode B. Clustering quality measures for each of the algorithms discussed for all datasets considered apart from the last one, which is noise.
}\label{table:6db_sorted}
\end{center}
\end{table*}

The algorithms were further evaluated on 2D Gaussian mixtures datasets with a different number of components ($K = \{ 2, 5, 7\}$). As before, each result reported represents the average performance of running the algorithm 50 times. In this case, $\lambda$ was chosen to be $0.006$ for both modes and in mode B the dataset was arbitrarily sorted according to its y coordinate.

\begin{figure}[htb]
\centering
  \begin{subfigure}[b]{.25\linewidth}
    \centering
    \includegraphics[width=.99\textwidth]{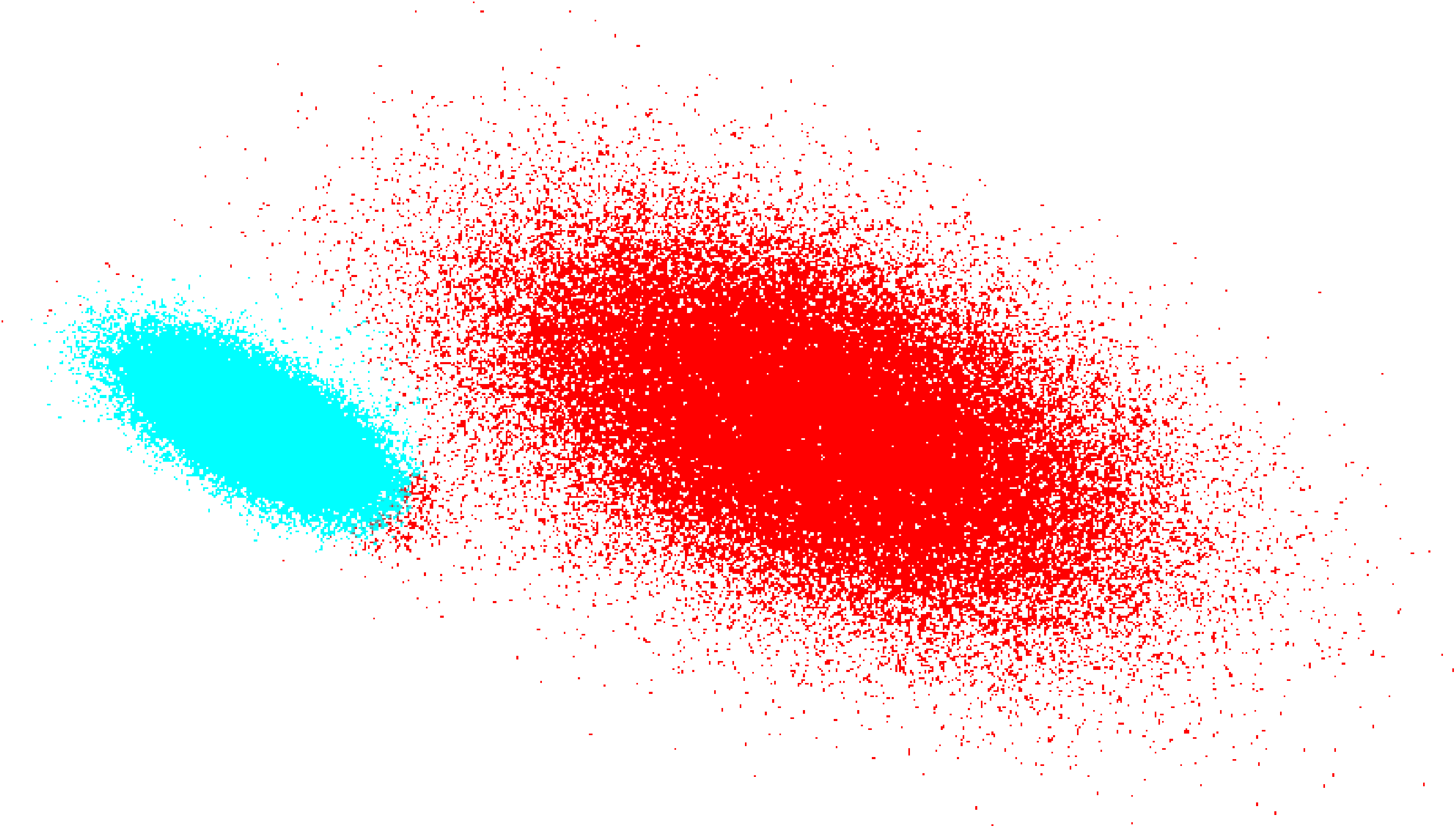}
  \end{subfigure}%
  \begin{subfigure}[b]{.25\linewidth}
    \centering
    \includegraphics[width=.99\textwidth]{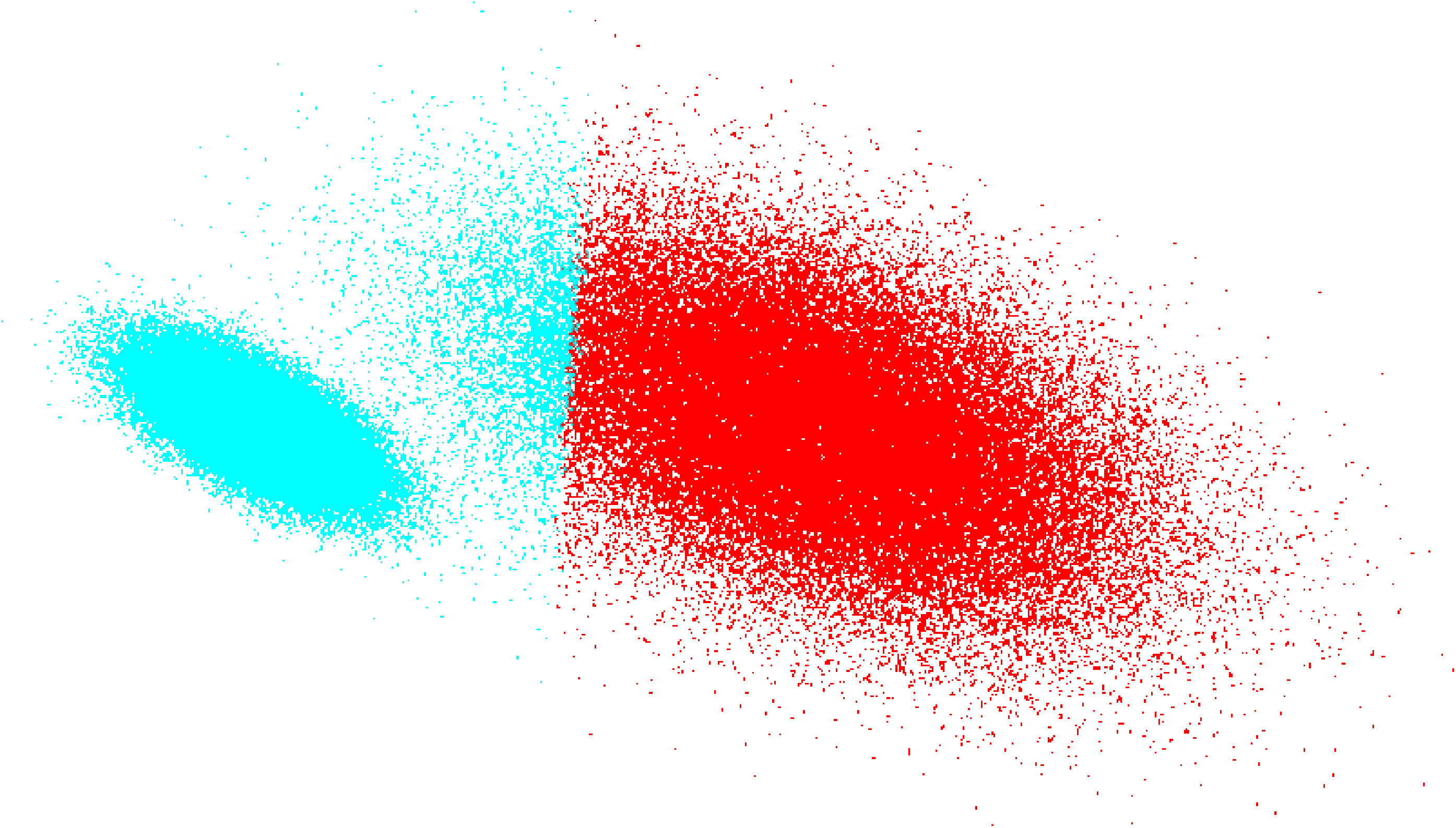}
  \end{subfigure}%
  \begin{subfigure}[b]{.25\linewidth}
    \centering
    \includegraphics[width=.99\textwidth]{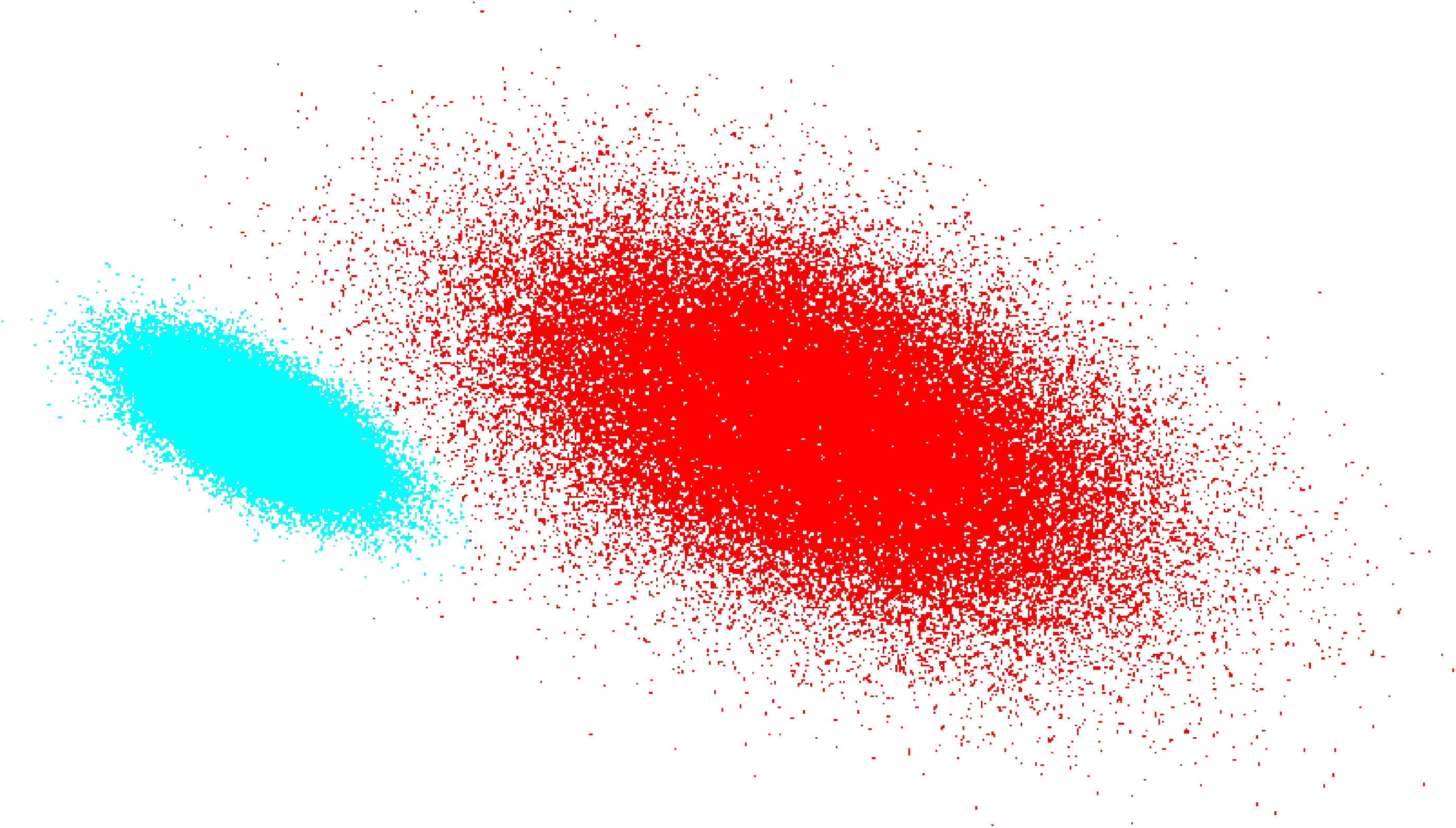}
  \end{subfigure}%
    \begin{subfigure}[b]{.25\linewidth}
    \centering
    \includegraphics[width=.99\textwidth]{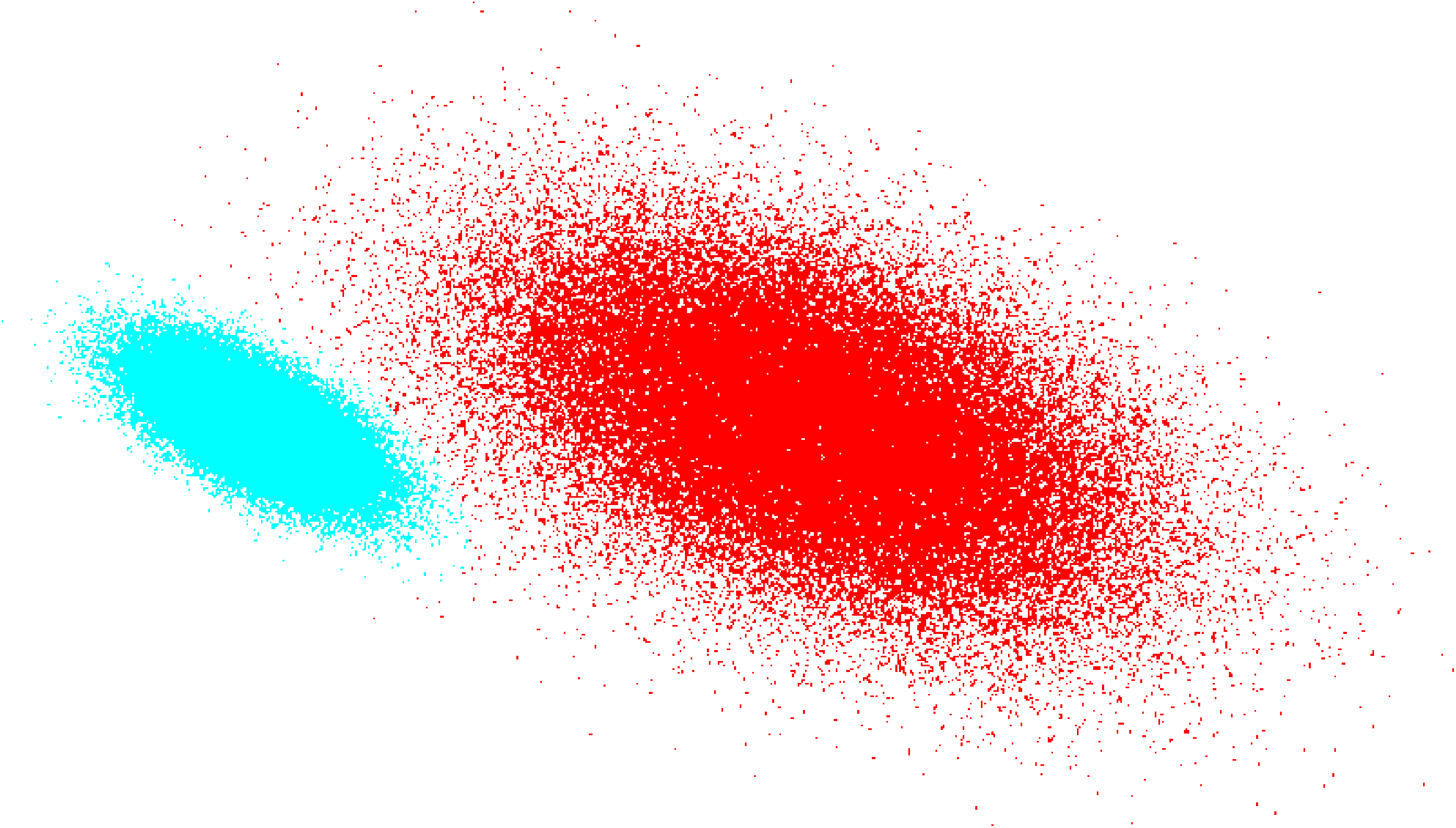}
  \end{subfigure}\\
  \begin{subfigure}[b]{.25\linewidth}
    \centering
    \includegraphics[width=.99\textwidth]{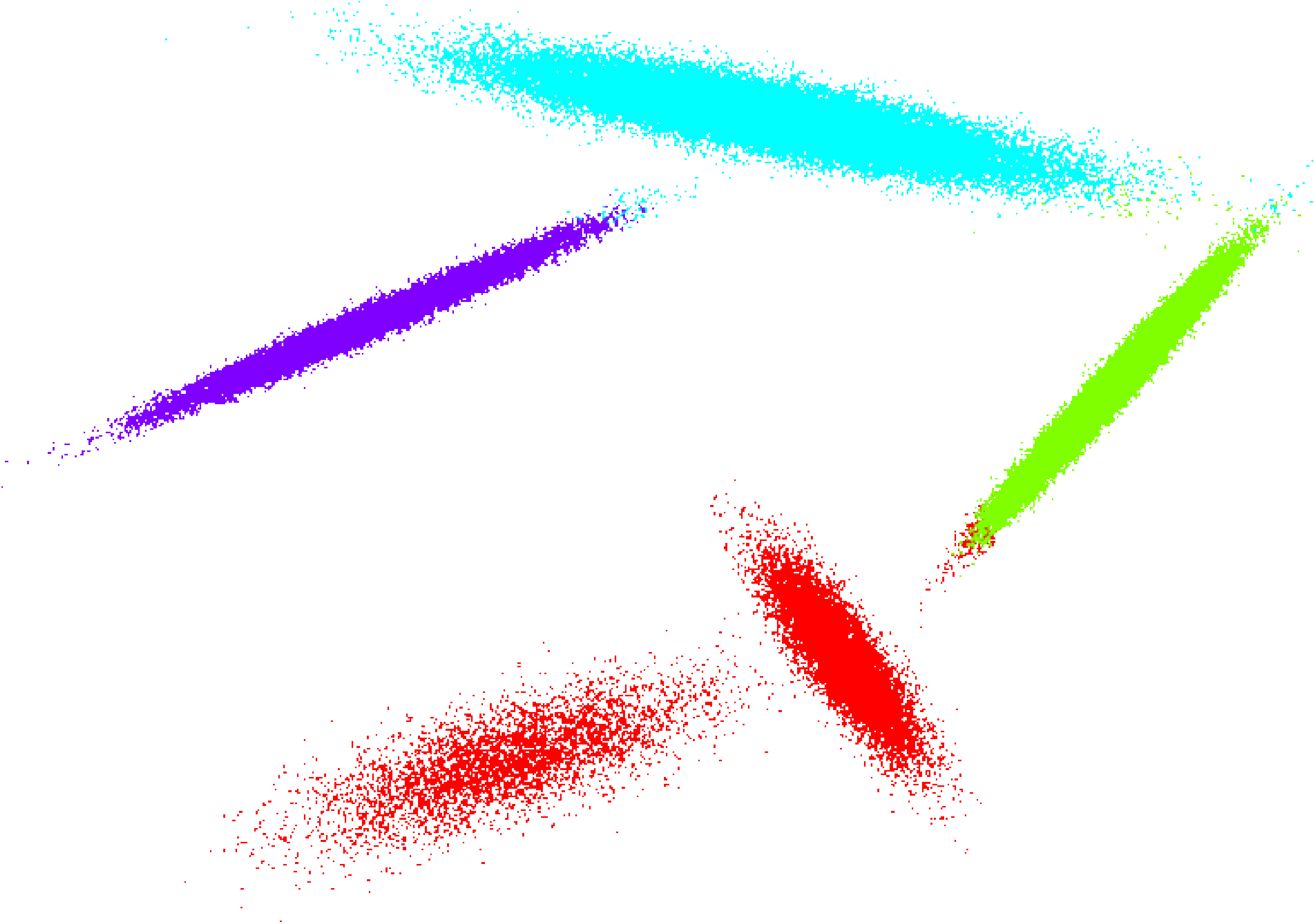}
  \end{subfigure}%
  \begin{subfigure}[b]{.25\linewidth}
    \centering
    \includegraphics[width=.99\textwidth]{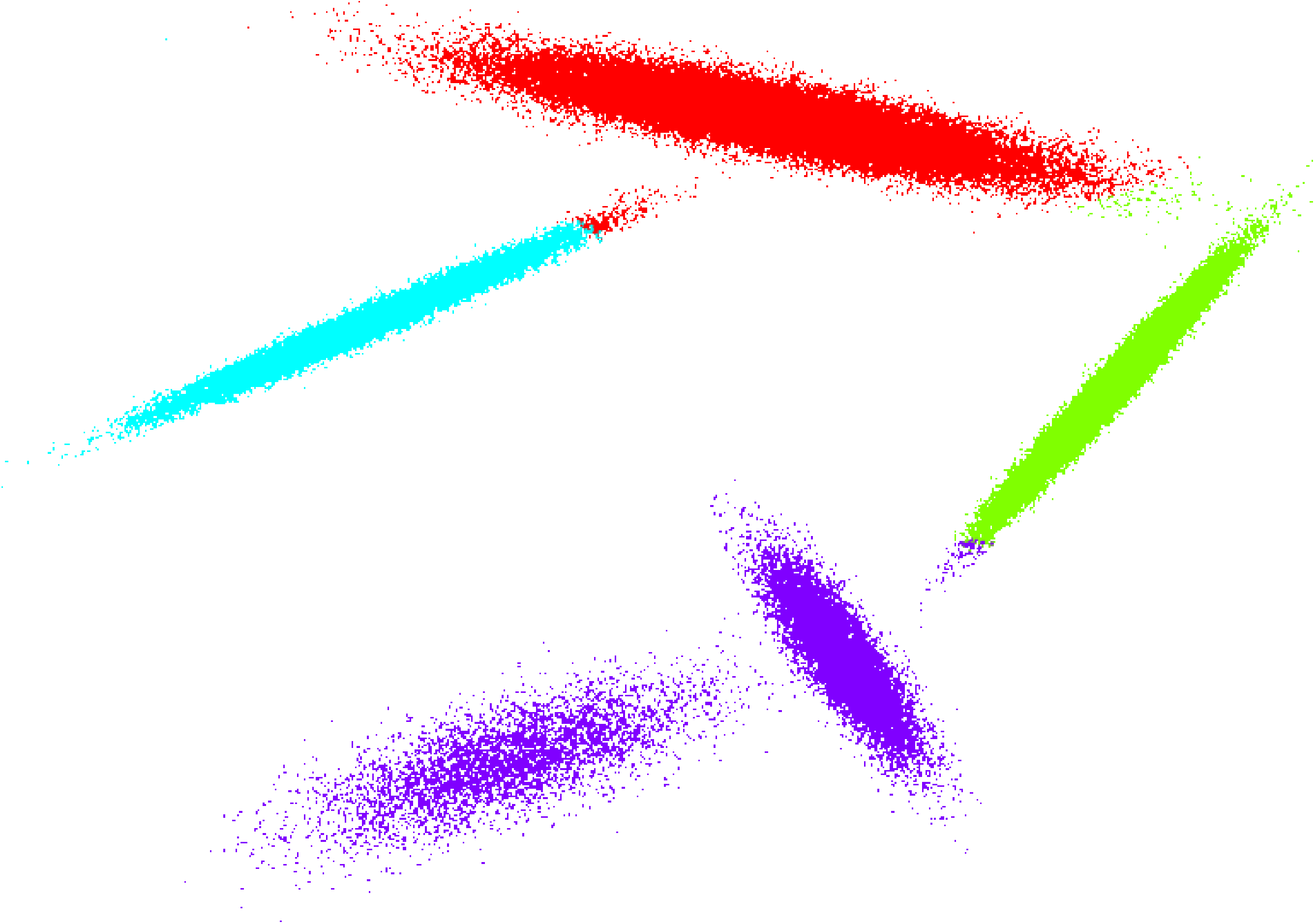}
  \end{subfigure}%
  \begin{subfigure}[b]{.25\linewidth}
    \centering
    \includegraphics[width=.99\textwidth]{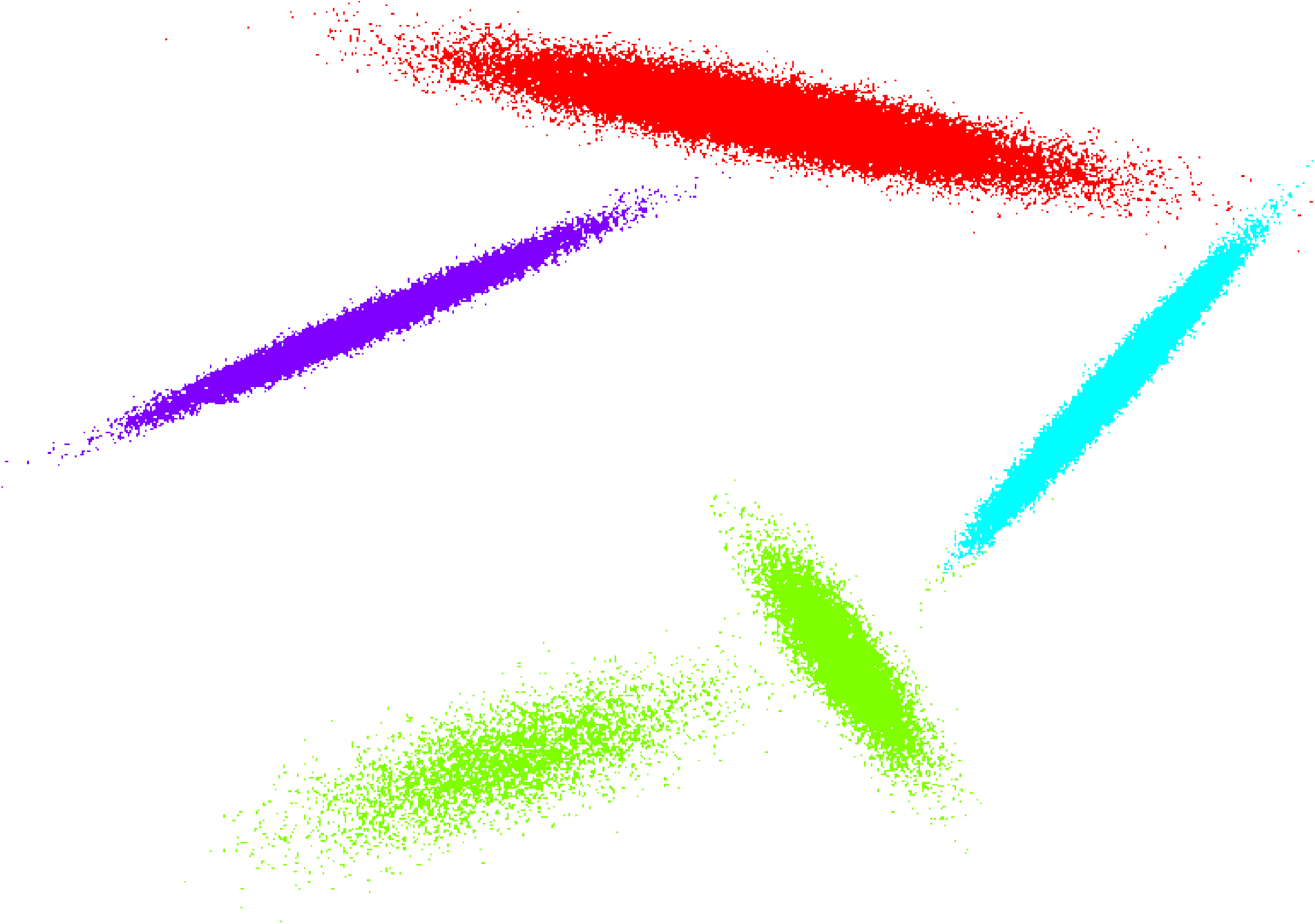}
  \end{subfigure}%
    \begin{subfigure}[b]{.25\linewidth}
    \centering
    \includegraphics[width=.99\textwidth]{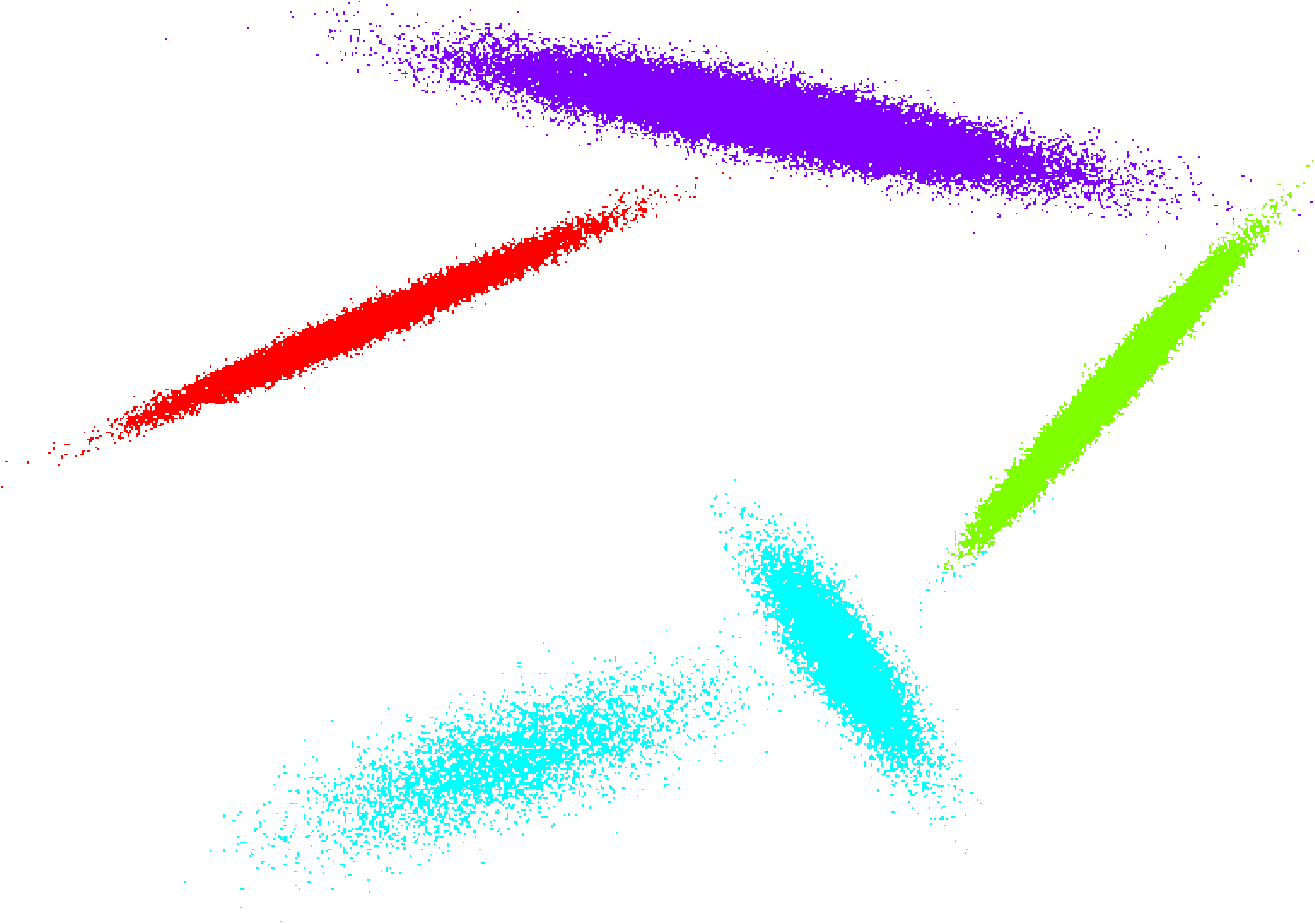}
  \end{subfigure}\\
    \begin{subfigure}[b]{.25\linewidth}
    \centering
    \includegraphics[width=.99\textwidth]{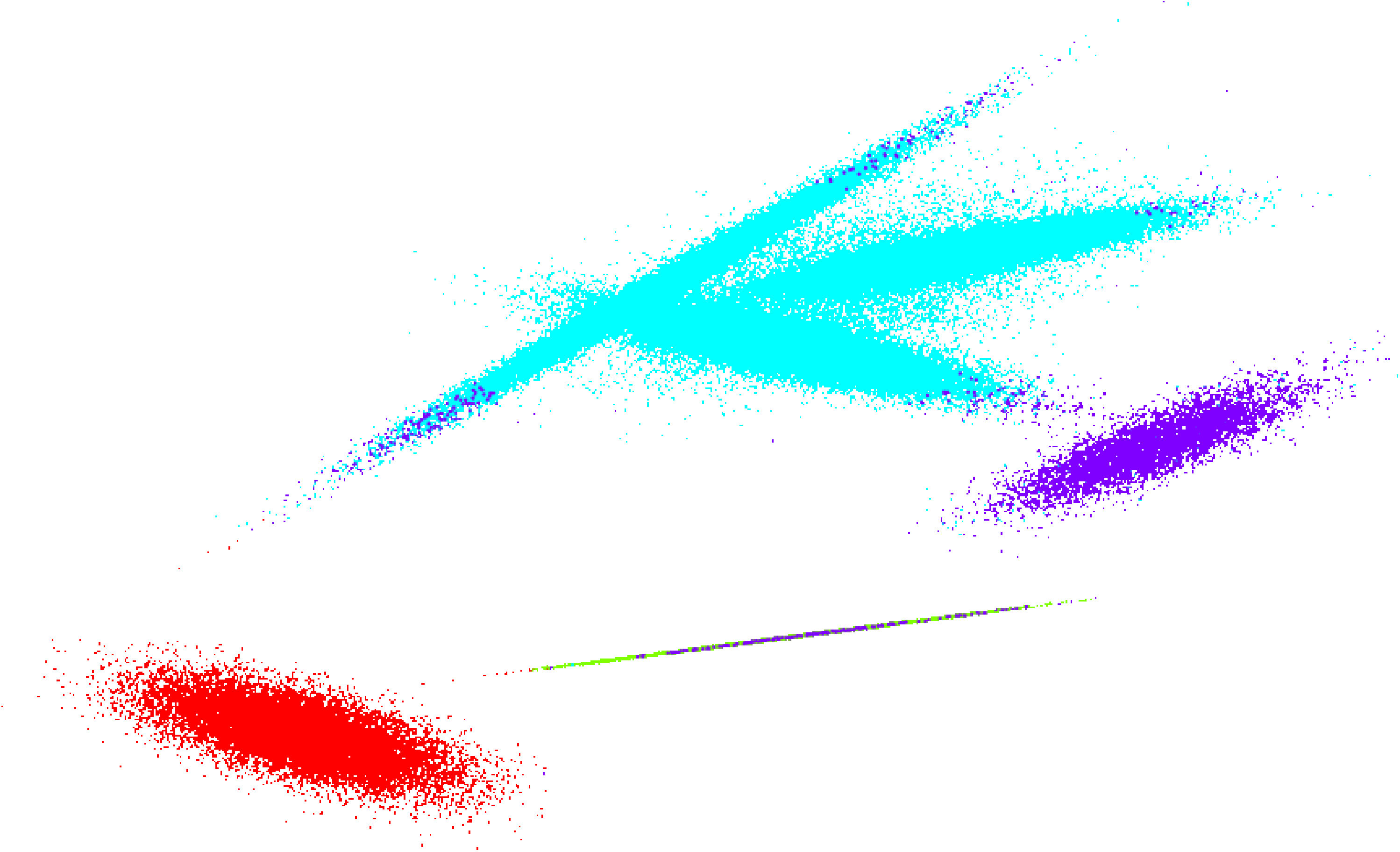}
    \caption{OPWG}
  \end{subfigure}%
  \begin{subfigure}[b]{.25\linewidth}
    \centering
    \includegraphics[width=.99\textwidth]{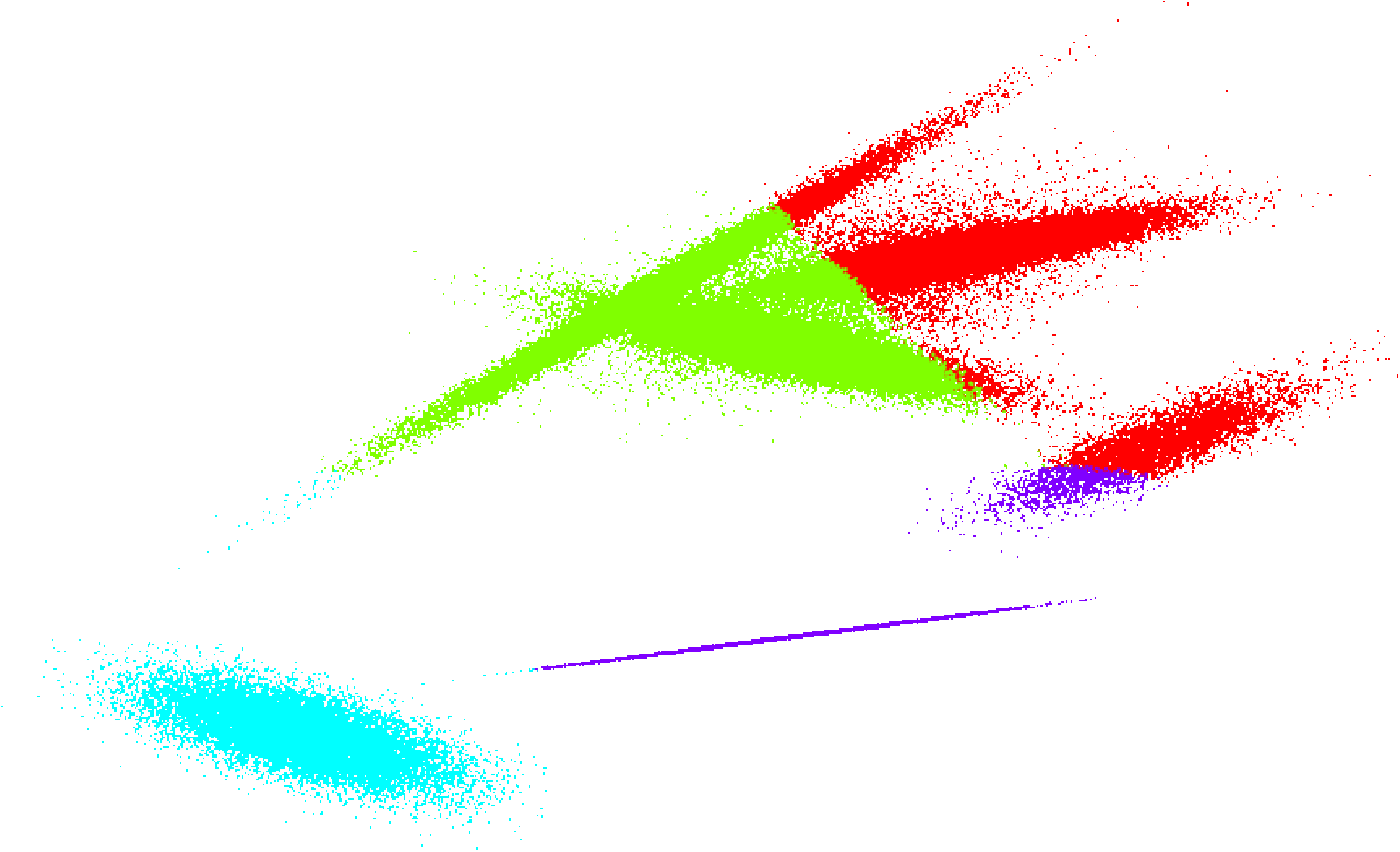}
    \caption{OFCM}
  \end{subfigure}%
  \begin{subfigure}[b]{.25\linewidth}
    \centering
    \includegraphics[width=.99\textwidth]{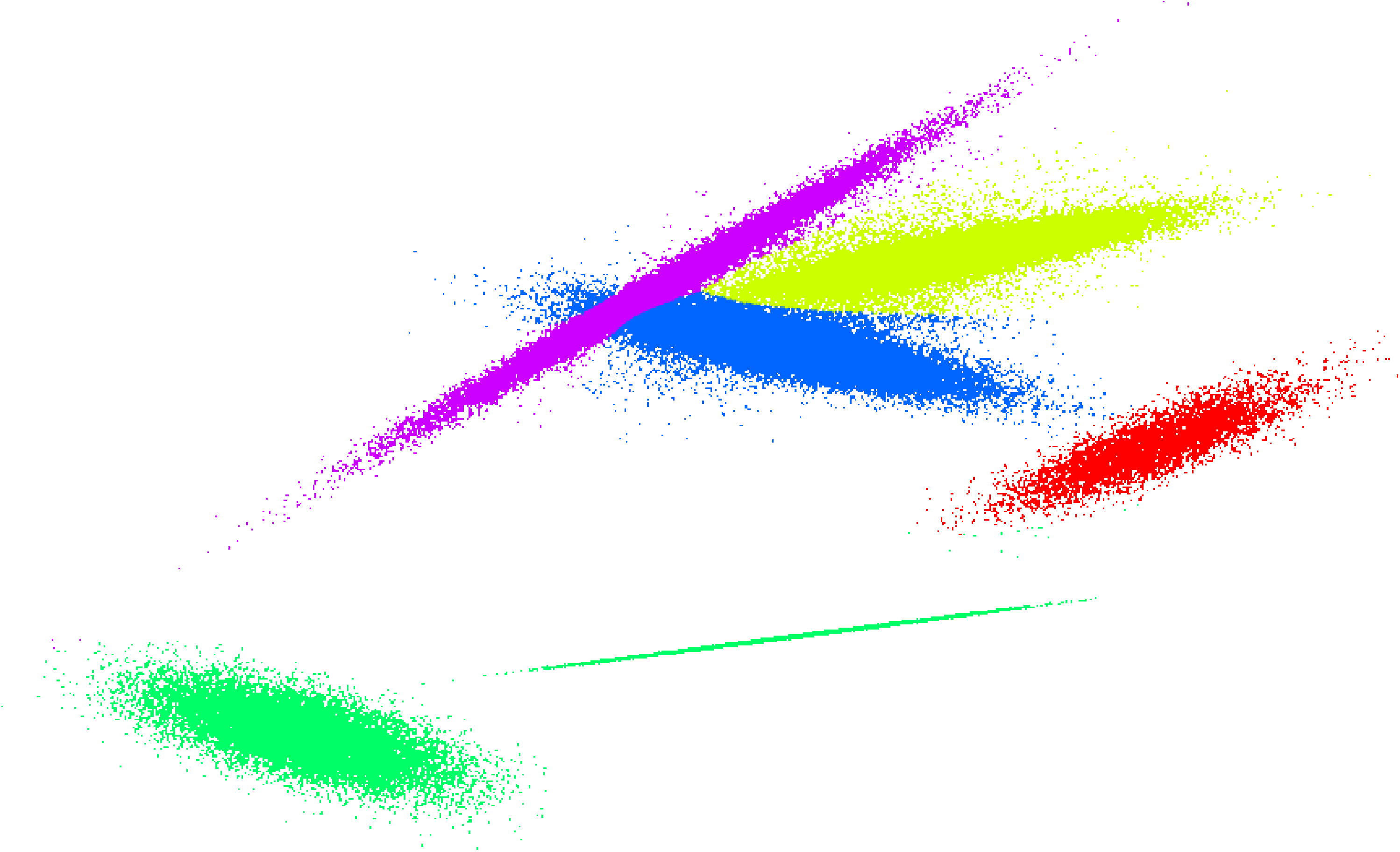}
    \caption{PGMM}
  \end{subfigure}%
    \begin{subfigure}[b]{.25\linewidth}
    \centering
    \includegraphics[width=.99\textwidth]{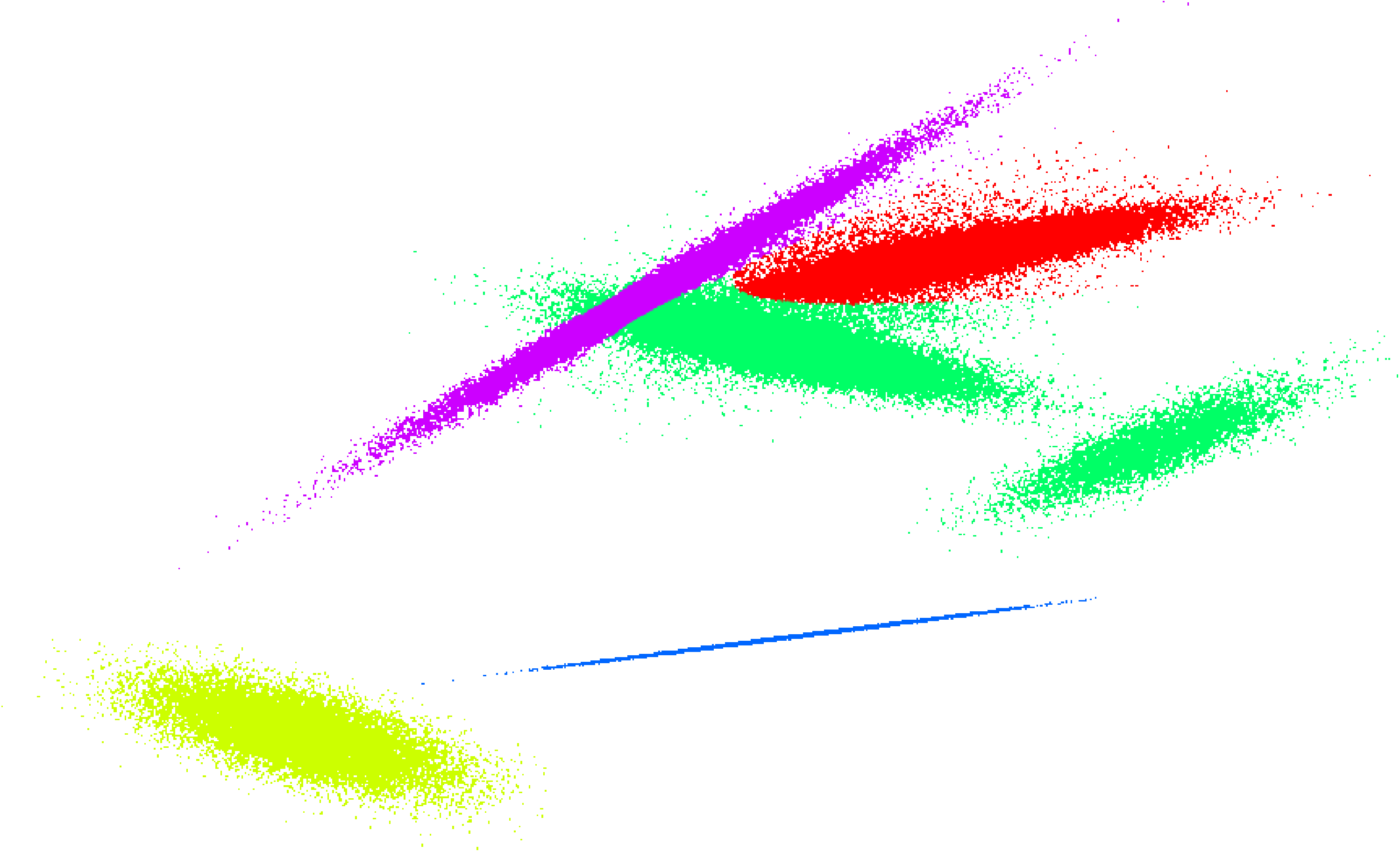}
    \caption{GMMM}
  \end{subfigure}\\
\caption
{Mode A - results for GMM datasets. Each algorithm was evaluated on three different datasets. For OPWG and PGMM, the initial number of clusters was 25. OFCM and GMM were initialized with the number of clusters found by OPWG and PGMM, respectively.}\label{fig:gmm_random}
\end{figure}

\begin{figure}[htb]
\centering
  \begin{subfigure}[b]{.25\linewidth}
    \centering
    \includegraphics[width=.99\textwidth]{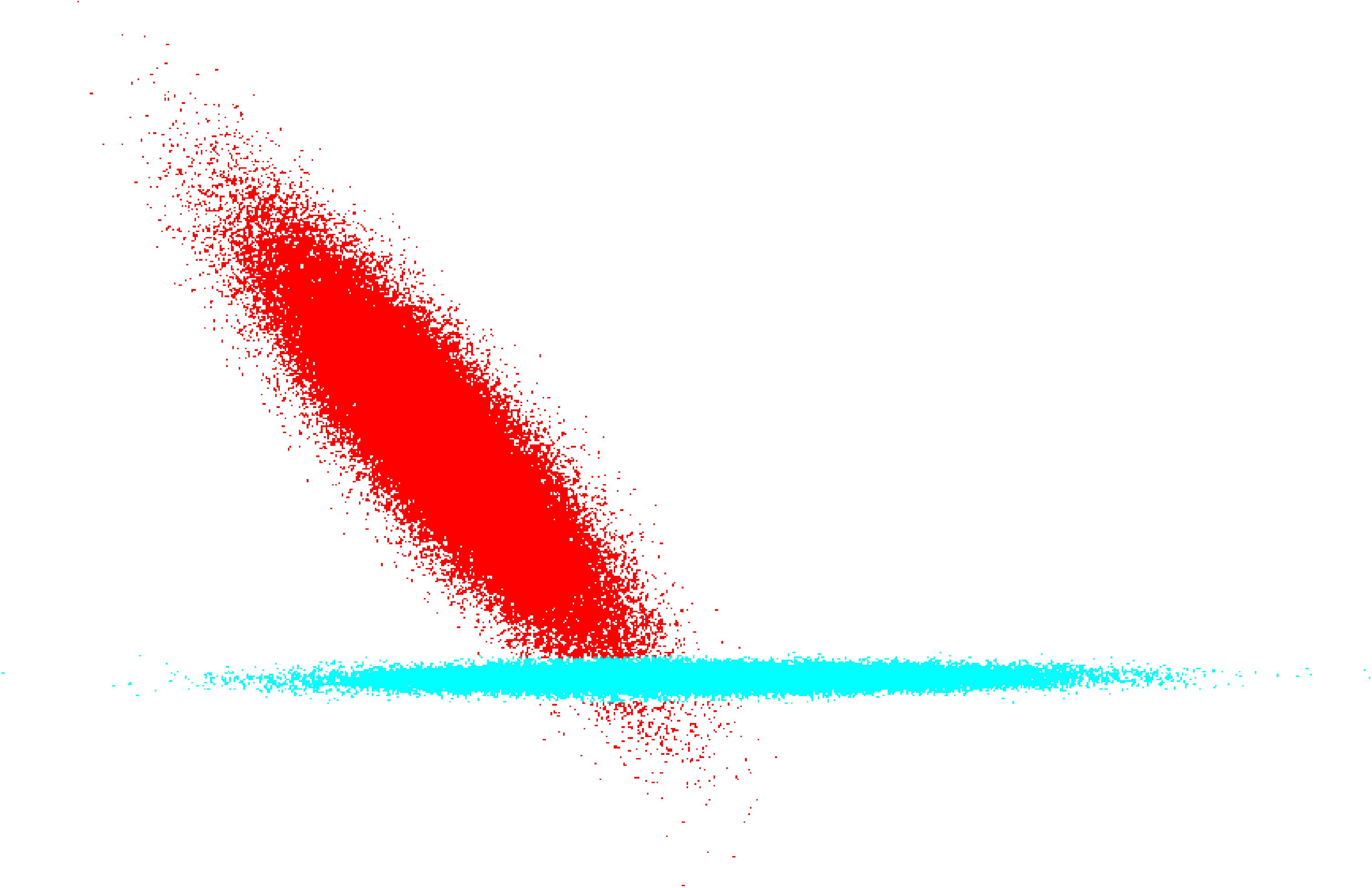}
  \end{subfigure}%
  \begin{subfigure}[b]{.25\linewidth}
    \centering
    \includegraphics[width=.99\textwidth]{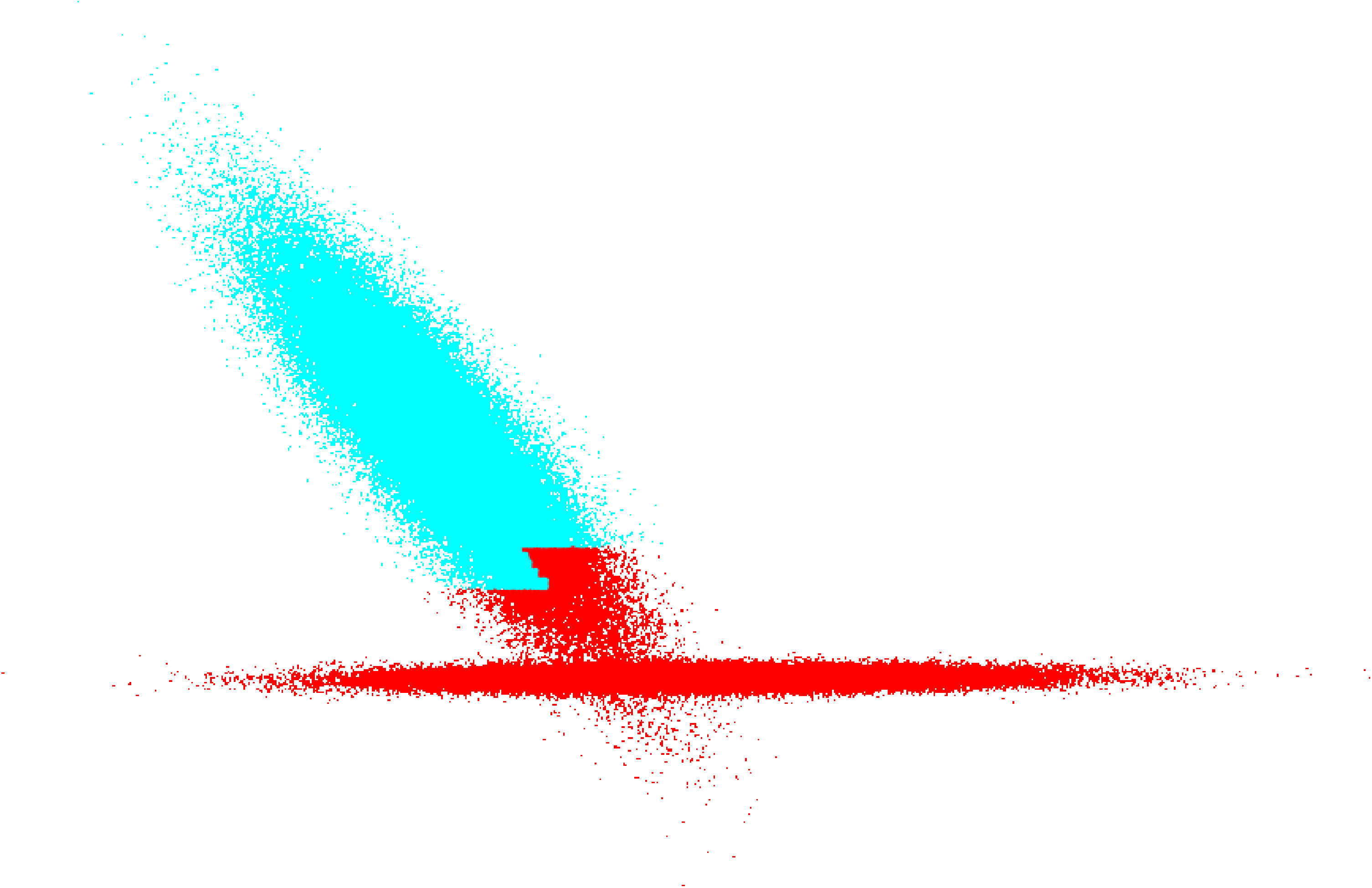}
  \end{subfigure}%
  \begin{subfigure}[b]{.25\linewidth}
    \centering
    \includegraphics[width=.99\textwidth]{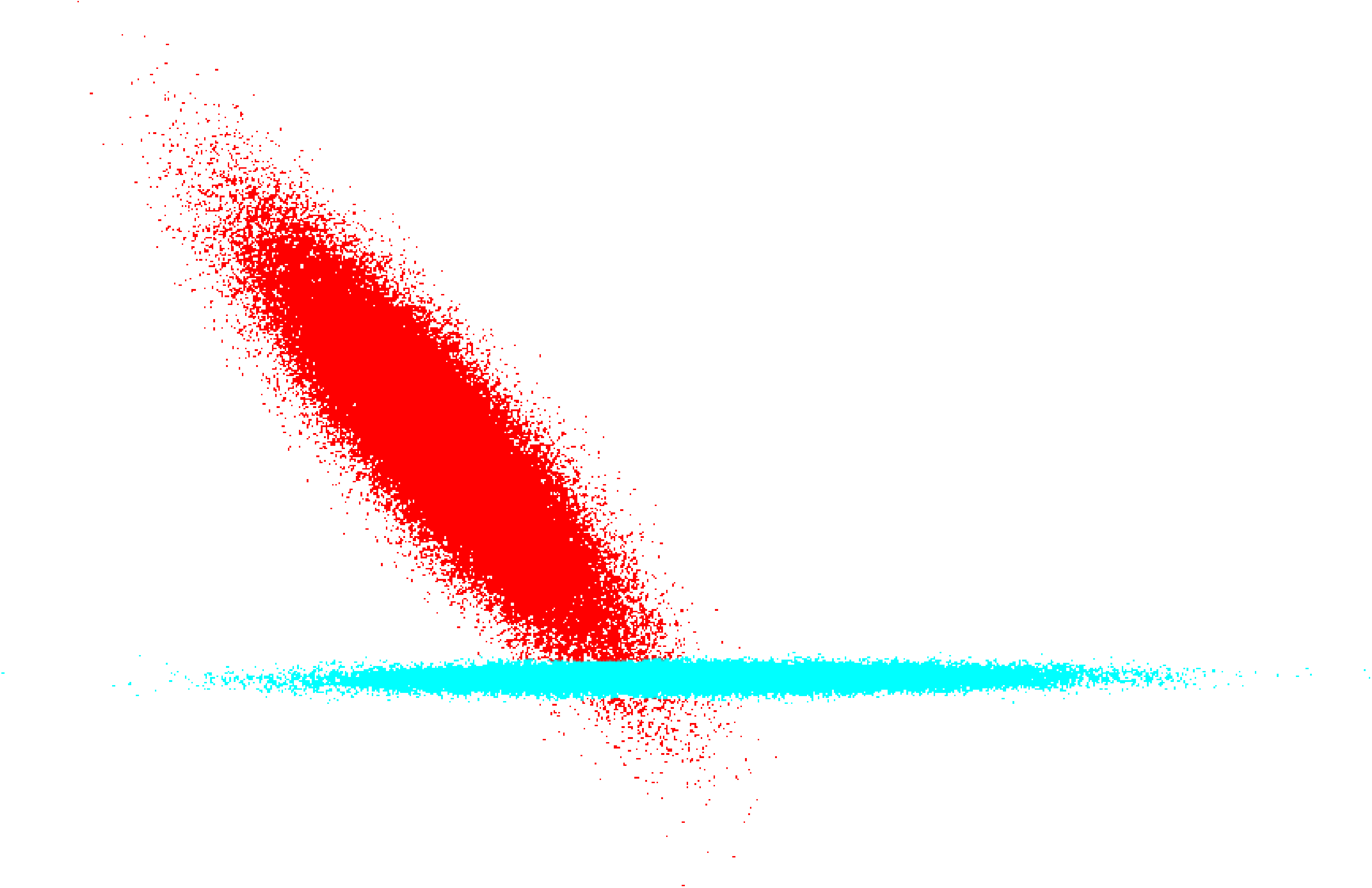}
  \end{subfigure}%
    \begin{subfigure}[b]{.25\linewidth}
    \centering
    \includegraphics[width=.99\textwidth]{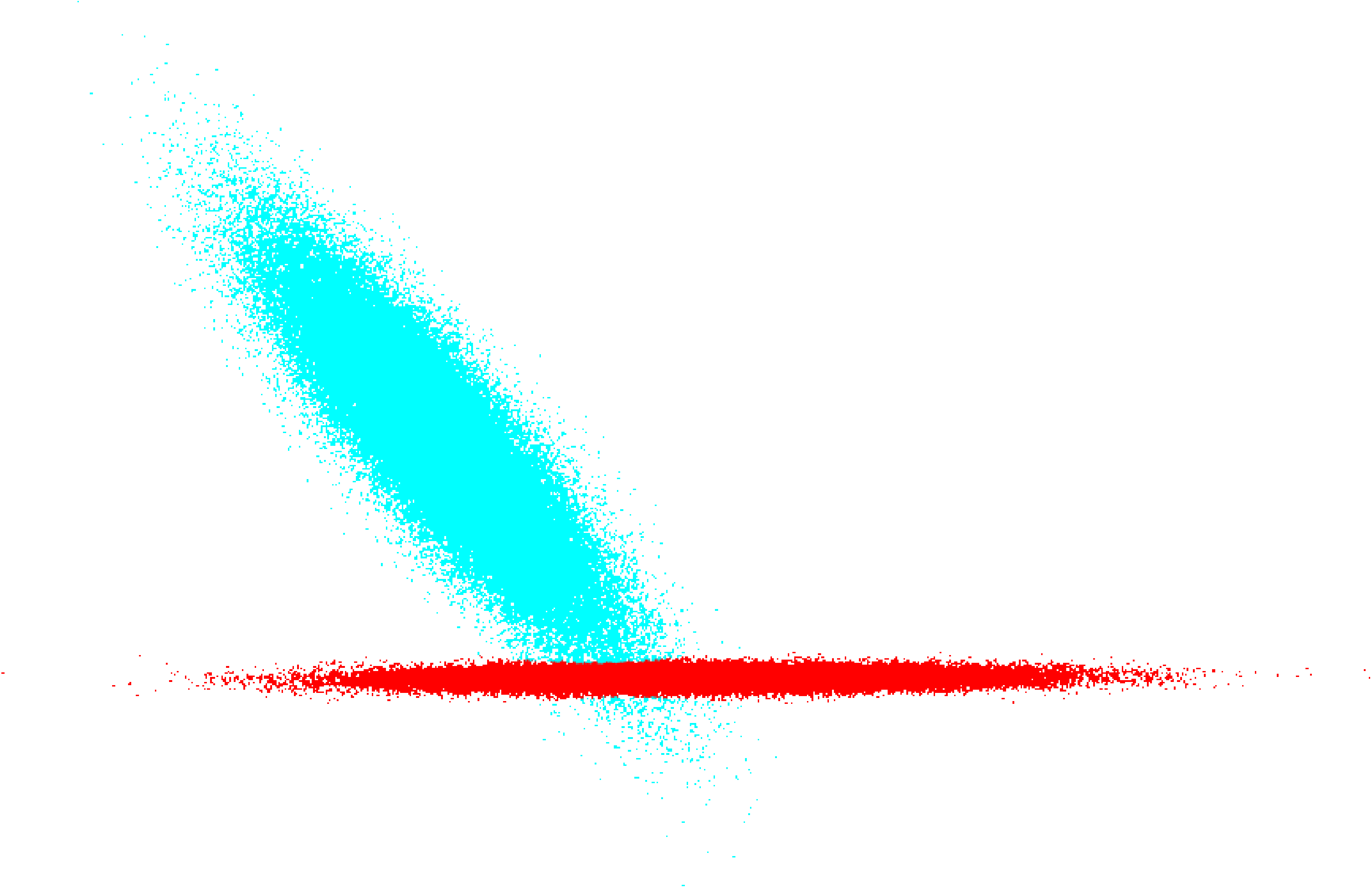}
  \end{subfigure}\\
  \begin{subfigure}[b]{.25\linewidth}
    \centering
    \includegraphics[width=.99\textwidth]{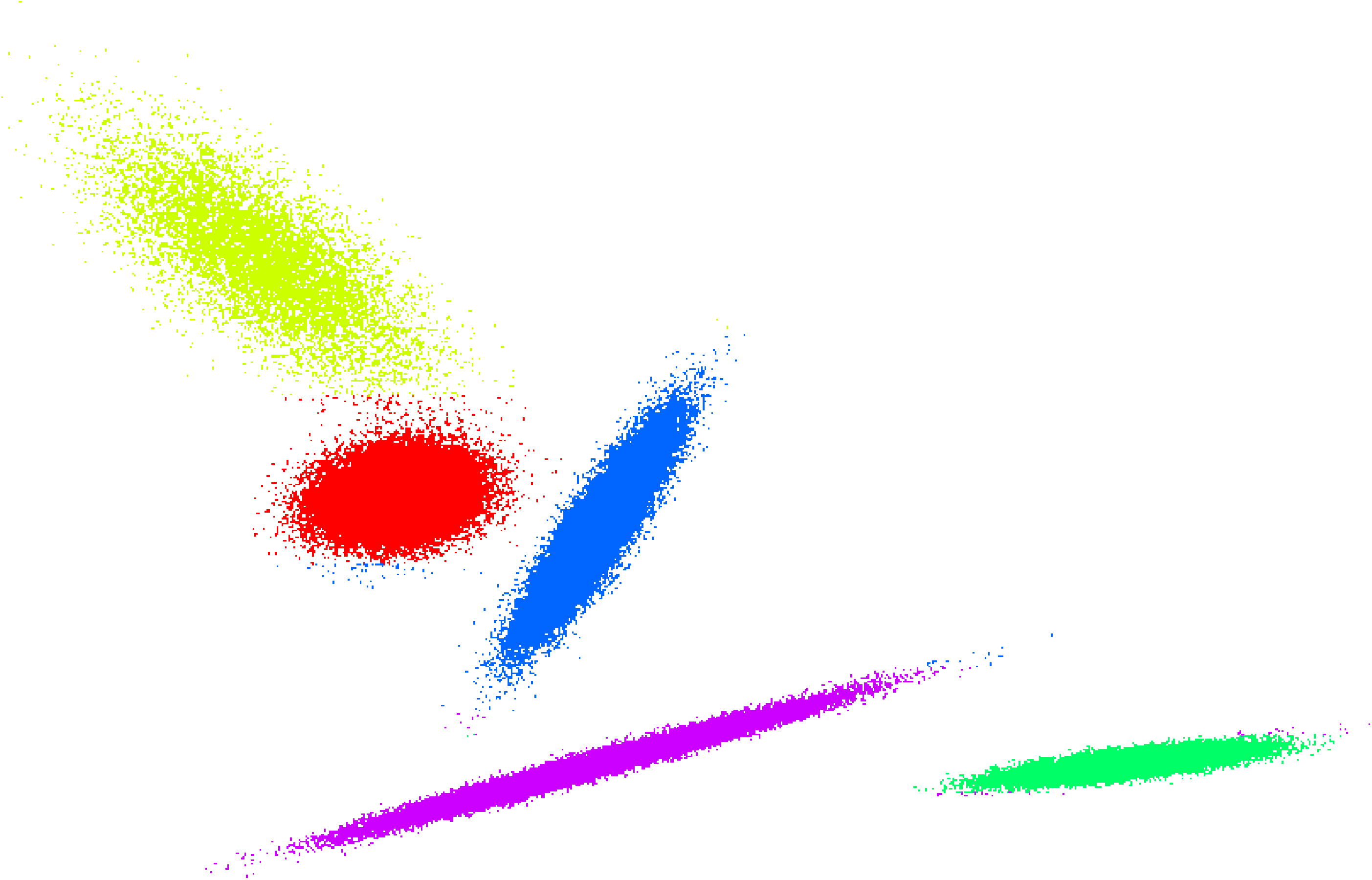}
  \end{subfigure}%
  \begin{subfigure}[b]{.25\linewidth}
    \centering
    \includegraphics[width=.99\textwidth]{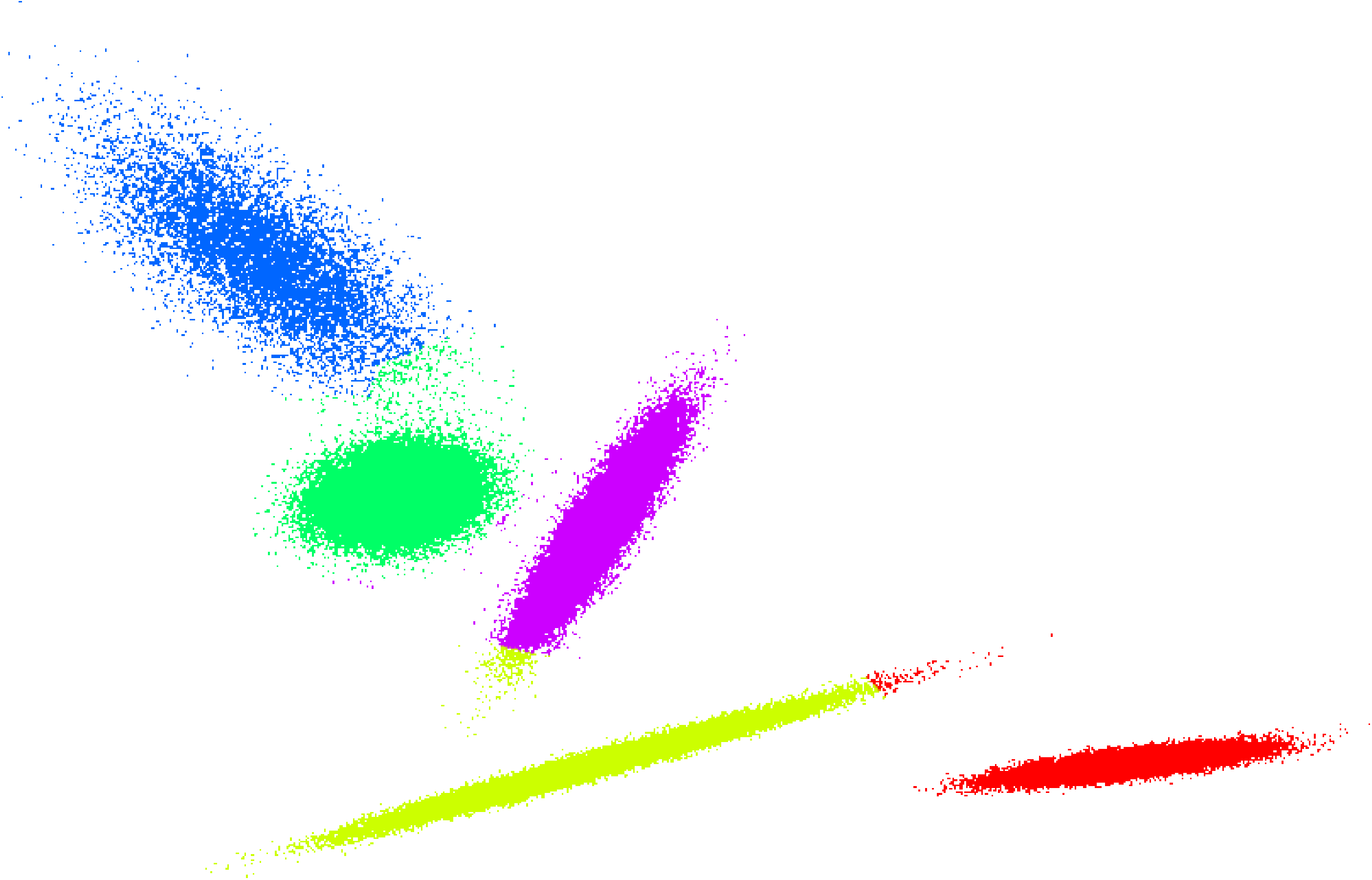}
  \end{subfigure}%
  \begin{subfigure}[b]{.25\linewidth}
    \centering
    \includegraphics[width=.99\textwidth]{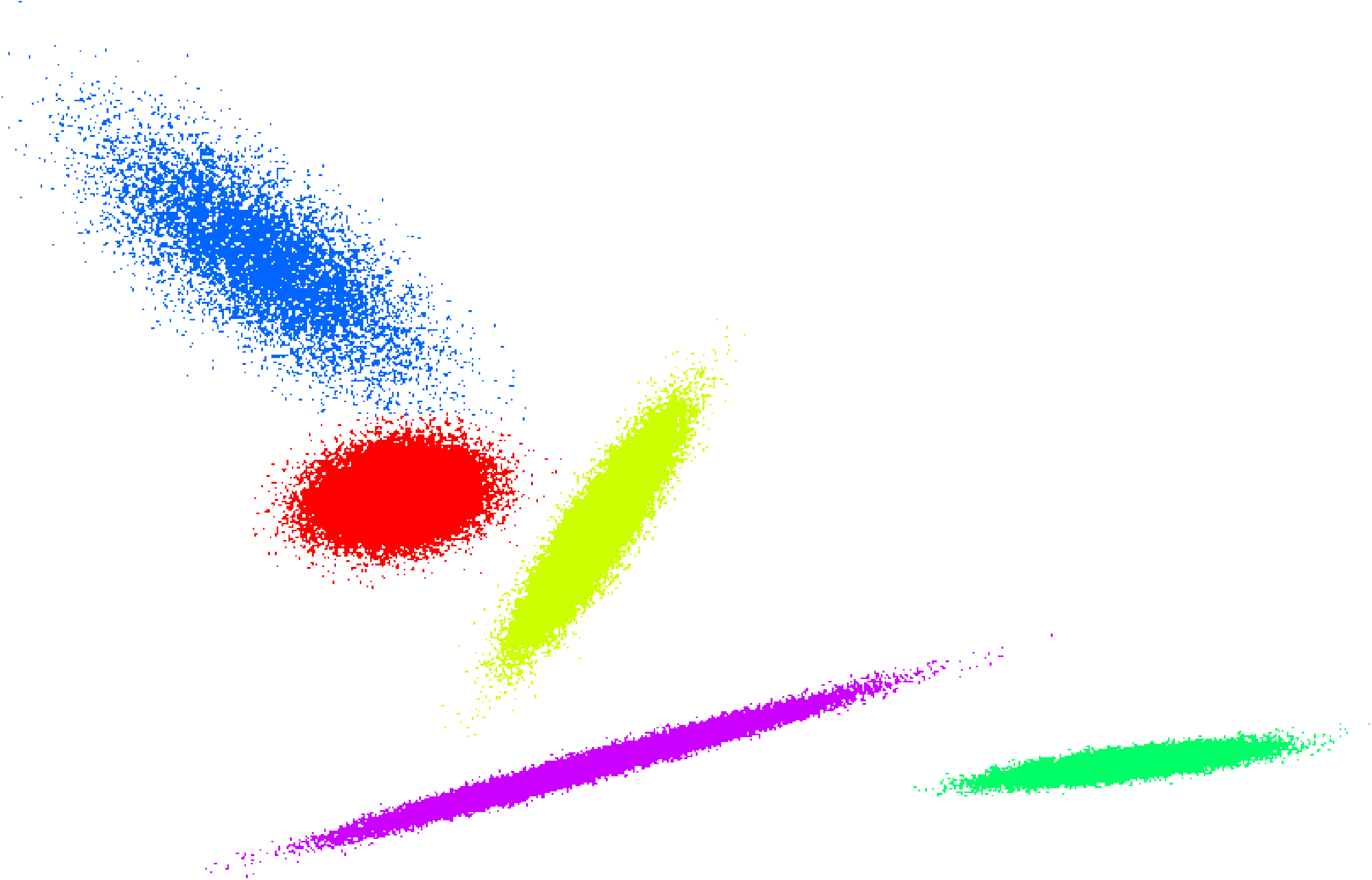}
  \end{subfigure}%
    \begin{subfigure}[b]{.25\linewidth}
    \centering
    \includegraphics[width=.99\textwidth]{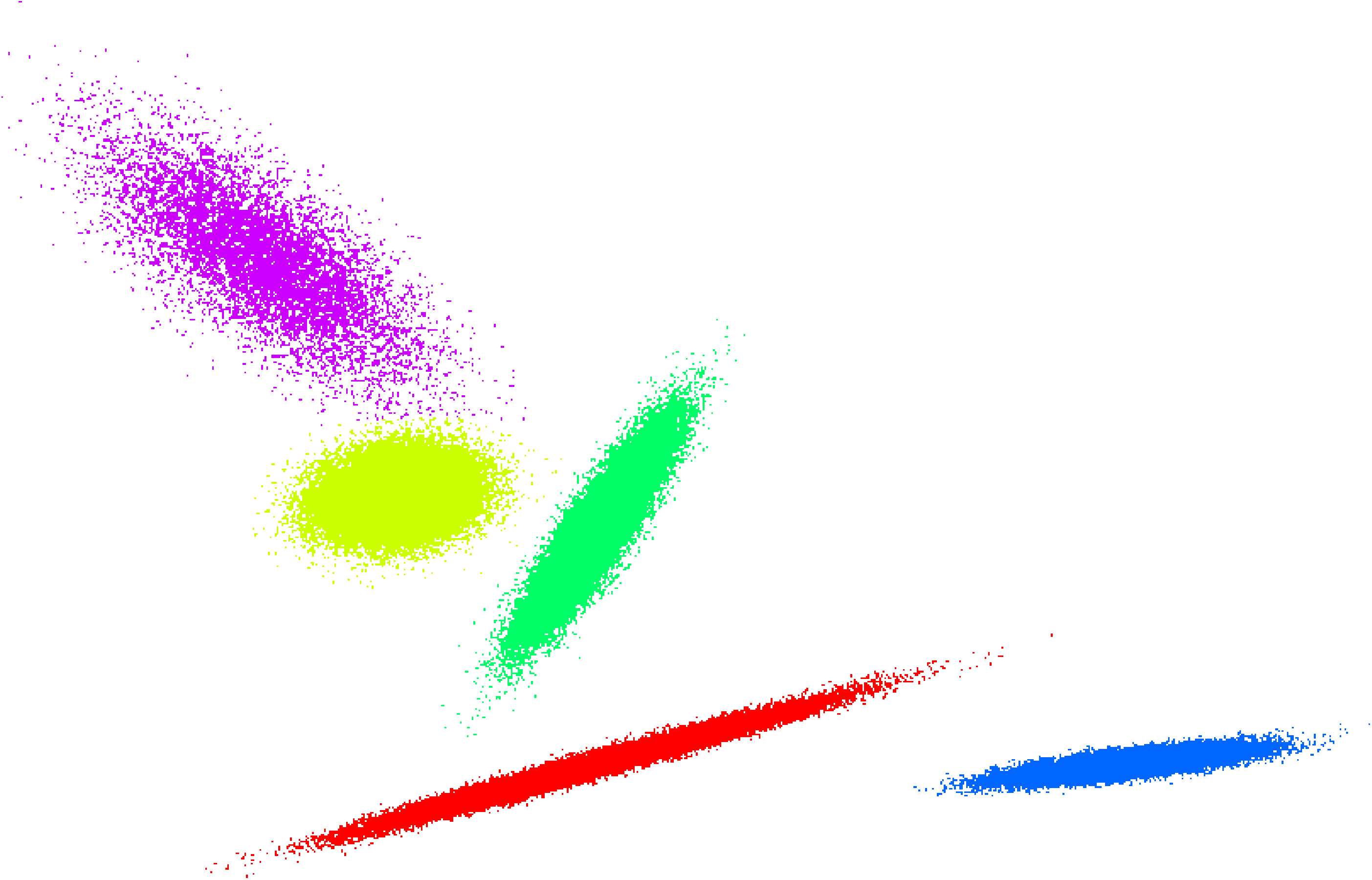}
  \end{subfigure}\\
    \begin{subfigure}[b]{.25\linewidth}
    \centering
    \includegraphics[width=.99\textwidth]{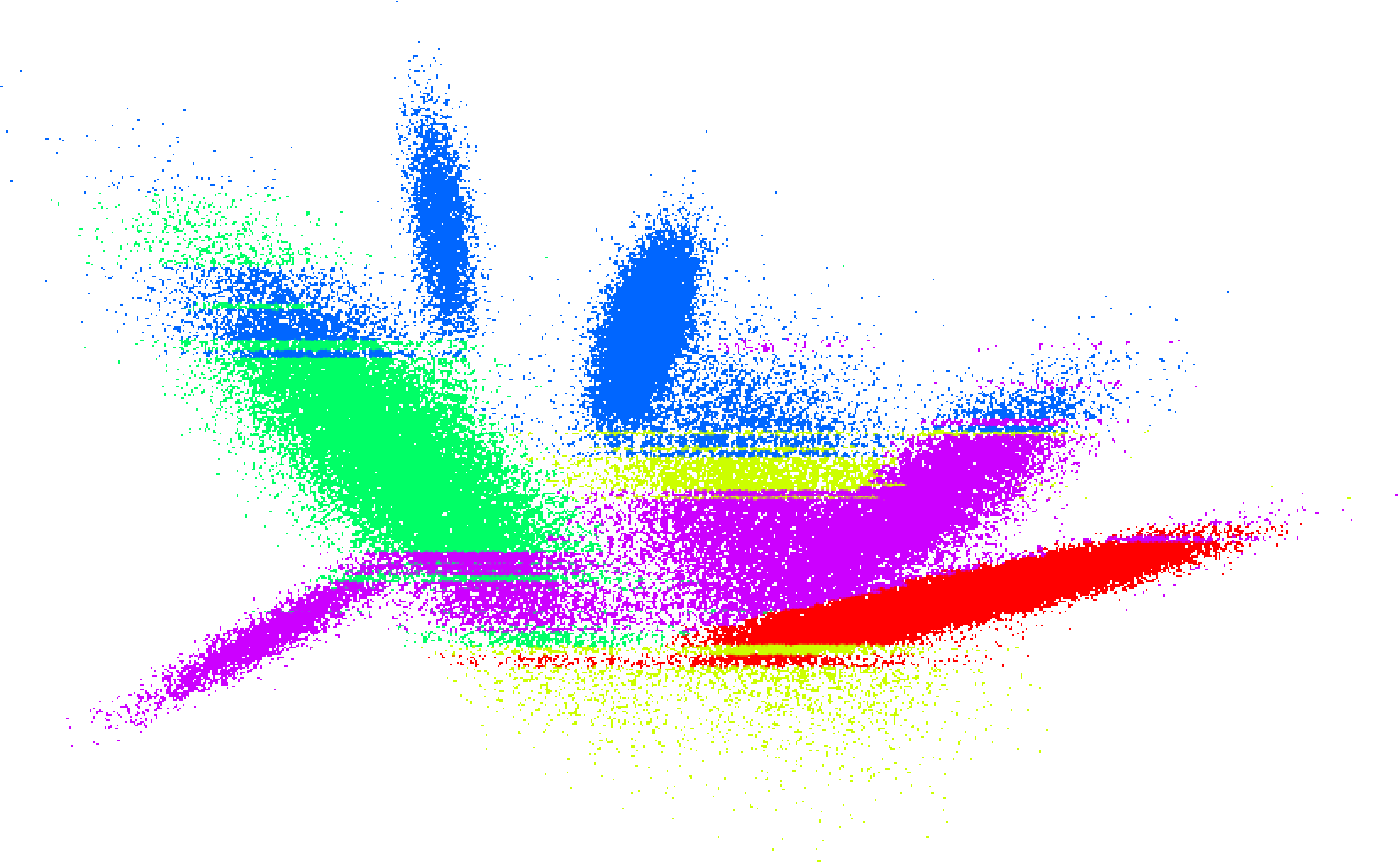}
    \caption{OPWG}
  \end{subfigure}%
  \begin{subfigure}[b]{.25\linewidth}
    \centering
    \includegraphics[width=.99\textwidth]{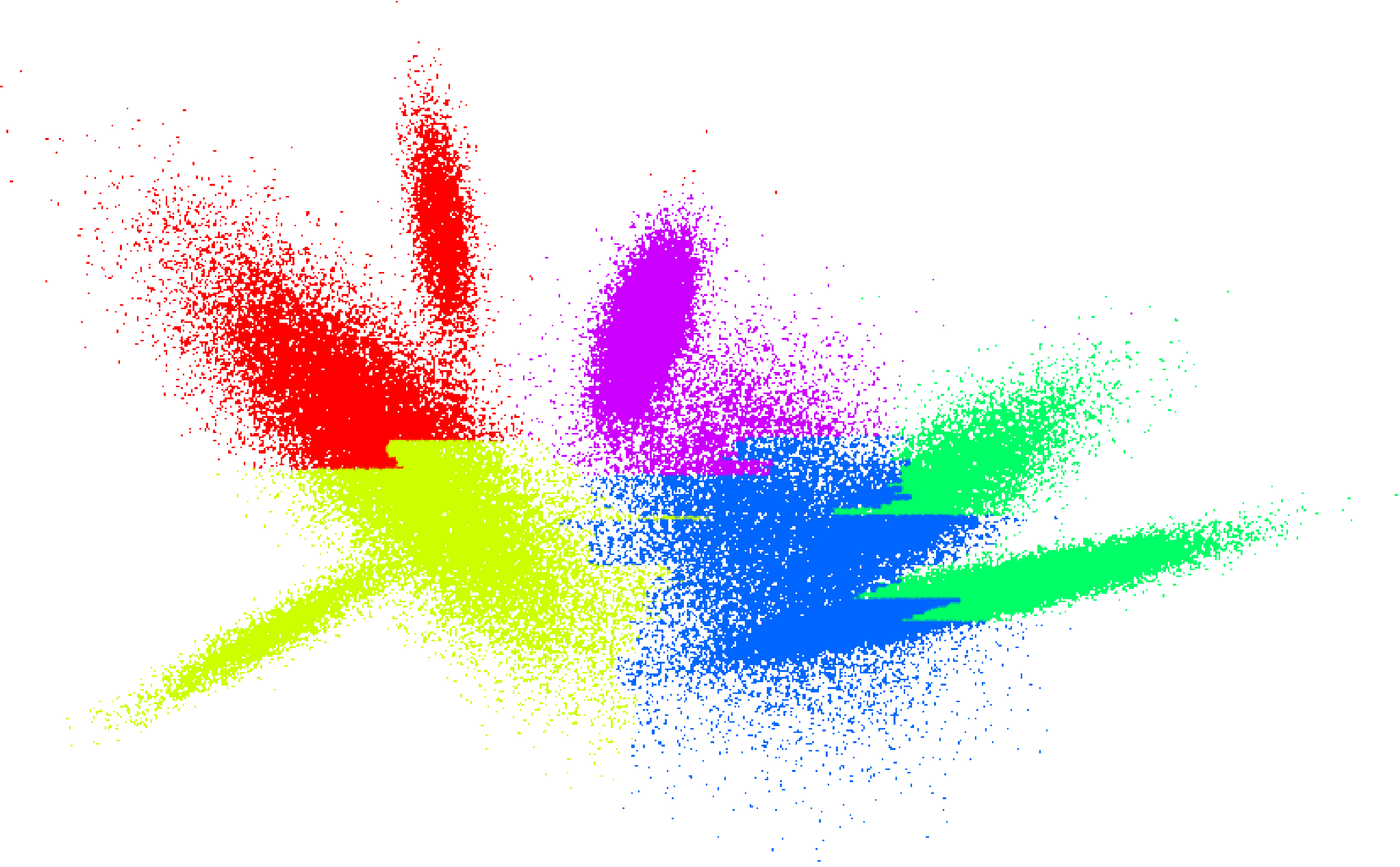}
    \caption{OFCM}
  \end{subfigure}%
  \begin{subfigure}[b]{.25\linewidth}
    \centering
    \includegraphics[width=.99\textwidth]{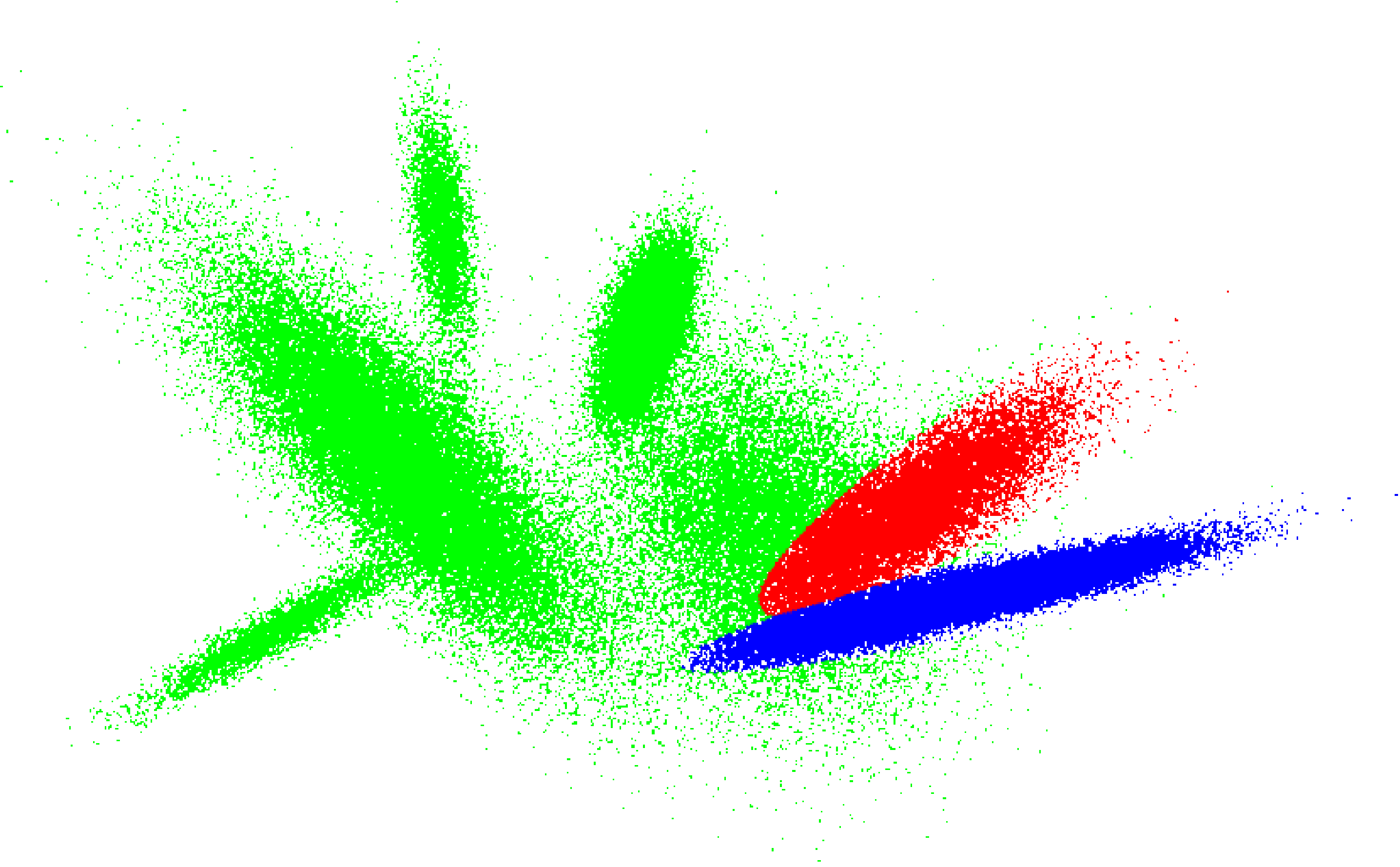}
    \caption{PGMM}
  \end{subfigure}%
    \begin{subfigure}[b]{.25\linewidth}
    \centering
    \includegraphics[width=.99\textwidth]{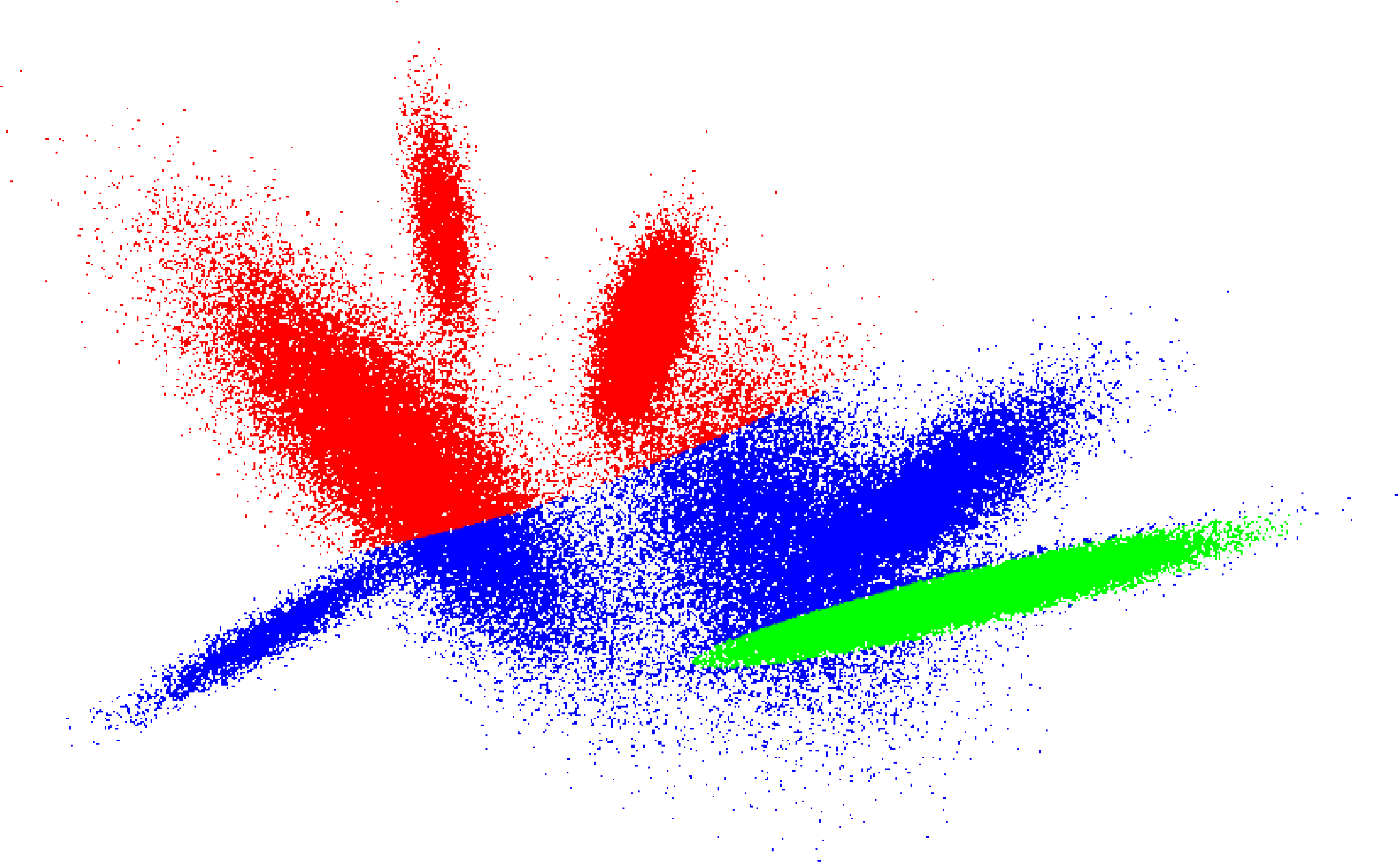}
    \caption{GMM}
  \end{subfigure}\\
\caption
{Mode B - results for GMM datasets. Each algorithm was run on three different datasets. For OPWG and PWG, the initial number of clusters was 25. OFCM and GMM were initialized with the number of clusters found by OPWG and PGMM, respectively.}\label{fig:gmm_sorted}
\end{figure}

The results for mode A are reported in Table \ref{table:gmm_random} and illustrated in Figure \ref{fig:gmm_random}. The results for mode B are reported in Table \ref{table:gmm_sorted} and illustrated in Figure \ref{fig:gmm_sorted}. For mode A, OPWG performs well comparing to the other algorithms. Its estimation of the number of clusters, $K= 2.32, 3.92, 4.2 $, is close to the estimation of PGMM that estimated $K=2.02, 4.00, 4.44$. Moreover, although it was initialized with a bigger number of clusters, $K_{max}=25$, it performs equal or better compared to the OFCM.

Surprisingly, for mode B, the average results of  OPWG are better than for mode A. In addition, in some cases, it even performs better than the algorithms that run of the full stream. A possible reason for this is that when running in online mode, the adaptive tendency of the algorithm makes it more robust to outliers compared to running on the full stream. Thus, the algorithm achieves more accurate clustering results. For this mode, the average number of clusters that OPWG found, $K=2.48, 5.1, 6 $, is again close to the average number of clusters that PGMM found - $K=2, 4.02, 4.54$. 

\begin{table}
\begin{center}
\begin{tabular}{lllllll}
\hline
            & \multicolumn{2}{c}{K=2}       & \multicolumn{2}{c}{K=5}       & \multicolumn{2}{c}{K=7}         \\
            & F1            & NMI           & F1            & NMI           & F1            & NMI              \\ \cline{2-7} 
OPWG  & 0.97          & 0.90          & 0.80          & 0.76          & 0.69          & 0.69           \\
OFCM & 0.92          & 0.84          & 0.81          & 0.78          & 0.73          & 0.75           \\
PGMM         & \textbf{0.99} & \textbf{0.95} & \textbf{0.90} & \textbf{0.88} & \textbf{0.78} & \textbf{0.80}  \\
GMM         & \textbf{0.99} & \textbf{0.95} & 0.88          & 0.87          & 0.77          & 0.79          \\ \hline
\end{tabular}
\caption {Mode A - numerical results for GMM datasets. Clustering quality measures (F1 and NMI) for each of the four algorithms where each batch contains random samples from the whole dataset.}\label{table:gmm_random}
\end{center}
\end{table}

\begin{table}[h!]
\begin{center}
\begin{tabular}{lllllllllll}
\hline
            & \multicolumn{2}{c}{K=2}       & \multicolumn{2}{c}{K=5}       & \multicolumn{2}{c}{K=7}          \\
            & F1            & NMI           & F1            & NMI           & F1            & NMI               \\ \cline{2-7} 
OPWG  & 0.94          & 0.85          & \textbf{0.93} & \textbf{0.88} & \textbf{0.86} & \textbf{0.82}    \\
OFCM & 0.88          & 0.79          & 0.86          & 0.83          & 0.79          & 0.79            \\
PGMM         & 0.97          & 0.91          & 0.89          & \textbf{0.88} & 0.79          & \textbf{0.82}   \\
GMM         & \textbf{0.99} & \textbf{0.93} & 0.87          & 0.87          & 0.79          & \textbf{0.82} \\ \hline
\end{tabular}
\caption {Mode B numerical results for GMM datasets. Clustering quality measures (F1 and NMI) for each of the four algorithms where not all clusters are represented in each batch.}\label{table:gmm_sorted}
\end{center}
\end{table}

\subsection{Algorithm Evaluation - Real Datasets}

We tested  OPWG on real datasets for the task of image segmentation. 
For this purpose, each image was converted to LAB color space and has been clustered based only on its LAB color features, i.e., segmentation by color only. 
Each batch consisted of four non-overlapping rows of the image, corresponding to reading four rows from a CCD image sensor. 
Minimal image size was $321 \times 482$. The algorithm was initialized with $K_{max}=25$ and the tuning parameter $\lambda$ was fixed to $0.03$ for the online stage and the best of $[0.006, 0.005, 0.004]$ for the post-processing stage. For both the online and the post-processing stages, we ran the algorithm with diagonal covariance matrices. This selection of diagonal covariance matrices is very important since full covariance matrices will result in large clusters or under segmentation. Our preference would be to do over segmentation and we thus selected diagonal matrices. 

Image segmentation is not a well-defined problem and different people may segment the images differently. Thus, OPWG was not evaluated numerically as in the synthetic datasets but rather compared to the popular segmentation algorithm Efficient Graph-Based Image Segmentation (EGBIS) \cite{Felzenszwalb:2004:EGI:981793.981796}.

\begin{figure*}[htb]
\centering
  \begin{subfigure}[b]{.20\linewidth}
    \centering
    \includegraphics[width=.99\textwidth]{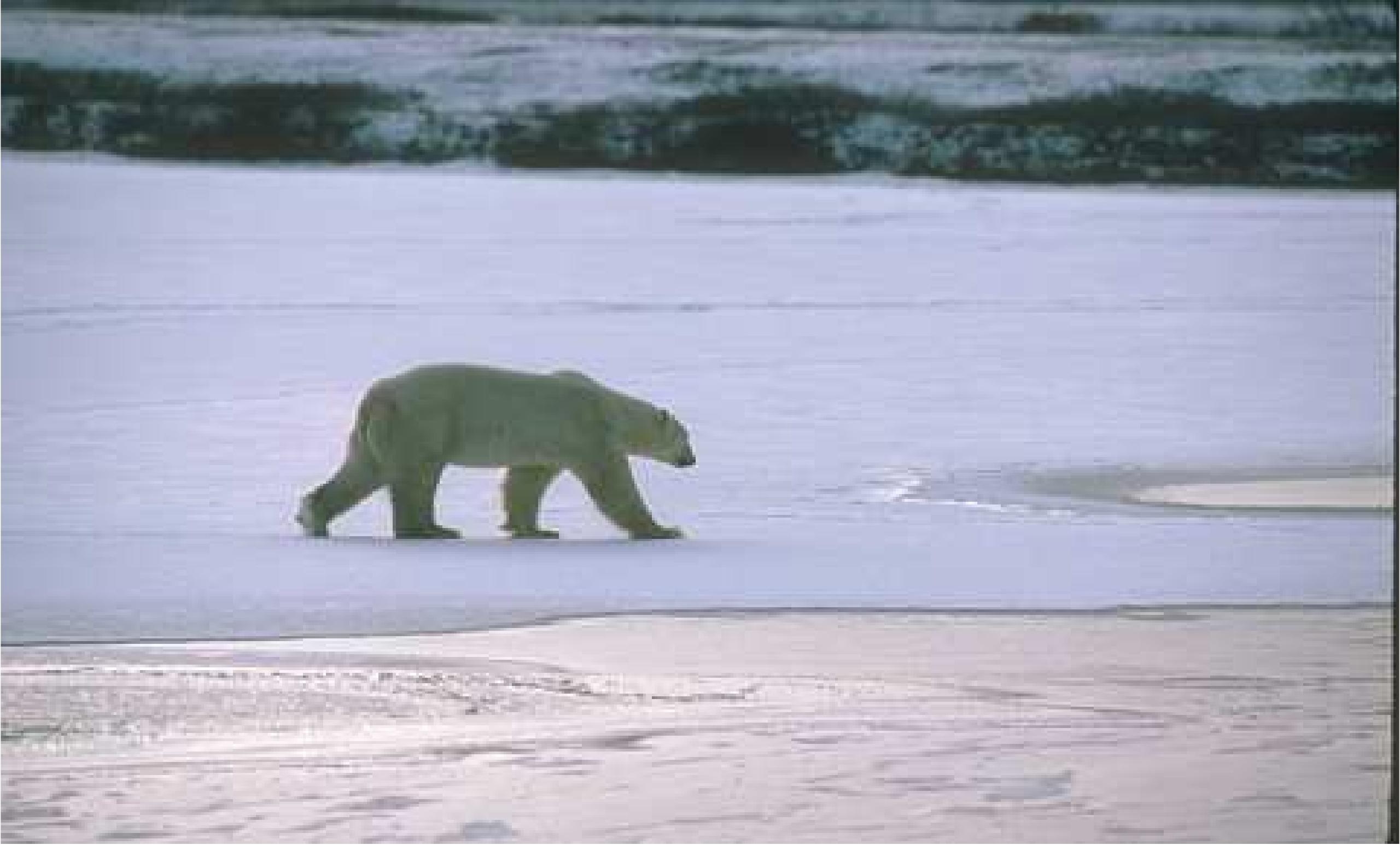}
  \end{subfigure}%
  \begin{subfigure}[b]{.20\linewidth}
    \centering
    \includegraphics[width=.99\textwidth]{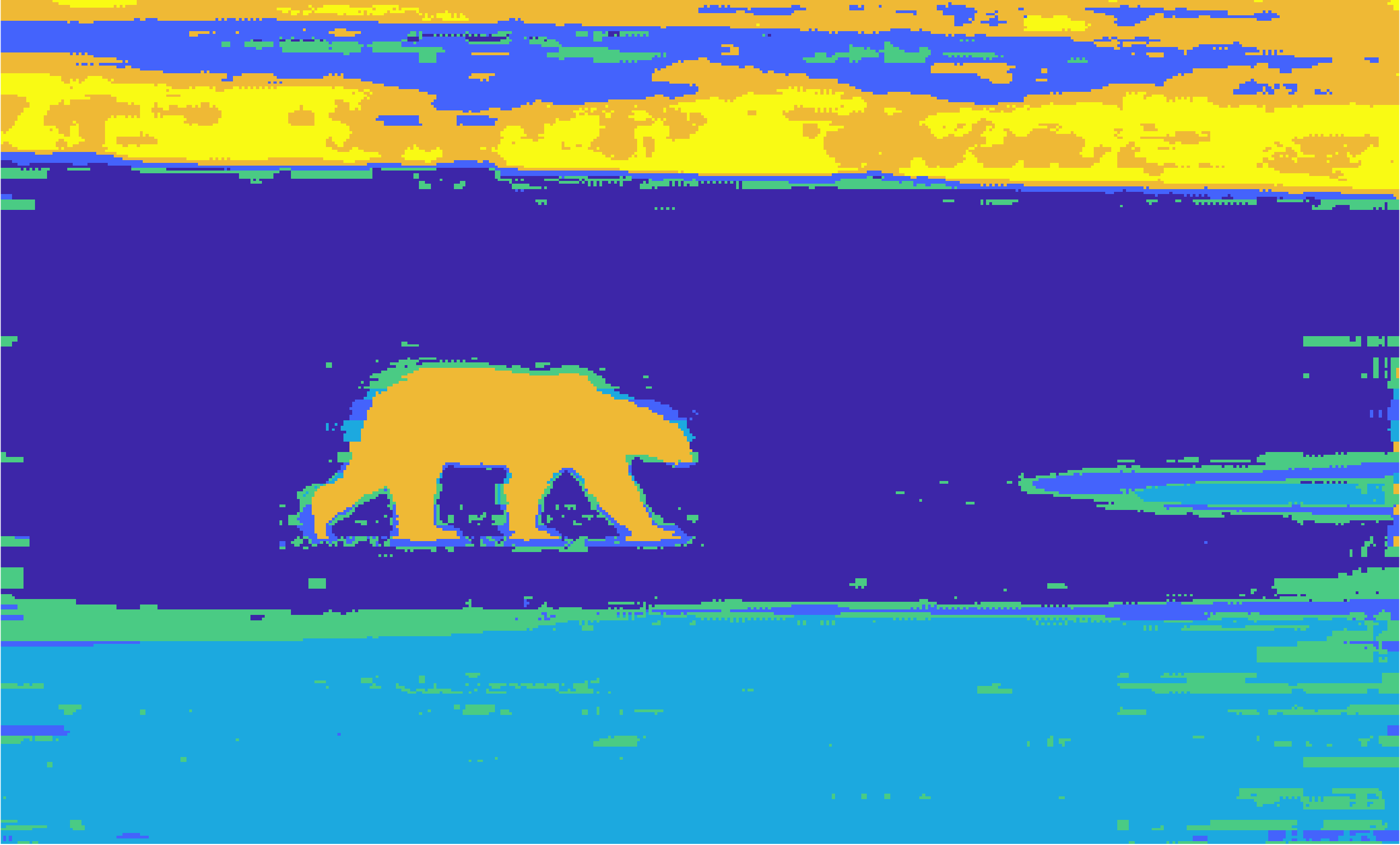}
  \end{subfigure}%
    \begin{subfigure}[b]{.20\linewidth}
    \centering
    \includegraphics[width=.99\textwidth]{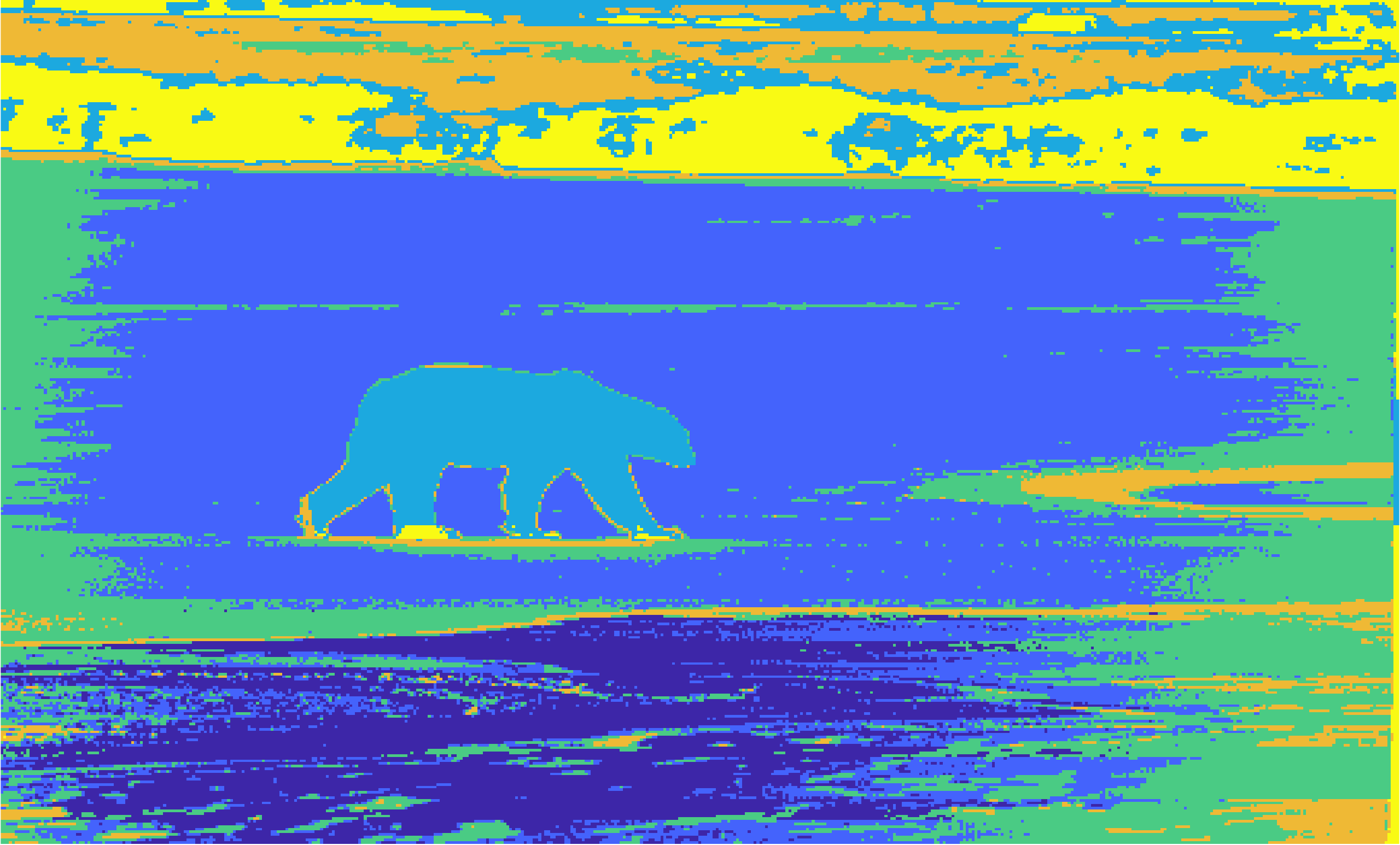}
  \end{subfigure}%
    \begin{subfigure}[b]{.20\linewidth}
    \centering
    \includegraphics[width=.99\textwidth]{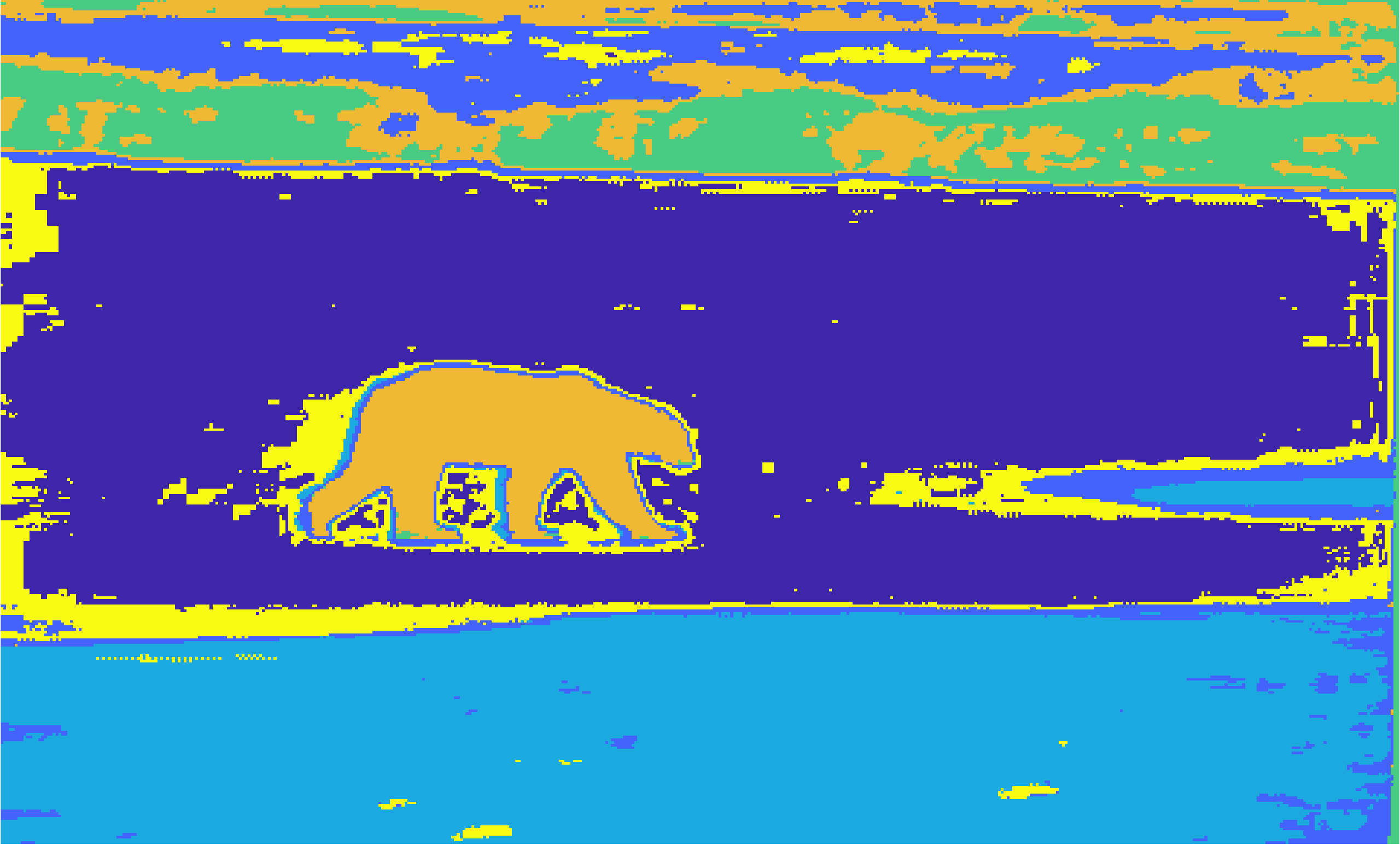}
  \end{subfigure}%
  \begin{subfigure}[b]{.20\linewidth}
    \centering
    \includegraphics[width=.99\textwidth]{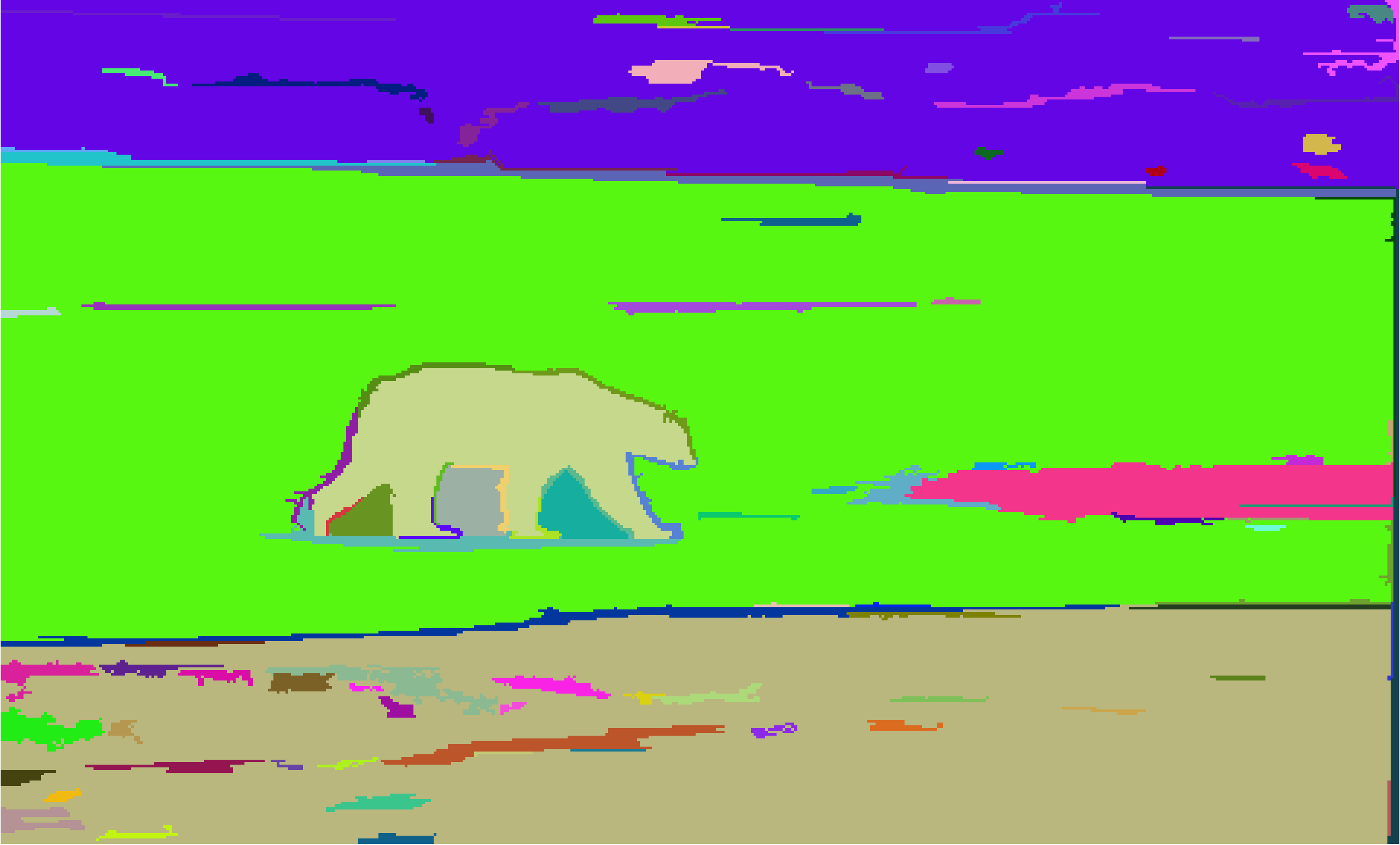}
  \end{subfigure}\\%
  
  
  \begin{subfigure}[b]{.20\linewidth}
    \centering
    \includegraphics[width=.99\textwidth]{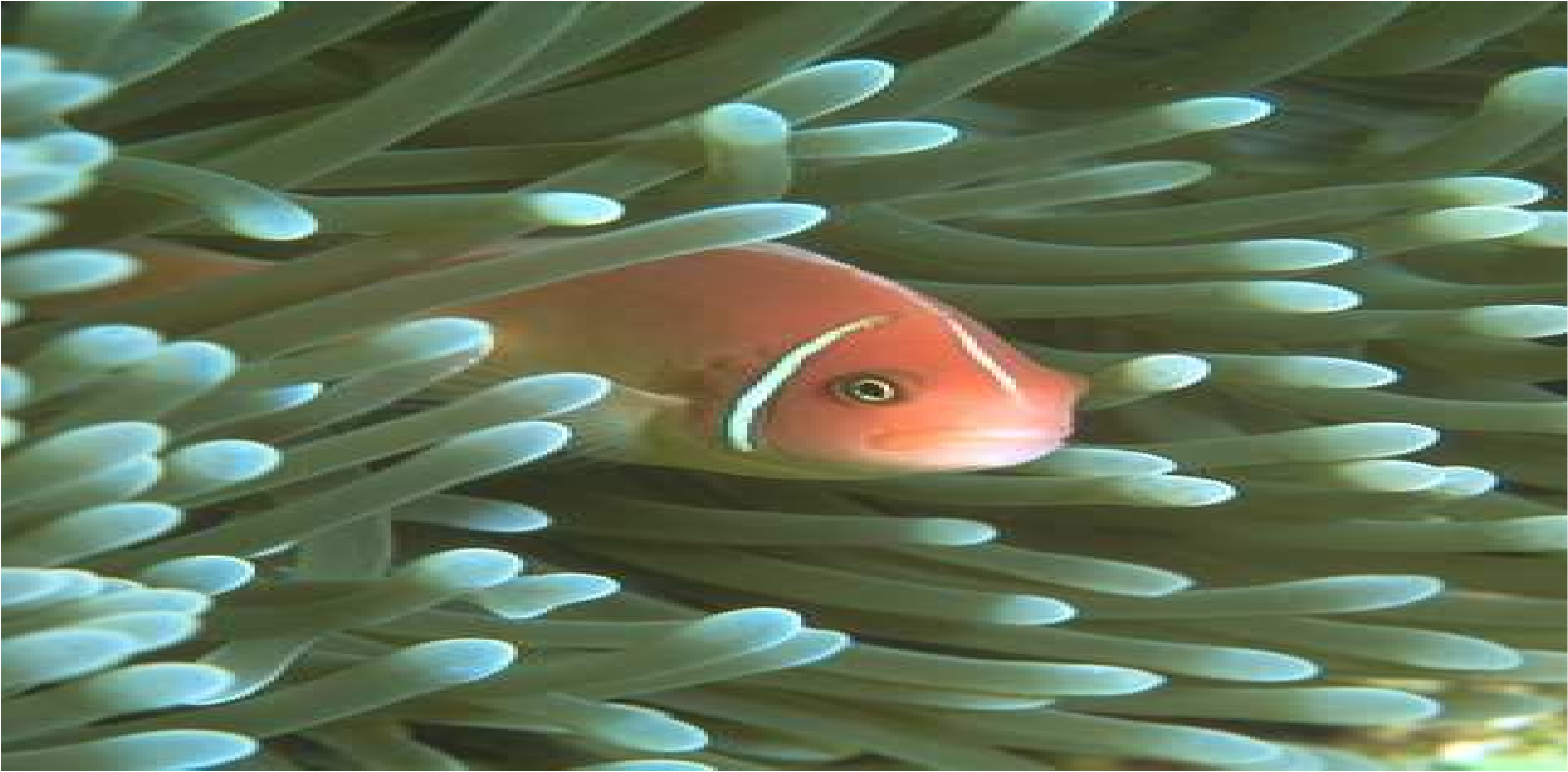}
  \end{subfigure}%
  \begin{subfigure}[b]{.20\linewidth}
    \centering
    \includegraphics[width=.99\textwidth]{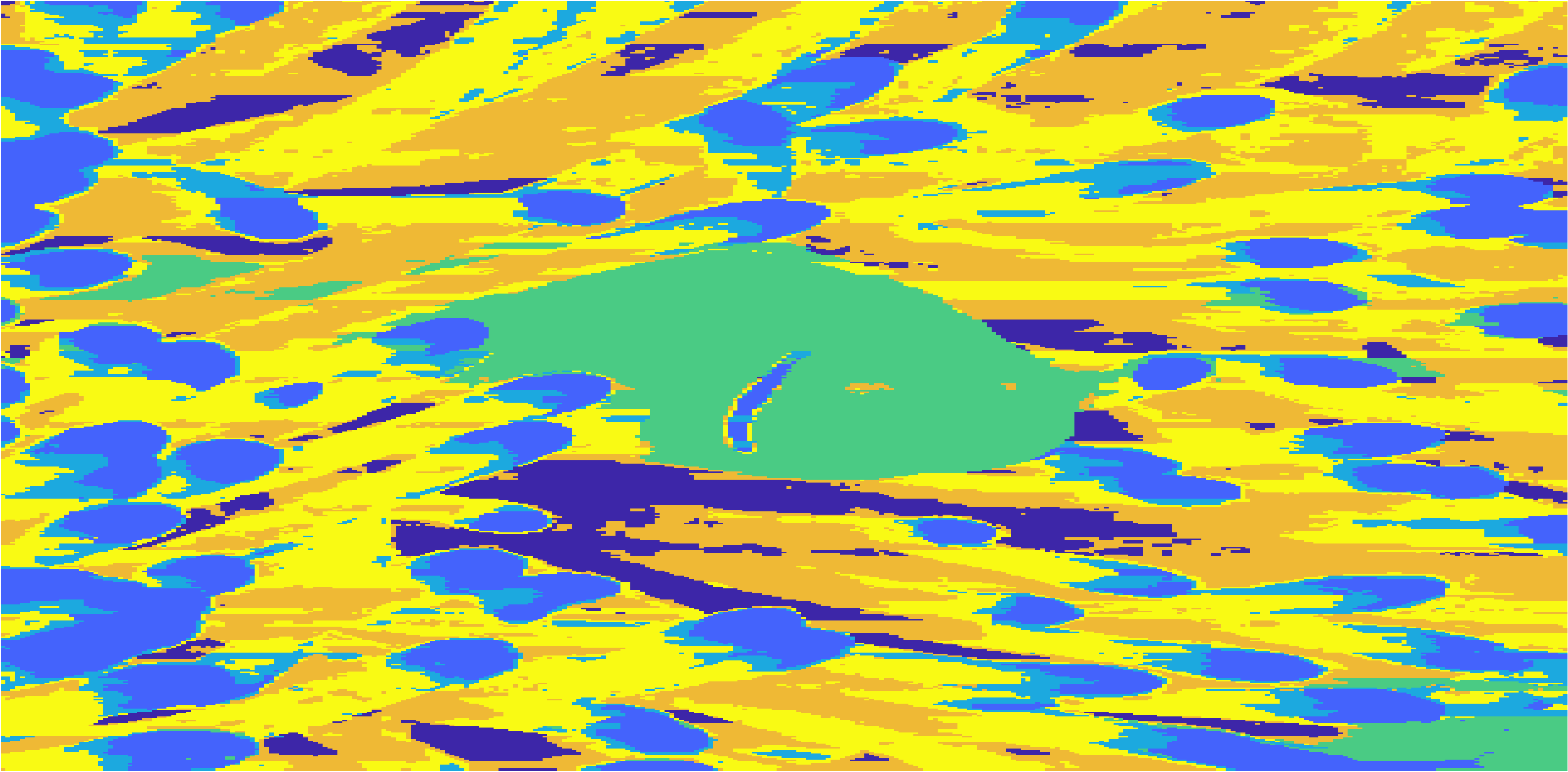}
  \end{subfigure}%
    \begin{subfigure}[b]{.20\linewidth}
    \centering
    \includegraphics[width=.99\textwidth]{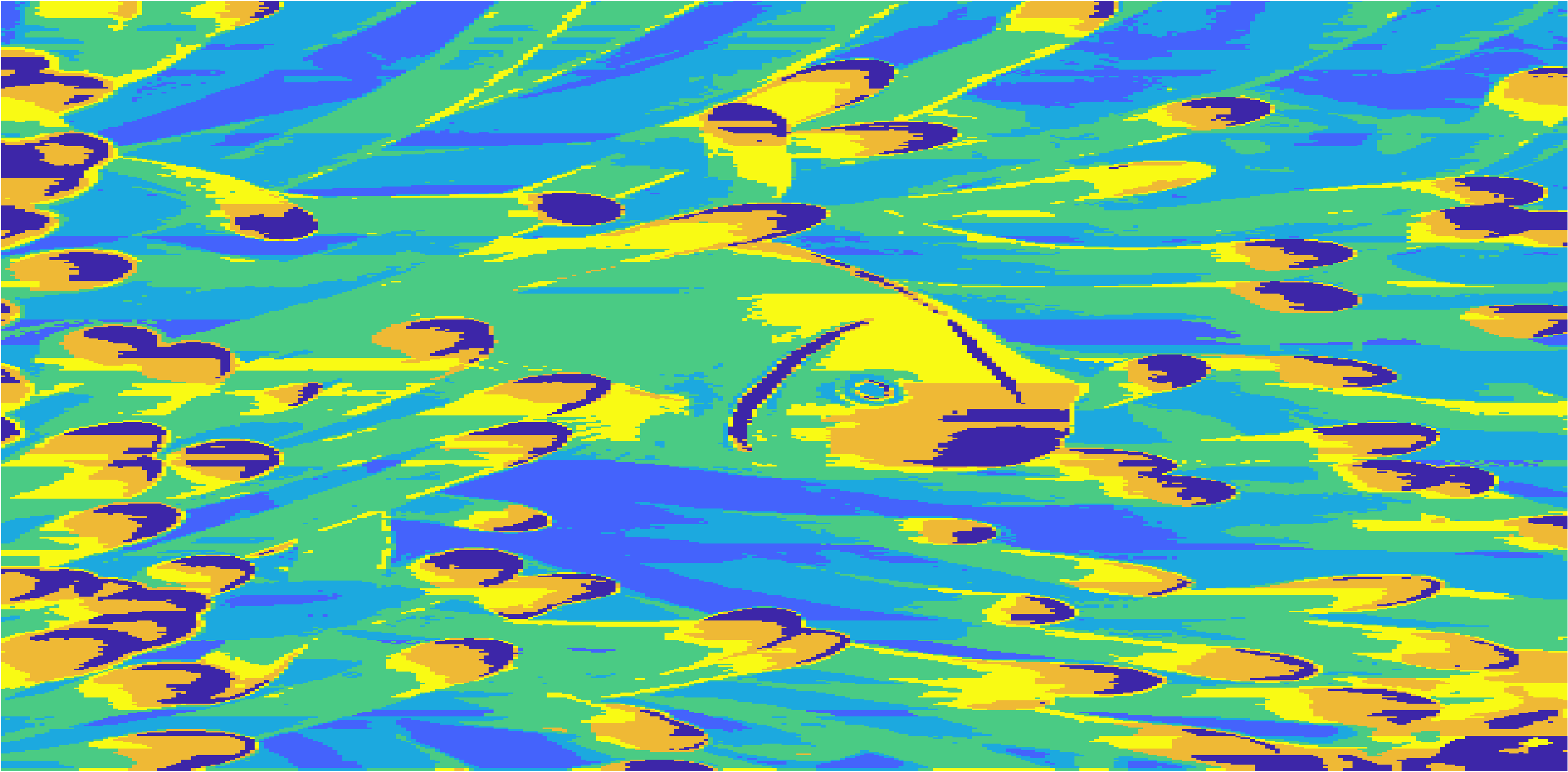}
  \end{subfigure}%
    \begin{subfigure}[b]{.20\linewidth}
    \centering
    \includegraphics[width=.99\textwidth]{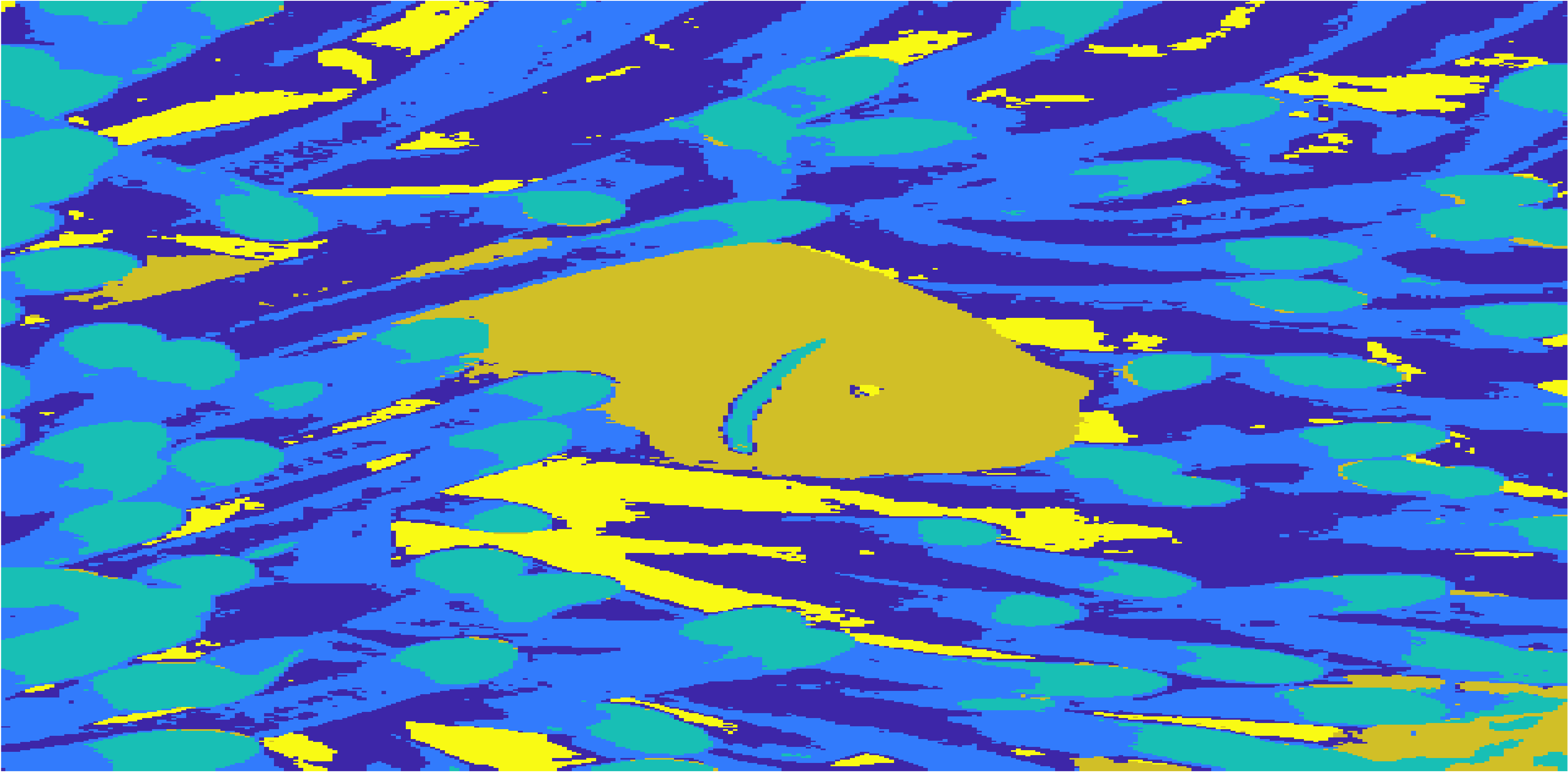}
  \end{subfigure}%
  \begin{subfigure}[b]{.20\linewidth}
    \centering
    \includegraphics[width=.99\textwidth]{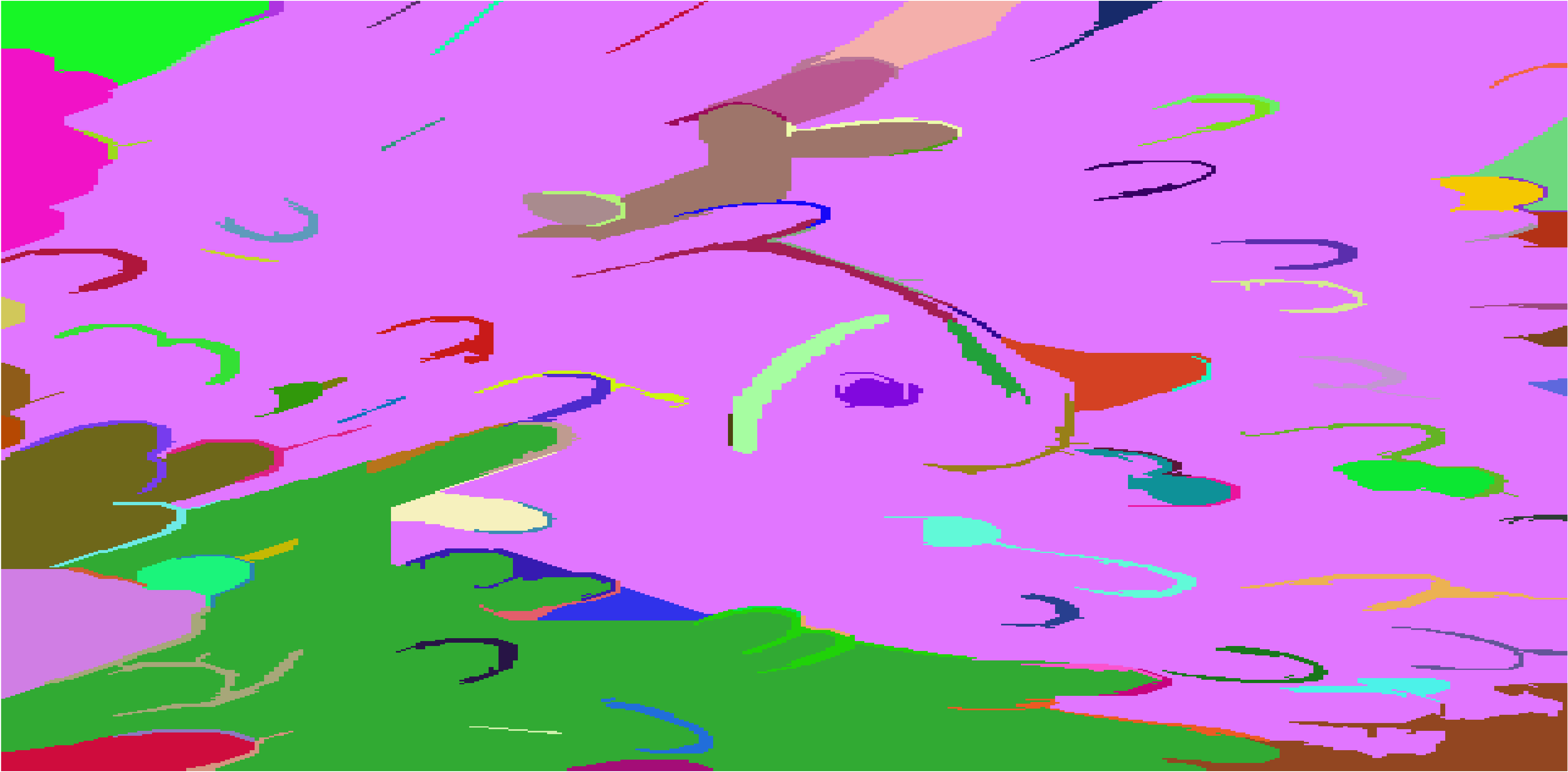}
  \end{subfigure}\\%


  \begin{subfigure}[b]{.20\linewidth}
    \centering
    \includegraphics[width=.99\textwidth]{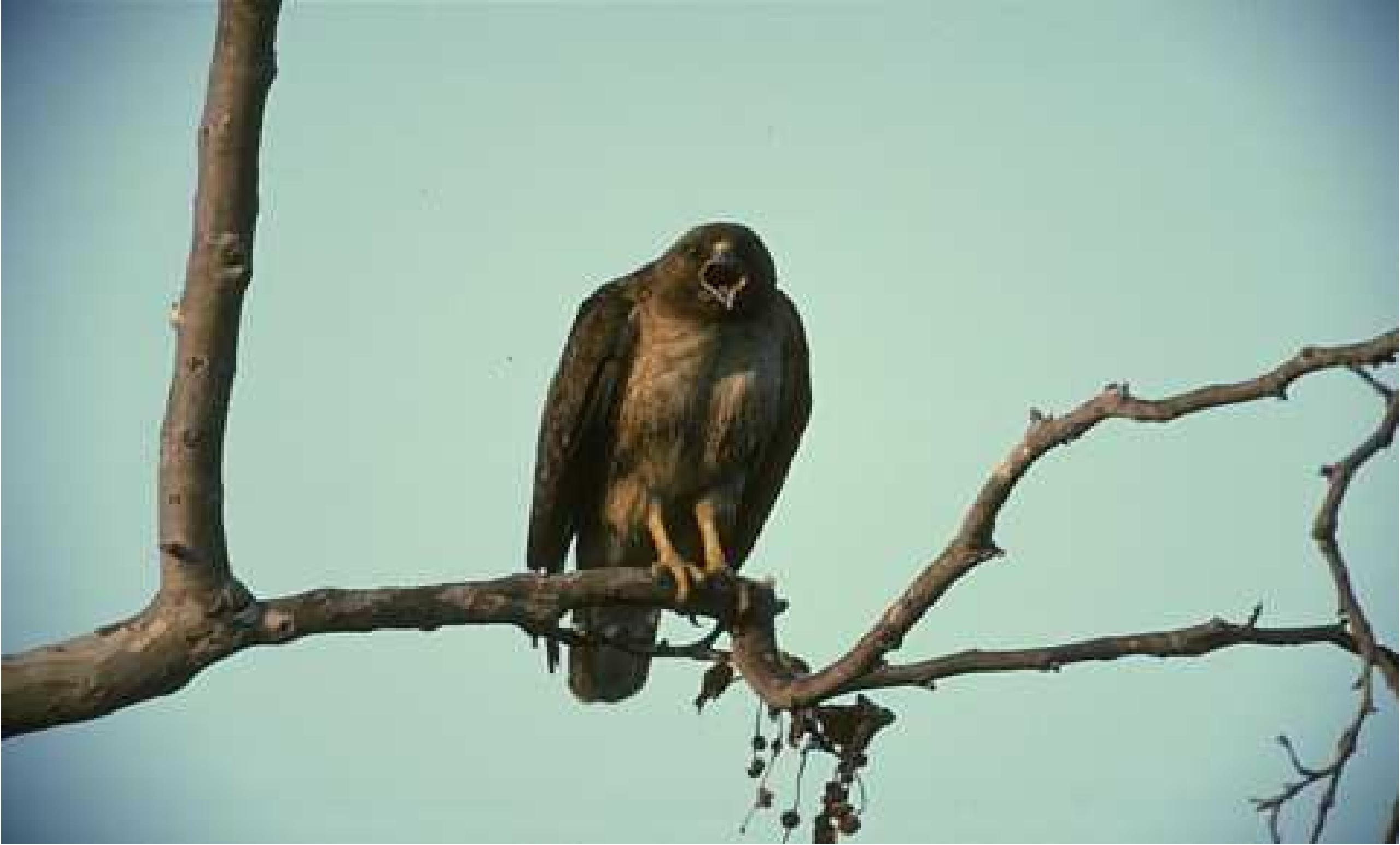}
  \end{subfigure}%
  \begin{subfigure}[b]{.20\linewidth}
    \centering
    \includegraphics[width=.99\textwidth]{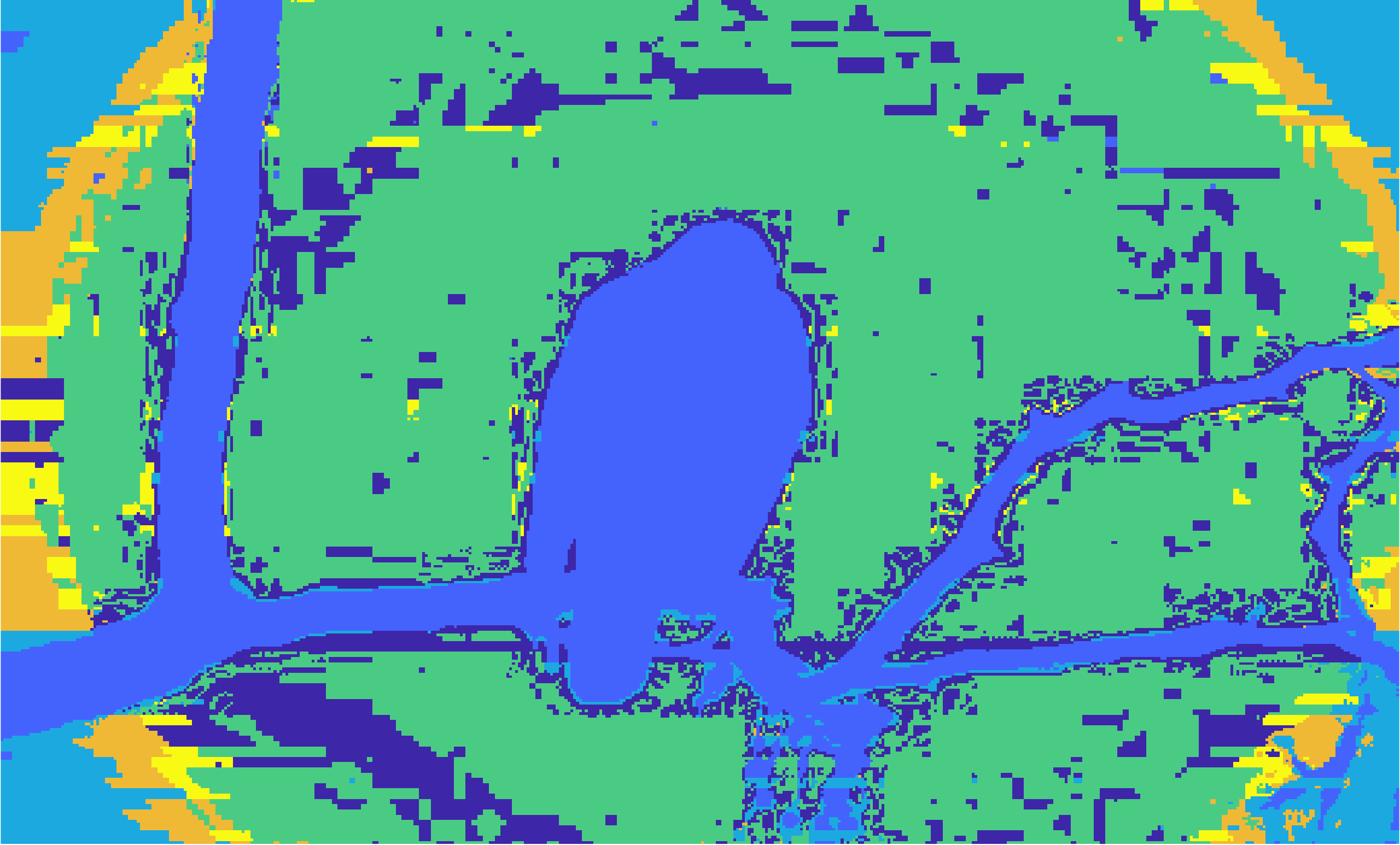}
  \end{subfigure}%
    \begin{subfigure}[b]{.20\linewidth}
    \centering
    \includegraphics[width=.99\textwidth]{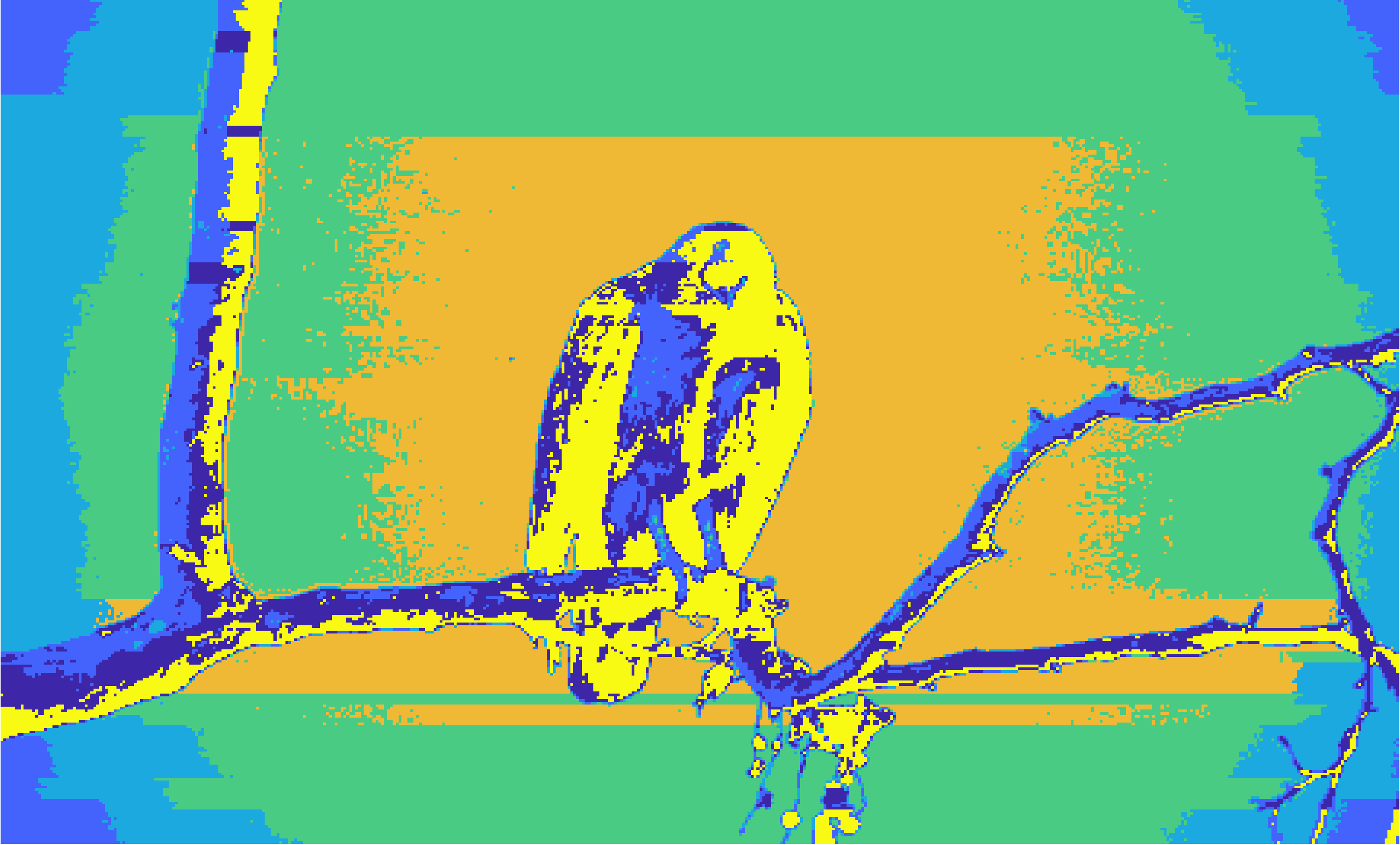}
  \end{subfigure}%
    \begin{subfigure}[b]{.20\linewidth}
    \centering
    \includegraphics[width=.99\textwidth]{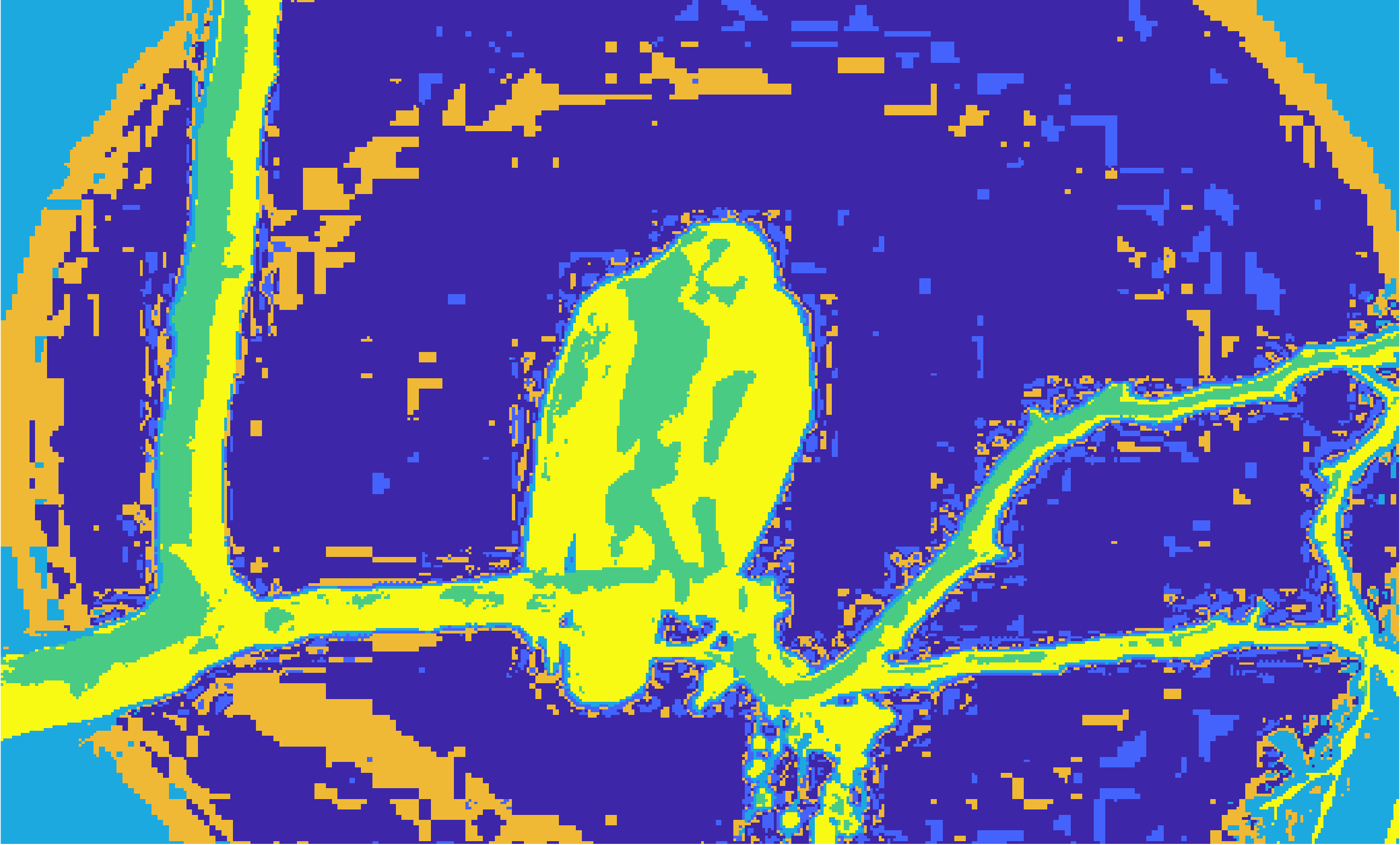}
  \end{subfigure}%
  \begin{subfigure}[b]{.20\linewidth}
    \centering
    \includegraphics[width=.99\textwidth]{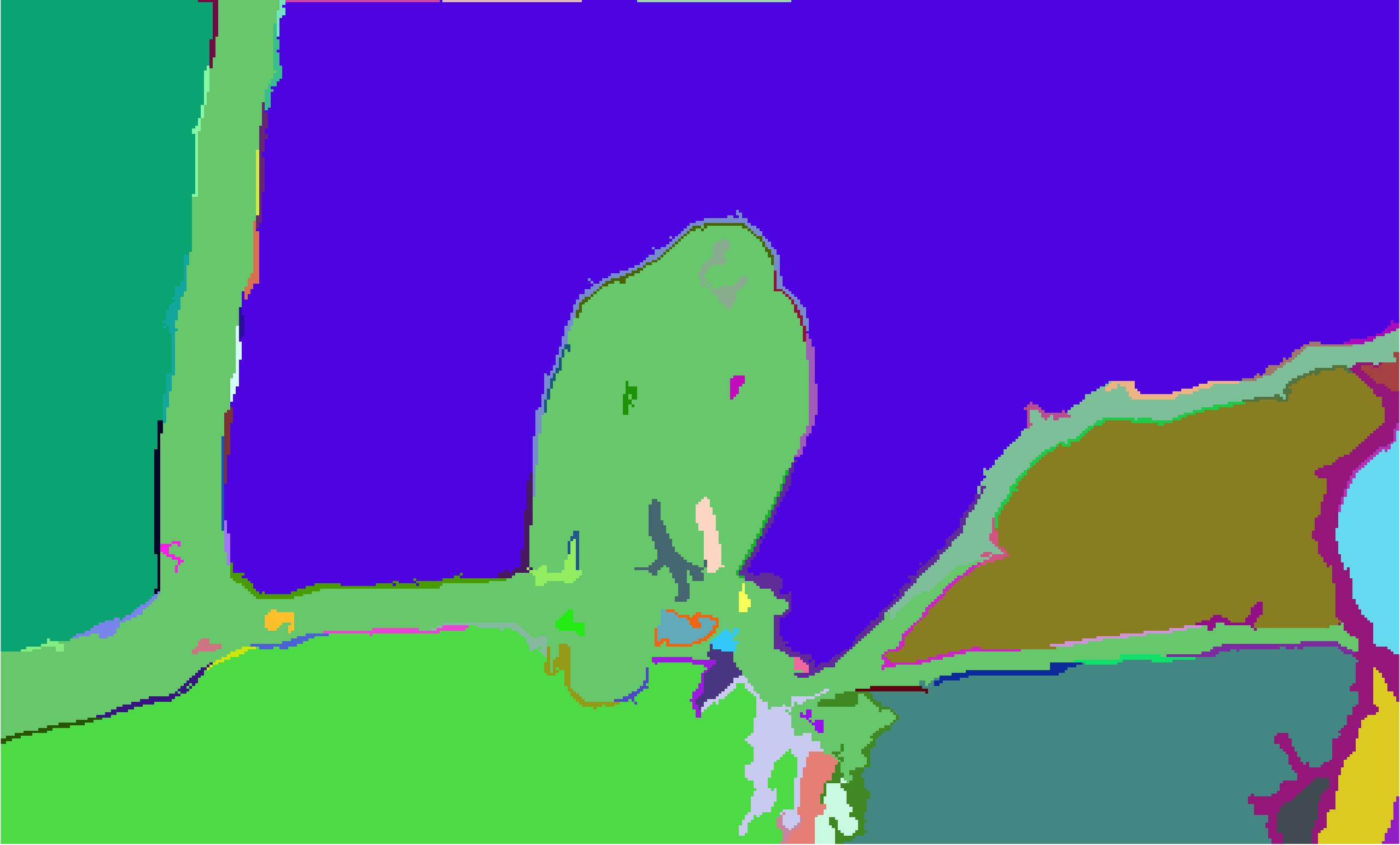}
  \end{subfigure}\\%

    \begin{subfigure}[b]{.20\linewidth}
    \centering
    \includegraphics[width=.99\textwidth]{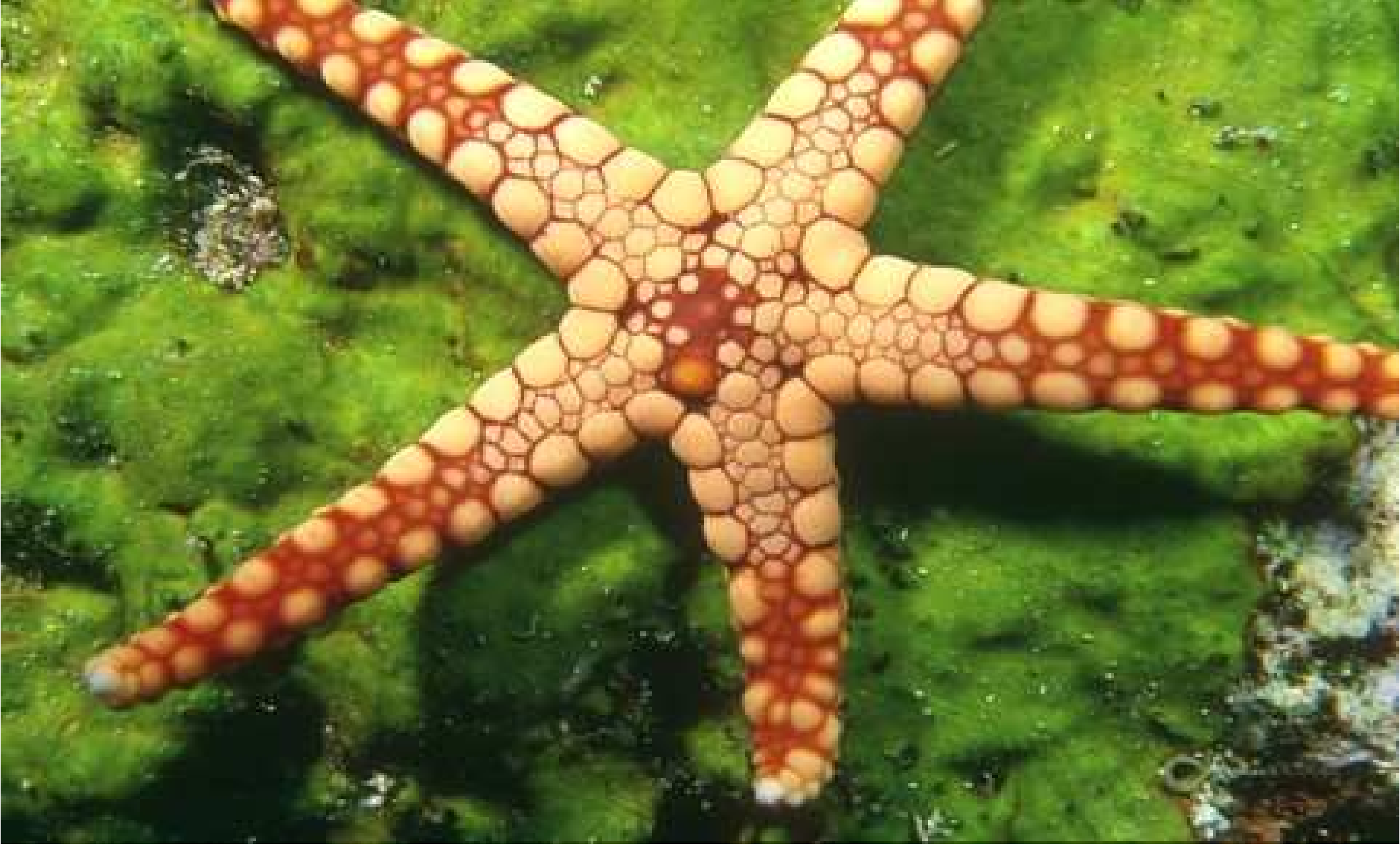}
  \end{subfigure}%
  \begin{subfigure}[b]{.20\linewidth}
    \centering
    \includegraphics[width=.99\textwidth]{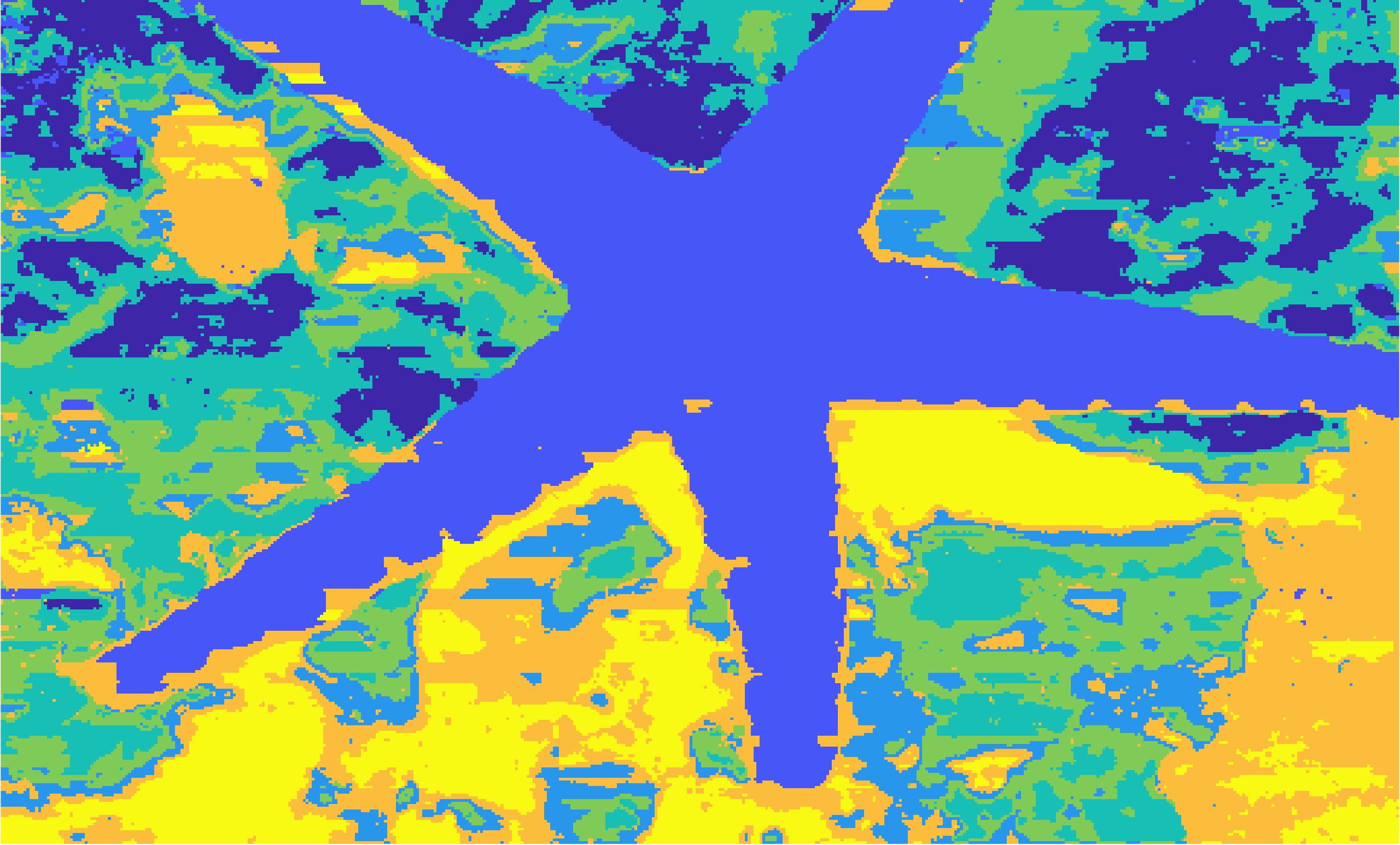}
  \end{subfigure}%
    \begin{subfigure}[b]{.20\linewidth}
    \centering
    \includegraphics[width=.99\textwidth]{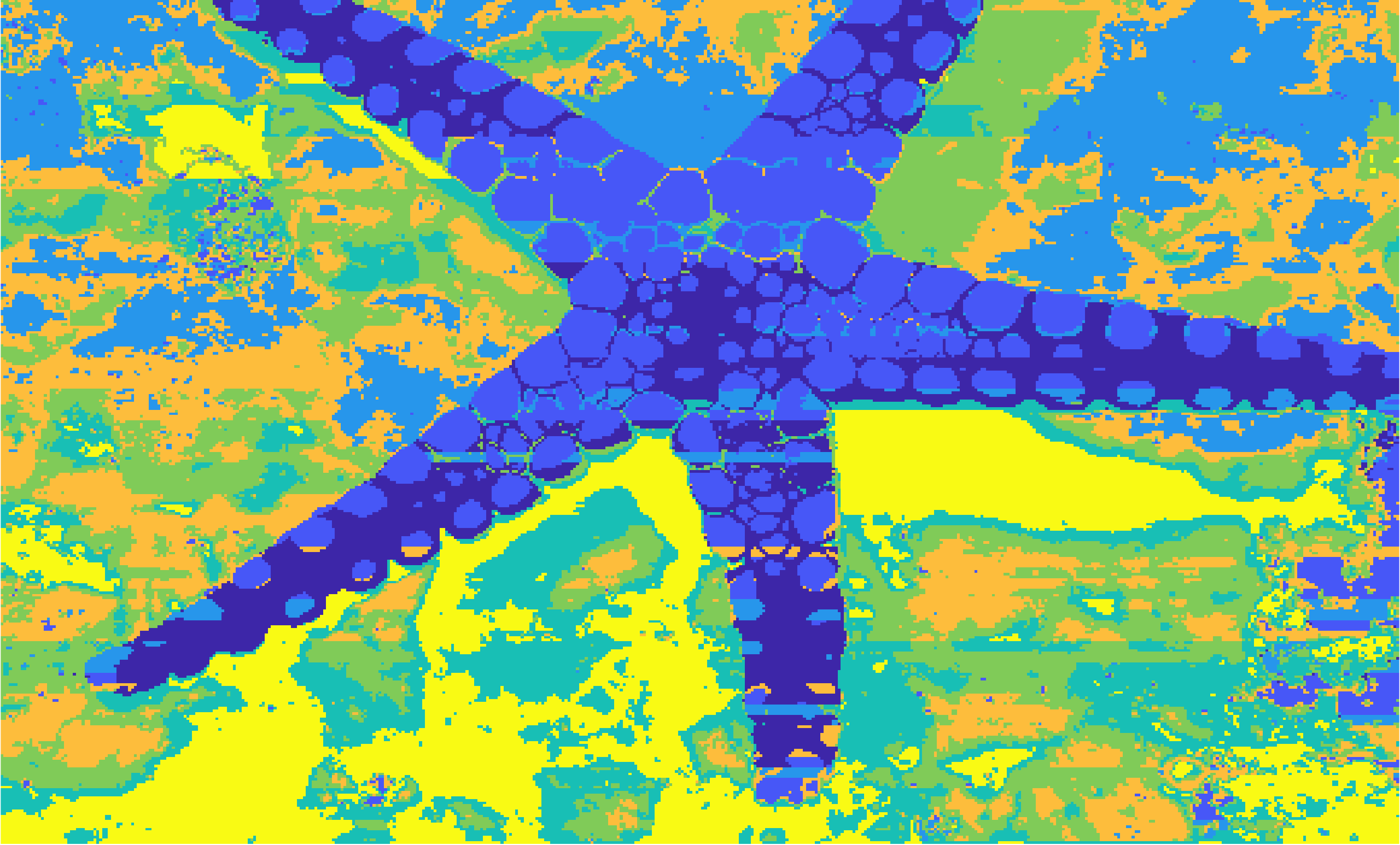}
  \end{subfigure}%
    \begin{subfigure}[b]{.20\linewidth}
    \centering
    \includegraphics[width=.99\textwidth]{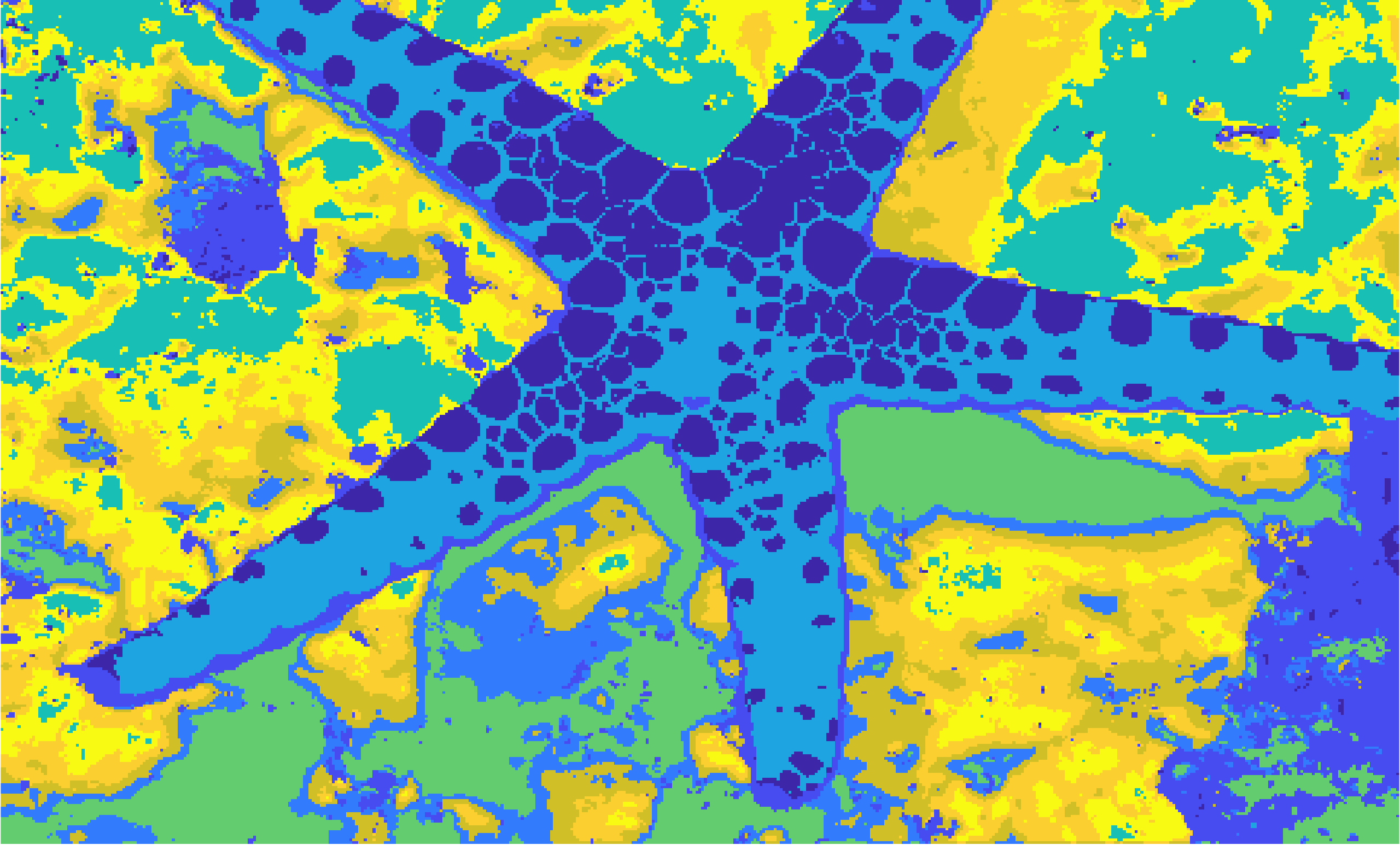}
  \end{subfigure}%
  \begin{subfigure}[b]{.20\linewidth}
    \centering
    \includegraphics[width=.99\textwidth]{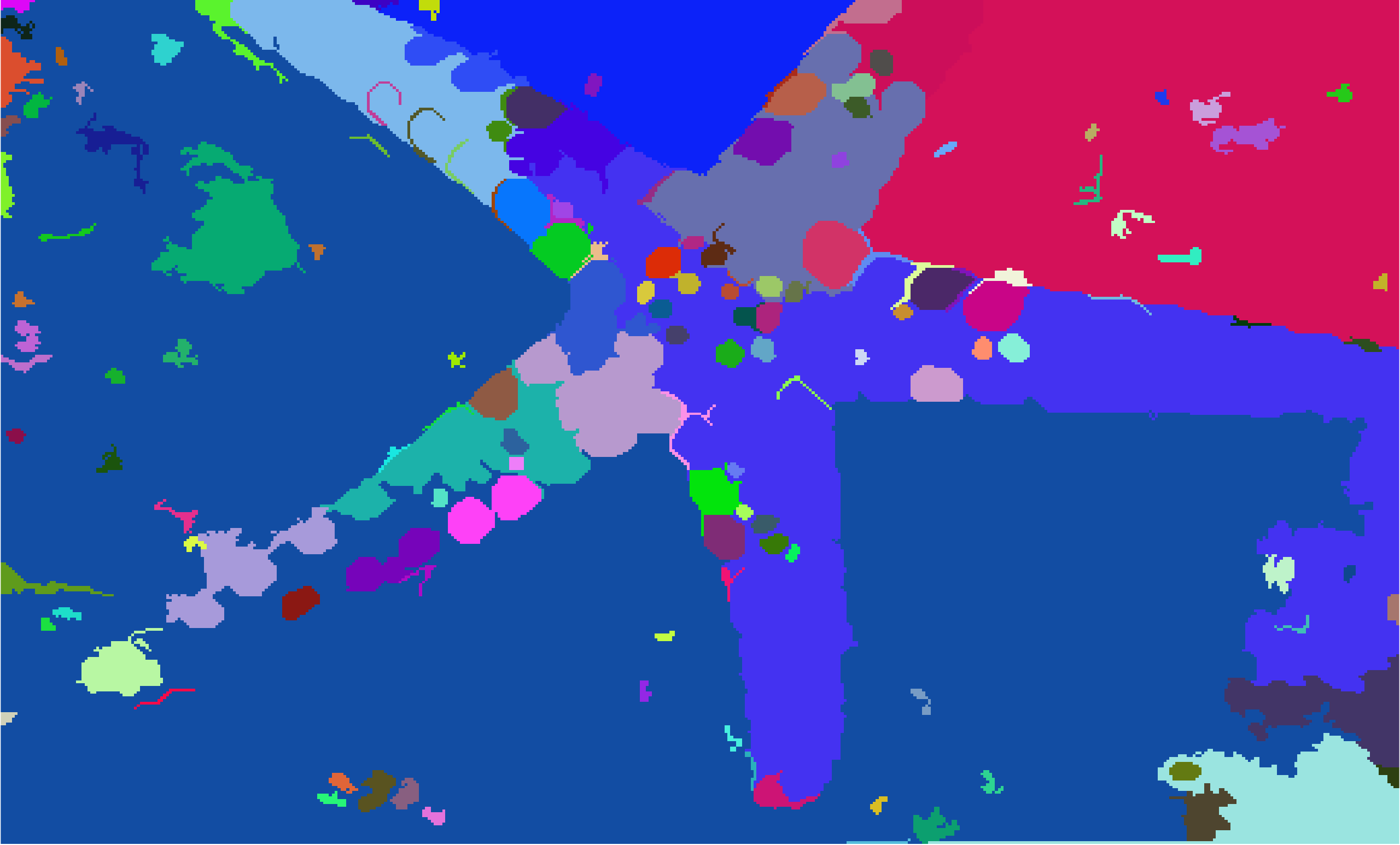}
  \end{subfigure}\\%
  

      \begin{subfigure}[b]{.20\linewidth}
    \centering
    \includegraphics[width=.99\textwidth]{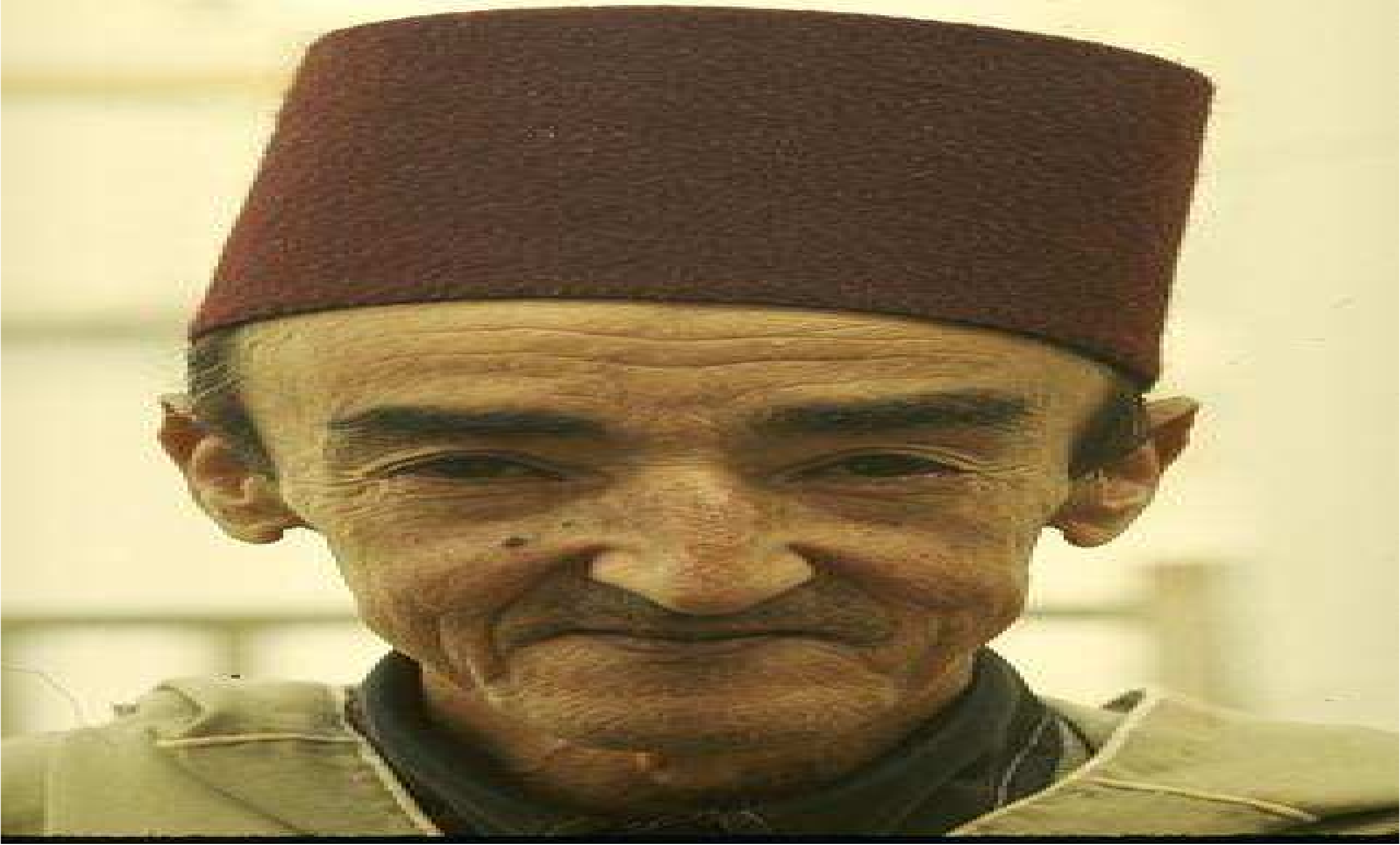}
  \end{subfigure}%
  \begin{subfigure}[b]{.20\linewidth}
    \centering
    \includegraphics[width=.99\textwidth]{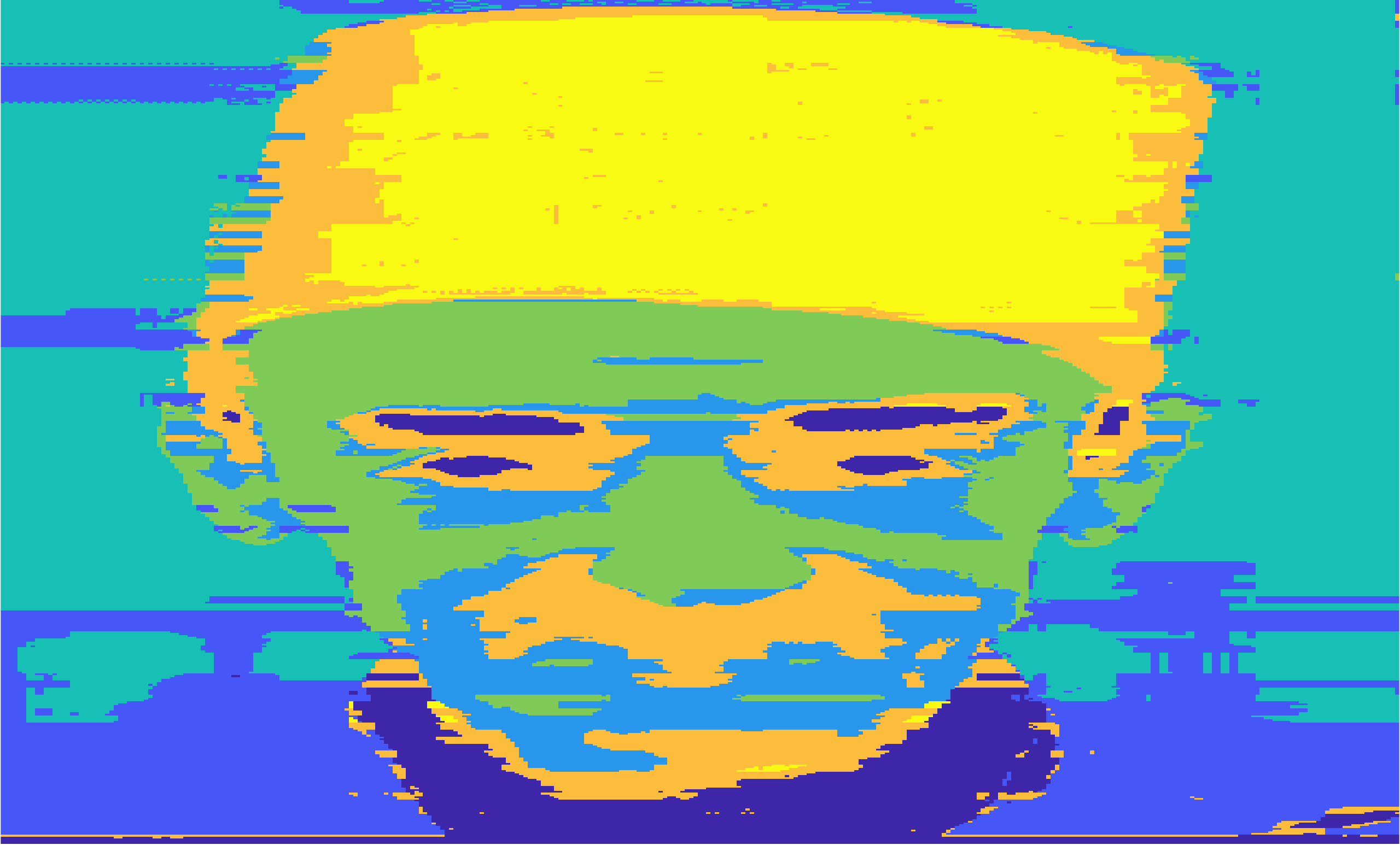}
  \end{subfigure}%
    \begin{subfigure}[b]{.20\linewidth}
    \centering
    \includegraphics[width=.99\textwidth]{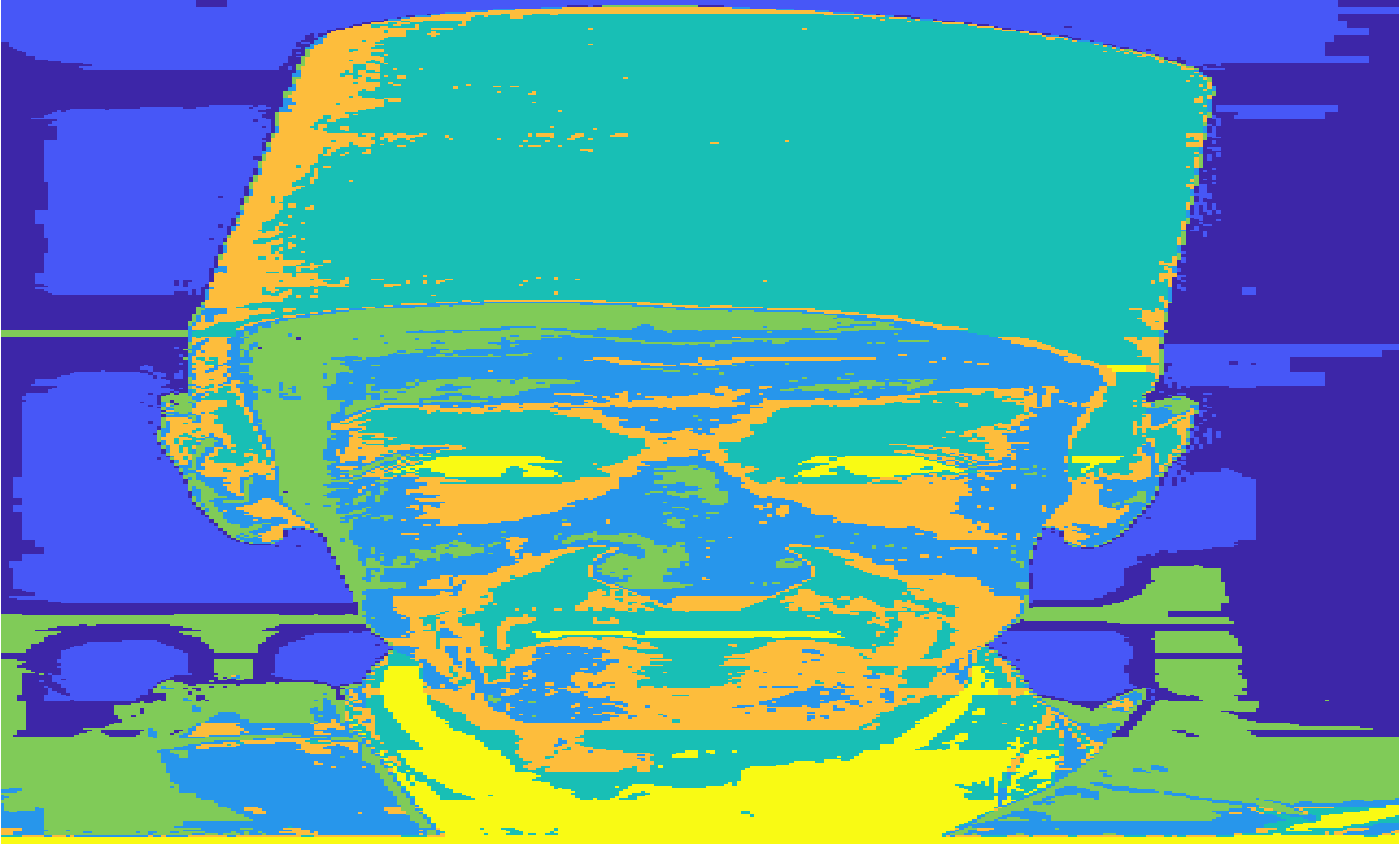}
  \end{subfigure}%
    \begin{subfigure}[b]{.20\linewidth}
    \centering
    \includegraphics[width=.99\textwidth]{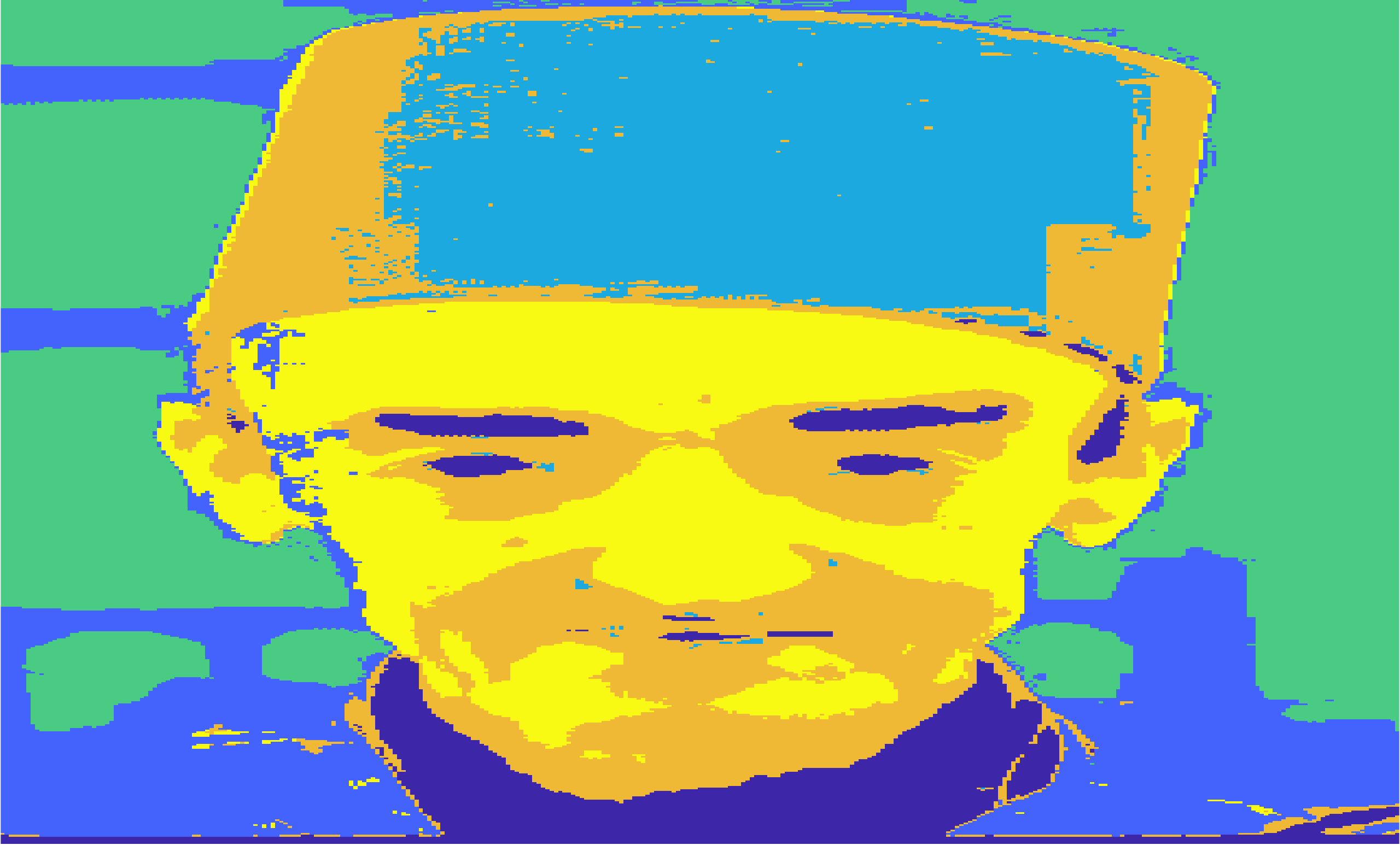}
  \end{subfigure}%
  \begin{subfigure}[b]{.20\linewidth}
    \centering
    \includegraphics[width=.99\textwidth]{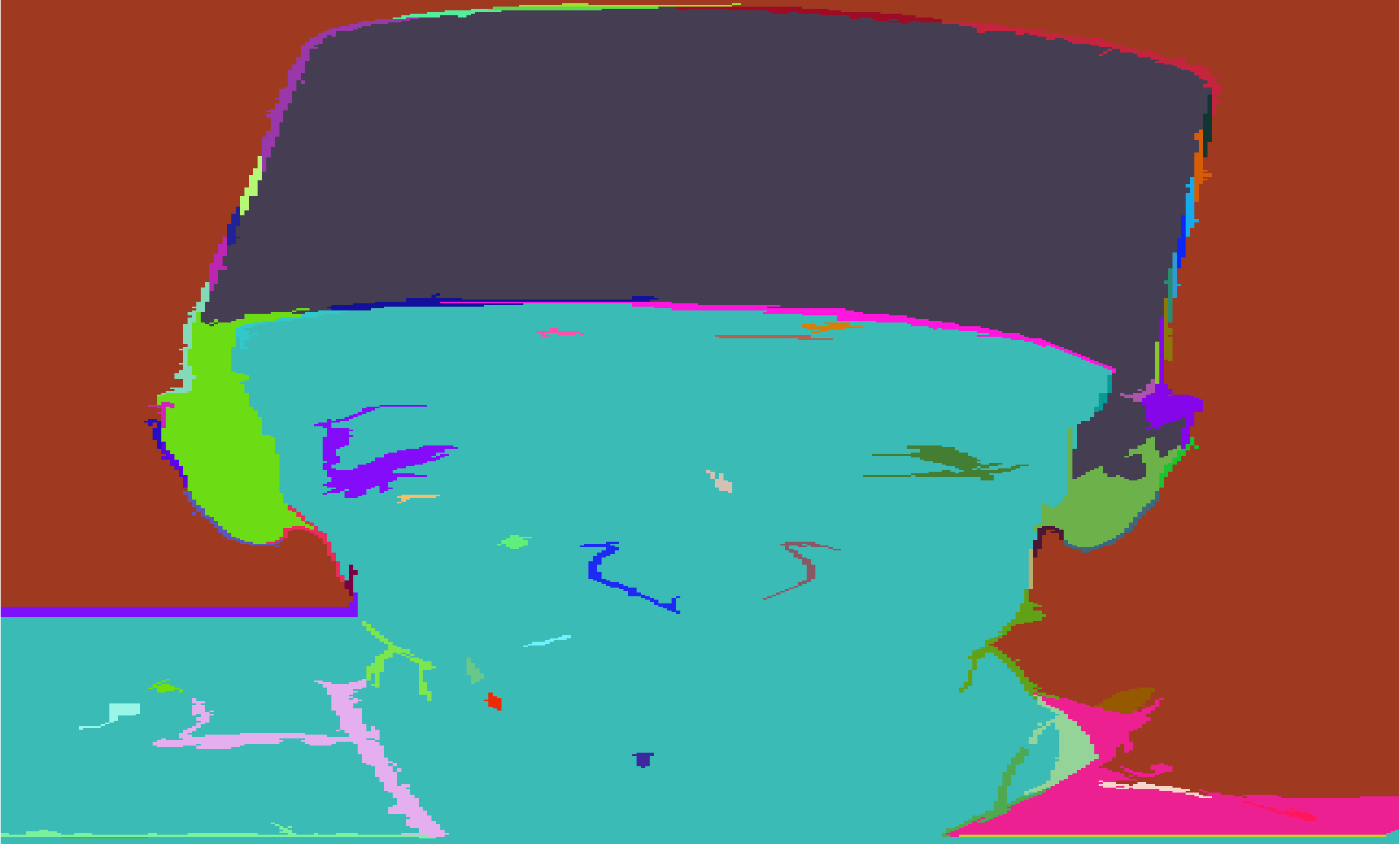}
  \end{subfigure}\\%
  
  
       \begin{subfigure}[b]{.20\linewidth}
    \centering
    \includegraphics[width=.99\textwidth]{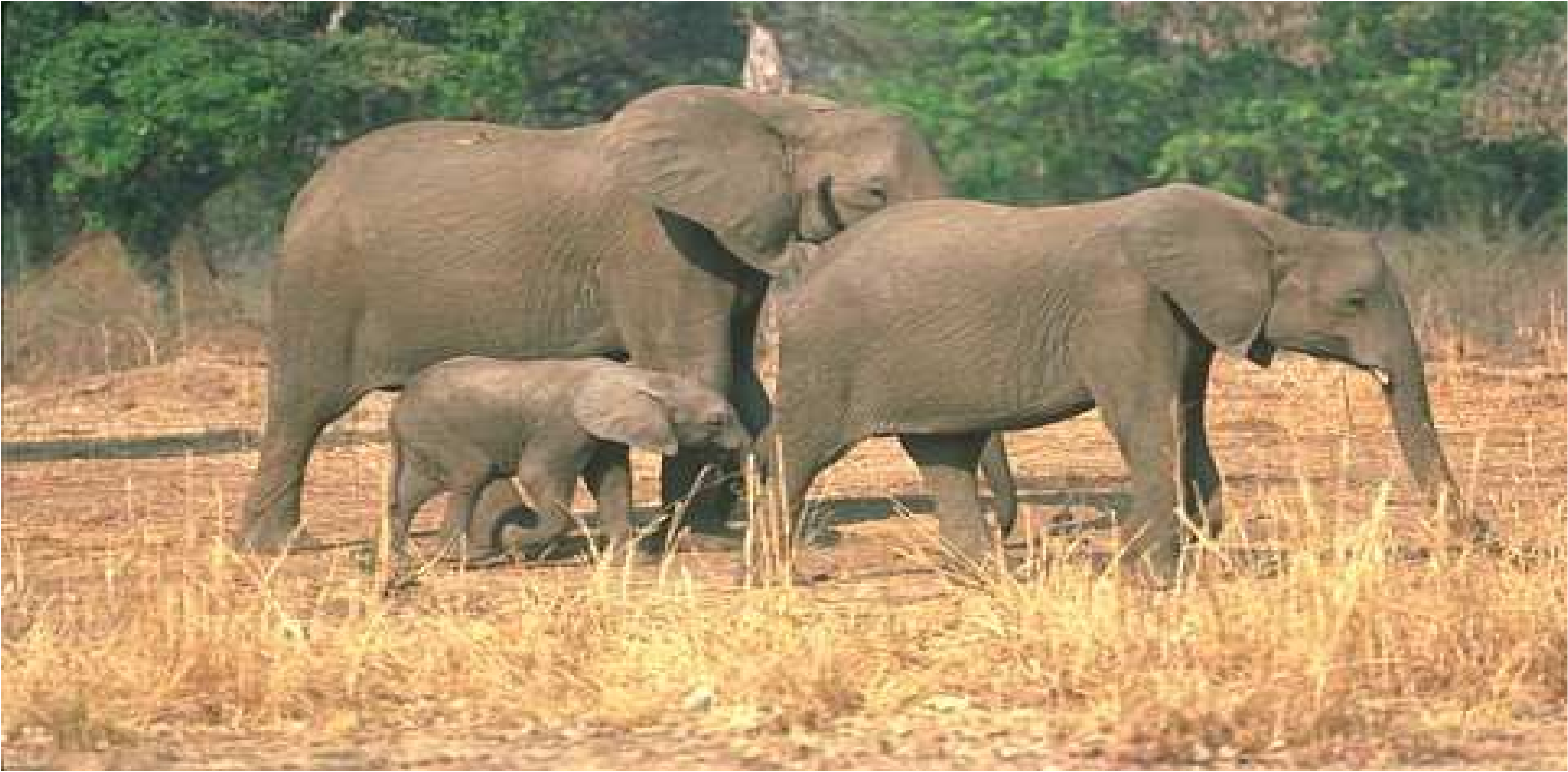}
  \end{subfigure}%
  \begin{subfigure}[b]{.20\linewidth}
    \centering
    \includegraphics[width=.99\textwidth]{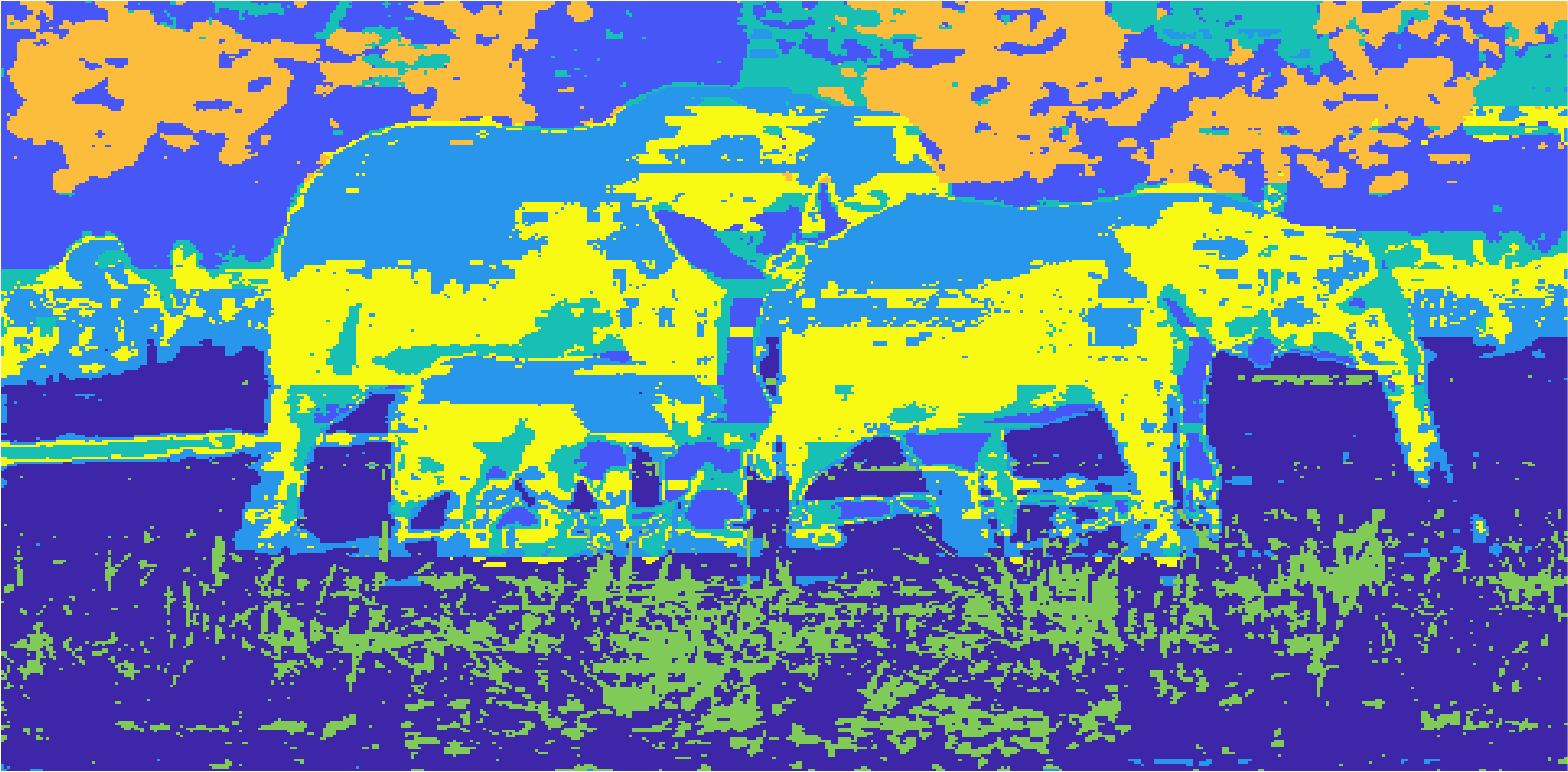}
  \end{subfigure}%
    \begin{subfigure}[b]{.20\linewidth}
    \centering
    \includegraphics[width=.99\textwidth]{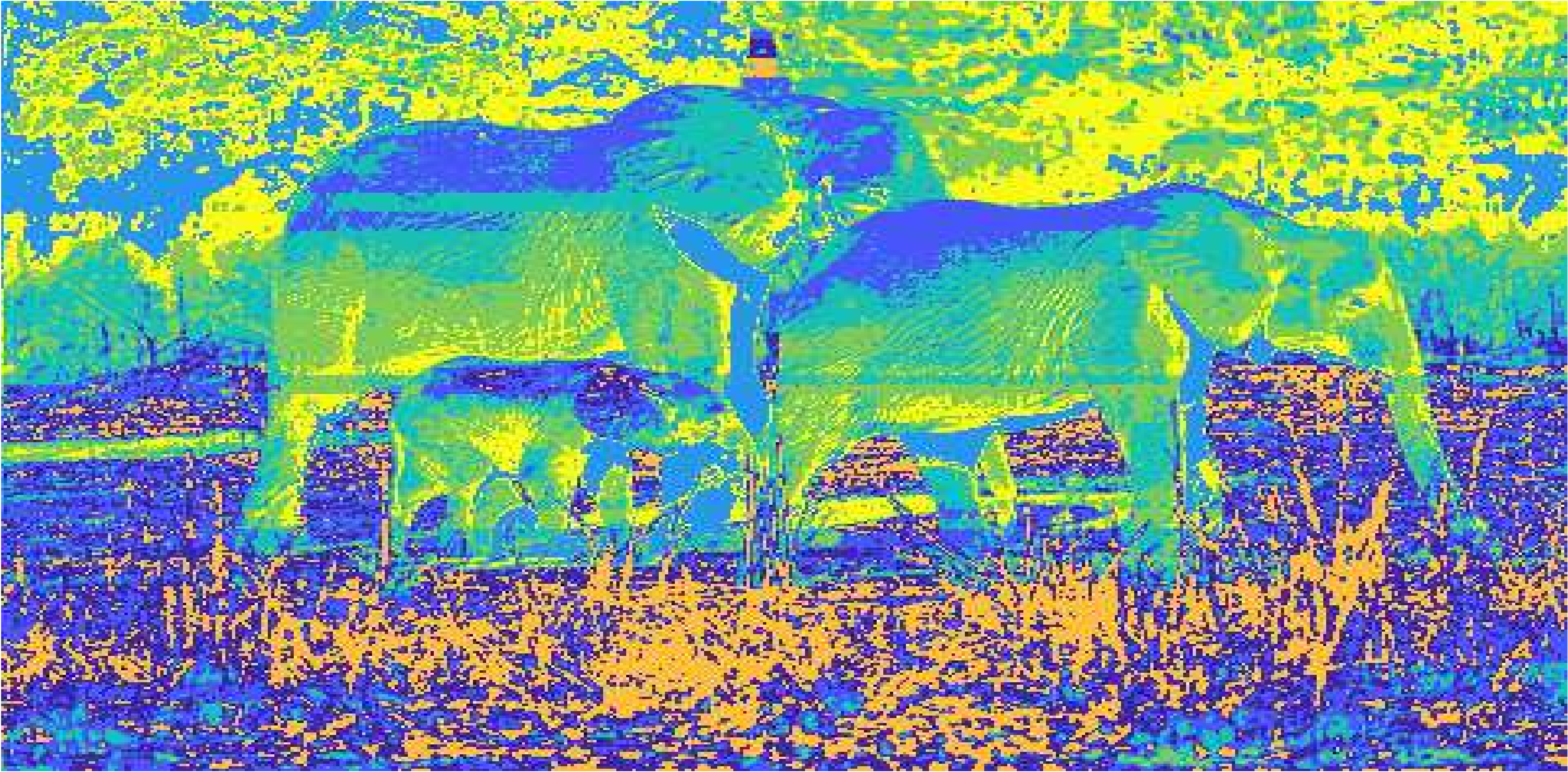}
  \end{subfigure}%
    \begin{subfigure}[b]{.20\linewidth}
    \centering
    \includegraphics[width=.99\textwidth]{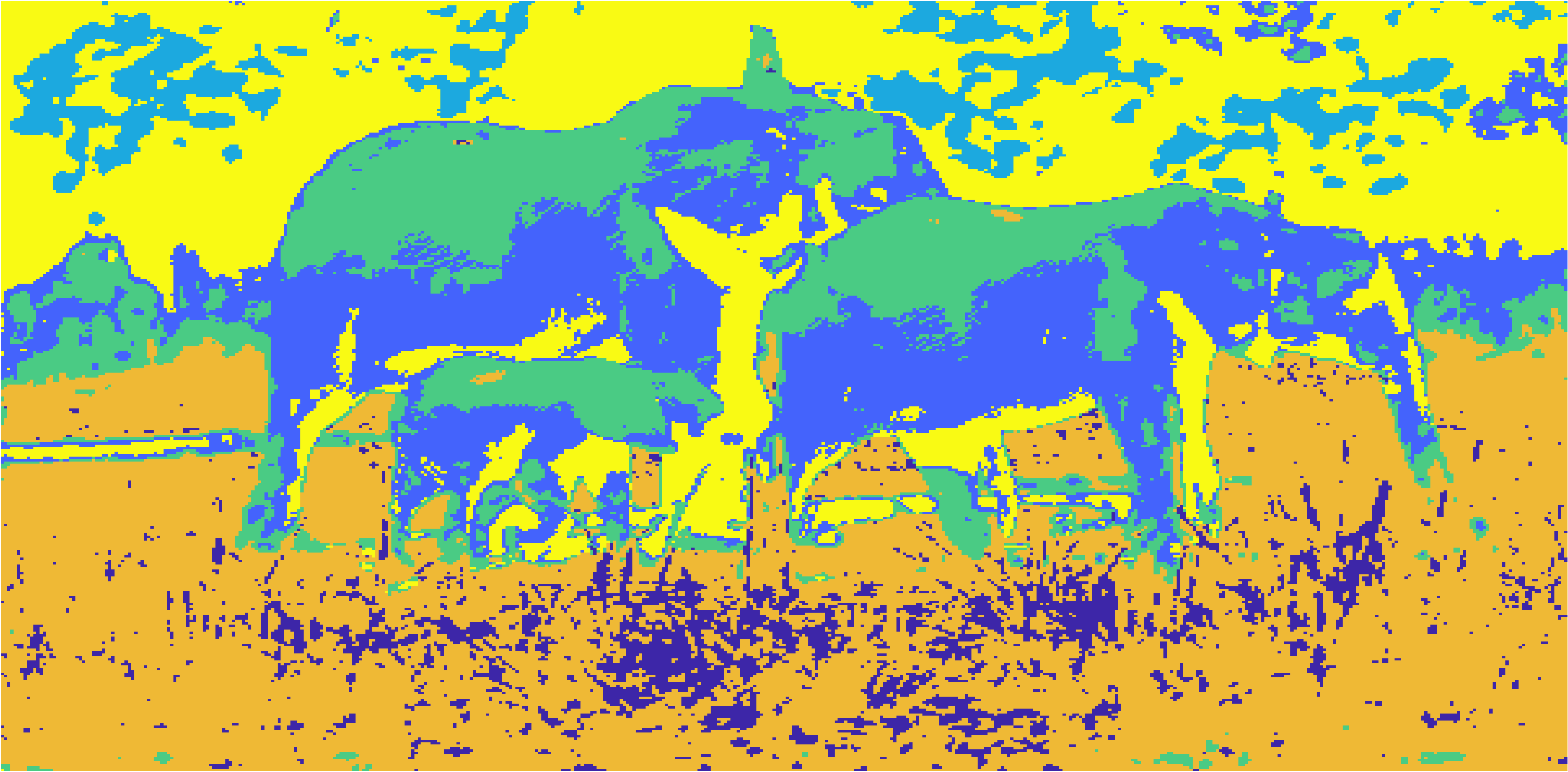}
  \end{subfigure}%
  \begin{subfigure}[b]{.20\linewidth}
    \centering
    \includegraphics[width=.99\textwidth]{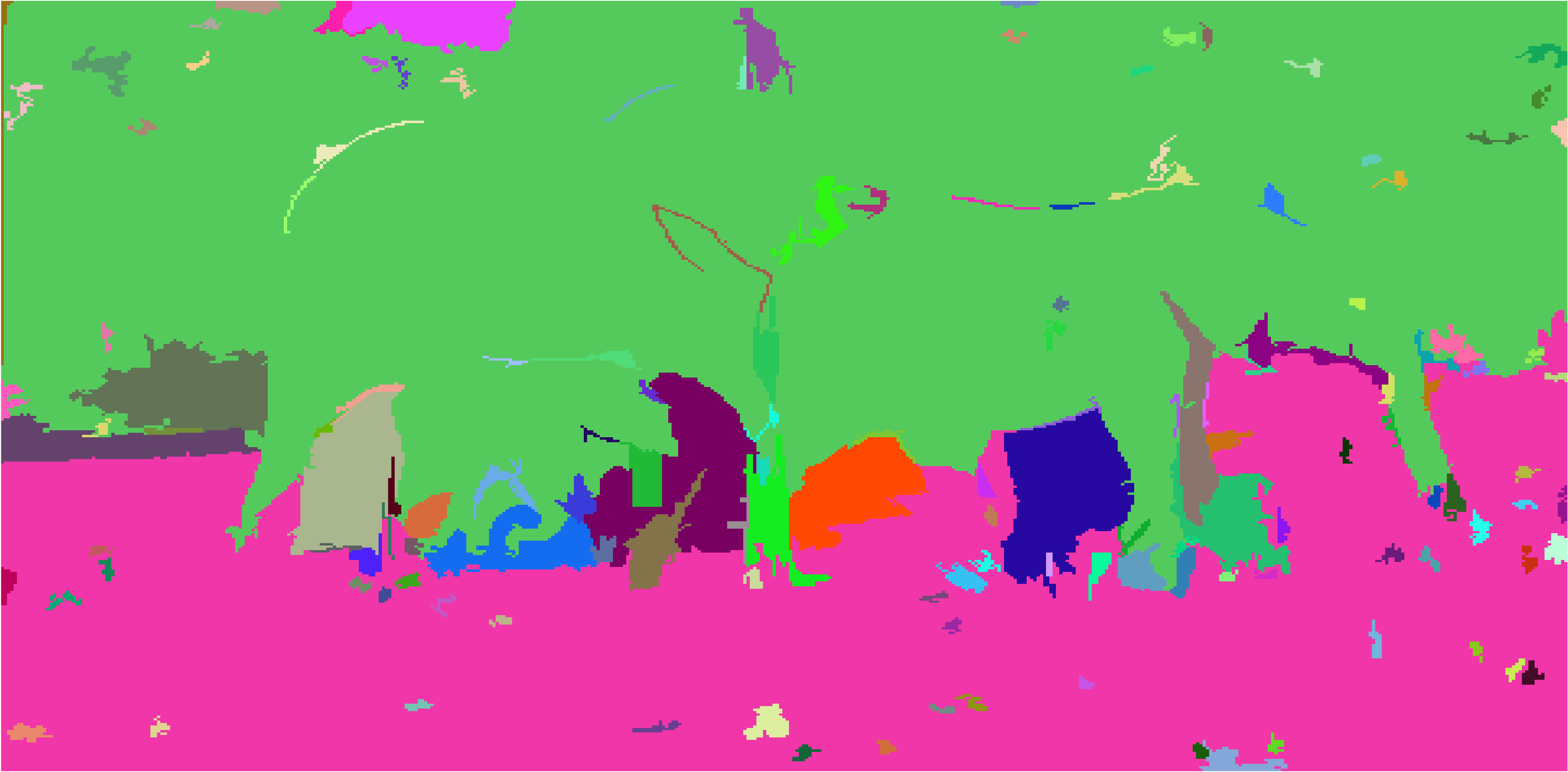}
  \end{subfigure}\\%


       \begin{subfigure}[b]{.20\linewidth}
    \centering
    \includegraphics[width=.99\textwidth]{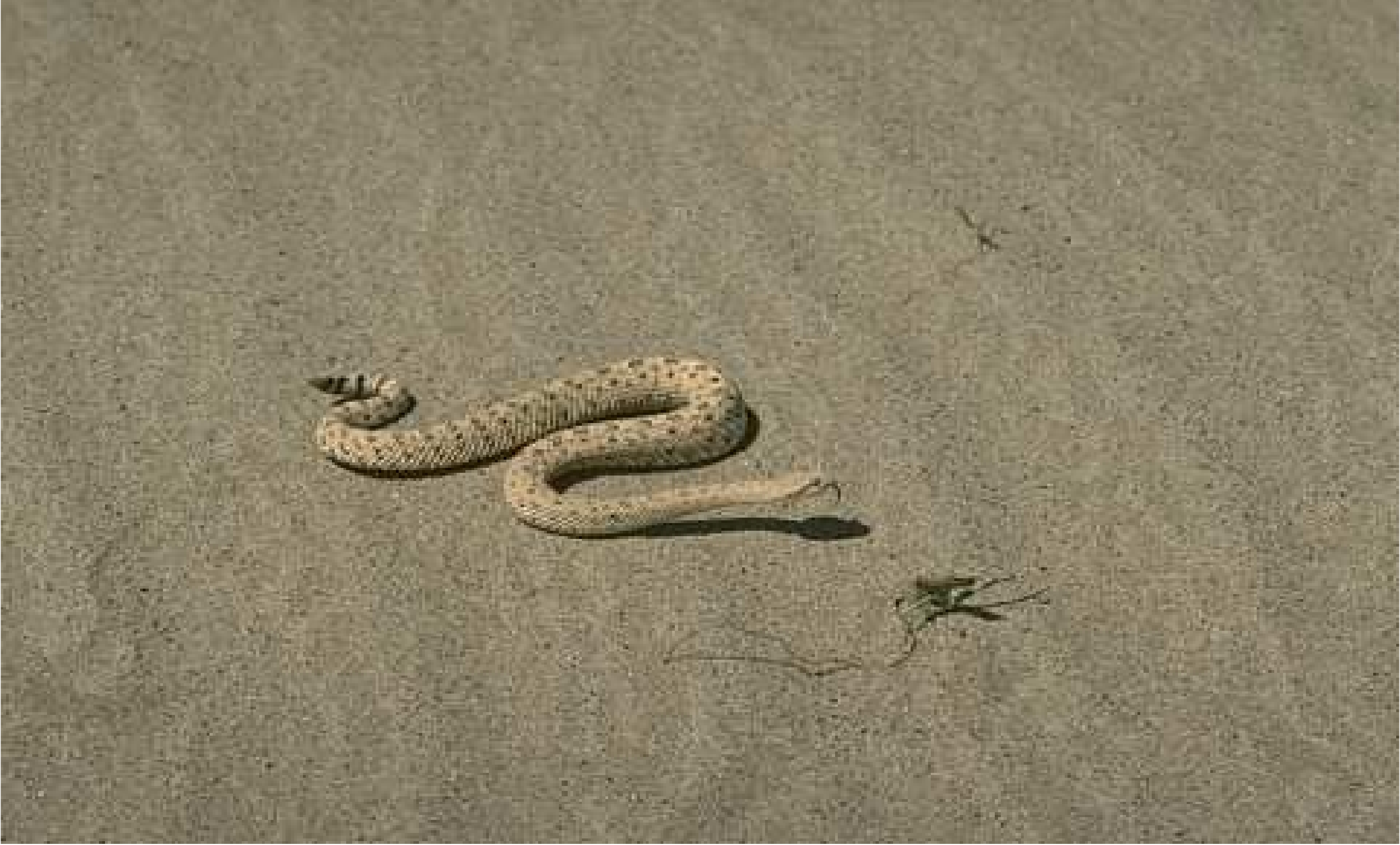}
  \end{subfigure}%
  \begin{subfigure}[b]{.20\linewidth}
    \centering
    \includegraphics[width=.99\textwidth]{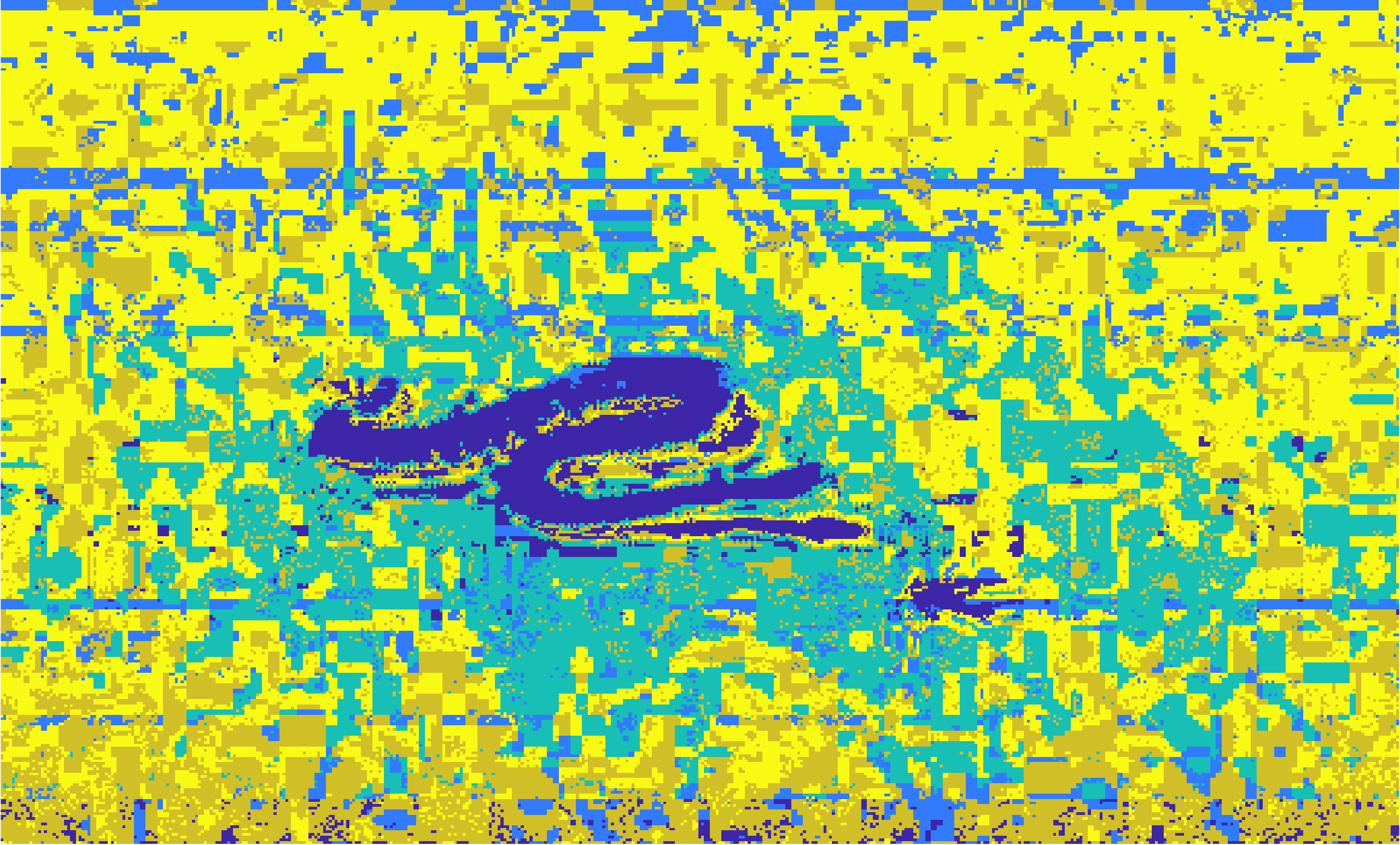}
  \end{subfigure}%
    \begin{subfigure}[b]{.20\linewidth}
    \centering
    \includegraphics[width=.99\textwidth]{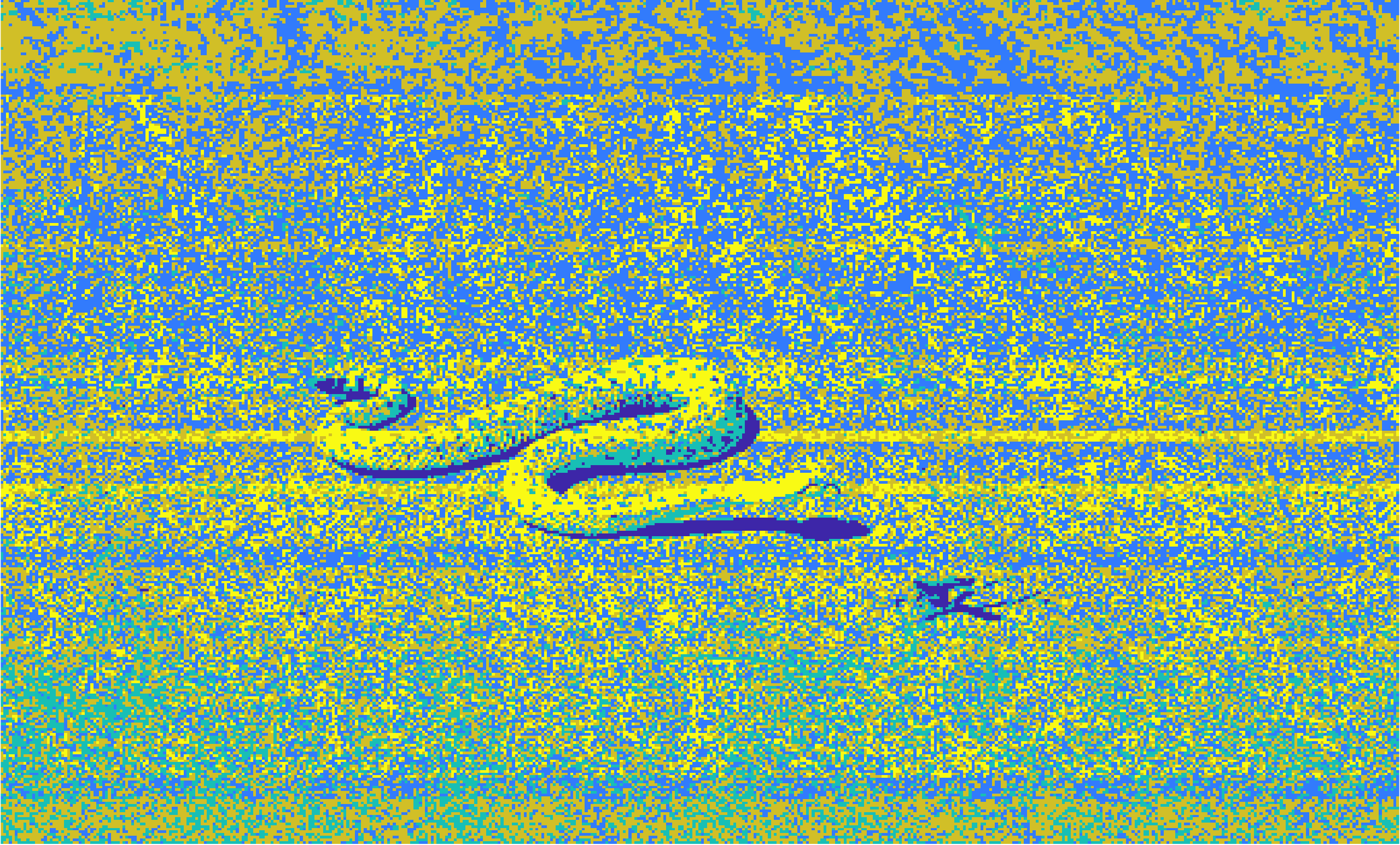}
  \end{subfigure}%
    \begin{subfigure}[b]{.20\linewidth}
    \centering
    \includegraphics[width=.99\textwidth]{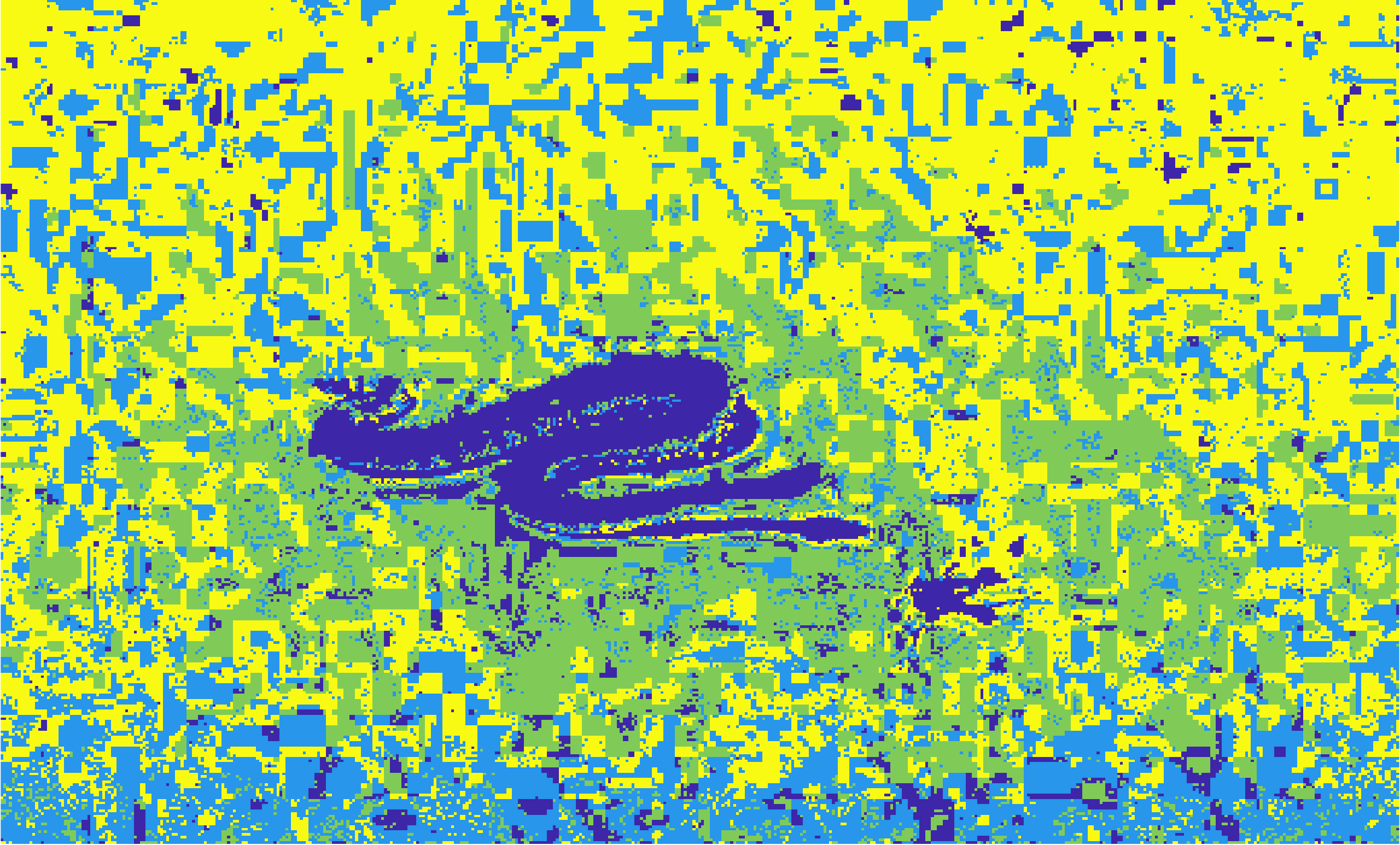}
  \end{subfigure}%
  \begin{subfigure}[b]{.20\linewidth}
    \centering
    \includegraphics[width=.99\textwidth]{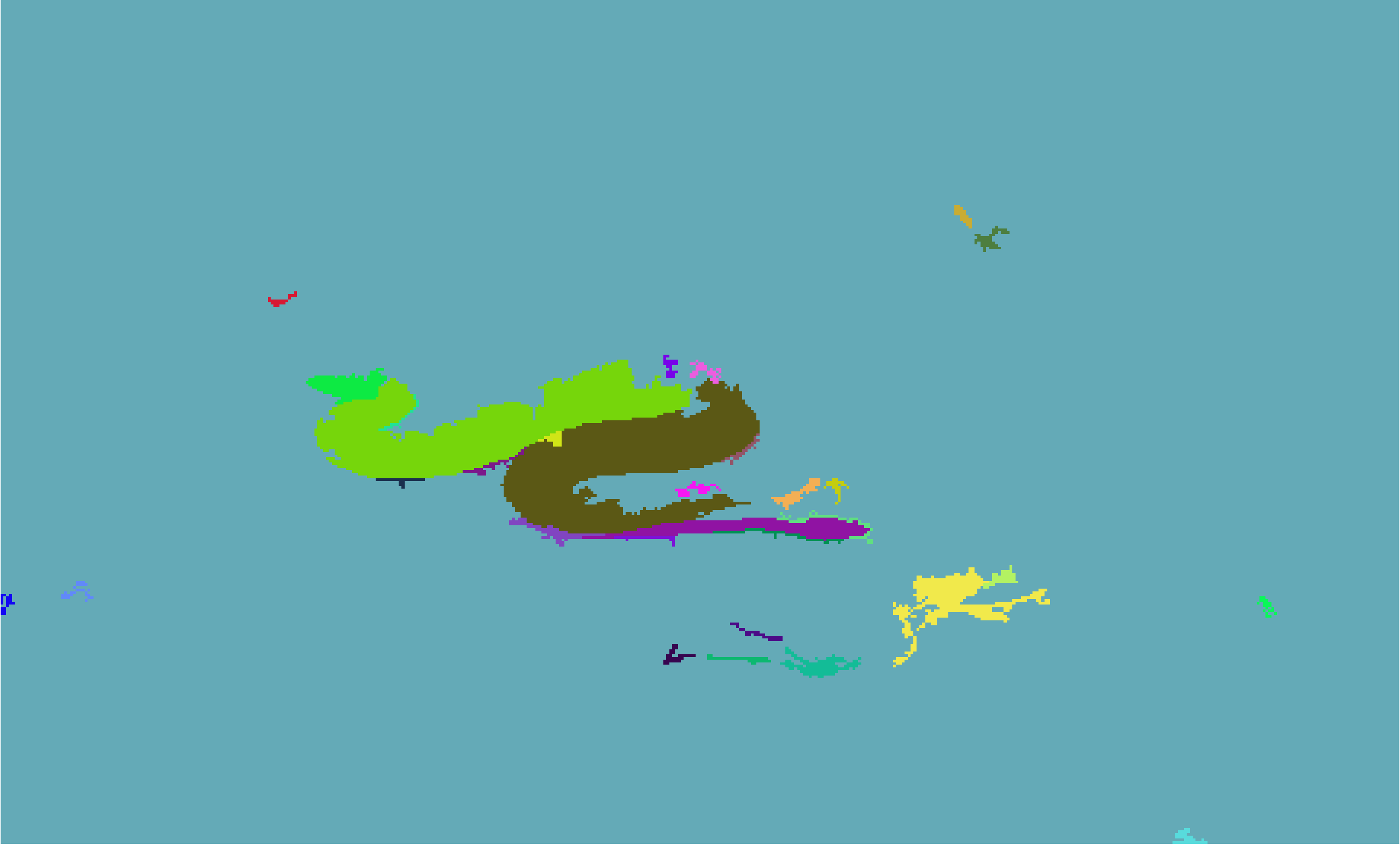}
  \end{subfigure}\\%


       \begin{subfigure}[b]{.20\linewidth}
    \centering
    \includegraphics[width=.99\textwidth]{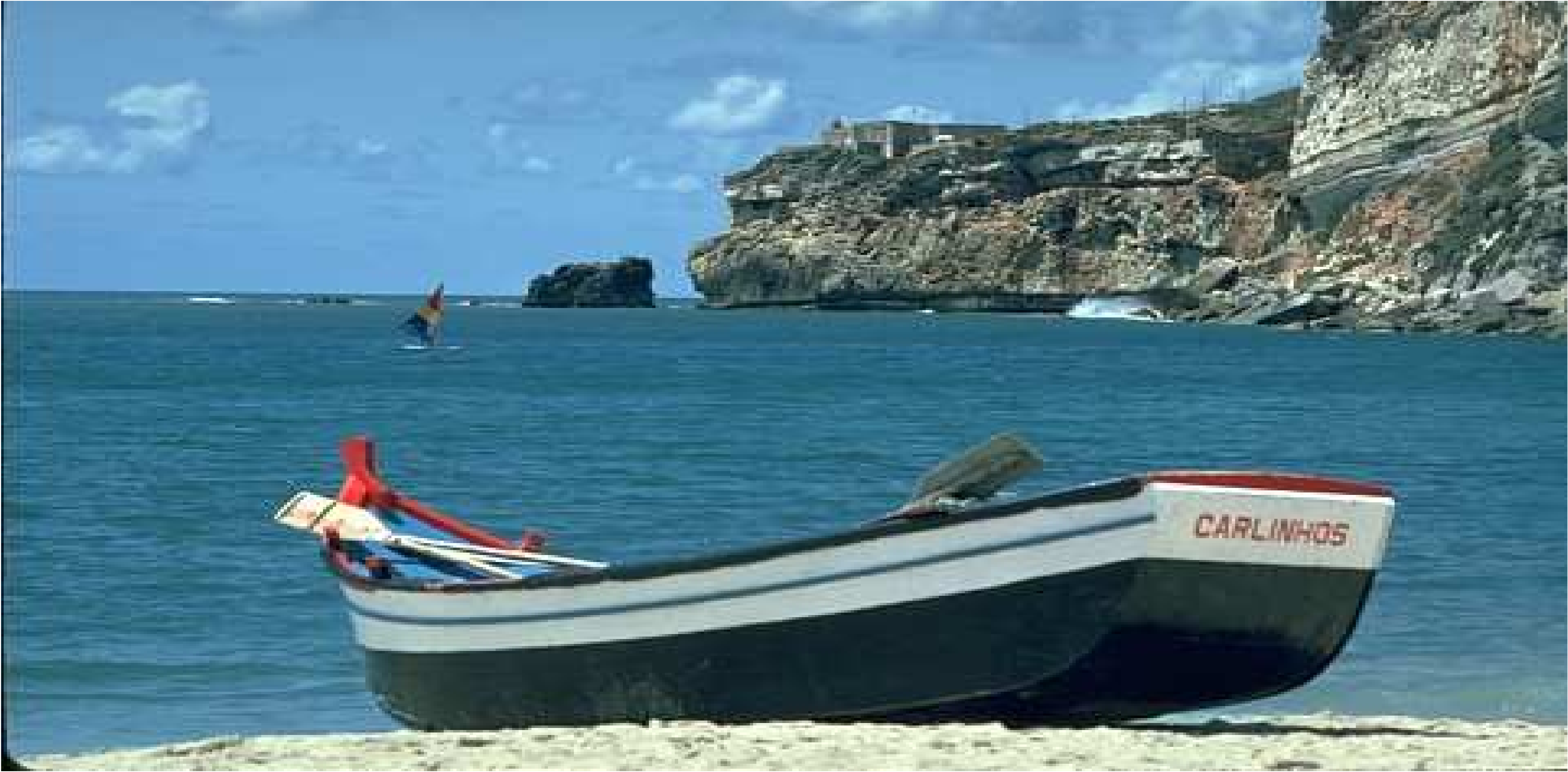}
  \end{subfigure}%
  \begin{subfigure}[b]{.20\linewidth}
    \centering
    \includegraphics[width=.99\textwidth]{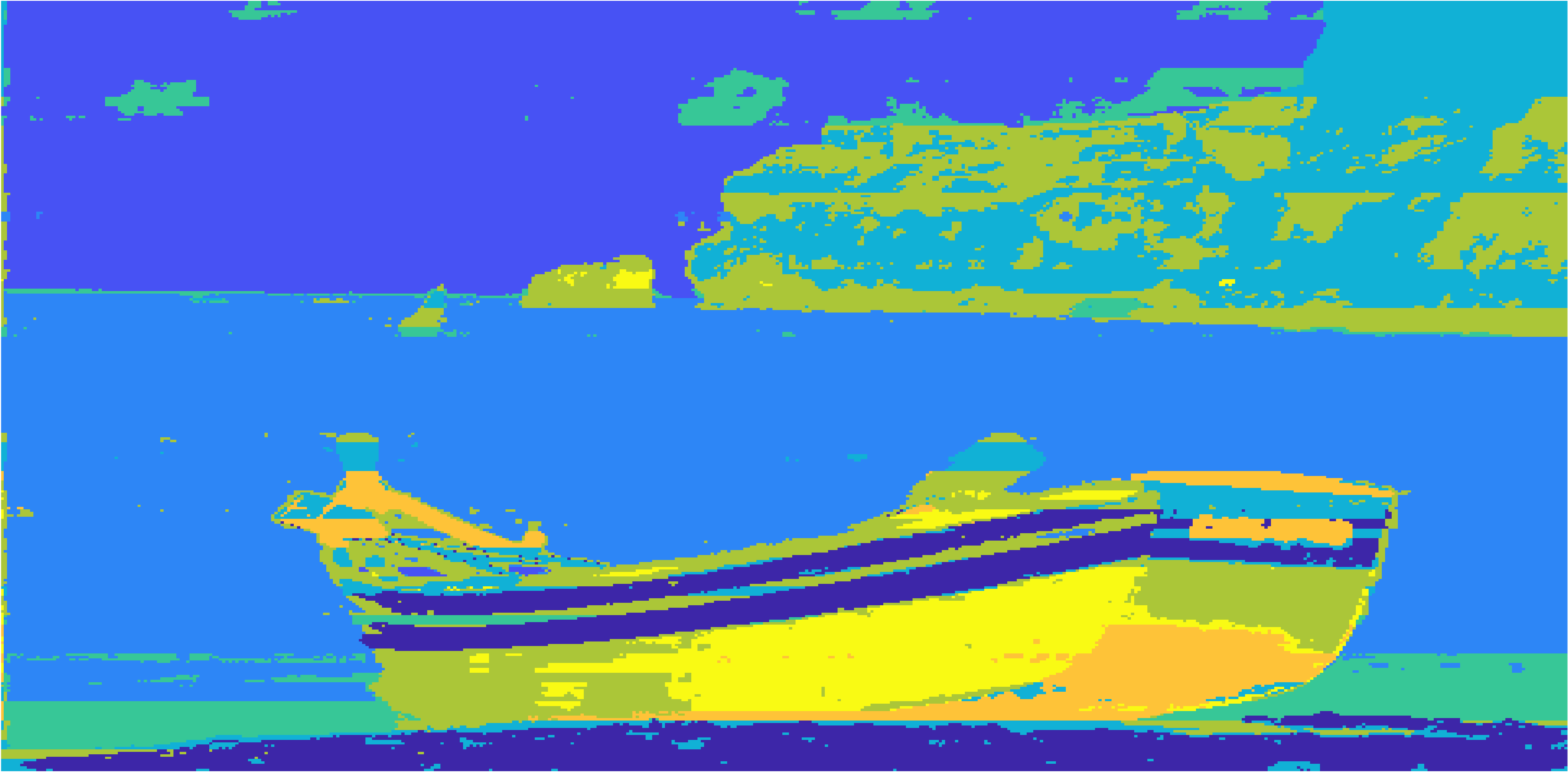}
  \end{subfigure}%
    \begin{subfigure}[b]{.20\linewidth}
    \centering
    \includegraphics[width=.99\textwidth]{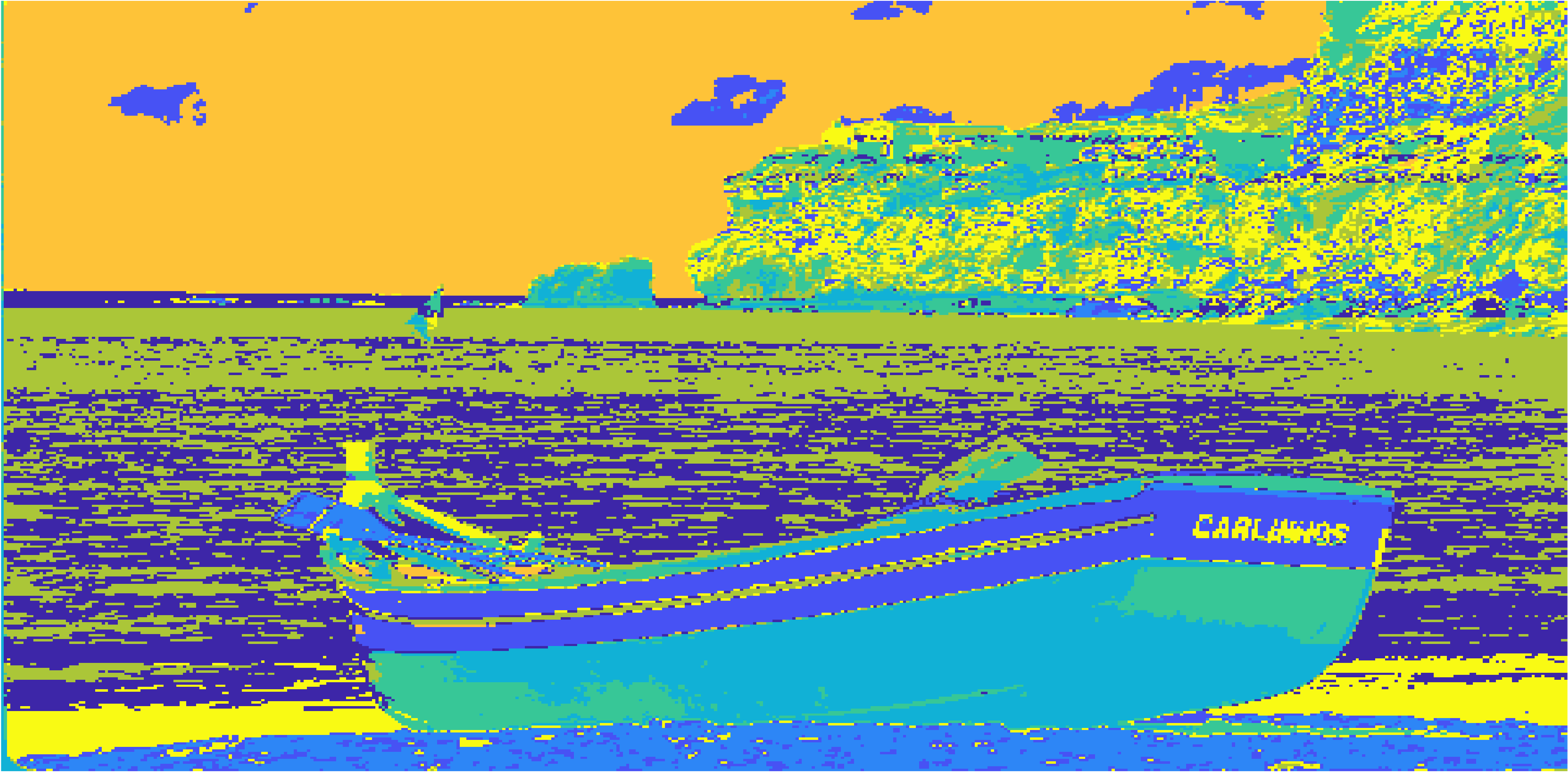}
  \end{subfigure}%
    \begin{subfigure}[b]{.20\linewidth}
    \centering
    \includegraphics[width=.99\textwidth]{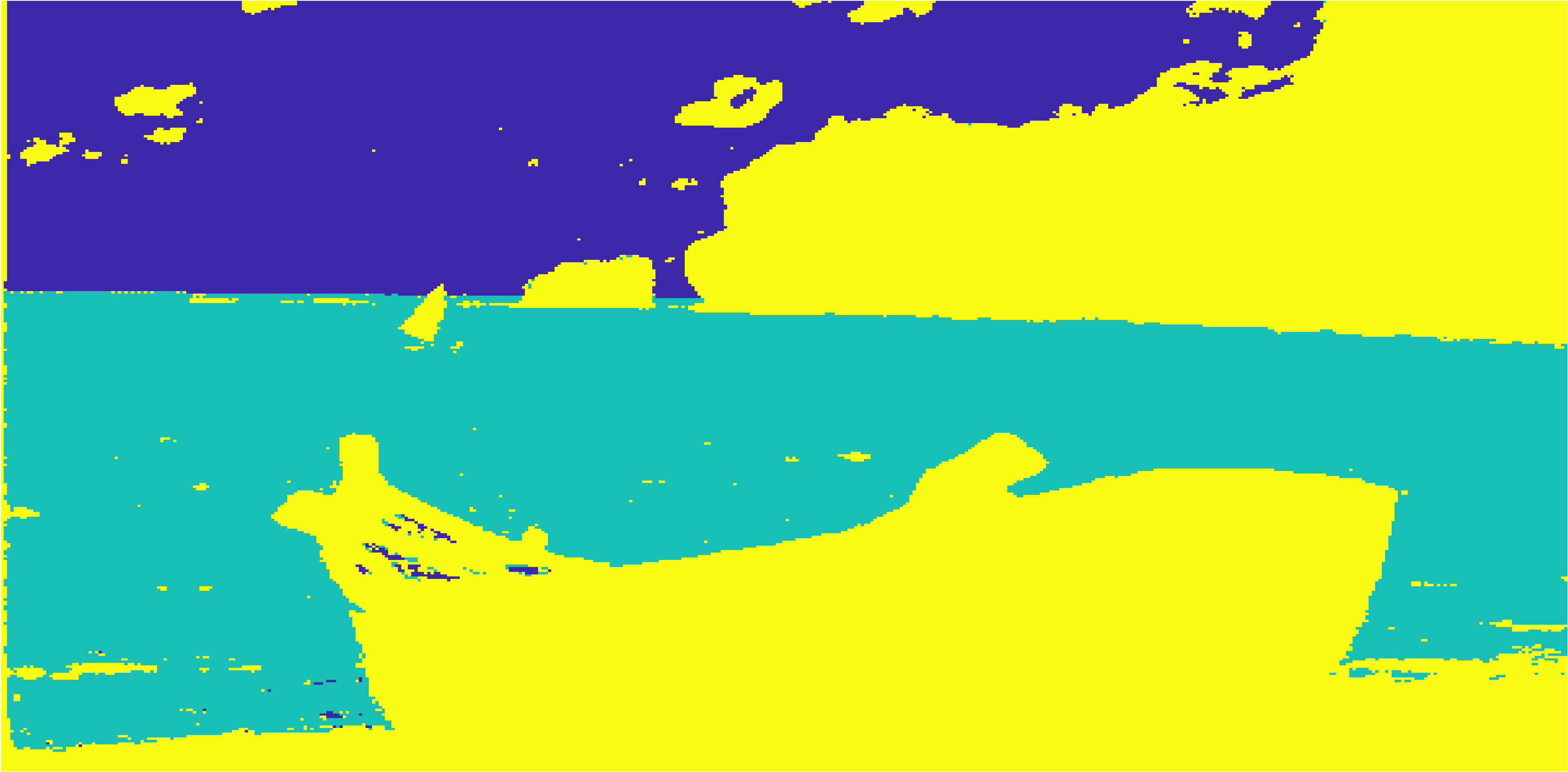}
  \end{subfigure}%
  \begin{subfigure}[b]{.20\linewidth}
    \centering
    \includegraphics[width=.99\textwidth]{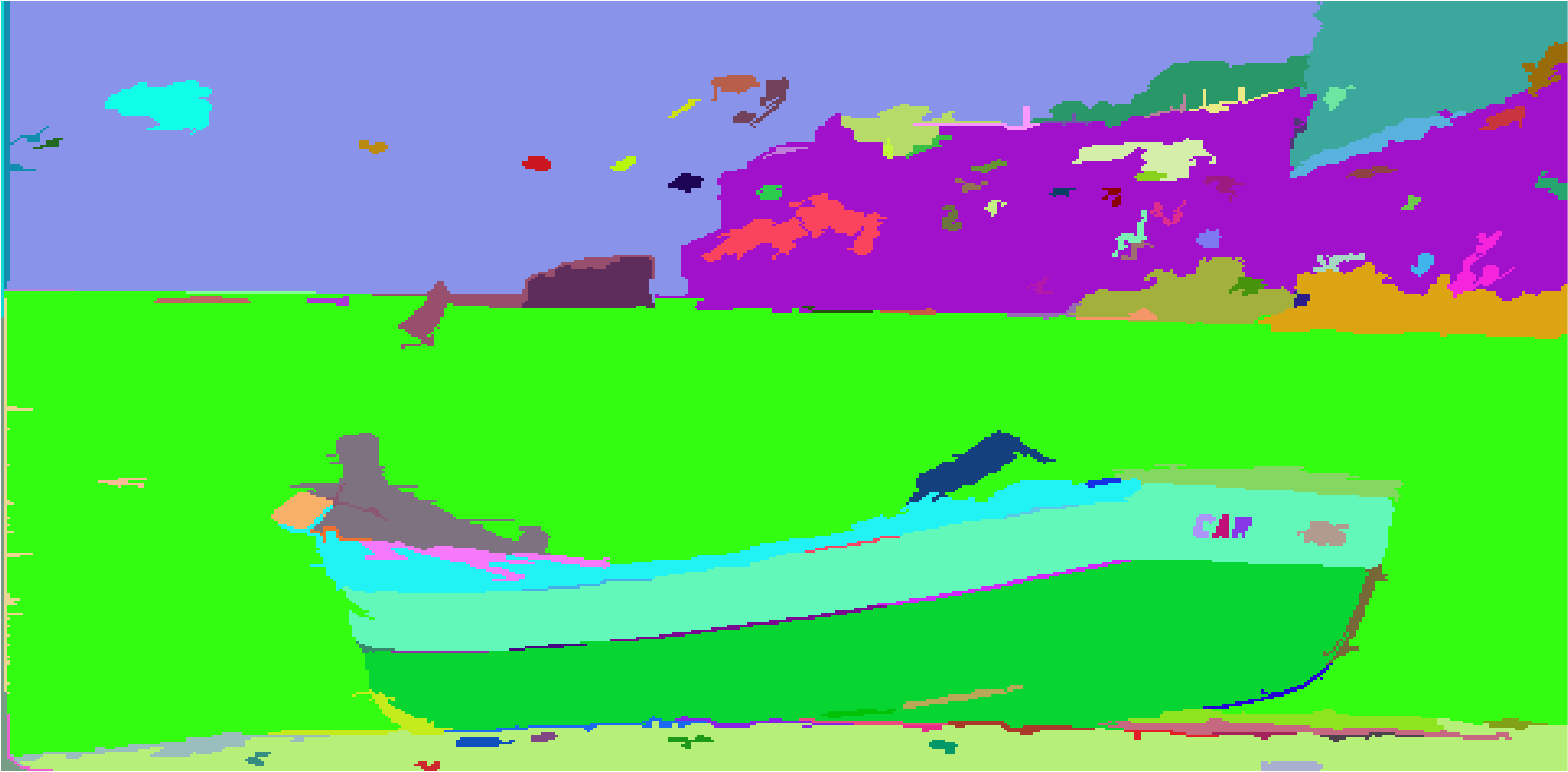}
  \end{subfigure}\\%


  \begin{subfigure}[b]{.20\linewidth}
    \centering
    \includegraphics[width=.99\textwidth]{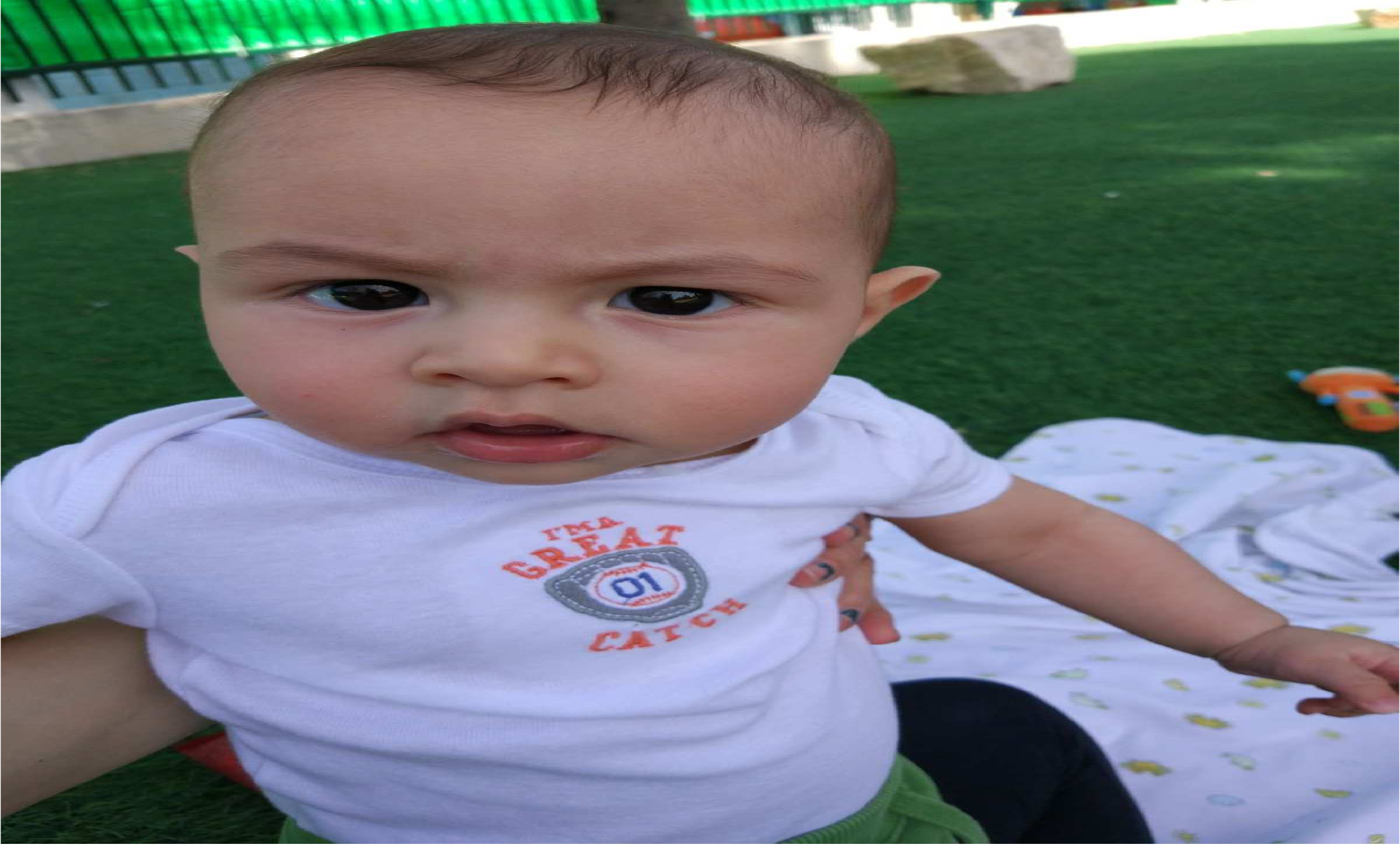}
  \end{subfigure}%
  \begin{subfigure}[b]{.20\linewidth}
    \centering
    \includegraphics[width=.99\textwidth]{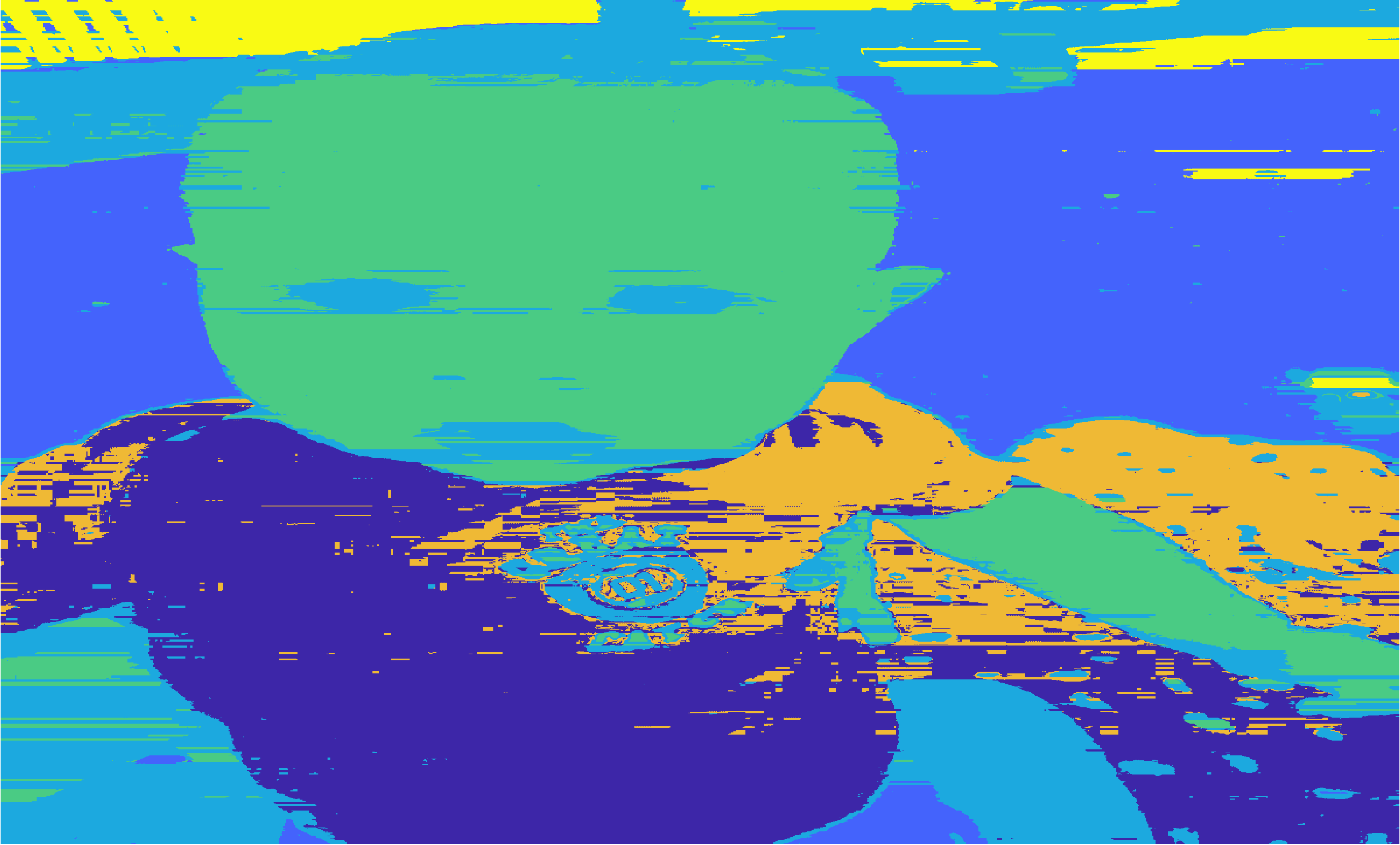}
  \end{subfigure}%
    \begin{subfigure}[b]{.20\linewidth}
    \centering
    \includegraphics[width=.99\textwidth]{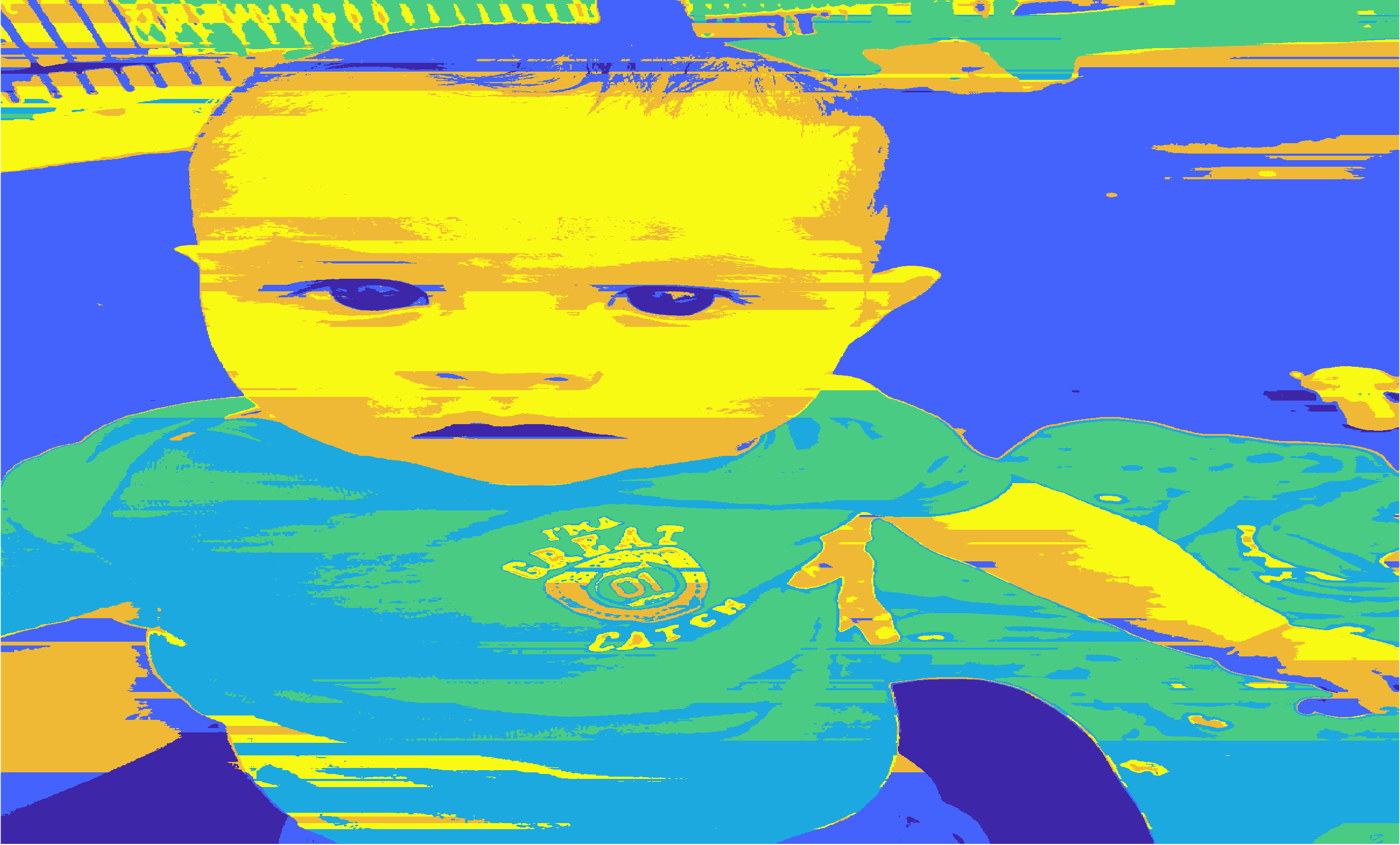}
  \end{subfigure}%
    \begin{subfigure}[b]{.20\linewidth}
    \centering
    \includegraphics[width=.99\textwidth]{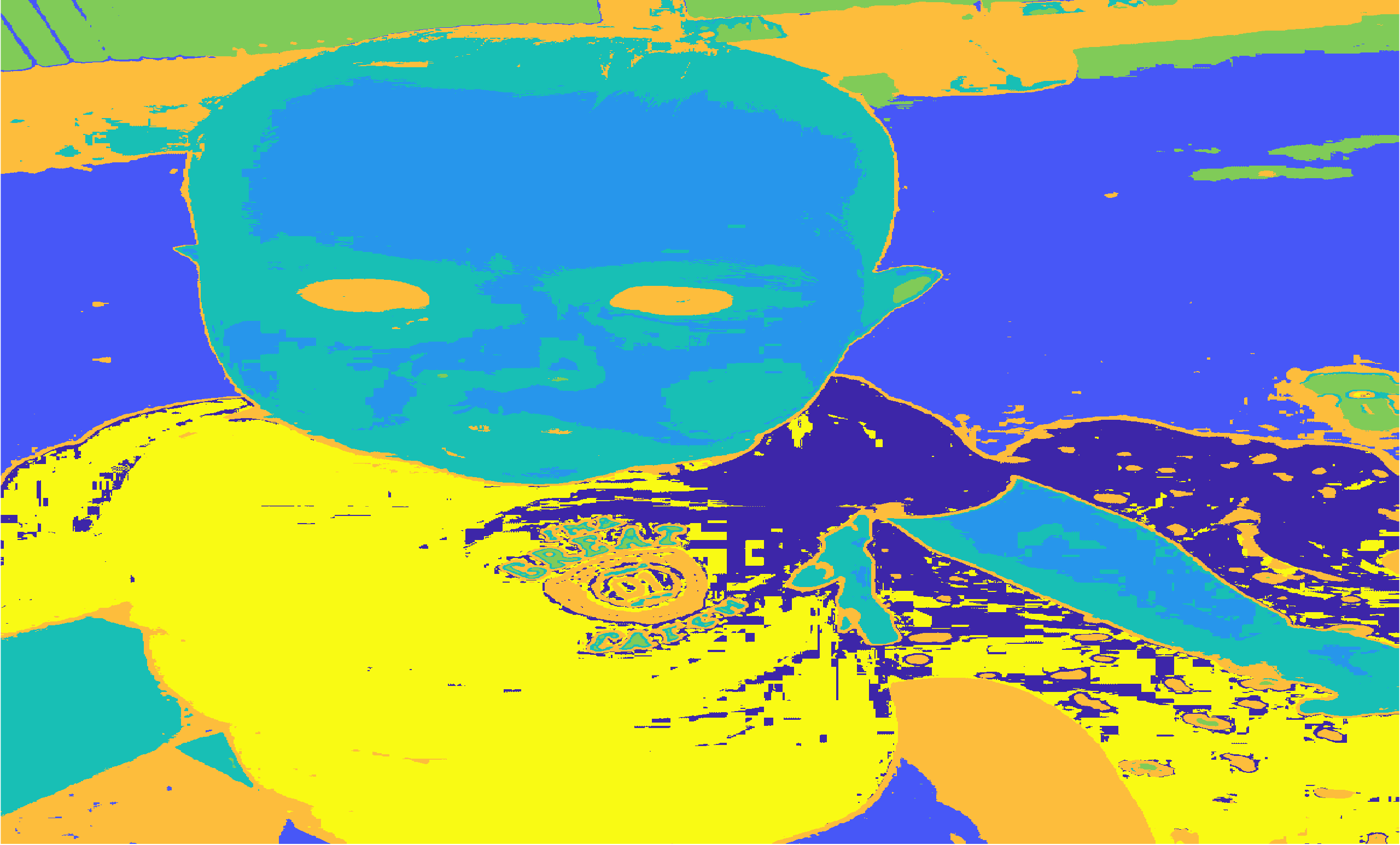}
  \end{subfigure}%
  \begin{subfigure}[b]{.20\linewidth}
    \centering
    \includegraphics[width=.99\textwidth]{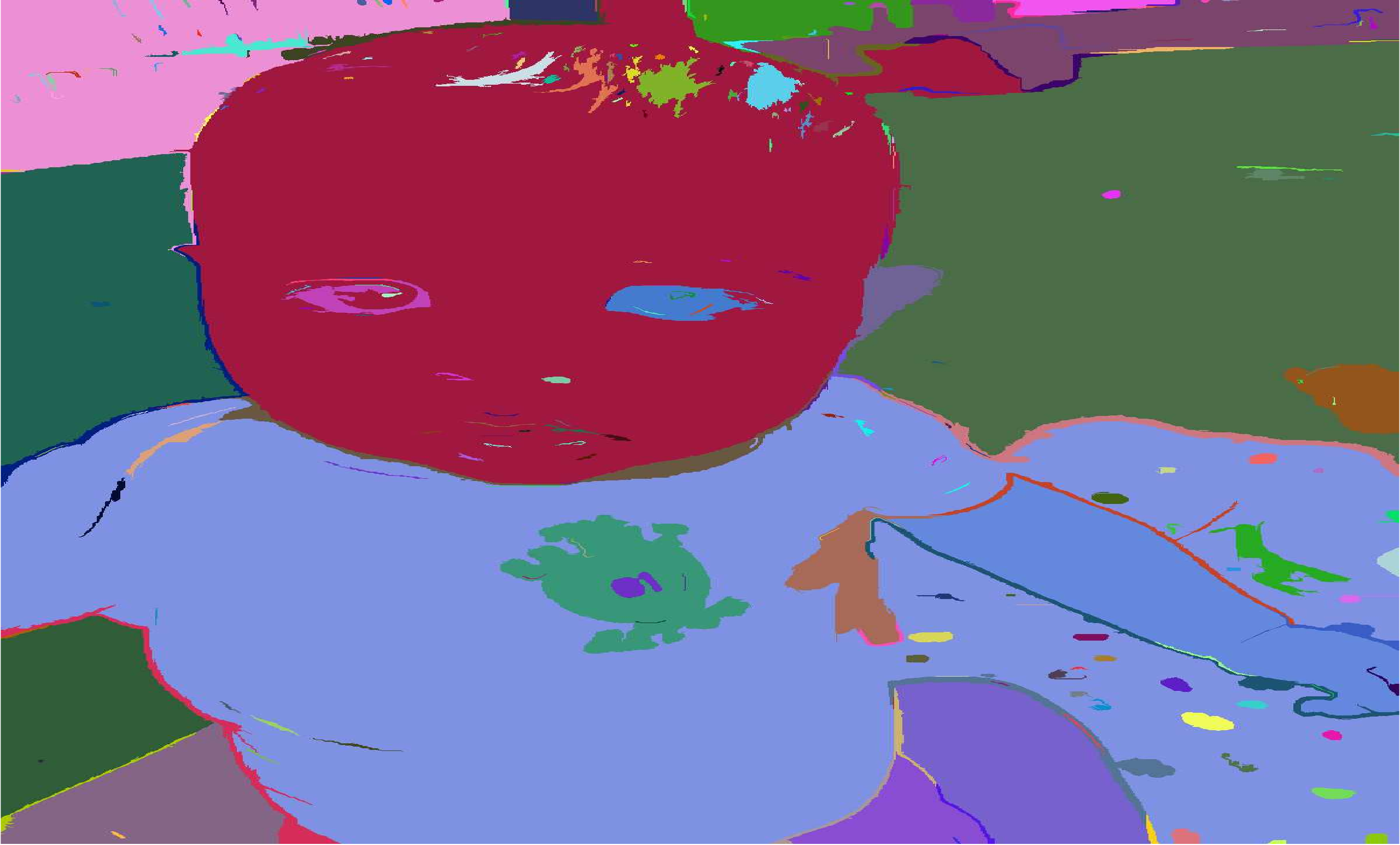}
  \end{subfigure}\\%

  
         \begin{subfigure}[b]{.20\linewidth}
    \centering
    \includegraphics[width=.99\textwidth]{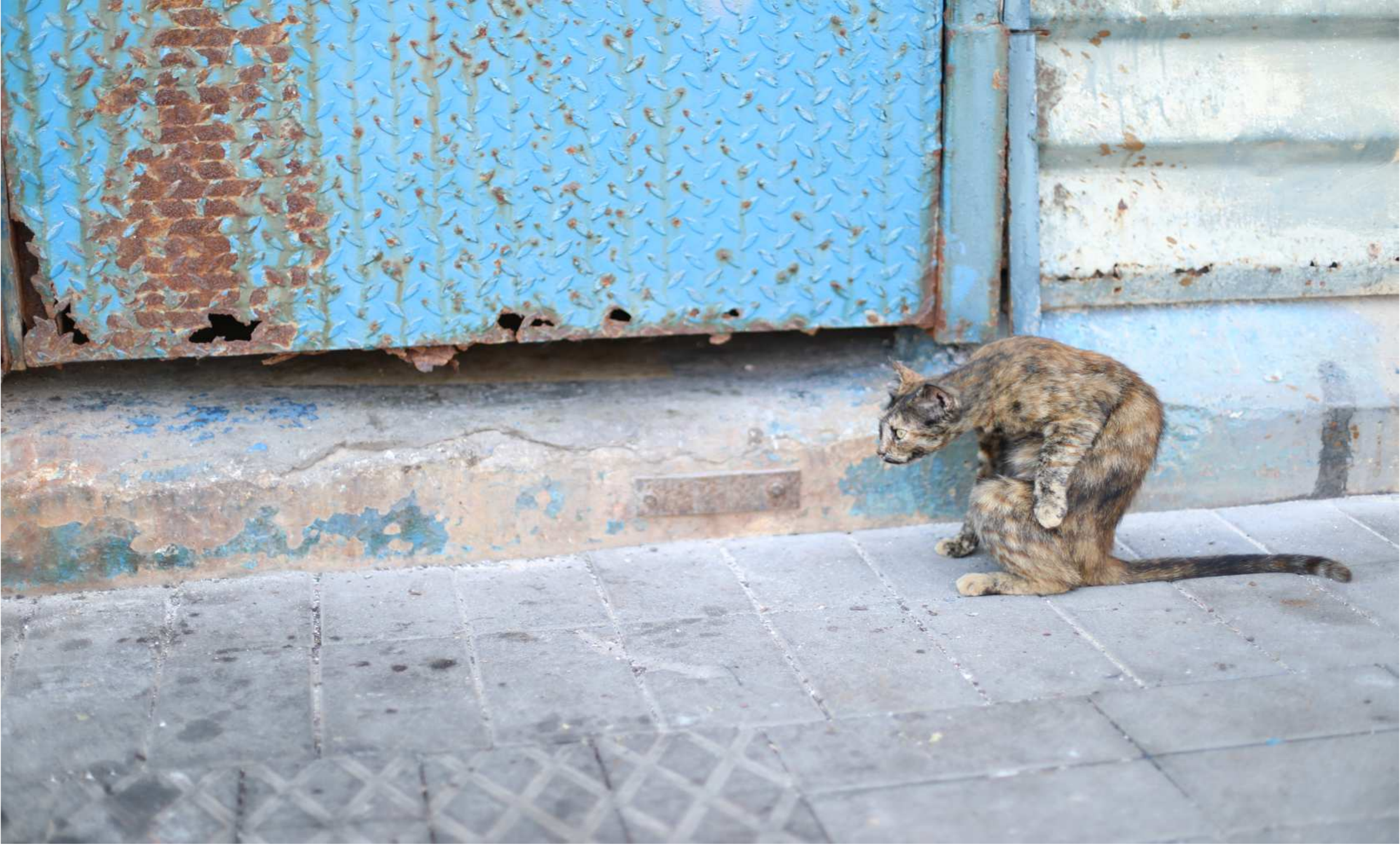}
    \caption{Image (RGB)}
  \end{subfigure}%
  \begin{subfigure}[b]{.20\linewidth}
    \centering
    \includegraphics[width=.99\textwidth]{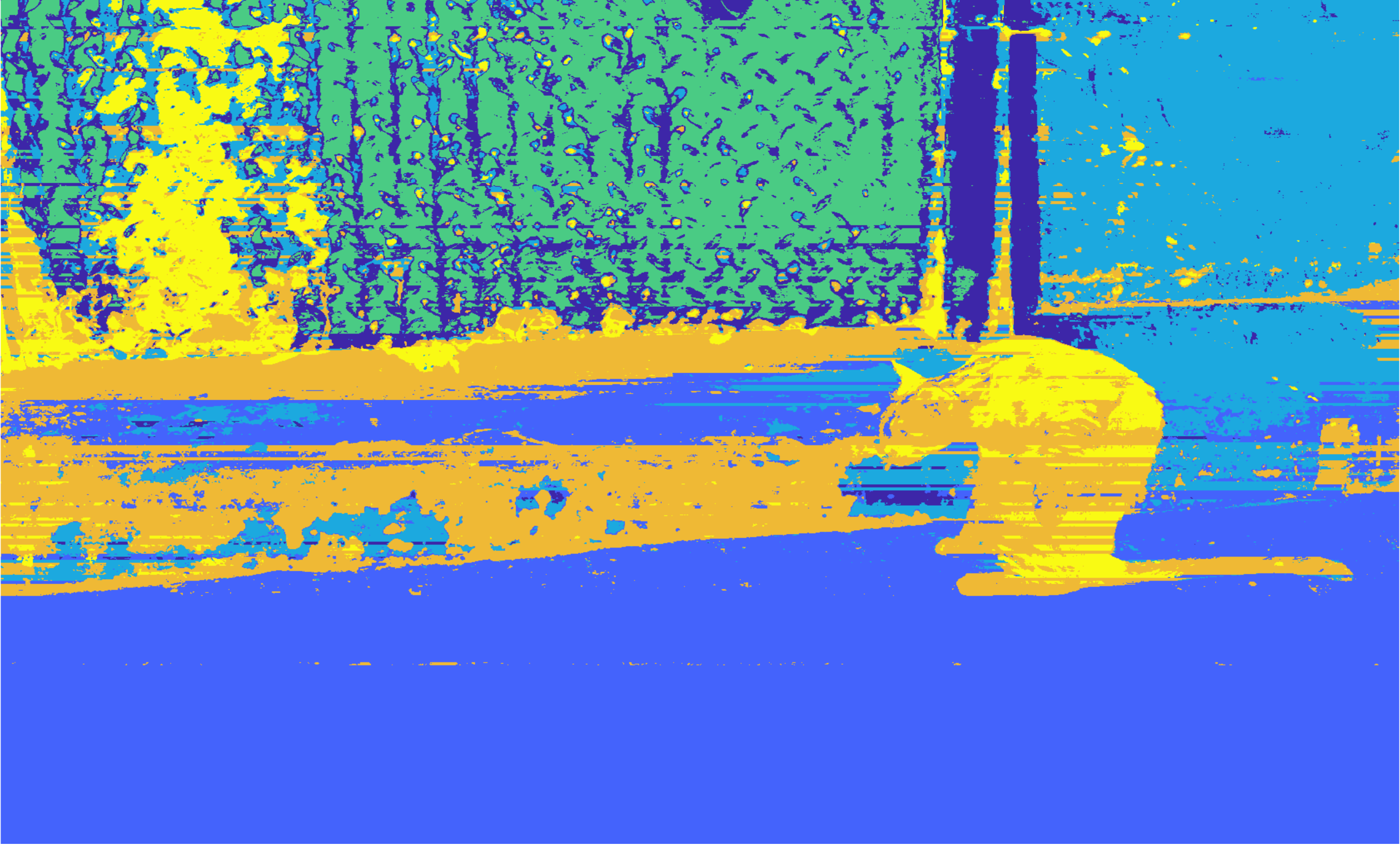}
    \caption{OPWG }\label{fig:1c}
  \end{subfigure}%
    \begin{subfigure}[b]{.20\linewidth}
    \centering
    \includegraphics[width=.99\textwidth]{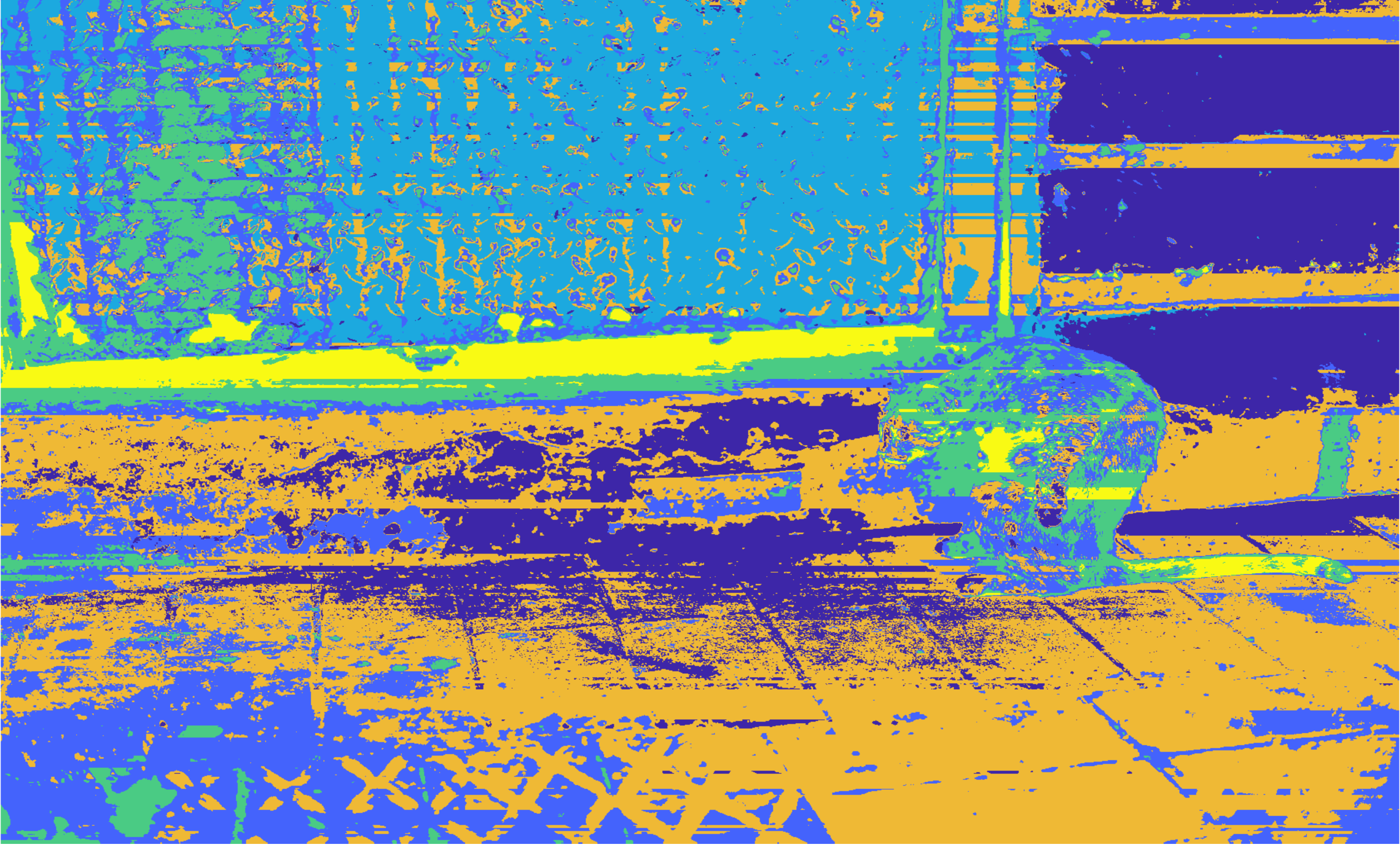}
    \caption{OFCM }\label{fig:1c}
  \end{subfigure}%
    \begin{subfigure}[b]{.20\linewidth}
    \centering
    \includegraphics[width=.99\textwidth]{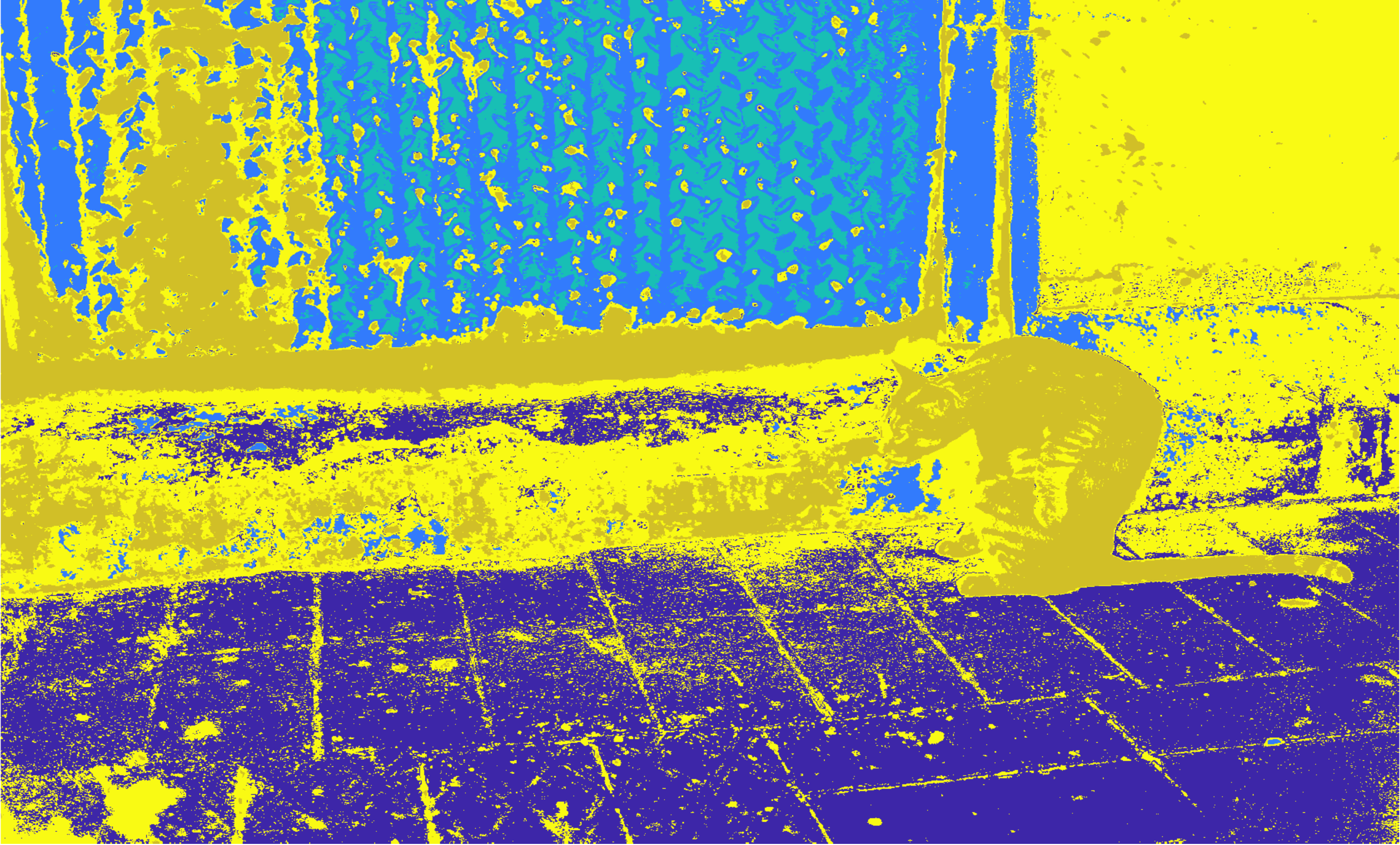}
    \caption{PGMM }\label{fig:1c}
  \end{subfigure}%
  \begin{subfigure}[b]{.20\linewidth}
    \centering
    \includegraphics[width=.99\textwidth]{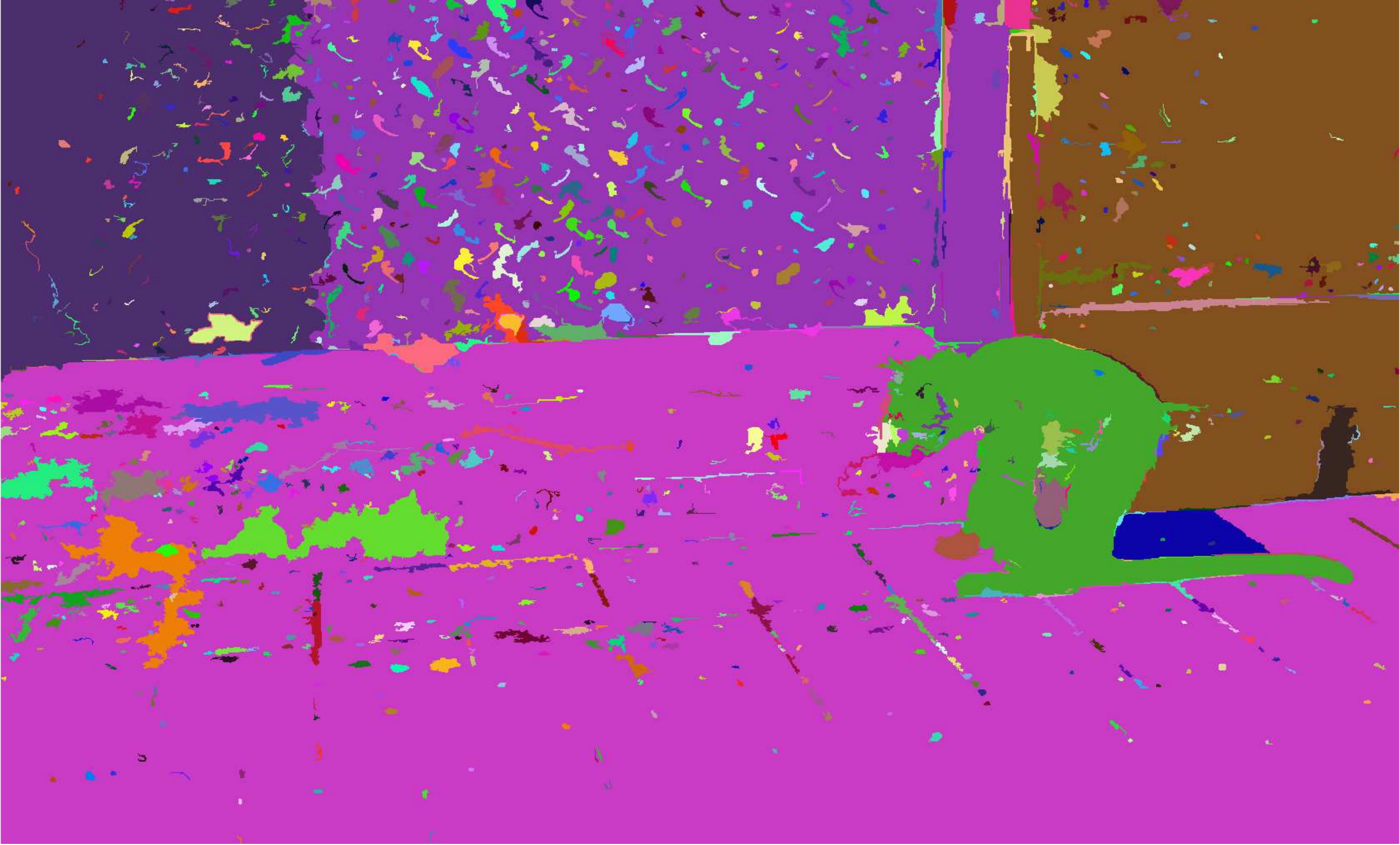}
    \caption{EGBIS }\label{fig:1d}
  \end{subfigure}\\%
  \caption{Color segmentation results for different images from BSD dataset. Note that the last two images are not from BSD dataset. Those images were taken by a mobile phone camera. Their resolution is $1600 \times 1200$ }
  \label{fig:image_sementation}
\end{figure*}

Figure \ref{fig:image_sementation} shows the evaluation results. As can be seen, even though the OPWG is solely based on the LAB features, it performs pretty well in segmenting the images. Comparing to the OFCM algorithm, the OPWG algorithm shows equal or better results although it was initialized with a bigger number of clusters -  $K_{max}=25$.   Even though it uses only four non-overlapping rows at each batch, the OPWG algorithm is able to achieve similar results to those of PGMM, which uses the full dataset. The OPWG algorithm sometimes even outperforms the EGBIS algorithm.

\section{Conclusions}
\label{sec:conc}
This paper proposes a penalized weighted GMM algorithm for online clustering. The algorithm is able to identify the underlying structure of the data and to estimate the number of clusters in a dynamic environment. Although the algorithm is initialized with an arbitrarily large number of clusters, results on both synthetic and real datasets show that the proposed algorithm produces partitions that are close to those you would get if you clustered the whole data at once.

Although the algorithm performs well, there is still work to be done. Better results may be achieved using different penalty functions. Moreover, since the algorithm was evaluated on datasets with low dimensionality, the effect of the "curse of dimensionality" on the algorithm is unknown. In addition, an extension of the algorithm to handle the property of data fading of streaming data should be considered.  Other directions maybe include an extension of the algorithm to deep learning architecture using RNN.

\pagebreak

{\small
\bibliographystyle{ieee}
\bibliography{bib}

}

\end{document}